%% file: main.tex
\documentclass{article}
\input{macros}

\title{Neural Ensemble Search\\ for Uncertainty Estimation and Dataset Shift}

\author{Sheheryar Zaidi$^1$$^*$\hspace{4mm} Arber Zela$^2$\thanks{Equal contribution.}\hspace{1.5mm} \hspace{4mm} Thomas Elsken$^{3, 2}$ \vspace{1mm} \\ \textbf{Chris Holmes$^1$\hspace{5mm} Frank Hutter$^{2, 3}$\hspace{4mm} Yee Whye Teh$^1$} \vspace{2mm} \\
 $^1$University of Oxford,
 $^2$University of Freiburg,
 $^3$Bosch Center for Artificial Intelligence \\
 $\texttt{\{szaidi, cholmes, y.w.teh\}@stats.ox.ac.uk}$,\\
 $\texttt{\{zelaa, fh\}@cs.uni-freiburg.de, thomas.elsken@de.bosch.com}$
}

\begin{document}
\maketitle

\input{text/0_abstract}
\input{text/1_introduction}

\input{text/2_related_work}
\input{text/3_diversity} 
\input{text/4_nes}
\input{text/5_experiments}

\input{text/6_conclusions}

\clearpage
\begin{ack}
AZ, TE and FH acknowledge support by the European Research Council (ERC) under the European Union Horizon 2020 research and innovation programme through grant no. 716721, and by BMBF grant DeToL. SZ acknowledges support from Aker Scholarship. CH wishes to acknowledge support from The Alan Turing Institute, The Medical Research Council UK, and the EPSRC Bayes4Health grant. The authors acknowledge support from ELLIS. We also thank Julien Siems for providing a parallel implementation of regularized evolution and Bobby He for useful comments on the manuscript.
\end{ack}

{\small
\bibliography{biblio}
\bibliographystyle{plain}
}

\clearpage
\input{text/B_supplementary}

\end{document}

%% file: macros.tex
\PassOptionsToPackage{numbers, compress}{natbib}

\usepackage[final]{neurips_2021}

\usepackage[utf8]{inputenc} %
\usepackage[T1]{fontenc}    %
\usepackage{hyperref}       %
\usepackage{url}            %
\usepackage{booktabs}       %
\usepackage{amsfonts}       %
\usepackage{nicefrac}       %
\usepackage{microtype}      %

\usepackage{amsthm}
\usepackage{amsmath}
\usepackage{amssymb}
\usepackage{graphicx}
\usepackage{subcaption}
\usepackage{color}
\usepackage{xcolor}
\usepackage{multirow}
\usepackage{enumitem}
\usepackage{wrapfig}
\usepackage{textcomp}
\usepackage{rotating}
\usepackage[ruled, linesnumbered, vlined]{algorithm2e} 
\SetArgSty{textnormal}

\usepackage{accents}
\usepackage{bm} 

\usepackage[graphicx]{realboxes}

\def\etal{\emph{et al.}\xspace}
\usepackage{tikz}
\usetikzlibrary{fadings}
\usetikzlibrary{patterns}
\usetikzlibrary{shadows.blur}
\usetikzlibrary{shapes}

\theoremstyle{definition}

\newtheorem{proposition}{Proposition}[section]

\DeclareMathOperator*{\argmin}{argmin}

\usepackage[normalem]{ulem}

\newcommand{\ens}{\small{\texttt{Ensemble}}}

\newcommand{\Dtrain}{\mathcal{D}_\text{train}}
\newcommand{\Dval}{\mathcal{D}_\text{val}}
\newcommand{\Dtest}{\mathcal{D}_\text{test}}
\newcommand{\Dvals}{\mathcal{D}_\text{val}^\text{shift}}
\newcommand{\Dtests}{\mathcal{D}_\text{test}^\text{shift}}

\newcommand{\bx}{\bm{x}}
\newcommand{\bbR}{\mathbb{R}}

\newcommand{\inputdim}{D}

\newcommand{\numclass}{C}
\newcommand{\budget}{K}
\newcommand{\loss}{\ell}
\newcommand{\Loss}{\mathcal{L}}
\newcommand{\oracens}{F_\text{OE}}
\newcommand{\archss}{\mathcal{A}}
\newcommand{\numdatapts}{N}
\newcommand{\esa}{\small{\texttt{ForwardSelect}}}
\newcommand{\topm}{\small{\texttt{TopM}}}
\newcommand{\stacking}{\small{\texttt{Stacking}}}
\newcommand{\quickgreedy}{\small{\texttt{QuickAndGreedy}}}

\newcommand{\pool}{\mathcal{P}}
\newcommand{\popsize}{P}
\newcommand{\popul}{\mathfrak{p}}
\newcommand{\numcand}{m}

\newcommand{\insixteen}{ImageNet-16-120 }

\usepackage{array}
\newcolumntype{L}[1]{>{\raggedright\let\newline\\\arraybackslash\hspace{0pt}}m{#1}}
\newcolumntype{C}[1]{>{\centering\let\newline\\\arraybackslash\hspace{0pt}}m{#1}}
\newcolumntype{R}[1]{>{\raggedleft\let\newline\\\arraybackslash\hspace{0pt}}m{#1}}

%% file: text/0_abstract.tex
\begin{abstract}

Ensembles of neural networks achieve superior performance compared to stand-alone networks in terms of accuracy, uncertainty calibration and robustness to dataset shift. \emph{Deep ensembles}, a state-of-the-art method for uncertainty estimation, only ensemble random initializations of a \emph{fixed} architecture. Instead, we propose two methods for automatically constructing ensembles with \emph{varying} architectures, which implicitly trade-off individual architectures' strengths against the ensemble's diversity and exploit architectural variation as a source of diversity. On a variety of classification tasks and modern architecture search spaces, we show that the resulting ensembles outperform deep ensembles not only in terms of accuracy but also uncertainty calibration and robustness to dataset shift. Our further analysis and ablation studies provide evidence of higher ensemble diversity due to architectural variation, resulting in ensembles that can outperform deep ensembles, even when having weaker average base learners. To foster reproducibility, our code is available: \url{https://github.com/automl/nes}

\end{abstract} 

%% file: text/1_introduction.tex
\section{Introduction}
\label{sec:introduction}

Some applications of deep learning rely only on point estimate predictions made by a neural network. However, many critical applications also require reliable predictive uncertainty estimates and robustness under the presence of dataset shift, that is, when the observed data distribution at deployment differs from the training data distribution.
Examples include medical imaging~\citep{esteva2017dermatologistlevel} and self-driving cars~\citep{Bojarski16}. Unfortunately, several studies have shown that neural networks are not always robust to dataset shift \citep{ovadia19, hendrycks2018benchmarking}, nor do they exhibit calibrated predictive uncertainty, resulting in incorrect predictions made with high confidence \citep{pmlr-v70-guo17a}.

\textit{Deep ensembles} \citep{lakshminarayanan2017} achieve state-of-the-art results for predictive uncertainty calibration and robustness to dataset shift. Notably, they have been shown to outperform various approximate Bayesian neural networks~\citep{lakshminarayanan2017, ovadia19, gustafsson2019evaluating}. Deep ensembles are constructed by training a \textit{fixed} architecture multiple times with different random initializations. Due to the multi-modal loss landscape \cite{fort2019deep, wilson2020bayesian}, \textit{randomization} by different initializations induces diversity among the base learners to yield a model with better uncertainty estimates than any of the individual base learners (i.e. ensemble members).  

\begin{wrapfigure}[28]{R}{.465\textwidth} 
\centering
\vspace{-6mm} 
\includegraphics[width=0.99\linewidth]{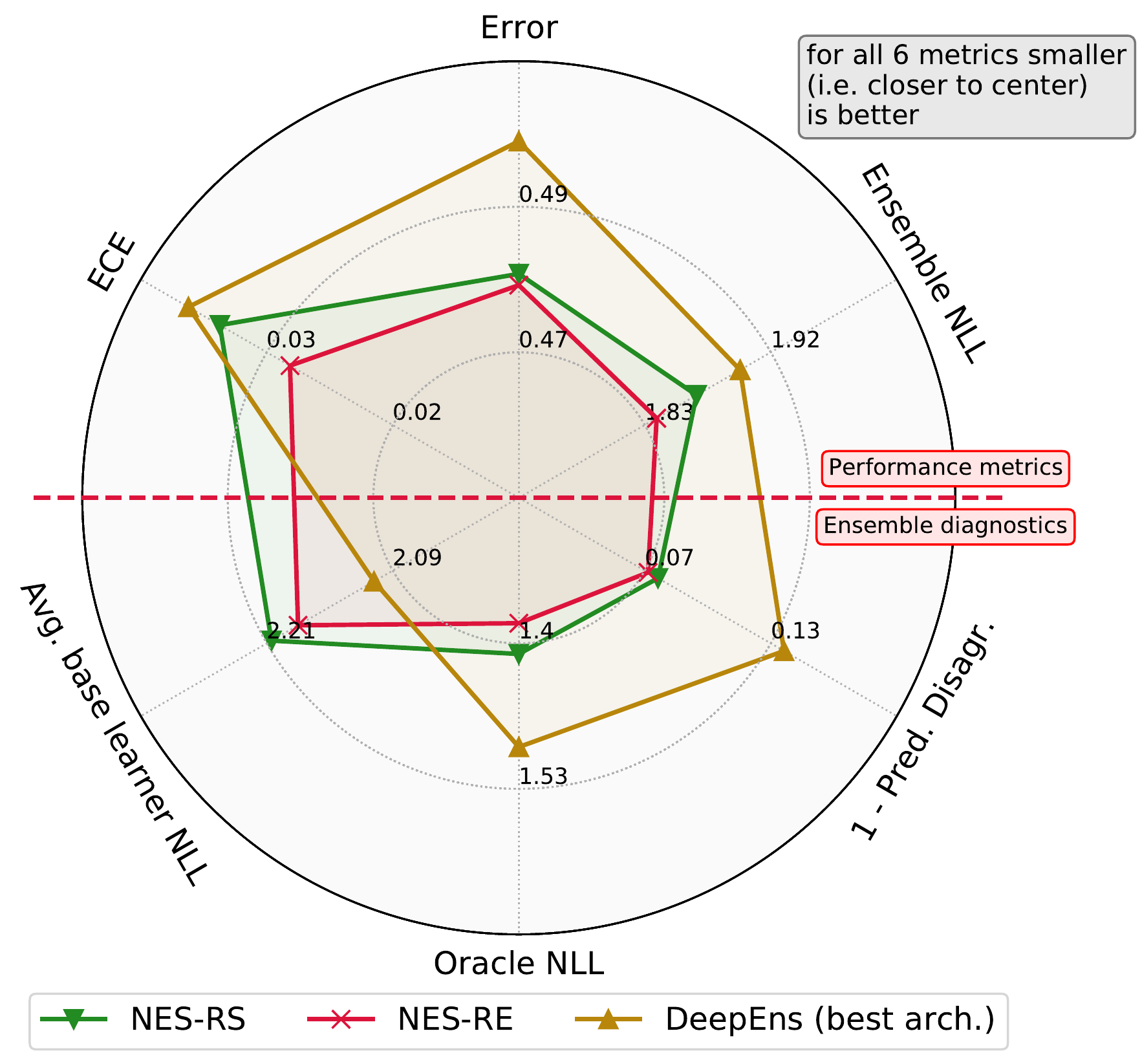}
\caption{A comparison of a deep ensemble with the best architecture (out of 15,625 possible architectures) and ensembles constructed by our method NES on \insixteen over NAS-Bench-201.
Performance metrics (smaller is better): error, negative log-likelihood (NLL) and expected calibration error (ECE). We also measure average base learner NLL and two metrics for ensemble diversity (see Section \ref{sec:diversity_of_ens}): oracle NLL and [$1 - $predictive disagreement]; small means more diversity for both metrics. NES ensembles outperform the deep ensemble, despite the latter having a significantly stronger average base learner.
}
\label{fig:radar_plot}
\end{wrapfigure}

Our work focuses on \textit{automatically} selecting \textit{varying} base learner architectures in the ensemble, exploiting architectural variation as a beneficial source of diversity missing in deep ensembles due to their \textit{fixed} architecture. Such architecture selection during ensemble construction allows a more “ensemble-aware” choice of architectures and is based on data rather than manual biases. As discussed in Section \ref{sec:related_work}, while ensembles with varying architectures has already been explored in the literature, variation in architectures is typically limited to just varying depth and/or width, in contrast to more complex variations, such as changes in the topology of the connections and operations used, as considered in our work. More generally, automatic ensemble construction is well-explored in AutoML \citep{auto-sklearn, bayes-opt-ens, olson_tpot_2016, mendoza-automl16a}. Our work builds on this by demonstrating that, in the context of \textit{uncertainty estimation}, automatically constructed ensembles with varying architectures outperform deep ensembles that use state-of-the-art, or even \textit{optimal}, architectures (Figure \ref{fig:radar_plot}). Studied under controlled settings, we assess the ensembles by various measures, including predictive performance, uncertainty estimation and calibration, base learner performance and two ensemble diversity metrics, showing that architectural variation is beneficial in ensembles. 

Note that, \textit{a priori}, it is not obvious how to find a set of diverse architectures that work well as an ensemble. On the one hand, optimizing the base learners’ architectures in isolation may yield multiple base learners with similar architectures (like a deep ensemble). On the other hand, selecting the architectures randomly may yield numerous base learners with poor architectures harming the ensemble. Moreover, as in neural architecture search (NAS), we face the challenge of needing to traverse vast architectural search spaces. We address these challenges in the problem of \emph{Neural Ensemble Search} (NES), an extension of NAS that aims to find a \emph{set} of complementary architectures that together form a strong ensemble. In summary, our contributions are as follows:

\begin{enumerate}[leftmargin=*]
    \item We present two NES algorithms for automatically constructing ensembles with varying base learner architectures. As a first step, we present NES with random search (NES-RS), which is simple and easily parallelizable. We further propose NES-RE inspired by regularized evolution \citep{real2019regularized}, which evolves a population of architectures yielding performant and robust ensembles.

    \item This work is the first to apply automatic ensemble construction over architectures to \textit{complex}, state-of-the-art neural architecture search spaces.

    \item In the context of uncertainty estimation and robustness to dataset shift, we demonstrate that ensembles constructed by NES improve upon state-of-the-art deep ensembles. We validate our findings over five datasets and two architecture search spaces.
\end{enumerate}

%% file: text/2_related_work.tex
\section{Related Work}
\label{sec:related_work}

\vspace{-5pt}
\textbf{Ensembles and uncertainty estimation.} Ensembles of neural networks \citep{hansen_ensembles, krogh-ens, dietterich_ensemble_methods} are commonly used to boost performance.
In practice, strategies for building ensembles include independently training multiple initializations of the same network, i.e. \textit{deep ensembles} \citep{lakshminarayanan2017}, training base learners on different bootstrap samples of the data \citep{zhou-many-bagging-2002}, training with diversity-encouraging losses \citep{Liu1999, lee-M-heads, diverse-ens-evol, webb2019joint, Jain2020, pearce20} and using checkpoints during the training trajectory of a network \citep{snapshot_ens, loshchilov-iclr17a}. Despite a variety of approaches, Ashukha \etal\citep{Ashukha2020Pitfalls} found many sophisticated ensembling techniques to be equivalent to a small-sized deep ensemble by test performance. 

Much recent interest in ensembles has been due to their state-of-the-art predictive uncertainty estimates, with extensive empirical studies \citep{ovadia19, gustafsson2019evaluating} observing that deep ensembles outperform other approaches for uncertainty estimation, notably including Bayesian neural networks \citep{pmlr-v37-blundell15, pmlr-v48-gal16, sgld11} and post-hoc calibration \citep{pmlr-v70-guo17a}. Although deep ensembles are not, technically speaking, equivalent to Bayesian neural networks and the relationship between the two is not well understood, diversity among base learners in a deep ensemble yields a model which is arguably closer to \textit{exact} Bayesian model averaging than other approximate Bayesian methods that only capture a single posterior mode in a multi-modal landscape \citep{wilson2020bayesian, fort2019deep}. Also, He \etal\citep{he2020bayesian} draw a rigorous link between Bayesian methods and deep ensembles for wide networks, and Pearce \etal\citep{pearce20} propose a technique for approximately Bayesian ensembling. Our primary baseline is deep ensembles as they provide state-of-the-art results in uncertainty estimation. 

\textbf{AutoML and ensembles of varying architectures.} Automatic ensemble construction is commonly used in AutoML \citep{auto-sklearn, automl_book}. Prior work includes use of Bayesian optimization to tune non-architectural hyperparameters of an ensemble's base learners \citep{bayes-opt-ens}, posthoc ensembling of fully-connected networks evaluated by Bayesian optimization \citep{mendoza-automl16a} and building ensembles by iteratively adding (sub-)networks to improve ensemble performance \citep{pmlr-v70-cortes17a, Macko2019ImprovingNA}. Various approaches, including ours, rely on ensemble selection \citep{rich_ens_select}. We also note that Simonyan \& Zisserman \citep{Simonyan15}, He \etal\citep{he2016deep} employ ensembles with varying architectures but \textit{without} automatic construction. Importantly, in contrast to our work, all aforementioned works limit architectural variation to only changing width/depth or fully-connected networks. Moreover, such ensembles have not been considered before in terms of uncertainty estimation, with prior work focusing instead on predictive performance \citep{frachon2020immunecs}. Another important part of AutoML is neural architecture search (NAS), the process of automatically designing \textit{single model} architectures~\citep{elsken_survey}, using strategies such as reinforcement learning~\citep{zoph-iclr17a}, evolutionary algorithms~\citep{real2019regularized} and gradient-based methods~\citep{liu2018darts}. We use the search spaces defined by Liu \etal\citep{liu2018darts} and Dong \& Yang \citep{dong20}, two of the most commonly used ones in recent literature.

Concurrent to our work, Wenzel \etal\citep{wenzel2020hyperparameter} consider ensembles with base learners having varying hyperparameters using an approach similar to NES-RS. However, they focus on non-architectural hyperparameters such as $L_2$ regularization strength and dropout rates, keeping the architecture fixed. As in our work, they also consider predictive uncertainty calibration and robustness to shift, finding similar improvements over deep ensembles.

%% file: text/3_diversity.tex
\vspace{-1pt}
\section{Visualizing Ensembles of Varying Architectures} \label{sec:diversity_of_ens}

\vspace{-5pt}
In this section, we discuss diversity in ensembles with varying architectures and visualize base learner predictions to add empirical evidence to the intuition that architectural variation results in more diversity. We also define two metrics for measuring diversity used later in Section \ref{sec:experiments}. 

\vspace{-5pt}
\subsection{Definitions and Set-up}\label{sec:definitions}

\vspace{-5pt}
Let $\Dtrain = \{(\bx_i, y_i) : i = 1, \dots, \numdatapts \}$ be the training dataset, where the input $\bx_i \in \bbR^\inputdim$ and, assuming a classification task, the output $y_i \in \{1, \dots, \numclass\}$. We use $\Dval$ and $\Dtest$ for the validation and test datasets, respectively. Denote by $f_\theta$ a neural network with weights $\theta$, so $f_\theta(\bx) \in \bbR^\numclass$ is the predicted probability vector over the classes for input $\bx$. Let $\loss(f_\theta(\bx), y)$ be the neural network's loss for data point $(\bx, y)$.
Given $M$ networks $f_{\theta_1}, \dots, f_{\theta_M}$, we construct the \textit{ensemble} $F$ of these networks by averaging the outputs, yielding $F(\bx) = \frac{1}{M} \sum_{i = 1}^M f_{\theta_i} (\bx)$.

In addition to the ensemble's loss $\loss(F(\bx), y)$, we will also consider the \textit{average base learner} loss and the \textit{oracle ensemble's} loss. The average base learner loss is simply defined as $\frac{1}{M} \sum_{i=1}^M \loss(f_{\theta_i}(\bx), y)$; we use this to measure the \textit{average base learner strength} later. Similar to prior work \citep{lee-M-heads, diverse-ens-evol}, the oracle ensemble $\oracens$ composed of base learners $f_{\theta_1}, \dots, f_{\theta_M}$ is defined to be the function which, given an input $\bx$, returns the prediction of the base learner with the smallest loss for $(\bx, y)$, that is,
\begin{align}
    \oracens(\bx) = f_{\theta_k} (\bx) \text{, \quad  where \quad} k \in \argmin_i \loss(f_{\theta_i}(\bx), y). \nonumber
\end{align}
The oracle ensemble can only be constructed if the true class $y$ is known. We use the oracle ensemble loss as one of the measures of \textit{diversity} in base learner predictions. Intuitively, if base learners make diverse predictions for $\bx$, the oracle ensemble is more likely to find some base learner with a small loss, whereas if all base learners make identical predictions, the oracle ensemble yields the same output as any (and all) base learners. Therefore, as a rule of thumb, all else being equal, smaller oracle ensemble loss indicates more diverse base learner predictions. 

\begin{proposition}\label{prop:loss-ineq}
Suppose $\loss$ is negative log-likelihood (NLL). Then, the oracle ensemble loss, ensemble loss, and average base learner loss satisfy the following inequality: 
\begin{align}
    \loss(\oracens(\bx), y) \leq \loss(F(\bx), y) \leq \frac{1}{M} \sum_{i=1}^M \loss (f_{\theta_i}(\bx), y). \nonumber
\end{align}
\end{proposition}

\begin{wrapfigure}[36]{R}{.35\textwidth}
    \begin{minipage}{\linewidth}
    \centering\captionsetup[subfigure]{justification=justified}
    \includegraphics[width=\linewidth]{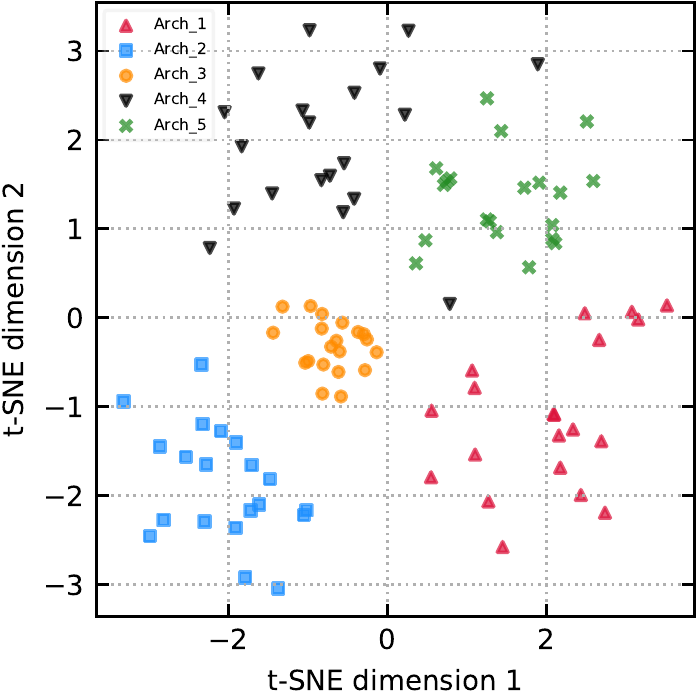}
    \subcaption{Five different architectures, each trained with 20 different initializations.}
    \label{fig:tsne_1}\par\vfill\vspace{4pt}
    \includegraphics[width=\linewidth]{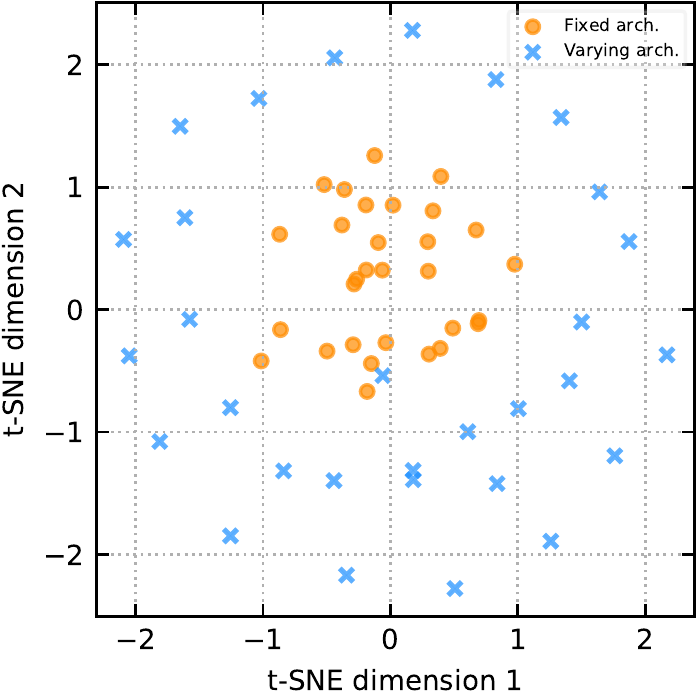}
    \subcaption{Predictions of base learners in two ensembles, one with fixed architecture and one with varying architectures.}
    \label{fig:tsne_2}
\end{minipage}
\caption{t-SNE visualization of base learner predictions. Each point corresponds to one network's (dimension-reduced) predictions.} 
\label{fig:t-sne}
\end{wrapfigure}

We refer to Appendix \ref{app:inequality} for a proof. Proposition \ref{prop:loss-ineq} suggests that it can be beneficial for ensembles to not only have strong average base learners (smaller upper bound), but also more diversity in their predictions (smaller lower bound). There is extensive theoretical work relating strong base learner performance and diversity with the generalization properties of ensembles~\citep{hansen_ensembles,ensemble_book, breimann-mlj01a,Jiang2017GeneralizedAD,bian2019does,goodfellow-16a}. In Section \ref{sec:experiments}, the two metrics we use for measuring diversity are oracle ensemble loss and (normalized) predictive disagreement, defined as the average pairwise predictive disagreement amongst the base learners, normalized by their average error~\citep{fort2019deep}.  

\vspace{-5pt}
\subsection{Visualizing Similarity in Base Learner Predictions}

\vspace{-5pt}
The fixed architecture used to build deep ensembles is typically chosen to be a strong stand-alone architecture, either hand-crafted or found by NAS. However, optimizing the base learner's architecture and \textit{then} constructing a deep ensemble can neglect diversity in favor of strong base learner performance. 
Having base learner architectures vary allows more diversity in their predictions. We provide empirical evidence for this intuition by visualizing the base learners' predictions. Fort \etal\citep{fort2019deep} found that base learners in a deep ensemble explore different parts of the function space by means of applying dimensionality reduction to their predictions. Building on this, we uniformly sample five architectures from the DARTS search space \citep{liu2018darts}, train 20 initializations of each architecture on CIFAR-10 and visualize the similarity among the networks' predictions on the test dataset using t-SNE \citep{van2008visualizing}. Experiment details are available in Section \ref{sec:experiments} and Appendix \ref{app:exp_details}.

As shown in Figure \ref{fig:tsne_1}, we observe clustering of predictions made by different initializations of a fixed architecture, suggesting that base learners with varying architectures explore different parts of the function space. Moreover, we also visualize the predictions of base learners of two ensembles, each of size $M = 30$, where one is a deep ensemble and the other has varying architectures (found by NES-RS as presented in Section \ref{sec:nes}). Figure \ref{fig:tsne_2} shows more diversity in the ensemble with varying architectures than in the deep ensemble. These qualitative findings can be quantified by measuring diversity: for the two ensembles shown in Figure \ref{fig:tsne_2}, we find the predictive disagreement to be $94.6\%$ for the ensemble constructed by NES and $76.7\%$ for the deep ensemble (this is consistent across independent runs). This indicates higher predictive diversity in the ensemble with varying architectures, in line with the t-SNE results.

\vspace{-2mm}

%% file: text/4_nes.tex
\section{Neural Ensemble Search}
\label{sec:nes}

In this section, we define \textit{neural ensemble search} (NES). In summary, a NES algorithm optimizes the architectures of base learners in an ensemble to minimize ensemble loss.

Given a network $f: \bbR^D \rightarrow \bbR^C$, let $\Loss (f, \mathcal{D}) = \sum_{(\bx, y) \in \mathcal{D}} \loss(f(\bx), y)$ be the loss of $f$ over dataset $\mathcal{D}$. Given a set of base learners $\{f_1, \dots, f_M\}$, let $\ens$ be the function which maps $\{f_1, \dots, f_M\}$ to the ensemble $F = \frac{1}{M} \sum_{i = 1}^M f_{i}$ as defined in Section \ref{sec:diversity_of_ens}
. To emphasize the architecture, we use the notation $f_{\theta, \alpha}$ to denote a network with architecture $\alpha \in \archss$ and weights $\theta$, where $\archss$ is a search space (SS) of architectures. A NES algorithm aims to solve the following optimization problem:
\begin{align}
    &\min_{\alpha_1, \dots, \alpha_M \in \archss}   \Loss \left( \ens(f_{\theta_1, \alpha_1}, \dots, f_{\theta_M, \alpha_M}), \Dval \right) \label{eq:nes-optim} \\
    &\text{s.t.\quad} \theta_i \in \argmin_{\theta} \Loss(f_{\theta, \alpha_i}, \Dtrain) \quad \text{\quad for } i = 1, \dots, M \nonumber
\end{align}
Eq. \ref{eq:nes-optim} is difficult to solve for at least two reasons. First, we are optimizing over $M$ architectures, so the search space is effectively $\archss^M$, compared to it being $\archss$ in typical NAS, making it more difficult to explore fully. Second, a larger search space also increases the risk of overfitting the ensemble loss to $\Dval$. A possible approach here is to consider the ensemble as a single large network to which we apply NAS, but joint training of an ensemble through a single loss has been empirically observed to underperform training base learners independently, specially for large networks \citep{webb2019joint}. 
Instead, our general approach to solve Eq. \ref{eq:nes-optim} consists of two steps: 
\begin{enumerate}[leftmargin=*]
    \item \textbf{Pool building}: build a \textit{pool} $\pool = \{f_{\theta_1,\alpha_1}, \dots, f_{\theta_\budget, \alpha_\budget}\}$ of size $\budget$ consisting of potential base learners, where each $f_{\theta_i, \alpha_i}$ is a network trained independently on $\Dtrain$. 
    \item \textbf{Ensemble selection}: select $M$ base learners (without replacement as discussed below) $f_{\theta_1^*,\alpha_1^*}, \dots, f_{\theta_M^*, \alpha_M^*}$ from $\pool$ to form an ensemble which minimizes loss on $\Dval$. (We set $K \geq M$.)
\end{enumerate}
Step 1 reduces the options for the base learner architectures, with the intention to make the search more feasible and focus on strong architectures; step 2 then selects a performant ensemble. This procedure also ensures that the ensemble's base learners are trained independently. We use ensemble selection without replacement \citep{rich_ens_select} for step 2. More specifically, this is forward step-wise selection; that is, given the set of networks $\pool$, we start with an empty ensemble and add the network from $\pool$ which minimizes ensemble loss on $\Dval$. We repeat this without replacement until the ensemble is of size $M$. $\esa(\pool, \Dval, M)$ denotes the resulting set of $M$ base learners selected from $\pool$. 

Note that selecting the ensemble from $\pool$ is a combinatorial optimization problem; a greedy approach such as $\esa$ is nevertheless effective as shown by Caruana \etal\citep{rich_ens_select}, while keeping computational overhead low, given the predictions of the networks on $\Dval$. We also experimented with various other ensemble selection algorithms, including weighted averaging, as discussed in Section \ref{sec:analysis_and_ablations} and Appendix \ref{app:esa_comparison}, finding $\esa$ to perform best.

We have not yet discussed the algorithm for building the pool in step 1; we present two approaches, NES-RS (Section \ref{sec:nes-rs}) and NES-RE (Section \ref{sec:nes-re}). NES-RS is a simple random search based algorithm, while NES-RE is based on regularized evolution \citep{real2019regularized}, a state-of-the-art NAS algorithm. Note that while  gradient-based NAS methods have recently become popular, they are not naively applicable in our setting as the base learner selection component $\esa$ is typically non-differentiable.

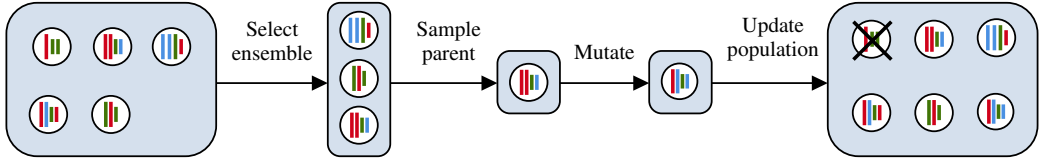
\begin{figure*}
    \centering
    \input{tikz_figure}    
    \caption{Illustration of one iteration of NES-RE. Network architectures are represented as colored bars of different lengths illustrating different layers and widths. Starting with the current population, ensemble selection is applied to select parent candidates, among which one is sampled as the parent. A mutated copy of the parent is added to the population, and the oldest member is removed.}
    \label{fig:nes_re}
    \vskip -0.2in
\end{figure*}

\vspace{-5pt}
\subsection{NES with Random Search} \label{sec:nes-rs}

\vspace{-5pt}
In NAS, random search (RS) is a competitive baseline on carefully designed architecture search spaces~\citep{li-uai19a, Yang2020NAS, Yu2020Evaluating}. Motivated by its success and simplicity, NES with random search (NES-RS) builds the pool $\pool$ by independently sampling architectures uniformly with replacement from the search space $\archss$ (and training them). Since the architectures of networks in $\pool$ vary, applying ensemble selection is a simple way to exploit diversity, yielding a performant ensemble. Importantly, NES-RS is easy to parallelize, exactly like deep ensembles. See Algorithm \ref{alg:nes-rs} in Appendix \ref{appsec:hypers} for pseudocode. 

\vspace{-5pt}
\subsection{NES with Regularized Evolution} \label{sec:nes-re}

\vspace{-5pt}
A more guided approach for building the pool $\pool$ is using regularized evolution (RE)~\citep{real2019regularized}. While RS has the benefit of simplicity by sampling architectures uniformly, the resulting pool might contain many weak architectures, leaving few strong architectures for $\esa$ to choose between. RE is an evolutionary algorithm used on NAS spaces. It explores the search space by evolving (via mutations) a \textit{population} of architectures. We first briefly describe RE as background before NES-RE. RE starts with a randomly initialized fixed-size population of architectures. At each iteration, a subset of size $\numcand$ of the population is sampled, from which the best network by validation loss is selected as the parent. A mutated copy (e.g. changing an operation in the network, see Appendix \ref{sec:app-implementation-nes-re} for examples of mutations) of the parent architecture, called the child, is trained and added to the population, and the oldest member of the population is removed, preserving the population size. This is iterated until the computational budget is reached, returning the \textit{history}, i.e. all the networks evaluated during the search, from which the best model is chosen by validation loss.

\IncMargin{12pt}{
\begin{algorithm}[t] 
\DontPrintSemicolon
\KwData{Search space $\mathcal{A}$; ensemble size $M$; comp. budget $\budget$; $\Dtrain, \Dval$; population size $\popsize$; number of parent candidates $\numcand$.}
Sample $\popsize$ architectures $\alpha_1, \dots, \alpha_{\popsize}$ independently and uniformly from $\archss$.\\
Train each architecture $\alpha_i$ using $\Dtrain$, and initialize $\popul = \pool = \{f_{\theta_1, \alpha_1}, \dots, f_{\theta_{\popsize}, \alpha_{\popsize}}\}$.\\
\While{$|\pool| < \budget$}{
    Select $m$ parent candidates $\{f_{\widetilde{\theta}_1,\widetilde{\alpha}_1}, \dots, f_{\widetilde{\theta}_\numcand,\widetilde{\alpha}_\numcand}\} = \esa(\popul, \Dval, \numcand)$. \label{lst:line4}\\
    Sample uniformly a parent architecture $\alpha$ from $\{\widetilde{\alpha}_1, \dots, \widetilde{\alpha}_\numcand\}$. \tcp*{$\alpha$ stays in $\popul$.} 
    Apply mutation to $\alpha$, yielding child architecture $\beta$.\\
    Train $\beta$ using $\Dtrain$ and add the trained network $f_{\theta, \beta}$ to $\popul$ and $\pool$.\\
    Remove the oldest member in $\popul$. \tcp*{as done in RE~\citep{real2019regularized}.} 
}
Select base learners $\{f_{\theta_1^*,\alpha_1^*}, \dots, f_{\theta_M^*, \alpha_M^*}\} = \esa(\pool, \Dval, M)$ by forward step-wise selection without replacement. \label{lst:line9}\\
\Return{ensemble $\ens(f_{\theta_1^*,\alpha_1^*}, \dots, f_{\theta_M^*, \alpha_M^*})$}
\caption{NES with Regularized Evolution}
\label{alg:nes-re}
\end{algorithm}
}

Building on RE for NAS, we propose NES-RE to build the pool of potential base learners. NES-RE starts by randomly initializing a population $\popul$ of size $\popsize$. At each iteration, we first apply $\esa$ to the population to select an ensemble of size $\numcand$, then we uniformly sample one base learner from this ensemble to be the parent. A mutated copy of the parent is added to $\popul$ and the oldest network is removed, as in usual regularized evolution. This process is repeated until the computational budget is reached, and the history is returned as the pool $\pool$. See Algorithm \ref{alg:nes-re} for pseudocode and Figure \ref{fig:nes_re} for an illustration.

Also, note the distinction between the \textit{population} and the \textit{pool} in NES-RE: the population is evolved, whereas the pool is the set of all networks evaluated during evolution (i.e., the history) and is used post-hoc for selecting the ensemble. Moreover, $\esa$ is used both for selecting $m$ parent candidates (line 4 in NES-RE) and choosing the final ensemble of size $M$ (line 9 in NES-RE). In general, $m \neq M$.

\vspace{-6pt}
\subsection{Ensemble Adaptation to Dataset Shift}\label{sec:ens-adaptation-shift}

\vspace{-5pt}
Deep ensembles offer (some) robustness to dataset shift relative to training data. In general, one may not know the type of shift that occurs at test time. By using an ensemble, diversity in base learner predictions prevents the model from relying on one base learner's predictions which may not only be incorrect but also overconfident. 

We assume that one does not have access to data points with test-time shift at training time, but one does have access to some validation data $\Dvals$ with a \textit{validation} shift, which encapsulates one's belief about test-time shift. Crucially, test and validation shifts are disjoint. To adapt NES-RS and NES-RE to return ensembles robust to shift, we propose using $\Dvals$ instead of $\Dval$ whenever applying $\esa$ to select the final ensemble. In Algorithms \ref{alg:nes-re} and \ref{alg:nes-rs}, this is in lines 9 and 3, respectively. Our experiments show that this is highly effective against shift.

Note that in line 4 of Algorithm \ref{alg:nes-re}, we can also replace $\Dval$ with $\Dvals$ when expecting test-time shift; we simply sample one of $\Dval, \Dvals$ uniformly at each iteration, in order to promote exploration of architectures that work well both in-distribution and during shift and reduce cost by avoiding running NES-RE once for each of $\Dval, \Dvals$. See Appendices \ref{appsubsec:cifarc_exp} and \ref{sec:app-implementation-nes-re} for further discussion.

%% file: tikz_figure.tex
\tikzset{every picture/.style={line width=0.75pt}} %

\resizebox{\textwidth}{!}{

\tikzset{every picture/.style={line width=0.75pt}} %

\begin{tikzpicture}[x=0.75pt,y=0.75pt,yscale=-1,xscale=1]

\draw    (410,49) -- (477,49) ;
\draw [shift={(480,49)}, rotate = 180] [fill={rgb, 255:red, 0; green, 0; blue, 0 }  ][line width=0.08]  [draw opacity=0] (8.93,-4.29) -- (0,0) -- (8.93,4.29) -- cycle    ;
\draw    (325,49) -- (377,49) ;
\draw [shift={(380,49)}, rotate = 180] [fill={rgb, 255:red, 0; green, 0; blue, 0 }  ][line width=0.08]  [draw opacity=0] (8.93,-4.29) -- (0,0) -- (8.93,4.29) -- cycle    ;
\draw  [fill={rgb, 255:red, 213; green, 224; blue, 237 }  ,fill opacity=1 ] (380,38) .. controls (380,34.13) and (383.13,31) .. (387,31) -- (408,31) .. controls (411.87,31) and (415,34.13) .. (415,38) -- (415,59) .. controls (415,62.87) and (411.87,66) .. (408,66) -- (387,66) .. controls (383.13,66) and (380,62.87) .. (380,59) -- cycle ;
\draw    (130,49) -- (197,49) ;
\draw [shift={(200,49)}, rotate = 180] [fill={rgb, 255:red, 0; green, 0; blue, 0 }  ][line width=0.08]  [draw opacity=0] (8.93,-4.29) -- (0,0) -- (8.93,4.29) -- cycle    ;
\draw  [fill={rgb, 255:red, 213; green, 224; blue, 237 }  ,fill opacity=1 ] (20,21.2) .. controls (20,11.7) and (27.7,4) .. (37.2,4) -- (120.3,4) .. controls (129.8,4) and (137.5,11.7) .. (137.5,21.2) -- (137.5,72.8) .. controls (137.5,82.3) and (129.8,90) .. (120.3,90) -- (37.2,90) .. controls (27.7,90) and (20,82.3) .. (20,72.8) -- cycle ;
\draw  [fill={rgb, 255:red, 213; green, 224; blue, 237 }  ,fill opacity=1 ] (200,11) .. controls (200,7.13) and (203.13,4) .. (207,4) -- (228,4) .. controls (231.87,4) and (235,7.13) .. (235,11) -- (235,83) .. controls (235,86.87) and (231.87,90) .. (228,90) -- (207,90) .. controls (203.13,90) and (200,86.87) .. (200,83) -- cycle ;
\draw    (235,49) -- (292,49) ;
\draw [shift={(295,49)}, rotate = 180] [fill={rgb, 255:red, 0; green, 0; blue, 0 }  ][line width=0.08]  [draw opacity=0] (8.93,-4.29) -- (0,0) -- (8.93,4.29) -- cycle    ;
\draw  [fill={rgb, 255:red, 213; green, 224; blue, 237 }  ,fill opacity=1 ] (295,38) .. controls (295,34.13) and (298.13,31) .. (302,31) -- (323,31) .. controls (326.87,31) and (330,34.13) .. (330,38) -- (330,59) .. controls (330,62.87) and (326.87,66) .. (323,66) -- (302,66) .. controls (298.13,66) and (295,62.87) .. (295,59) -- cycle ;
\draw  [fill={rgb, 255:red, 213; green, 224; blue, 237 }  ,fill opacity=1 ] (480,21.2) .. controls (480,11.7) and (487.7,4) .. (497.2,4) -- (580.3,4) .. controls (589.8,4) and (597.5,11.7) .. (597.5,21.2) -- (597.5,72.8) .. controls (597.5,82.3) and (589.8,90) .. (580.3,90) -- (497.2,90) .. controls (487.7,90) and (480,82.3) .. (480,72.8) -- cycle ;
\draw  [color={rgb, 255:red, 0; green, 0; blue, 0 }  ,draw opacity=1 ][fill={rgb, 255:red, 255; green, 255; blue, 255 }  ,fill opacity=1 ] (68.95,28.58) .. controls (68.95,22.83) and (73.66,18.17) .. (79.47,18.17) .. controls (85.29,18.17) and (90,22.83) .. (90,28.58) .. controls (90,34.34) and (85.29,39) .. (79.47,39) .. controls (73.66,39) and (68.95,34.34) .. (68.95,28.58) -- cycle ;
\draw [color={rgb, 255:red, 208; green, 2; blue, 27 }  ,draw opacity=1 ][line width=1.5]    (78.34,21.55) -- (78.34,35.62) ;
\draw [color={rgb, 255:red, 65; green, 117; blue, 5 }  ,draw opacity=1 ][line width=1.5]    (81.18,24.36) -- (81.18,32.81) ;
\draw [color={rgb, 255:red, 74; green, 144; blue, 226 }  ,draw opacity=1 ][line width=1.5]    (84.03,24.36) -- (84.03,32.81) ;
\draw [color={rgb, 255:red, 208; green, 2; blue, 27 }  ,draw opacity=1 ][line width=1.5]    (75.49,21.55) -- (75.49,35.62) ;

\draw  [color={rgb, 255:red, 0; green, 0; blue, 0 }  ,draw opacity=1 ][fill={rgb, 255:red, 255; green, 255; blue, 255 }  ,fill opacity=1 ] (35,28.58) .. controls (35,22.83) and (39.71,18.17) .. (45.53,18.17) .. controls (51.34,18.17) and (56.05,22.83) .. (56.05,28.58) .. controls (56.05,34.34) and (51.34,39) .. (45.53,39) .. controls (39.71,39) and (35,34.34) .. (35,28.58) -- cycle ;
\draw [color={rgb, 255:red, 65; green, 117; blue, 5 }  ,draw opacity=1 ][line width=1.5]    (48.92,24.36) -- (48.92,32.81) ;
\draw [color={rgb, 255:red, 65; green, 117; blue, 5 }  ,draw opacity=1 ][line width=1.5]    (45.87,24.36) -- (45.87,32.81) ;
\draw [color={rgb, 255:red, 208; green, 2; blue, 27 }  ,draw opacity=1 ][line width=1.5]    (42.39,21.55) -- (42.39,35.62) ;

\draw  [color={rgb, 255:red, 0; green, 0; blue, 0 }  ,draw opacity=1 ][fill={rgb, 255:red, 255; green, 255; blue, 255 }  ,fill opacity=1 ] (101.95,28.58) .. controls (101.95,22.83) and (106.66,18.17) .. (112.47,18.17) .. controls (118.29,18.17) and (123,22.83) .. (123,28.58) .. controls (123,34.34) and (118.29,39) .. (112.47,39) .. controls (106.66,39) and (101.95,34.34) .. (101.95,28.58) -- cycle ;
\draw [color={rgb, 255:red, 74; green, 144; blue, 226 }  ,draw opacity=1 ][line width=1.5]    (111.34,21.55) -- (111.34,35.62) ;
\draw [color={rgb, 255:red, 208; green, 2; blue, 27 }  ,draw opacity=1 ][line width=1.5]    (117.87,24.36) -- (117.87,32.81) ;
\draw [color={rgb, 255:red, 74; green, 144; blue, 226 }  ,draw opacity=1 ][line width=1.5]    (107.65,21.55) -- (107.65,35.62) ;
\draw [color={rgb, 255:red, 65; green, 117; blue, 5 }  ,draw opacity=1 ][line width=1.5]    (114.7,21.55) -- (114.7,35.62) ;

\draw  [color={rgb, 255:red, 0; green, 0; blue, 0 }  ,draw opacity=1 ][fill={rgb, 255:red, 255; green, 255; blue, 255 }  ,fill opacity=1 ] (33,66.58) .. controls (33,60.83) and (37.71,56.17) .. (43.53,56.17) .. controls (49.34,56.17) and (54.05,60.83) .. (54.05,66.58) .. controls (54.05,72.34) and (49.34,77) .. (43.53,77) .. controls (37.71,77) and (33,72.34) .. (33,66.58) -- cycle ;
\draw [color={rgb, 255:red, 74; green, 144; blue, 226 }  ,draw opacity=1 ][line width=1.5]    (42.39,59.55) -- (42.39,73.62) ;
\draw [color={rgb, 255:red, 65; green, 117; blue, 5 }  ,draw opacity=1 ][line width=1.5]    (45.23,62.36) -- (45.23,70.81) ;
\draw [color={rgb, 255:red, 208; green, 2; blue, 27 }  ,draw opacity=1 ][line width=1.5]    (48.08,62.36) -- (48.08,70.81) ;
\draw [color={rgb, 255:red, 208; green, 2; blue, 27 }  ,draw opacity=1 ][line width=1.5]    (39.54,59.55) -- (39.54,73.62) ;

\draw  [color={rgb, 255:red, 0; green, 0; blue, 0 }  ,draw opacity=1 ][fill={rgb, 255:red, 255; green, 255; blue, 255 }  ,fill opacity=1 ] (68,66.58) .. controls (68,60.83) and (72.71,56.17) .. (78.53,56.17) .. controls (84.34,56.17) and (89.05,60.83) .. (89.05,66.58) .. controls (89.05,72.34) and (84.34,77) .. (78.53,77) .. controls (72.71,77) and (68,72.34) .. (68,66.58) -- cycle ;
\draw [color={rgb, 255:red, 208; green, 2; blue, 27 }  ,draw opacity=1 ][line width=1.5]    (78.53,59.5) -- (78.53,73.58) ;
\draw [color={rgb, 255:red, 65; green, 117; blue, 5 }  ,draw opacity=1 ][line width=1.5]    (81.37,62.32) -- (81.37,70.76) ;
\draw [color={rgb, 255:red, 65; green, 117; blue, 5 }  ,draw opacity=1 ][line width=1.5]    (75.68,59.5) -- (75.68,73.58) ;

\draw  [color={rgb, 255:red, 0; green, 0; blue, 0 }  ,draw opacity=1 ][fill={rgb, 255:red, 255; green, 255; blue, 255 }  ,fill opacity=1 ] (207,19.42) .. controls (207,13.66) and (211.71,9) .. (217.53,9) .. controls (223.34,9) and (228.05,13.66) .. (228.05,19.42) .. controls (228.05,25.17) and (223.34,29.83) .. (217.53,29.83) .. controls (211.71,29.83) and (207,25.17) .. (207,19.42) -- cycle ;
\draw [color={rgb, 255:red, 74; green, 144; blue, 226 }  ,draw opacity=1 ][line width=1.5]    (216.39,12.38) -- (216.39,26.45) ;
\draw [color={rgb, 255:red, 208; green, 2; blue, 27 }  ,draw opacity=1 ][line width=1.5]    (222.92,15.19) -- (222.92,23.64) ;
\draw [color={rgb, 255:red, 74; green, 144; blue, 226 }  ,draw opacity=1 ][line width=1.5]    (212.7,12.38) -- (212.7,26.45) ;
\draw [color={rgb, 255:red, 65; green, 117; blue, 5 }  ,draw opacity=1 ][line width=1.5]    (219.76,12.38) -- (219.76,26.45) ;

\draw  [color={rgb, 255:red, 0; green, 0; blue, 0 }  ,draw opacity=1 ][fill={rgb, 255:red, 255; green, 255; blue, 255 }  ,fill opacity=1 ] (207,46.42) .. controls (207,40.66) and (211.71,36) .. (217.53,36) .. controls (223.34,36) and (228.05,40.66) .. (228.05,46.42) .. controls (228.05,52.17) and (223.34,56.83) .. (217.53,56.83) .. controls (211.71,56.83) and (207,52.17) .. (207,46.42) -- cycle ;
\draw [color={rgb, 255:red, 208; green, 2; blue, 27 }  ,draw opacity=1 ][line width=1.5]    (217.53,39.33) -- (217.53,53.41) ;
\draw [color={rgb, 255:red, 65; green, 117; blue, 5 }  ,draw opacity=1 ][line width=1.5]    (220.37,42.15) -- (220.37,50.59) ;
\draw [color={rgb, 255:red, 65; green, 117; blue, 5 }  ,draw opacity=1 ][line width=1.5]    (214.68,39.33) -- (214.68,53.41) ;

\draw  [color={rgb, 255:red, 0; green, 0; blue, 0 }  ,draw opacity=1 ][fill={rgb, 255:red, 255; green, 255; blue, 255 }  ,fill opacity=1 ] (206.95,72.58) .. controls (206.95,66.83) and (211.66,62.17) .. (217.47,62.17) .. controls (223.29,62.17) and (228,66.83) .. (228,72.58) .. controls (228,78.34) and (223.29,83) .. (217.47,83) .. controls (211.66,83) and (206.95,78.34) .. (206.95,72.58) -- cycle ;
\draw [color={rgb, 255:red, 208; green, 2; blue, 27 }  ,draw opacity=1 ][line width=1.5]    (216.34,65.55) -- (216.34,79.62) ;
\draw [color={rgb, 255:red, 65; green, 117; blue, 5 }  ,draw opacity=1 ][line width=1.5]    (219.18,68.36) -- (219.18,76.81) ;
\draw [color={rgb, 255:red, 74; green, 144; blue, 226 }  ,draw opacity=1 ][line width=1.5]    (222.03,68.36) -- (222.03,76.81) ;
\draw [color={rgb, 255:red, 208; green, 2; blue, 27 }  ,draw opacity=1 ][line width=1.5]    (213.49,65.55) -- (213.49,79.62) ;

\draw  [color={rgb, 255:red, 0; green, 0; blue, 0 }  ,draw opacity=1 ][fill={rgb, 255:red, 255; green, 255; blue, 255 }  ,fill opacity=1 ] (302,48.58) .. controls (302,42.83) and (306.71,38.17) .. (312.53,38.17) .. controls (318.34,38.17) and (323.05,42.83) .. (323.05,48.58) .. controls (323.05,54.34) and (318.34,59) .. (312.53,59) .. controls (306.71,59) and (302,54.34) .. (302,48.58) -- cycle ;
\draw [color={rgb, 255:red, 208; green, 2; blue, 27 }  ,draw opacity=1 ][line width=1.5]    (311.39,41.55) -- (311.39,55.62) ;
\draw [color={rgb, 255:red, 65; green, 117; blue, 5 }  ,draw opacity=1 ][line width=1.5]    (314.23,44.36) -- (314.23,52.81) ;
\draw [color={rgb, 255:red, 74; green, 144; blue, 226 }  ,draw opacity=1 ][line width=1.5]    (317.08,44.36) -- (317.08,52.81) ;
\draw [color={rgb, 255:red, 208; green, 2; blue, 27 }  ,draw opacity=1 ][line width=1.5]    (308.54,41.55) -- (308.54,55.62) ;

\draw  [color={rgb, 255:red, 0; green, 0; blue, 0 }  ,draw opacity=1 ][fill={rgb, 255:red, 255; green, 255; blue, 255 }  ,fill opacity=1 ] (387,48) .. controls (387,42.48) and (391.48,38) .. (397,38) .. controls (402.52,38) and (407,42.48) .. (407,48) .. controls (407,53.52) and (402.52,58) .. (397,58) .. controls (391.48,58) and (387,53.52) .. (387,48) -- cycle ;
\draw [color={rgb, 255:red, 74; green, 144; blue, 226 }  ,draw opacity=1 ][line width=1.5]    (395.92,41.24) -- (395.92,54.76) ;
\draw [color={rgb, 255:red, 65; green, 117; blue, 5 }  ,draw opacity=1 ][line width=1.5]    (398.62,43.95) -- (398.62,52.05) ;
\draw [color={rgb, 255:red, 74; green, 144; blue, 226 }  ,draw opacity=1 ][line width=1.5]    (401.32,43.95) -- (401.32,52.05) ;
\draw [color={rgb, 255:red, 208; green, 2; blue, 27 }  ,draw opacity=1 ][line width=1.5]    (393.22,41.24) -- (393.22,54.76) ;

\draw  [color={rgb, 255:red, 0; green, 0; blue, 0 }  ,draw opacity=1 ][fill={rgb, 255:red, 255; green, 255; blue, 255 }  ,fill opacity=1 ] (528.95,24.42) .. controls (528.95,18.66) and (533.66,14) .. (539.47,14) .. controls (545.29,14) and (550,18.66) .. (550,24.42) .. controls (550,30.17) and (545.29,34.83) .. (539.47,34.83) .. controls (533.66,34.83) and (528.95,30.17) .. (528.95,24.42) -- cycle ;
\draw [color={rgb, 255:red, 208; green, 2; blue, 27 }  ,draw opacity=1 ][line width=1.5]    (538.34,17.38) -- (538.34,31.45) ;
\draw [color={rgb, 255:red, 65; green, 117; blue, 5 }  ,draw opacity=1 ][line width=1.5]    (541.18,20.19) -- (541.18,28.64) ;
\draw [color={rgb, 255:red, 74; green, 144; blue, 226 }  ,draw opacity=1 ][line width=1.5]    (544.03,20.19) -- (544.03,28.64) ;
\draw [color={rgb, 255:red, 208; green, 2; blue, 27 }  ,draw opacity=1 ][line width=1.5]    (535.49,17.38) -- (535.49,31.45) ;

\draw  [color={rgb, 255:red, 0; green, 0; blue, 0 }  ,draw opacity=1 ][fill={rgb, 255:red, 255; green, 255; blue, 255 }  ,fill opacity=1 ] (493.95,24.42) .. controls (493.95,18.66) and (498.66,14) .. (504.47,14) .. controls (510.29,14) and (515,18.66) .. (515,24.42) .. controls (515,30.17) and (510.29,34.83) .. (504.47,34.83) .. controls (498.66,34.83) and (493.95,30.17) .. (493.95,24.42) -- cycle ;
\draw [color={rgb, 255:red, 65; green, 117; blue, 5 }  ,draw opacity=1 ][line width=1.5]    (507.86,20.19) -- (507.86,28.64) ;
\draw [color={rgb, 255:red, 65; green, 117; blue, 5 }  ,draw opacity=1 ][line width=1.5]    (504.82,20.19) -- (504.82,28.64) ;
\draw [color={rgb, 255:red, 208; green, 2; blue, 27 }  ,draw opacity=1 ][line width=1.5]    (501.33,17.38) -- (501.33,31.45) ;

\draw  [color={rgb, 255:red, 0; green, 0; blue, 0 }  ,draw opacity=1 ][fill={rgb, 255:red, 255; green, 255; blue, 255 }  ,fill opacity=1 ] (563.95,23.58) .. controls (563.95,17.83) and (568.66,13.17) .. (574.47,13.17) .. controls (580.29,13.17) and (585,17.83) .. (585,23.58) .. controls (585,29.34) and (580.29,34) .. (574.47,34) .. controls (568.66,34) and (563.95,29.34) .. (563.95,23.58) -- cycle ;
\draw [color={rgb, 255:red, 74; green, 144; blue, 226 }  ,draw opacity=1 ][line width=1.5]    (573.34,16.55) -- (573.34,30.62) ;
\draw [color={rgb, 255:red, 208; green, 2; blue, 27 }  ,draw opacity=1 ][line width=1.5]    (579.87,19.36) -- (579.87,27.81) ;
\draw [color={rgb, 255:red, 74; green, 144; blue, 226 }  ,draw opacity=1 ][line width=1.5]    (569.65,16.55) -- (569.65,30.62) ;
\draw [color={rgb, 255:red, 65; green, 117; blue, 5 }  ,draw opacity=1 ][line width=1.5]    (576.7,16.55) -- (576.7,30.62) ;

\draw  [color={rgb, 255:red, 0; green, 0; blue, 0 }  ,draw opacity=1 ][fill={rgb, 255:red, 255; green, 255; blue, 255 }  ,fill opacity=1 ] (494,65.58) .. controls (494,59.83) and (498.71,55.17) .. (504.53,55.17) .. controls (510.34,55.17) and (515.05,59.83) .. (515.05,65.58) .. controls (515.05,71.34) and (510.34,76) .. (504.53,76) .. controls (498.71,76) and (494,71.34) .. (494,65.58) -- cycle ;
\draw [color={rgb, 255:red, 74; green, 144; blue, 226 }  ,draw opacity=1 ][line width=1.5]    (503.39,58.55) -- (503.39,72.62) ;
\draw [color={rgb, 255:red, 65; green, 117; blue, 5 }  ,draw opacity=1 ][line width=1.5]    (506.23,61.36) -- (506.23,69.81) ;
\draw [color={rgb, 255:red, 208; green, 2; blue, 27 }  ,draw opacity=1 ][line width=1.5]    (509.08,61.36) -- (509.08,69.81) ;
\draw [color={rgb, 255:red, 208; green, 2; blue, 27 }  ,draw opacity=1 ][line width=1.5]    (500.54,58.55) -- (500.54,72.62) ;

\draw  [color={rgb, 255:red, 0; green, 0; blue, 0 }  ,draw opacity=1 ][fill={rgb, 255:red, 255; green, 255; blue, 255 }  ,fill opacity=1 ] (529,65.58) .. controls (529,59.83) and (533.71,55.17) .. (539.53,55.17) .. controls (545.34,55.17) and (550.05,59.83) .. (550.05,65.58) .. controls (550.05,71.34) and (545.34,76) .. (539.53,76) .. controls (533.71,76) and (529,71.34) .. (529,65.58) -- cycle ;
\draw [color={rgb, 255:red, 208; green, 2; blue, 27 }  ,draw opacity=1 ][line width=1.5]    (539.53,58.5) -- (539.53,72.58) ;
\draw [color={rgb, 255:red, 65; green, 117; blue, 5 }  ,draw opacity=1 ][line width=1.5]    (542.37,61.32) -- (542.37,69.76) ;
\draw [color={rgb, 255:red, 65; green, 117; blue, 5 }  ,draw opacity=1 ][line width=1.5]    (536.68,58.5) -- (536.68,72.58) ;

\draw  [color={rgb, 255:red, 0; green, 0; blue, 0 }  ,draw opacity=1 ][fill={rgb, 255:red, 255; green, 255; blue, 255 }  ,fill opacity=1 ] (564,65) .. controls (564,59.48) and (568.48,55) .. (574,55) .. controls (579.52,55) and (584,59.48) .. (584,65) .. controls (584,70.52) and (579.52,75) .. (574,75) .. controls (568.48,75) and (564,70.52) .. (564,65) -- cycle ;
\draw [color={rgb, 255:red, 74; green, 144; blue, 226 }  ,draw opacity=1 ][line width=1.5]    (572.92,58.24) -- (572.92,71.76) ;
\draw [color={rgb, 255:red, 65; green, 117; blue, 5 }  ,draw opacity=1 ][line width=1.5]    (575.62,60.95) -- (575.62,69.05) ;
\draw [color={rgb, 255:red, 74; green, 144; blue, 226 }  ,draw opacity=1 ][line width=1.5]    (578.32,60.95) -- (578.32,69.05) ;
\draw [color={rgb, 255:red, 208; green, 2; blue, 27 }  ,draw opacity=1 ][line width=1.5]    (570.22,58.24) -- (570.22,71.76) ;

\draw  [color={rgb, 255:red, 0; green, 0; blue, 0 }  ,draw opacity=1 ][line width=1.5]  (494.39,14.61) -- (515.61,35.82)(515.61,14.61) -- (494.39,35.82) ;

\draw (142,13) node [anchor=north west][inner sep=0.75pt]  [font=\footnotesize] [align=left] {\begin{minipage}[lt]{38.101216pt}\setlength\topsep{0pt}
\begin{center}
Select\\ensemble
\end{center}

\end{minipage}};
\draw (246,13) node [anchor=north west][inner sep=0.75pt]  [font=\footnotesize] [align=left] {\begin{minipage}[lt]{30.387500000000003pt}\setlength\topsep{0pt}
\begin{center}
Sample\\parent
\end{center}

\end{minipage}};
\draw (335,26) node [anchor=north west][inner sep=0.75pt]  [font=\footnotesize] [align=left] {\begin{minipage}[lt]{27.6675pt}\setlength\topsep{0pt}
\begin{center}
Mutate
\end{center}

\end{minipage}};
\draw (421,11) node [anchor=north west][inner sep=0.75pt]  [font=\footnotesize] [align=left] {\begin{minipage}[lt]{40.383500000000005pt}\setlength\topsep{0pt}
\begin{center}
Update\\population
\end{center}

\end{minipage}};

\end{tikzpicture}
}

%% file: text/5_experiments.tex
\section{Experiments}\label{sec:experiments}

\subsection{Comparison to Baselines: Uncertainty Estimation \& Robustness to Dataset Shift}\label{subsec:baselines}
\vspace{-7pt}
We compare NES to deep ensembles on different choices of architecture search space (DARTS~\citep{liu2018darts} and NAS-Bench-201 \citep{dong20} search spaces) and dataset (Fashion-MNIST, CIFAR-10, CIFAR-100, \insixteen and Tiny ImageNet). The search spaces are \textit{cell-based}, containing a rich variety of convolutional architectures with differing cell topologies, number of connections and operations (see Appendix \ref{app:arch_search_space_descriptions} for visualizations). For CIFAR-10/100 and Tiny ImageNet, we also consider dataset shifts proposed by Hendrycks \& Dietterich \citep{hendrycks2018benchmarking}. We use NLL, classification error and expected calibration error (ECE) \citep{pmlr-v70-guo17a, naeini15} as our metrics, for which small values are better. Note that NLL and ECE evaluate predictive uncertainty. Experimental/implementation details are in Appendix \ref{app:exp_details}, additional experiments are in Appendix \ref{app:add_exp}. All evaluations are on the test dataset. Each paragraph below highlights one of our main findings.

\textbf{Baselines.} One of our main baselines is deep ensembles with fixed, optimized architectures. We consider various optimized architectures, indicated as ``DeepEns (\texttt{arch})'': 
    \begin{itemize}[leftmargin=*, itemsep=0pt, topsep=3pt]
        \item On the DARTS search space, we consider the architectures found by:
        \begin{enumerate}[leftmargin=*]
            \item the DARTS algorithm (DeepEns (\texttt{DARTS})),
            \item regularized evolution (DeepEns (\texttt{AmoebaNet})),
            \item random search, \textit{with the same number of networks trained} as NES algorithms (DeepEns (\texttt{RS})).
        \end{enumerate}
        \item On the NAS-Bench-201 search space, in addition to DeepEns (\texttt{RS}), we consider:
        \begin{enumerate}[leftmargin=*]
            \item the architecture found by GDAS\footnote{We did not consider the DARTS algorithm on NAS-Bench-201, since it returns degenerate architectures with poor performance on this space \citep{dong20}. Whereas, GDAS yields state-of-the-art performance on this space.} \citep{dong-cvpr19a} (DeepEns (\texttt{GDAS})),
            \item the best architecture in the search space by validation loss (DeepEns (\texttt{best arch.})).\footnote{This is feasible, because all architectures in this search space were evaluated and are available.}
        \end{enumerate}
    \end{itemize}
We also compare to \textit{anchored ensembles} \citep{pearce20}, a recent technique for approximately Bayesian ensembles, which explicitly regularizes each base learner towards a fresh initialization sample and aims to induce more ensemble diversity. We use the DARTS architecture, and our implementation is described in Appendix \ref{app:anchored_ensembles}.

\begin{figure}[t!]
    \centering
    \captionsetup[subfigure]{justification=centering}
    \begin{subfigure}[t]{0.49\textwidth}
        \centering
        \includegraphics[width=.32\linewidth]{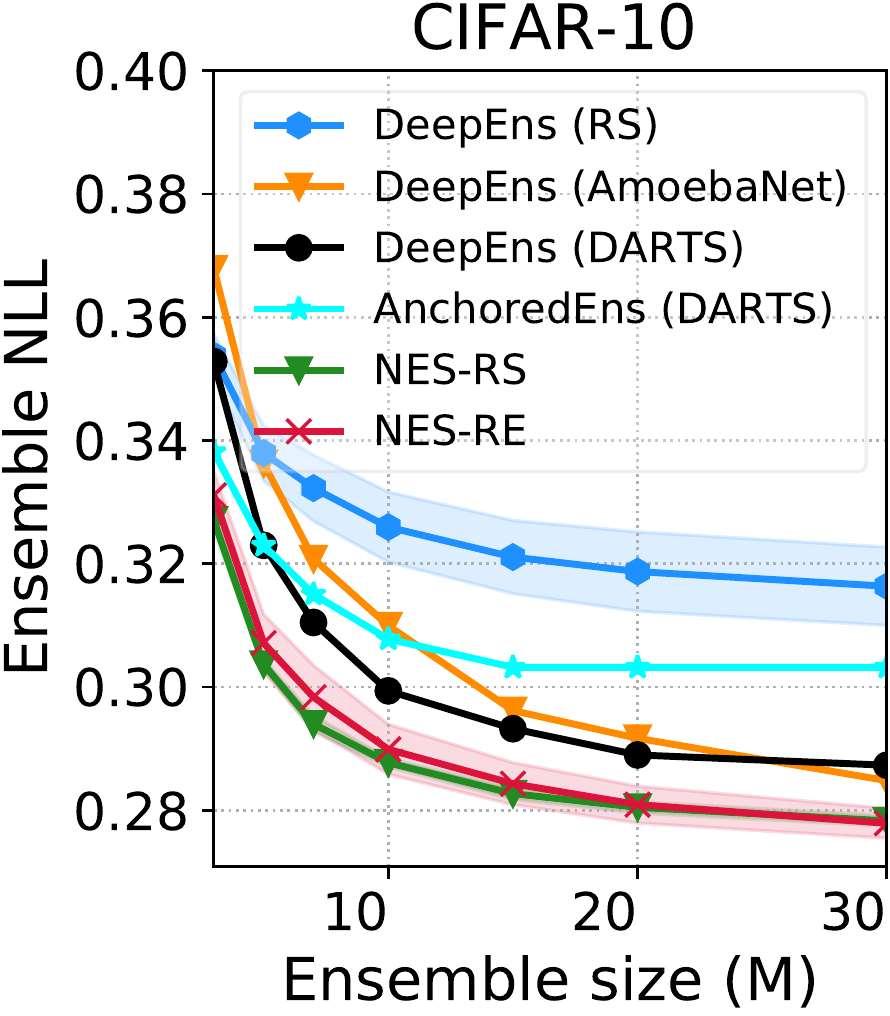}
        \includegraphics[width=.32\linewidth]{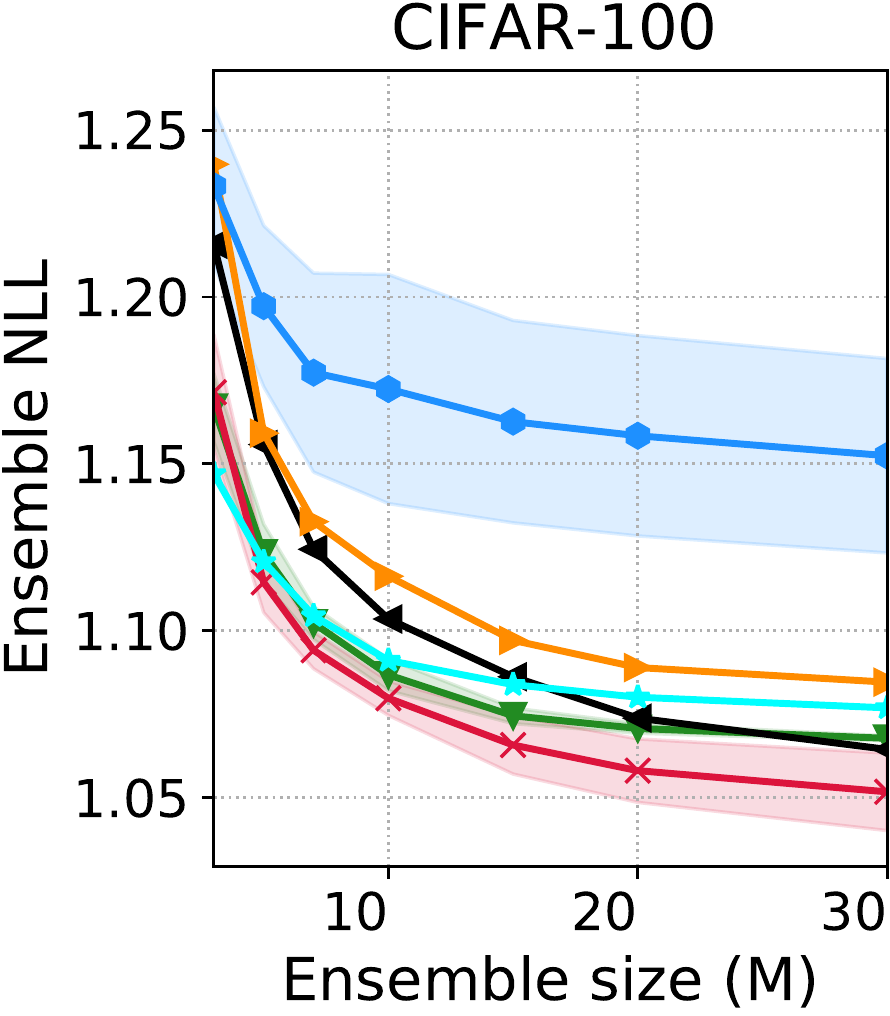}
        \includegraphics[width=.32\linewidth]{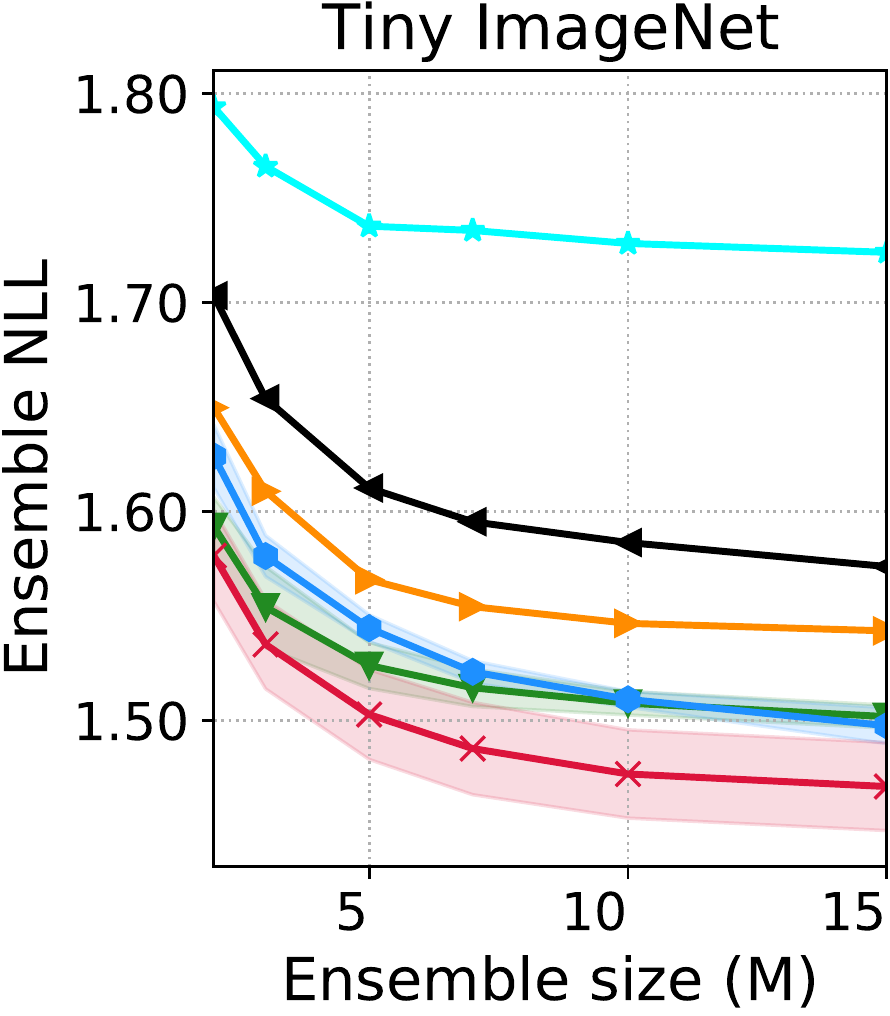}
        \subcaption{No data shift}
        \label{fig:test_loss_M_noshift}
    \end{subfigure}%
    ~\hspace{5.5pt} %
    \begin{subfigure}[t]{0.49\textwidth}
        \centering
        \includegraphics[width=.32\linewidth]{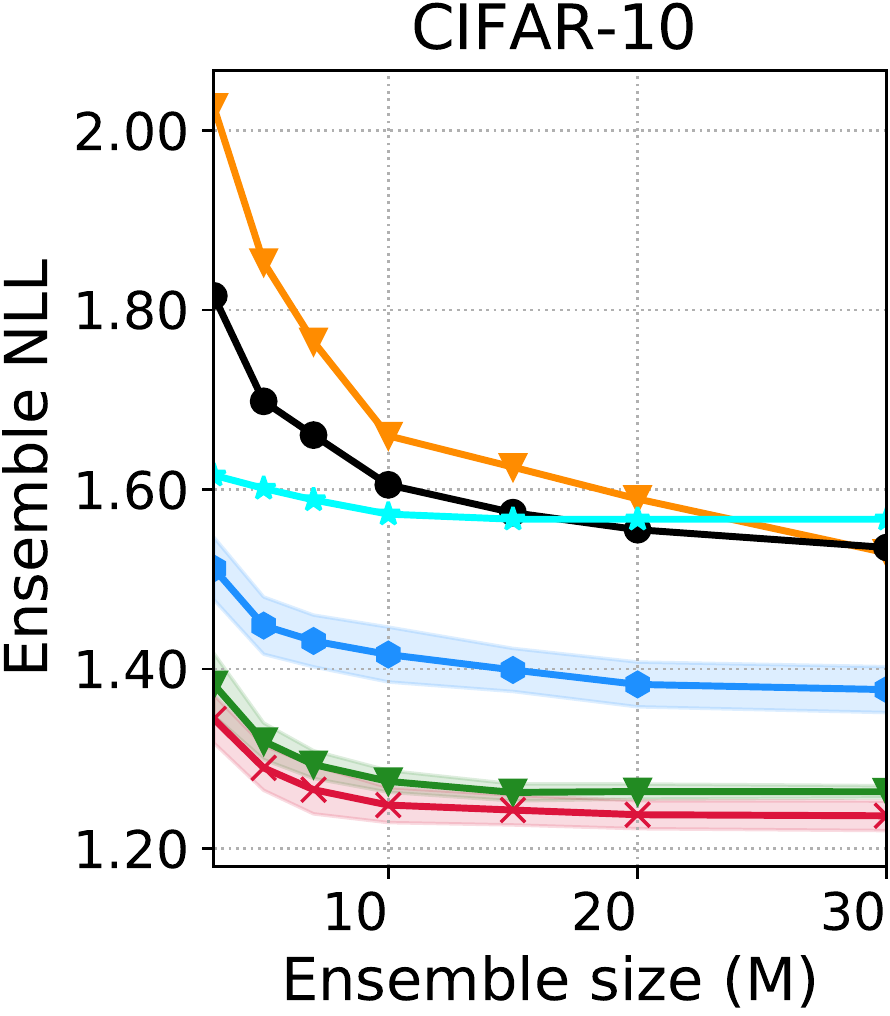}
        \includegraphics[width=.32\linewidth]{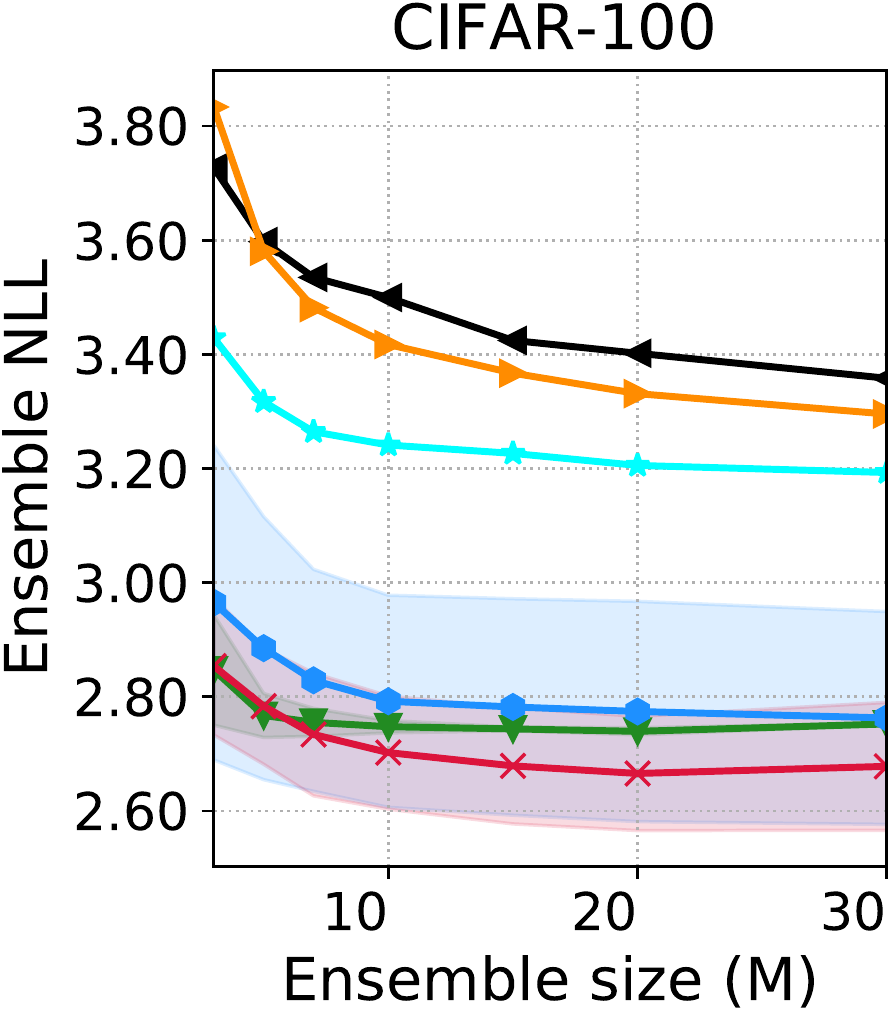}
        \includegraphics[width=.32\linewidth]{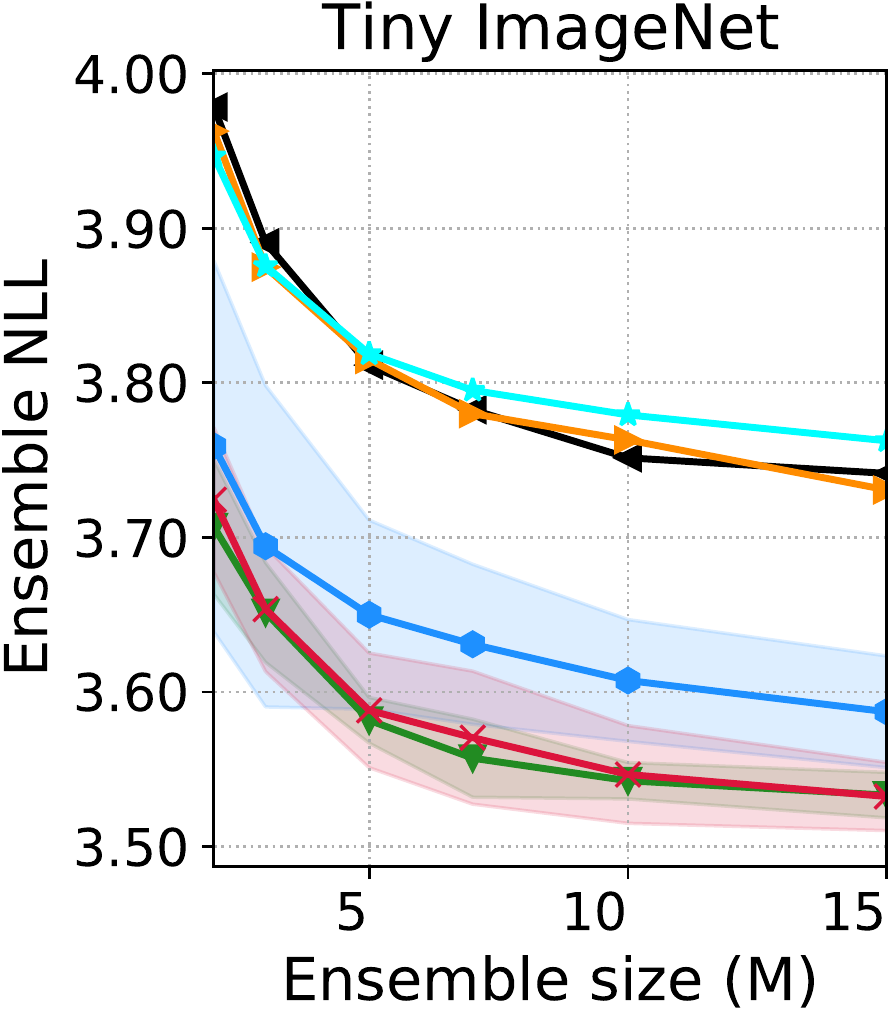}
        \subcaption{Dataset shift: severity 5 (out of 5)}
        \label{fig:test_loss_M_shift}
    \end{subfigure}
    
    \caption{NLL vs. ensemble sizes on CIFAR-10, CIFAR-100 and Tiny ImageNet with and without dataset shifts~\citep{hendrycks2018benchmarking} over the DARTS search space. Mean NLL shown with 95\% confidence intervals.} 
    \label{fig:test_loss_M}
    \vskip -0.2in
\end{figure}

\begin{wraptable}[25]{R}{.49\textwidth} 
\vspace{-14pt}
\caption{Classification error of ensembles for different shift severities. Best values and all values within $95\%$ confidence interval are bold faced. Note that NAS-Bench-201 comes with each architecture trained with \textit{three} random initializations; therefore we set $M = 3$ in that case.} 
\label{tbl:error}
\centering
\subfloat[$M = 10$; DARTS search space.]{
\resizebox{0.98\linewidth}{!}{%
\begin{tabular}{@{}cccccccc@{}}
\toprule
\multirow{3}{*}{\textbf{Dataset}} &
  \multirow{3}{*}{\textbf{\begin{tabular}[c]{@{}c@{}}Shift\\ Severity\end{tabular}}} &
  \multicolumn{6}{c}{\textbf{Classif. error} (\%), $\archss = $ DARTS search space} \\ \cmidrule(l){3-8} 
 &
  &
  \begin{tabular}[c]{@{}c@{}}DeepEns\\ (RS)\end{tabular} &
  \begin{tabular}[c]{@{}c@{}}DeepEns\\ (Amoe.)\end{tabular} &
  \begin{tabular}[c]{@{}c@{}}DeepEns\\ (DARTS)\end{tabular} &
  \begin{tabular}[c]{@{}c@{}}AnchoredEns\\ (DARTS)\end{tabular} &
  NES-RS &
  NES-RE \\ \midrule
  \multirow{3}{*}{CIFAR-10}  & 0 & $10.8$\tiny{$\pm 0.2 $} & $9.7$  & $10.0$ & $10.2$ & $\mathbf{ 9.4 }$\tiny{$\pm 0.1 $}  & $\mathbf{ 9.4 }$\tiny{$\pm 0.2 $}  \\  
                      & 3 & $25.1$\tiny{$\pm 0.7 $} & $25.6$ & $26.3$ & $28.6$ & $23.2$\tiny{$\pm 0.2 $}            & $\mathbf{ 22.9 }$\tiny{$\pm 0.2 $} \\  
                      & 5 & $41.1$\tiny{$\pm 0.9 $} & $42.7$ & $42.9$ & $44.9$ & $38.0$\tiny{$\pm 0.2 $}            & $\mathbf{ 37.4 }$\tiny{$\pm 0.4 $} \\ \hline
\multirow{3}{*}{CIFAR-100} & 0 & $33.2$\tiny{$\pm 1.2 $} & $31.6$ & $30.9$ & $30.9$ & $\mathbf{ 30.7 }$\tiny{$\pm 0.1 $} & $\mathbf{ 30.4 }$\tiny{$\pm 0.4 $} \\  
                      & 3 & $54.8$\tiny{$\pm 1.6 $} & $54.2$ & $55.1$ & $55.2$ & $\mathbf{ 49.8 }$\tiny{$\pm 0.1 $} & $\mathbf{ 49.1 }$\tiny{$\pm 1.0 $} \\  
                      & 5 & $64.3$\tiny{$\pm 3.2 $} & $68.5$ & $68.5$ & $68.9$ & $\mathbf{ 62.4 }$\tiny{$\pm 0.2 $} & $\mathbf{ 61.4 }$\tiny{$\pm 1.4 $} \\ \hline
\multirow{3}{*}{\begin{tabular}[c]{@{}c@{}}Tiny\\ ImageNet\end{tabular}} &
  0 &
  $\mathbf{ 37.5 }$\tiny{$\pm 0.3 $} &
  $38.5$ &
  $39.1$ &
  $42.8$ &
  $\mathbf{ 37.4 }$\tiny{$\pm 0.2 $} &
  $\mathbf{ 37.0 }$\tiny{$\pm 0.6 $} \\ 
                      & 3 & $53.8$\tiny{$\pm 0.6 $} & $55.4$ & $54.6$ & $58.7$ & $53.0$\tiny{$\pm 0.2 $}            & $\mathbf{ 52.7 }$\tiny{$\pm 0.2 $} \\ 
                      & 5 & $70.5$\tiny{$\pm 0.4 $} & $71.7$ & $71.6$ & $73.7$ & $\mathbf{ 69.9 }$\tiny{$\pm 0.2 $} & $70.2$\tiny{$\pm 0.1 $}  \\ \bottomrule
\end{tabular}}
}\\ \vspace{5pt}
\subfloat[$M = 3$; NAS-Bench-201 search space.]{
\resizebox{0.98\linewidth}{!}{ 
\begin{tabular}{@{}ccccccc@{}}
\toprule
\multirow{3}{*}{\textbf{Dataset}} &
  \multirow{3}{*}{\textbf{\begin{tabular}[c]{@{}c@{}}Shift\\ Severity\end{tabular}}} &
  \multicolumn{5}{c}{\textbf{Classif. error} (\%), $\archss = $ NAS-Bench-201 search space} \\ \cmidrule(l){3-7} 
 &
  &
  \multicolumn{1}{c}{\begin{tabular}[c]{@{}c@{}}DeepEns\\ (GDAS)\end{tabular}} &
  \multicolumn{1}{c}{\begin{tabular}[c]{@{}c@{}}DeepEns\\ (best arch.)\end{tabular}} &
  \multicolumn{1}{c}{\begin{tabular}[c]{@{}c@{}}DeepEns\\ (RS)\end{tabular}} &
  \multicolumn{1}{c}{NES-RS} &
  \multicolumn{1}{c}{NES-RE} \\ \midrule
\multirow{3}{*}{CIFAR-10}  & 0 & $8.4$  & $\mathbf{ 7.2 }$ & $7.8$\tiny{$\pm 0.2 $}  & $7.7$\tiny{$\pm 0.1 $}             & $7.6$\tiny{$\pm 0.1 $}             \\
                      & 3 & $28.7$ & $27.1$           & $28.3$\tiny{$\pm 0.3 $} & $\mathbf{ 22.0 }$\tiny{$\pm 0.2 $} & $22.5$\tiny{$\pm 0.1 $}            \\
                      & 5 & $47.8$ & $46.3$           & $37.1$\tiny{$\pm 0.0$}                  & $\mathbf{ 32.5 }$\tiny{$\pm 0.2 $} & $33.0$\tiny{$\pm 0.5 $}            \\ \midrule
\multirow{3}{*}{CIFAR-100} & 0 & $29.9$ & $26.4$           & $26.3$\tiny{$\pm 0.4 $} & $\mathbf{ 23.3 }$\tiny{$\pm 0.3 $} & $\mathbf{ 23.5 }$\tiny{$\pm 0.2 $} \\
                      & 3 & $60.3$ & $54.5$           & $57.0$\tiny{$\pm 0.9 $} & $\mathbf{ 46.6 }$\tiny{$\pm 0.3 $} & $\mathbf{ 46.7 }$\tiny{$\pm 0.5 $} \\
                      & 5 & $75.3$ & $69.9$           & $64.5$\tiny{$\pm 0.0 $} & $\mathbf{ 59.7 }$\tiny{$\pm 0.2 $} & $60.0$\tiny{$\pm 0.6 $}            \\ \midrule
\insixteen            & 0 & $49.9$ & $49.9$           & $50.5$\tiny{$\pm 0.6 $} & $\mathbf{ 48.1 }$\tiny{$\pm 1.0 $} & $\mathbf{ 47.9 }$\tiny{$\pm 0.4 $} \\ \bottomrule
\end{tabular}}
}
\end{wraptable}

\textbf{NES shows improved predictive uncertainty (NLL) and robustness to dataset shift (Figure \ref{fig:test_loss_M}).} Figure \ref{fig:test_loss_M_noshift} shows the NLL achieved by NES-RS, NES-RE and the baselines as functions of the ensemble size $M$ without dataset shift. We find that NES algorithms consistently outperform deep ensembles, with NES-RE usually outperforming NES-RS. Next, we evaluate the robustness of the ensembles to dataset shift in Figure \ref{fig:test_loss_M_shift}. In our setup, all base learners are trained on $\Dtrain$ without data augmentation of shifted examples. However, as explained in Section \ref{sec:ens-adaptation-shift}, we use a shifted validation dataset, $\Dvals$, and evaluate on a shifted test dataset, $\Dtests$. The types of shifts appearing in $\Dvals$ are disjoint from those in $\Dtests$. The severity of the shift varies between 1-5. We refer to Appendix \ref{app:exp_details} and Hendrycks \& Dietterich \citep{hendrycks2018benchmarking} for details. 
The fixed architecture used in the baseline DeepEns (RS) is selected based on its loss over $\Dvals$, but the DARTS and AmoebaNet are architectures from the literature. As shown in Figure \ref{fig:test_loss_M_shift}, ensembles picked by NES-RS and NES-RE are significantly more robust to dataset shift than the baselines, highlighting the effectiveness of applying $\esa$ with $\Dvals$. Unsurprisingly, AnchoredEns (DARTS) and DeepEns (DARTS/AmoebaNet) perform poorly compared to the other methods, as they are not optimized to deal with dataset shift here. The results of our experiments on Fashion-MNIST are in Appendix \ref{appsubsec:fmnist_exp}. In line with the results in this section, both NES algorithms outperform deep ensembles with NES-RE performing best.

\begin{figure} 
	\begin{minipage}{0.49\linewidth} 
        \centering
        \includegraphics[width=.32\linewidth]{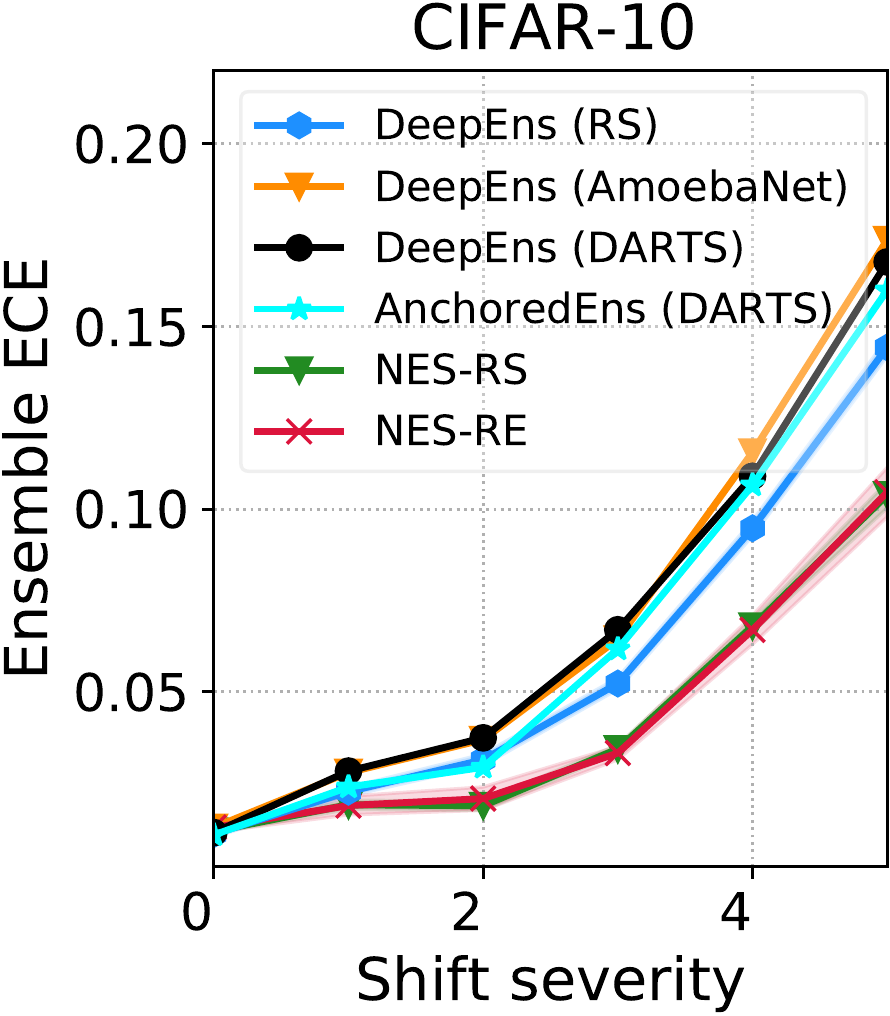}
        \includegraphics[width=.32\linewidth]{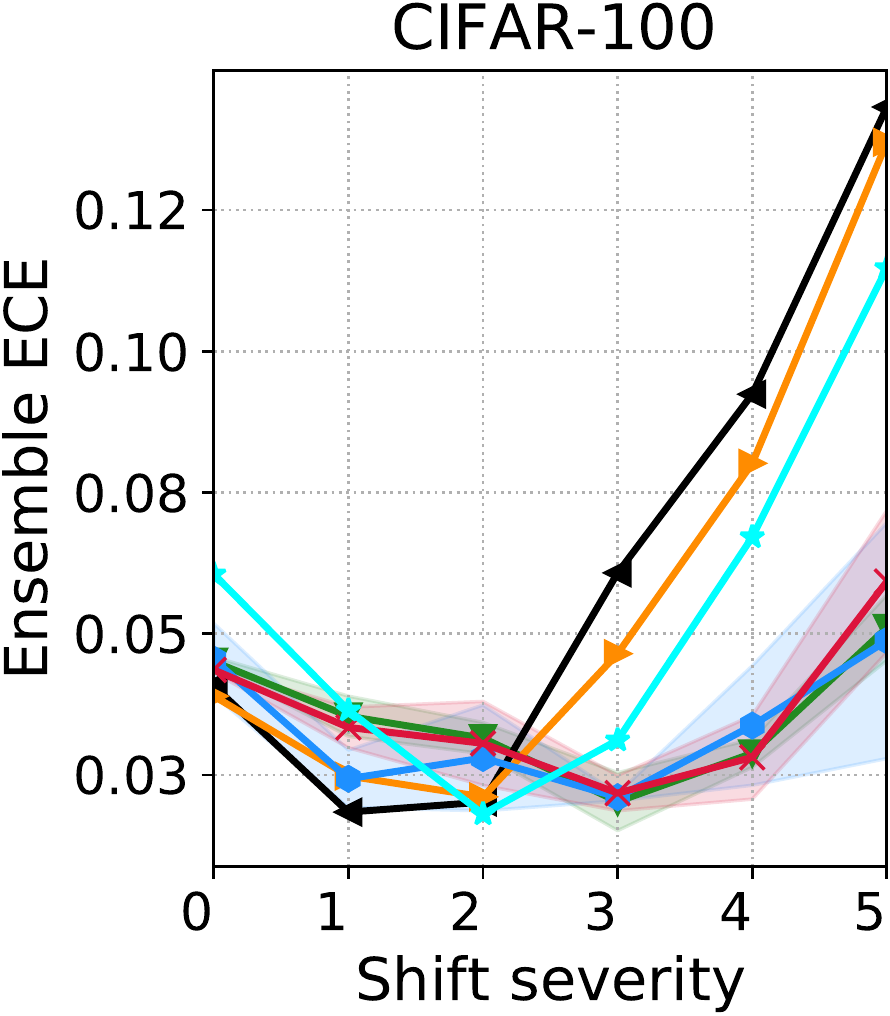}
        \includegraphics[width=.32\linewidth]{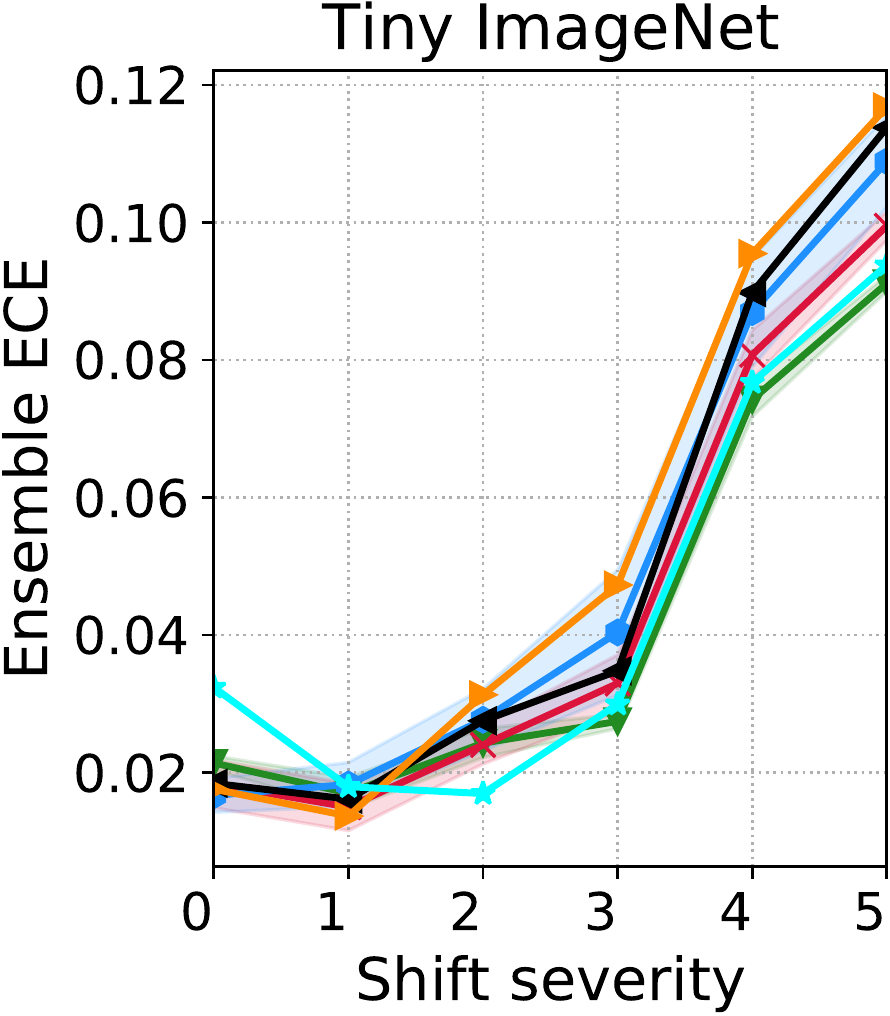}
        \captionof{figure}{ECE vs. dataset shift severity on CIFAR-10, CIFAR-100 and Tiny ImageNet over the DARTS search space. No dataset shift is indicated as severity 0. Ensemble size is $M = 10$.}
        \label{fig:ece_sev}
	\end{minipage}\hspace{6pt}
	\begin{minipage}{0.49\linewidth}
        \centering
        \includegraphics[width=.32\linewidth]{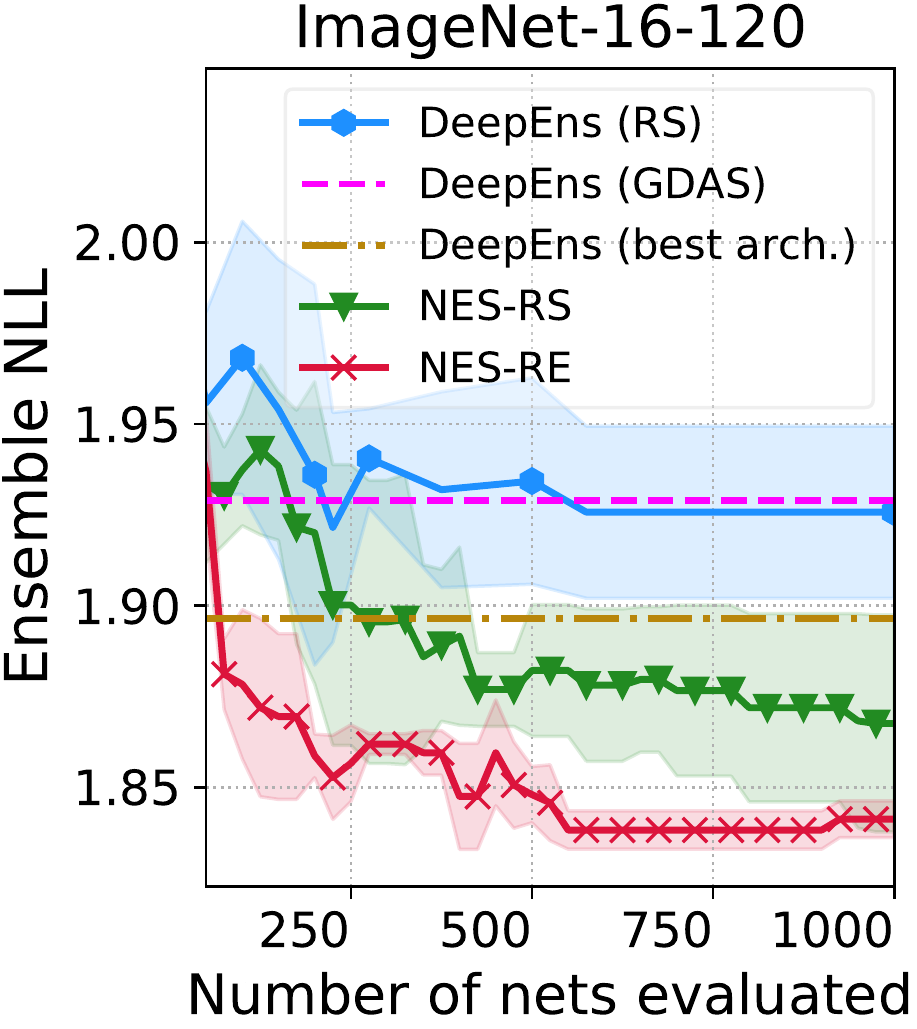}
        \includegraphics[width=.32\linewidth]{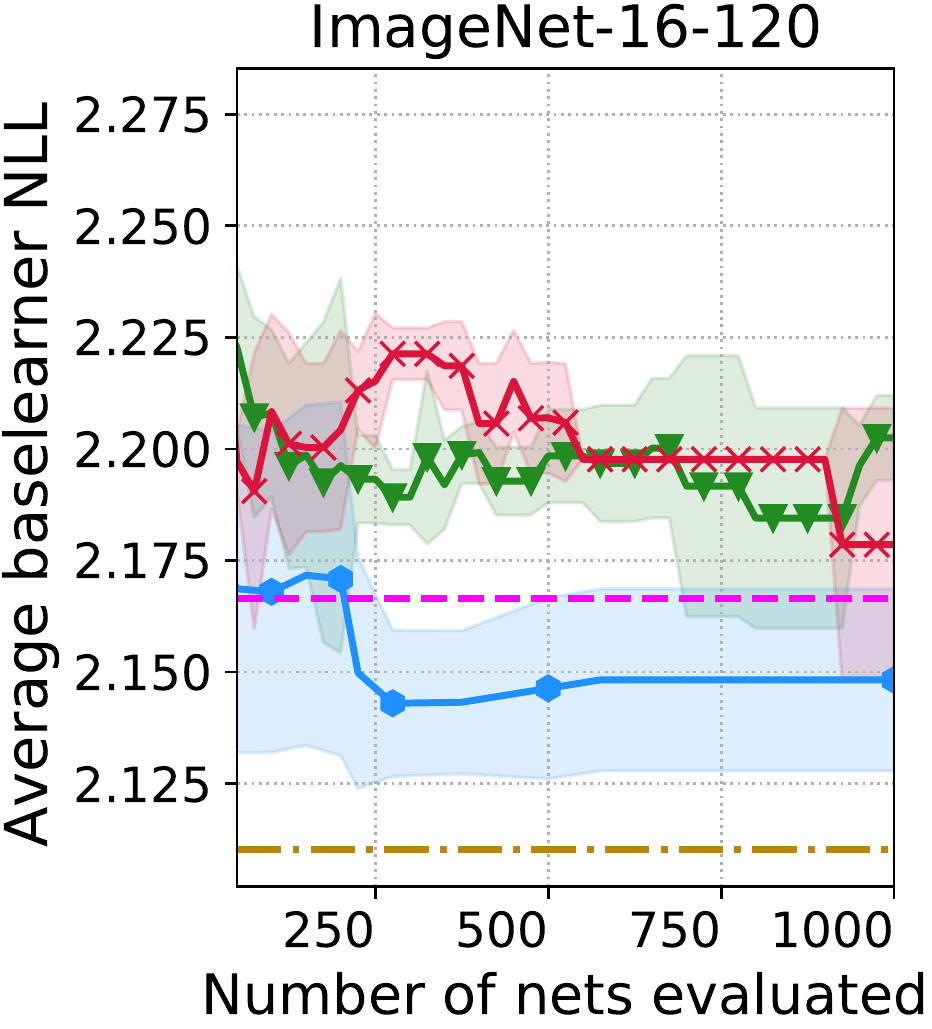}
        \includegraphics[width=.32\linewidth]{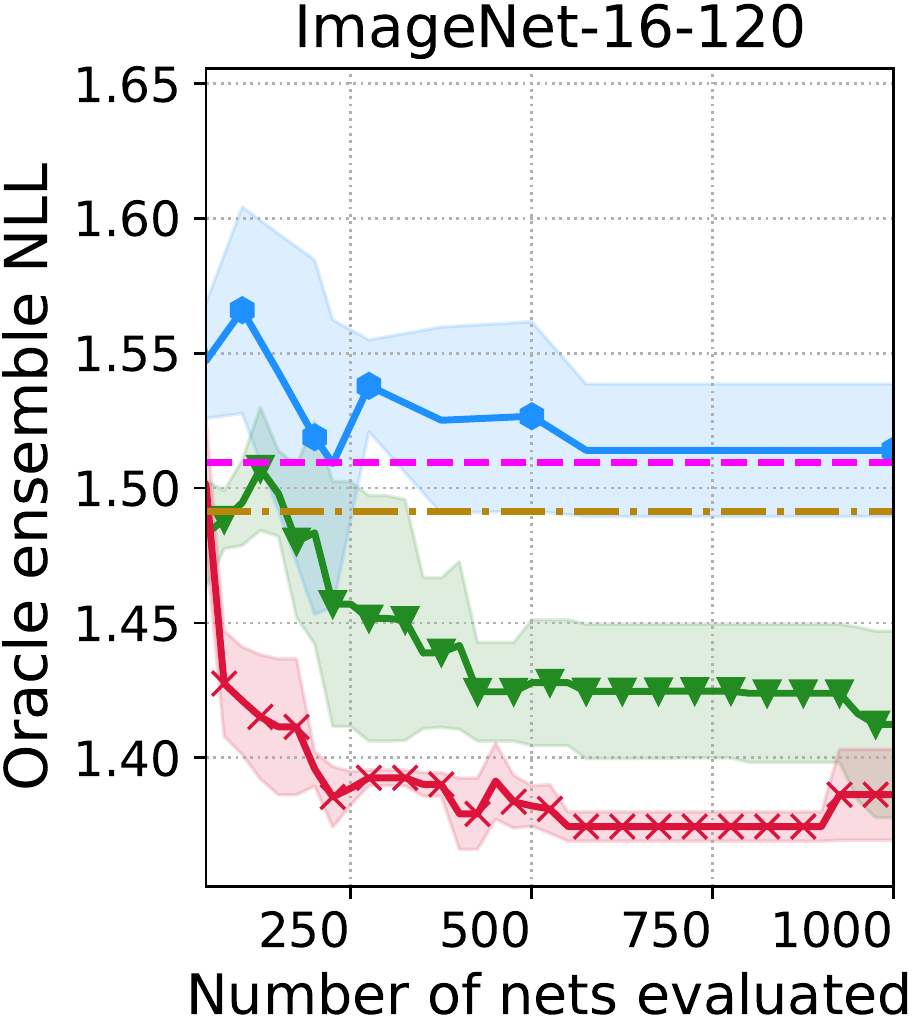}
        \caption{Ensemble, average base learner and oracle ensemble NLL versus budget $\budget$ on \insixteen over the NAS-Bench-201 search space. Ensemble size is $M = 3$.}
        \label{fig:nb201_imagenet}
	\end{minipage}\hfill
	\vskip -0.2in
\end{figure}

\textbf{Better uncertainty calibration versus dataset shift and classification error (Figure \ref{fig:ece_sev}, Table \ref{tbl:error}).} We also assess the ensembles by error and ECE. In short, ECE measures the mismatch between the model's confidence and accuracy. Figure \ref{fig:ece_sev} shows the ECE achieved by the ensembles at varying dataset shift severities, noting that uncertainty calibration is especially important when models are used during dataset shift. Ensembles found with NES tend to exhibit better uncertainty calibration and are either competitive with or outperform anchored and deep ensembles for most shift severities. Notably, on CIFAR-10, ECE is reduced by up to $40\%$ relative to the baselines. In terms of classification error, we find that ensembles constructed by NES outperform deep ensembles, with reductions of up to $7$ percentage points in error, shown in Table \ref{tbl:error}. As with NLL, NES-RE tends to outperforms NES-RS. 

\begin{figure} 
	\begin{minipage}{0.50\linewidth} 
        \setlength{\tabcolsep}{1.2pt}
        \renewcommand{\arraystretch}{1.3} 
        \captionof{table}{Diversity and base learner strength. Predictive disagreement (larger means more diverse) and oracle ensemble NLL (smaller means more diverse) are defined in Section \ref{sec:diversity_of_ens}. Despite the stronger base learners, deep ensembles tend to be less diverse than ensembles constructed by NES. The results are consistent across datasets and shift severities (Appendix \ref{app:add_exp}). Best values and all values within $95\%$ confidence interval are bold faced.}
        \resizebox{\linewidth}{!}{%
        \begin{tabular}{clcccccc}
        \hline
        \multirow{2}{*}{\textbf{Dataset}} &
          \multirow{2}{*}{\textbf{Metric}} &
          \multicolumn{6}{c}{\textbf{Method} (with $M=10$)} \\ \cline{3-8} 
         &
           &
          \begin{tabular}[c]{@{}c@{}}DeepEns\\ (RS)\end{tabular} &
          \begin{tabular}[c]{@{}c@{}}DeepEns\\ (Amoe.)\end{tabular} &
          \begin{tabular}[c]{@{}c@{}}DeepEns\\ (DARTS)\end{tabular} &
          \begin{tabular}[c]{@{}c@{}}AnchoredEns\\ (DARTS)\end{tabular} &
          NES-RS &
          NES-RE \\ \hline
        \multirow{3}{*}{CIFAR-10} &
          Pred. Disagr. &
          $0.823$\tiny{$\pm 0.023 $} &
          $\mathbf{ 0.947 }$ &
          $0.932$ &
          $0.842$ &
          $\mathbf{ 0.948 }$\tiny{$\pm 0.004 $} &
          $0.943$\tiny{$\pm 0.009 $} \\
         &
          Oracle NLL &
          $0.125$\tiny{$\pm 0.007 $} &
          $0.103$ &
          $0.093$ &
          $0.113$ &
          $\mathbf{ 0.086 }$\tiny{$\pm 0.001 $} &
          $0.088$\tiny{$\pm 0.002 $} \\
         &
          Avg. bsl. NLL &
          $0.438$\tiny{$\pm 0.005 $} &
          $0.552$ &
          $0.513$ &
          $\mathbf{ 0.411 }$ &
          $0.485$\tiny{$\pm 0.003 $} &
          $0.493$\tiny{$\pm 0.010 $} \\ \hline
        \multirow{3}{*}{CIFAR-100} &
          Pred. Disagr. &
          $0.831$\tiny{$\pm 0.027 $} &
          $\mathbf{ 0.946 }$ &
          $0.935$ &
          $0.839$ &
          $0.934$\tiny{$\pm 0.006 $} &
          $0.943$\tiny{$\pm 0.014 $} \\
         &
          Oracle NLL &
          $0.635$\tiny{$\pm 0.040 $} &
          $0.502$ &
          $0.509$ &
          $0.583$ &
          $\mathbf{ 0.498 }$\tiny{$\pm 0.008 $} &
          $\mathbf{ 0.487 }$\tiny{$\pm 0.014 $} \\
         &
          Avg. bsl. NLL &
          $1.405$\tiny{$\pm 0.028 $} &
          $1.552$ &
          $1.491$ &
          $\mathbf{ 1.290 }$ &
          $1.467$\tiny{$\pm 0.022 $} &
          $1.487$\tiny{$\pm 0.036 $} \\ \hline
        \multirow{3}{*}{\begin{tabular}[c]{@{}c@{}}Tiny\\ ImageNet\end{tabular}} &
          Pred. Disagr. &
          $0.742$\tiny{$\pm 0.008 $} &
          $0.737$ &
          $0.749$ &
          $0.662$ &
          $\mathbf{ 0.772 }$\tiny{$\pm 0.005 $} &
          $\mathbf{ 0.768 }$\tiny{$\pm 0.005 $} \\
         &
          Oracle NLL &
          $0.956$\tiny{$\pm 0.009 $} &
          $0.987$ &
          $1.009$ &
          $1.203$ &
          $0.929$\tiny{$\pm 0.007 $} &
          $\mathbf{ 0.910 }$\tiny{$\pm 0.015 $} \\
         &
          Avg. bsl. NLL &
          $1.743$\tiny{$\pm 0.005 $} &
          $1.764$ &
          $1.813$ &
          $1.871$ &
          $1.755$\tiny{$\pm 0.007 $} &
          $\mathbf{ 1.728 }$\tiny{$\pm 0.012 $} \\ \hline
        \end{tabular}
        }
        \label{tbl:diversity}
	\end{minipage}\hspace{6pt}
	\begin{minipage}{0.48\linewidth}%
        \centering
            \begin{minipage}{.45\linewidth}
                \includegraphics[width=\linewidth]{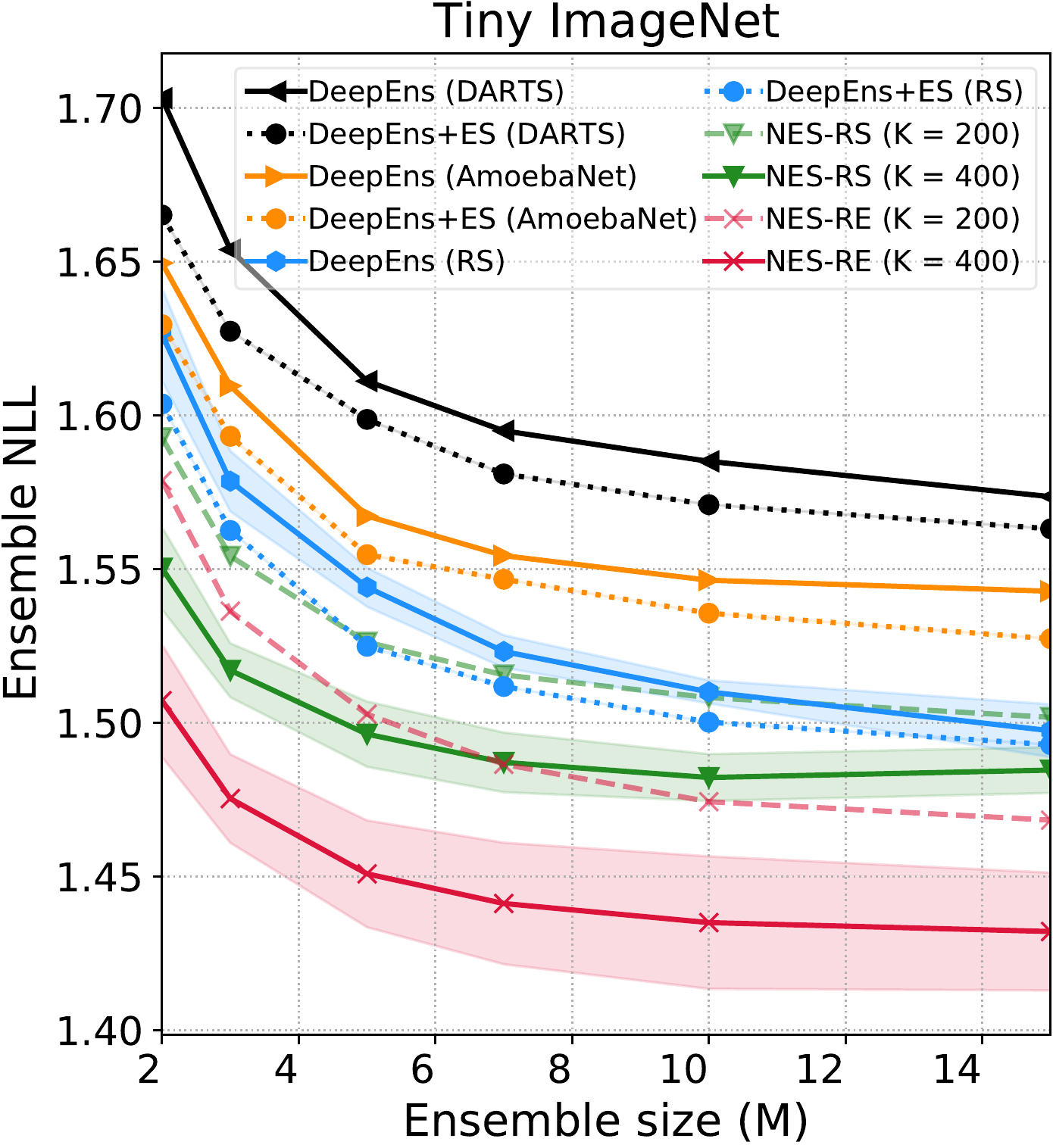}
            \end{minipage}\hfill
            \begin{minipage}{.53\linewidth}
                \resizebox{\linewidth}{!}{%
                \setlength{\tabcolsep}{1.8pt}
                \begin{tabular}{@{}lcc@{}}
                \toprule
                \multicolumn{1}{c}{\multirow{2}{*}{\textbf{Method}}} &
                \multicolumn{2}{c}{\textbf{\# nets trained}} \\ \cmidrule(l){2-3} 
                \multicolumn{1}{c}{} & {Arch.} & \multicolumn{1}{l}{{Ensemble}} \\ \midrule
                DeepEns (DARTS)           & 32 & $10$ \\
                DeepEns + ES (DARTS)      & 32 & $200$ \\
                DeepEns (AmoebaNet)       & 25200 & $10$ \\
                DeepEns + ES (AmoebaNet)  & 25200 & $200$ \\
                DeepEns (RS)              & $200$ & $10$ \\
                DeepEns + ES (RS)         & $200$ & $200$ \\
                NES-RS $(K=200)$          & \multicolumn{2}{c}{$200$} \\
                NES-RS $(K=400)$          & \multicolumn{2}{c}{$400$} \\
                NES-RE $(K=200)$          & \multicolumn{2}{c}{$200$} \\
                NES-RE $(K=400)$          & \multicolumn{2}{c}{$400$} \\ \bottomrule 
                \end{tabular}
                } \vspace{2mm}
            \end{minipage}
        \caption{Comparison of NES to deep ensembles with ensemble selection on Tiny ImageNet. \textsc{Left}: NLL vs ensemble size. \textsc{Right}: Cost for Tiny ImageNet experiments reported in terms of the number of networks trained when $M=10$. The ``arch'' column indicates the number of networks trained to first select an architecture, and the ``ensemble'' column contains the number of networks trained to build the ensemble.} 
        \label{fig:deepens_es_ablation_main_paper}
        \vskip -0.1in 
	\end{minipage}\hfill
	\vskip -0.15in
\end{figure}

\textbf{Ensembles found by NES tend to be more diverse (Table \ref{tbl:diversity}).} 
We measure ensemble diversity using the two metrics predictive disagreement (larger means more diverse) and oracle ensemble NLL (smaller means more diverse) as defined in Section \ref{sec:diversity_of_ens} and average base learner NLL. Table \ref{tbl:diversity} contrasts NES with the baselines in terms of these metrics. In terms of both diversity metrics, ensembles constructed by NES tend to be more diverse than anchored and deep ensembles. Ranking of the methods is largely consistent for different ensemble sizes $M$ (see Appendix \ref{app:add_exp}). Unsurprisingly, note that the average base learner in NES is not always the best: by optimizing the fixed architecture first and \textit{then} ensembling it, deep ensembles end up with a strong average base learner at the expense of less ensemble diversity. Despite higher average base learner NLL, NES ensembles perform better (recall Figure \ref{fig:test_loss_M}), highlighting once again the importance of diversity. 

\textbf{NES outperforms the deep ensemble of the best architecture in the NAS-Bench-201 search space (Figure \ref{fig:nb201_imagenet}).} Next, we compare NES to deep ensembles over the NAS-Bench-201 search space, which has two benefits: we demonstrate that our findings are not specific to the DARTS search space, and NAS-Bench-201 is an exhaustively evaluated search space for which all architectures' trained weights are available (three initializations per architecture), allowing us to compare NES to the deep ensemble of the \textit{best} architecture by validation loss. Results shown in Figure \ref{fig:nb201_imagenet} compare the losses of the ensemble, average base learner and oracle ensemble versus the number of networks evaluated $\budget$. Interestingly, although DeepEns (best arch.) has a significantly stronger average base learner than the other methods, its lack of diversity, as indicated by higher oracle ensemble loss (Figure \ref{fig:nb201_imagenet}) and lower predictive disagreement (Figure \ref{fig:radar_plot}), yields a weaker ensemble than both NES algorithms. Also, NES-RE outperforms NES-RS with a 6.6x speedup as shown in Figure \ref{fig:nb201_imagenet}-left.

\vspace{-5pt}
\subsection{Analysis and Ablations}
\label{sec:analysis_and_ablations}
\vspace{-5pt}
\textbf{Why does NES work? What if deep ensembles use ensemble selection over initializations? (Figure \ref{fig:deepens_es_ablation_main_paper}).} NES algorithms differ from deep ensembles in two important ways: the ensembles use varying architectures and NES utilizes ensemble selection ($\esa$) to pick the base learners. On Tiny ImageNet over the DARTS search space, we conduct an experiment to explore whether the improvement offered by NES over deep ensembles is only due to ensemble selection. The baselines ``DeepEns + ES'' operate as follows: we optimize a fixed architecture for the base learners, train $K$ random initializations of it to form a pool and apply $\esa$ to select an ensemble of size $M$. Figure \ref{fig:deepens_es_ablation_main_paper}-left shows that both NES algorithms (each shown for two computational budgets $K = 200, 400$) outperform all DeepEns + ES baselines. Figure \ref{fig:deepens_es_ablation_main_paper}-right contains the cost of each method in terms of the number of networks trained. Note that DeepEns + ES (RS) is the most competitive of the deep ensemble baselines, and, at an equal budget of 400, it is outperformed by both NES algorithms. However, even at half the cost (200), NES-RE outperforms DeepEns + ES (RS) while NES-RS performs competitively. As expected, deep ensembles \textit{with} ensemble selection consistently perform better than \textit{without} ensemble selection at the expense of higher computational cost, but do not close the gap with NES algorithms. We also re-emphasize that comparisons between NES and deep ensembles in our experiments always fix the base learner training routine and method for composing the ensemble, so variations in ensemble performance are only due to the architecture choices. Therefore, to summarize, combined with the finding that NES outperforms deep ensembles even when NES' average base learner is weaker (as in e.g. Figure \ref{fig:nb201_imagenet} and Table \ref{tbl:diversity}), we find architectural variation in ensembles to be important for NES' performance gains. 

\begin{wrapfigure}[20]{R}{.25\textwidth}
    \vspace{-7mm}
    \centering
    \includegraphics[width=.99\linewidth]{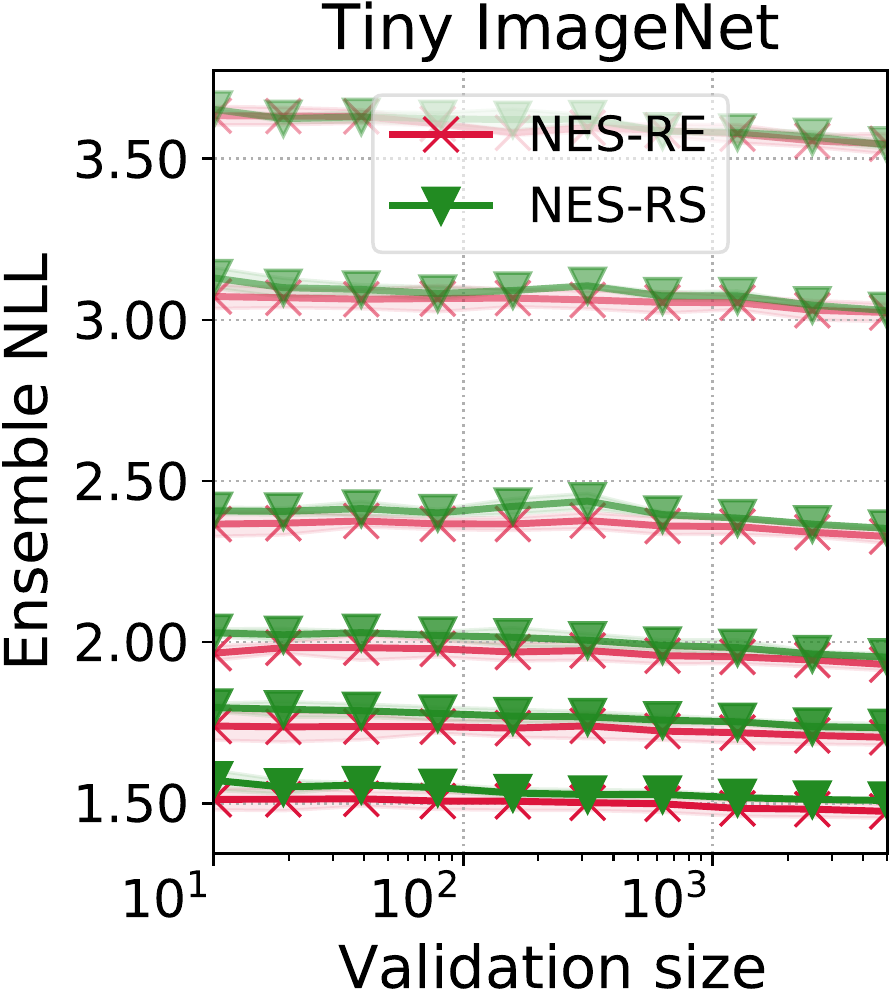}
    \caption{Test performance of NES algorithms with varying validation data sizes. Each curve corresponds to one particular dataset shift severity (0-5 with 0 being no shift). The more transparent curves correspond to higher shift severities.}
    \label{fig:val_sensitivity}
\end{wrapfigure}

\textbf{Computational cost and parallelizability.} The primary computational cost involved in NES is training $\budget$ networks to build the pool $\pool$. In our experiments on the DARTS search space, we set $\budget$ to be $400$ for CIFAR-10/100 and $200$ for Tiny ImageNet (except Figure \ref{fig:deepens_es_ablation_main_paper} which additionally considers $\budget = 400$). Figure \ref{fig:deepens_es_ablation_main_paper}-right gives an example of costs for each method. The cost of a deep ensemble with an optimized architecture stems from the initial architecture search and the subsequent training of $M$ random initializations to build ensemble. We refer to Appendix \ref{app:exp_details} for details and discussion of computation cost, including training times. We note that apart from DeepEns (DARTS) and AnchoredEns (DARTS), NES has a lower computational cost than the baselines in our experiments. Similar to deep ensembles, training the pool for NES-RS is embarrassingly parallel. Our implementation of NES-RE is also parallelized as described in Appendix \ref{app:exp_details}.

\textbf{Comparison of different ensemble selection algorithms.} Both NES algorithms utilize $\esa$ as the ensemble selection algorithm (ESA). We experimented with various other choices of ESAs, including weighted averaging and explicit diversity regularization, as shown in Figure \ref{fig:esa_ablation}. A detailed description of each ESA is in Appendix \ref{app:esa_comparison}. In summary, our choice of $\esa$ performs better than or at par with all ESAs considered. Moreover, weighted averaging using stacking and Bayesian model averaging has a very minor impact since the weights end up being close to uniform. Explicit diversity regularization, as shown in Figure \ref{fig:esa_div_comparison}, appears to slightly improve performance in some cases,
provided that the diversity regularization strength hyperparameter is appropriately tuned.

\begin{figure}
\begin{minipage}{0.49\linewidth} 
    \centering
    \captionsetup[subfigure]{justification=centering}
    \begin{subfigure}[t]{0.49\linewidth}
        \centering
        \includegraphics[width=.99\linewidth]{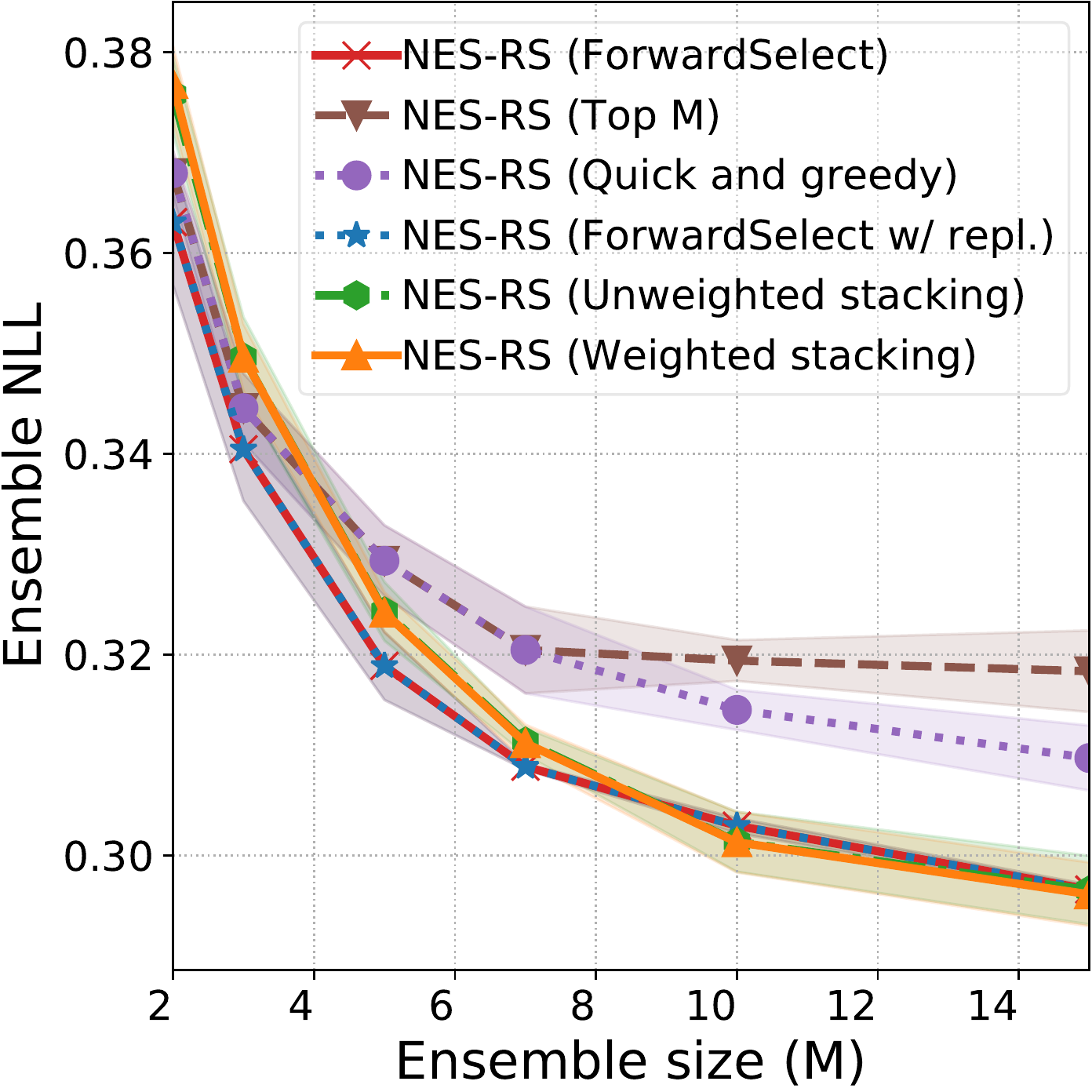}
        \subcaption{Ensemble selection algorithms.}
        \label{fig:esa_comparison}
    \end{subfigure}
    \begin{subfigure}[t]{0.49\textwidth}
        \centering
        \includegraphics[width=.99\linewidth]{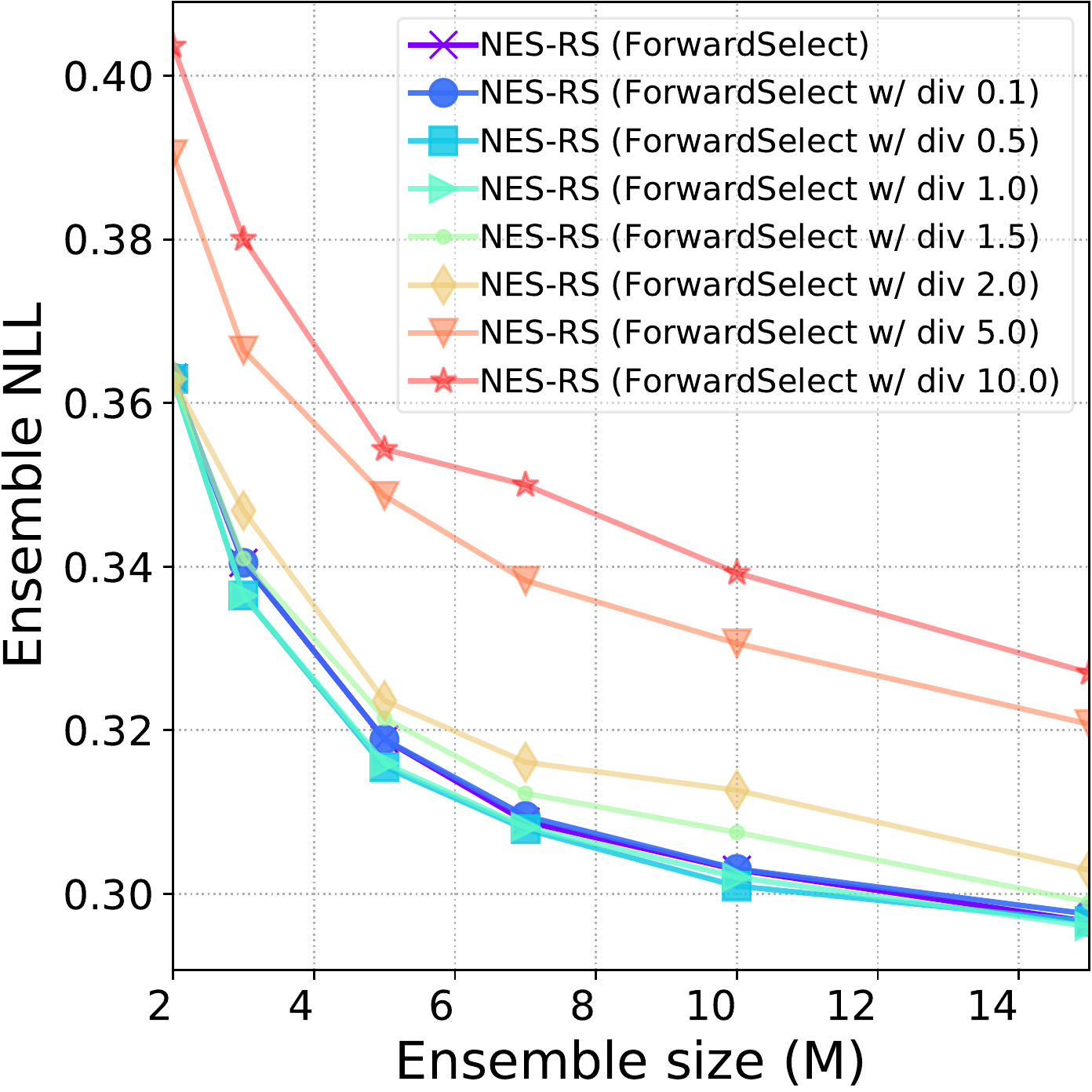}
        \subcaption{Ensemble selection with explicit diversity regularization.}
        \label{fig:esa_div_comparison}
    \end{subfigure}
    \caption{NES-RS on CIFAR-10.}
    \label{fig:esa_ablation}
\end{minipage}\hspace{6pt}
\begin{minipage}{0.49\linewidth}
    \centering
    \includegraphics[width=.99\linewidth]{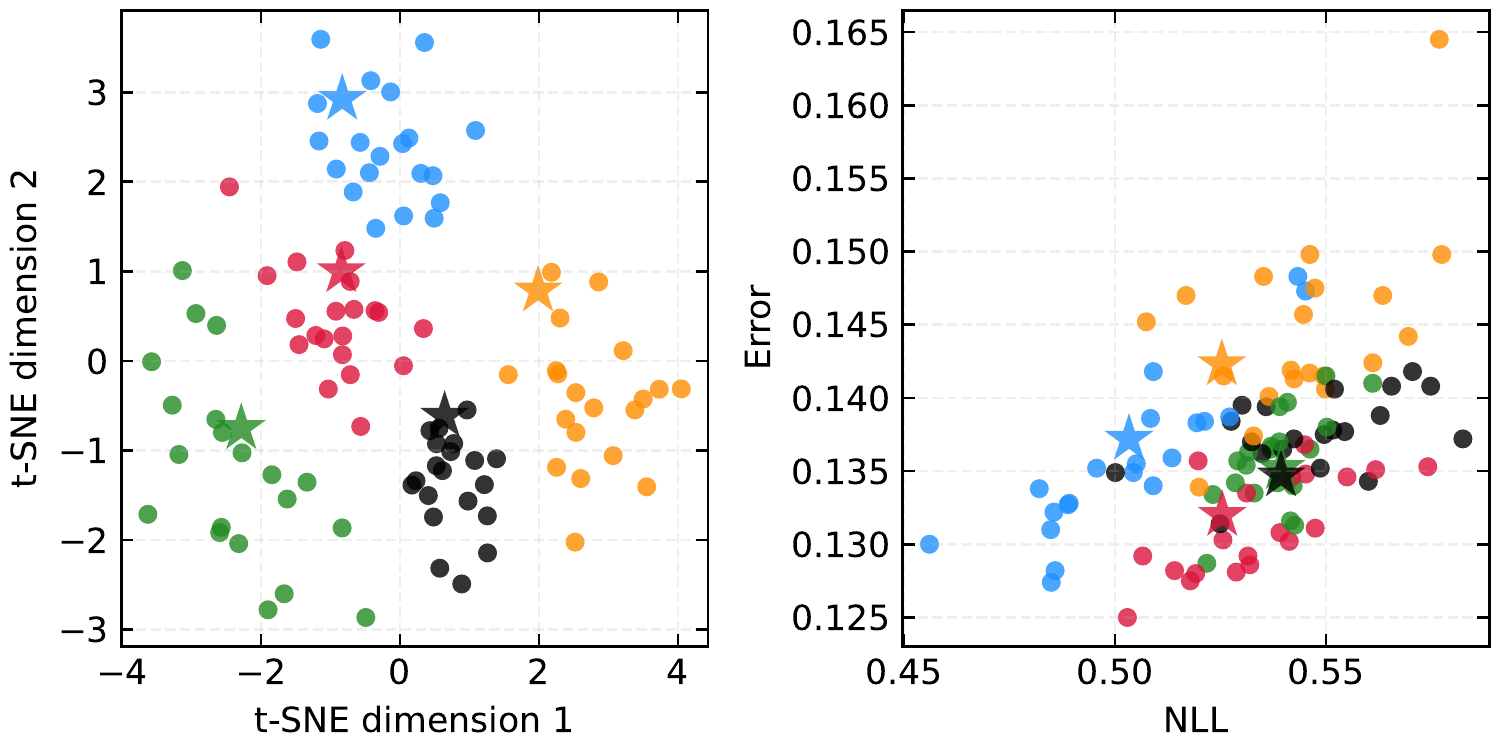}
    \caption{{\small\textsc{Left:}} t-SNE of the test predictions of 20 single random mutations (circles) from 5 parent architecture (stars). 
    {\small \textsc{Right:}} NLL and error achieved by these architectures.}
    \label{fig:mutations_tsne}
\end{minipage}
\vspace{-1em}
\end{figure}

\textbf{NES is insensitive to the size of validation dataset $\Dval$.} We study the sensitivity of NES to the size of $\Dval$. Specifically, we measure test loss of the ensembles selected using $\Dval$ of different sizes (with as few as 10 validation samples). The results in Figure \ref{fig:val_sensitivity} indicate that NES is insensitive to validation size for different levels of dataset shift severity, achieving almost the same loss when using 10$\times$ fewer validation data. 
We provide more details in Appendix \ref{app:val-data-sensitivity-overfitting} and we also discuss why overfitting to $\Dval$ is averted during ensemble selection.

\textbf{Mutations of an architecture are similar in function space and by performance.} To explore how NES-RE traverses the search space, we analyzed how mutations affect an architecture as follows. We sampled five random architectures from the DARTS space, and for each one, we applied a single random mutation twenty times, yielding a “family” of parent-children architectures, which are then trained on CIFAR-10. Figure~\ref{fig:mutations_tsne}-left shows the result of t-SNE applied to the test predictions of these architectures, where each color corresponds to one “family”, of which the “star” is the parent architecture and the circles are the children architectures. The clustering demonstrates that architectures which differ by only a single mutation are similar in function space after training. Figure~\ref{fig:mutations_tsne}-right shows the NLL and error achieved by these architectures. Again, similar clustering shows that architectures differing by a single mutation also perform similarly w.r.t. NLL and error. This confirms that mutations allow for locally exploring the search space.

\textbf{Further experiments.} We also provide additional experiments in the Appendix.  This includes a comparison of ensembles built by averaging logits vs. probabilities (Appendix~\ref{app:logits_vs_probabilities}), a comparison of NES to ensembles with other hyperparameters being varied (either width/depth or training hyperparameters similar to Wenzel \etal\citep{wenzel2020hyperparameter}) (Appendix \ref{app:hyperdeepens}) and using a weight-sharing model~\citep{bender_icml:2018} as a proxy to accelerate the search phase in NES (Appendix \ref{app:one_shot_nes}).

%% file: text/6_conclusions.tex
\vspace{-5pt}
\section{Conclusion, Limitations \& Broader Impact} \label{sec:conclusions}

\vspace{-5pt}
We presented Neural Ensemble Search for automatically constructing ensembles with varying architectures and demonstrated that the resulting ensembles outperform state-of-the-art deep ensembles in terms of uncertainty estimation and robustness to dataset shift. Our work highlights the benefit of ensembling varying architectures. In future work, we aim to address the limitation posed by the computational cost of NES due to building the pool $\pool$. An interesting approach in this direction could be the use of differentiable NAS methods to simultaneously optimize base learner architectures within a one-shot model and reduce cost~\citep{liu2018darts, Xu2020PC-DARTS:, chen2021drnas}. More generally, we also hope to explore what other hyperparameters can be varied to improve ensemble performance and how best to select them. 

\vspace{-1pt}
Our work can readily be applied to many existing ensemble-based deep learning systems to improve their predictive performance. This work also focuses on improving uncertainty estimation in neural networks, which is a key problem of growing importance with implications for safe deployment of such systems. We are not aware of any direct negative societal impacts of our work, since NES is task-agnostic and its impact depends on its applications.

%% file: text/B_supplementary.tex
\appendix
{\large{\textbf{Supplementary Material for Neural Ensemble Search for Uncertainty Estimation and Dataset Shift}}}

\vspace{-5pt}
\section{Proof of Proposition \ref{prop:loss-ineq}}\label{app:inequality}

\vspace{-5pt}
Taking the loss function to be NLL, we have $\loss(f(\bx), y)) = - \log {[f(\bx)]_y}$, where $[f(\bx)]_y$ is the probability assigned by the network $f$ of $\bx$ belonging to the true class $y$, i.e. indexing the predicted probabilities $f(\bx)$ with the true target $y$. Note that $t \mapsto -\log{t}$ is a convex and decreasing function.

We first prove $\loss(\oracens(\bx), y) \leq \loss(F(\bx), y)$. Recall, by definition of $\oracens$, we have $\oracens(\bx) = f_{\theta_k} (\bx)$ where $k \in \argmin_i \loss(f_{\theta_i}(\bx), y)$, therefore $[\oracens(\bx)]_y = [f_{\theta_k}(\bx)]_y \geq [f_{\theta_i}(\bx)]_y$ for all $i = 1, \dots, \numclass$. That is, $f_{\theta_k}$ assigns the highest probability to the correct class $y$ for input $\bx$. Since $-\log$ is a decreasing function, we have
\begin{align}
    \loss(F(\bx), y) &= -\log { \left(\frac{1}{M} \sum_{i = 1}^M [f_{\theta_i}(\bx)]_y \right) } \nonumber \\
    &\geq -\log {\left( [f_{\theta_k}(\bx)]_y \right) } = \loss(\oracens(\bx), y). \nonumber
\end{align}

We apply Jensen's inequality in its finite form  for the second inequality. Jensen's inequality states that for a real-valued, convex function $\varphi$ with its domain being a subset of $\bbR$ and numbers $t_1, \dots, t_n$ in its domain, $\varphi(\frac{1}{n} \sum_{i=1}^n t_i) \leq \frac{1}{n}\sum_{i=1}^n \varphi(t_i)$. Noting that $-\log$ is a convex function, $\loss(F(\bx), y) \leq \frac{1}{M} \sum_{i=1}^M \loss (f_{\theta_i}(\bx), y)$ follows directly.

\vspace{-5pt}
\section{Experimental and Implementation Details}\label{app:exp_details}

\vspace{-5pt}
We describe details of the experiments shown in Section \ref{sec:experiments} and Appendix \ref{app:add_exp} and include Algorithms \ref{alg:nes-rs} and \ref{alg:esa} describing NES-RS and $\esa$, respectively. Note that unless stated otherwise, all sampling over a discrete set is done uniformly in the discussion below.

\IncMargin{0em}{
\begin{algorithm}
\KwData{Search space $\mathcal{A}$; ensemble size $M$; comp. budget $\budget$; $\Dtrain, \Dval$.}
Sample $\budget$ architectures $\alpha_1, \dots, \alpha_\budget$ independently and uniformly from $\archss$.\\
Train each architecture $\alpha_i$ using $\Dtrain$, yielding a pool of networks $\pool = \{f_{\theta_1, \alpha_1}, \dots, f_{\theta_K, \alpha_K}\}$.\\
Select base learners $\{f_{\theta_1^*,\alpha_1^*}, \dots, f_{\theta_M^*, \alpha_M^*}\} = \esa(\pool, \Dval, M)$ by forward step-wise selection without replacement. \label{lst:line3}\\
\Return{ensemble $\ens(f_{\theta_1^*,\alpha_1^*}, \dots, f_{\theta_M^*, \alpha_M^*})$}
\caption{NES with Random Search}
\label{alg:nes-rs}
\end{algorithm}}

\vspace{-9pt}
\IncMargin{0em}{
\begin{algorithm}
\DontPrintSemicolon
\KwData{Pool of base learners $\pool$; ensemble size $M$; $\Dval$. Assume $|\pool| \geq M$.}
Initialize an empty set of base learners $E = \{\}$.\\
\While{$|E| < M$}{
    Add $f_{\theta, \alpha}$ to $E$, where $f_{\theta, \alpha} \in \argmin_{f \in \pool} \Loss(\ens(E \cup \{f\}), \Dval)$ \\ %
    Remove $f_{\theta, \alpha}$ from $\pool$. \tcp*{without replacement.} 
}
\Return{$E$} \tcp*{set of selected base learners.}
\caption{$\esa$ (forward step-wise selection without replacement \citep{rich_ens_select})}
\label{alg:esa}
\end{algorithm}}

\vspace{-5pt}
\subsection{Architecture Search Spaces}\label{app:arch_search_space_descriptions}

\begin{figure}[t]
    \centering
    \captionsetup[subfigure]{justification=centering}
    \begin{subfigure}[t]{0.8\textwidth}
        \centering
        \includegraphics[width=.95\linewidth]{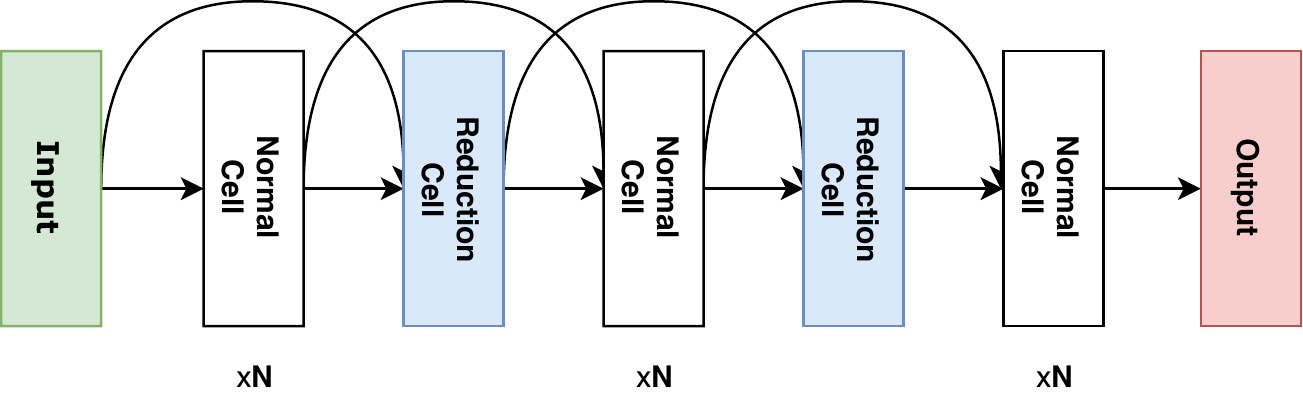}
        \subcaption{Macro architecture.}
        \label{fig:macro_network}
    \end{subfigure}\\%
    \begin{subfigure}[t]{0.89\textwidth}
        \centering
        \includegraphics[width=.49\linewidth]{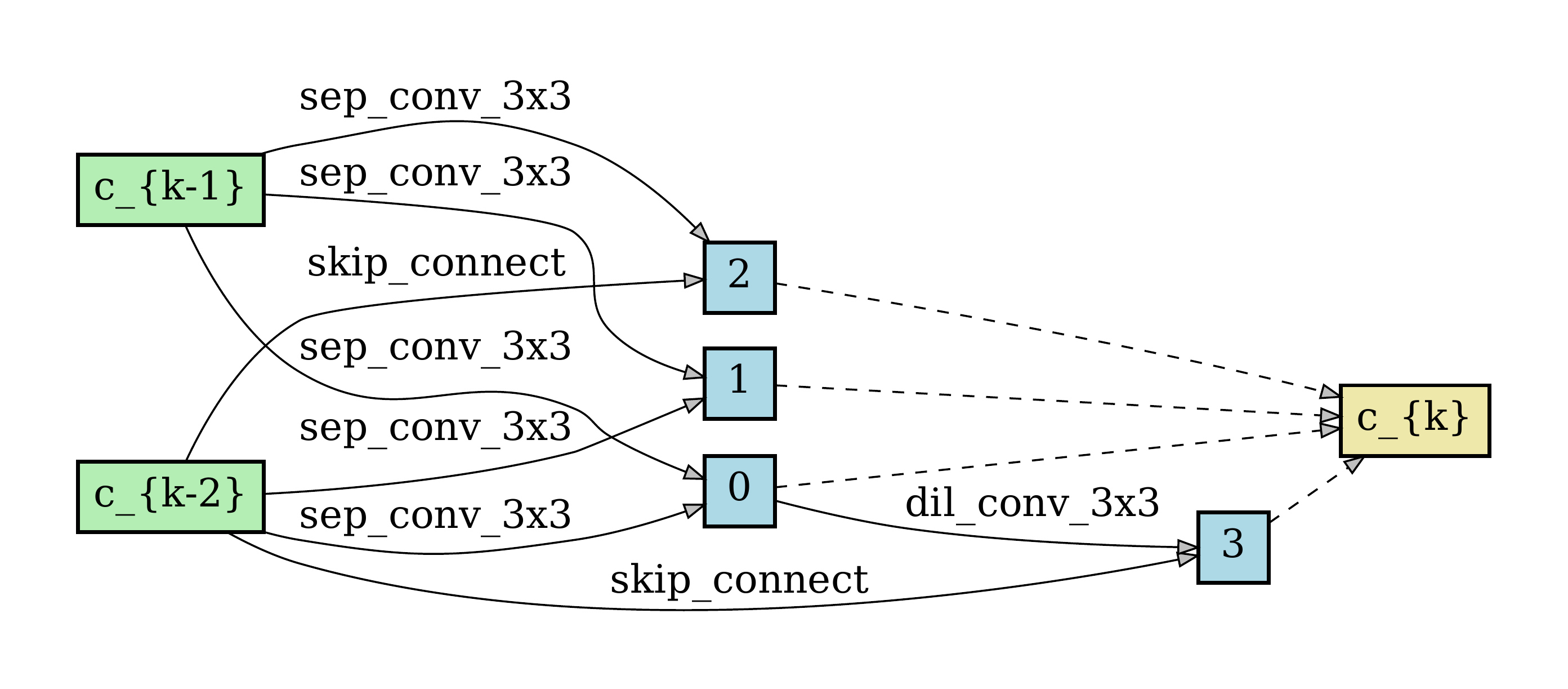}
        \includegraphics[width=.49\linewidth]{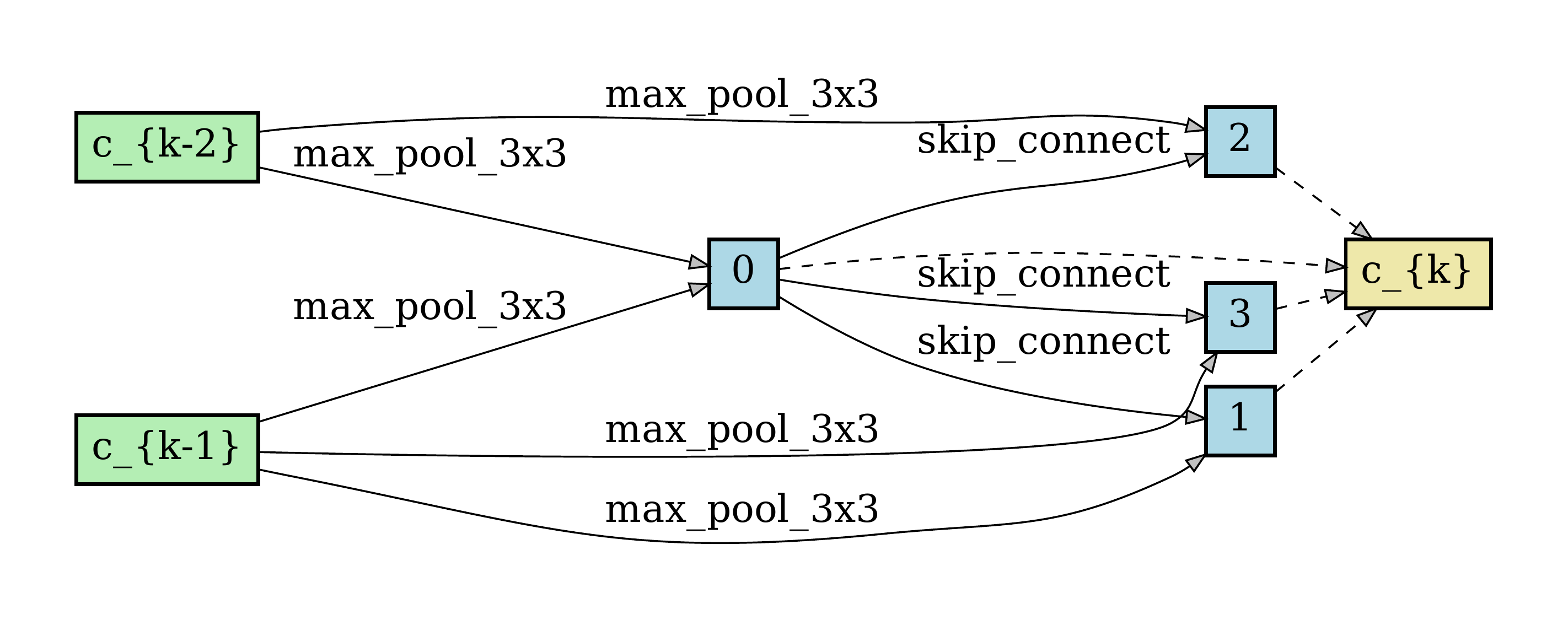}
        \subcaption{Normal (left) and reduction (right) cells.}
        \label{fig:normal_reduction_cells}
    \end{subfigure}
    \caption{Illustration of the DARTS search space: (a) The macro architecture is a stack of normal and reduction cells. Normal cells are repeated $N$ times between reduction cells ($N=2$ in our experiments, i.e. 8 cells in total). (b) The cells for the DARTS architecture are depicted as directed acyclic graphs. Briefly, cells are composed of 2 input nodes (green), 4 intermediate nodes (blue) and one output node (red). Each edge is an operation which is applied to the preceding node's tensor and the output is summed element-wise in the succeeding node (an intermediate node). The output node is a concatenation of all intermediate nodes.}
    \vspace{-9pt}
    \label{fig:darts_cell_space}
\end{figure}

\paragraph{DARTS search space. }
The first architecture search space we consider in our experiments is the one from DARTS~\citep{liu2018darts}. We search for two types of \textit{cells}: \emph{normal} cells, which preserve the spatial dimensions, and \emph{reduction} cells, which reduce the spatial dimensions. See Figure \ref{fig:normal_reduction_cells} for an illustration. The cells are stacked in a macro architecture, where they are repeated and connected using skip connections (shown in Figure~\ref{fig:macro_network}). Each cell is a directed acyclic graph, where nodes represent feature maps in the computational graph and edges between them correspond to operation choices (e.g. a convolution operation). The cell parses inputs from the previous and previous-previous cells in its 2 input nodes. Afterwards it contains 5 nodes: 4 intermediate nodes that aggregate the information coming from 2 previous nodes in the cell and finally an output node that concatenates the output of all intermediate nodes across the channel dimension. AmoebaNet contains one more intermediate node, making that a deeper architecture.
The set of possible operations (eight in total in DARTS) that we use for each edge in the cells is the same as DARTS, but we leave out the ``zero'' operation since that is not necessary for non-differentiable approaches such as random search and evolution. Specifically, this leaves us with the following set of seven operations: $3 \times 3$ and  $5 \times 5$ separable convolutions, $3 \times 3$ and  $5 \times 5$ dilated separable convolutions, $3 \times 3$ average pooling, $3 \times 3$ max pooling and identity. Randomly of architectures is done by sampling the structure of the cell and the operations at each edge.  The total number of architectures contained in this space is $\approx10^{18}$. We refer the reader to Liu \etal\citep{liu2018darts} for more details.

\vspace{-5pt}
\paragraph{NAS-Bench-201 search space. } NAS-Bench-201~\citep{dong20} is a tabular NAS benchmark, i.e. all architectures in the cell search space are trained and evaluated beforehand so one can query their performance (and weights) from a table quickly. Since this space is exhaustively evaluated, its size is also limited to only \emph{normal} cells containing 4 nodes in total (1 input, 2 intermediate and 1 output node) and 5 operation choices on every edge connecting two nodes. This means that there are only 15,625 possible architecture configurations in this space. The networks are constructed by stacking 5 cells with in-between fixed residual blocks for reducing the spacial resolution. Each of them is trained for 200 epochs 3 times with 3 different seeds on 3 image classification datasets. For more details, please refer to Dong \& Yang \citep{dong20}.

\vspace{-5pt}
\subsection{Datasets}\label{app:datasets}

\vspace{-5pt}
\paragraph{Fashion-MNIST~\citep{fmnist}.} Fashion-MNIST consists of a training set of 60k 28$\times$28 grayscale images and a test set of 10k images. The number of total labels is 10 classes. We split the 60k training set images to 50k used to train the networks and 10k used only for validation.

\vspace{-5pt}
\paragraph{CIFAR-10/100~\citep{cifar}.} CIFAR-10 and CIFAR-100 both consist of 60k 32$\times$32 colour images with 10 and 100 classes, respectively. We use 10k of the 60k training images as the validation set. We use the 10k original test set for final evaluation.

\vspace{-5pt}
\paragraph{Tiny ImageNet~\citep{tiny}.} Tiny Imagenet has 200 classes and each class has 500 training, 50 validation and 50 test colour images with 64$\times$64 resolution. Since the original test labels are not available, we split the 10k validation examples into 5k for testing and 5k for validation.

\vspace{-5pt}
\paragraph{\insixteen~\citep{dong20}} This variant of the \insixteen~\citep{chrabaszcz2017downsampled} contains 151.7k train, 3k validation and 3k test ImageNet images downsampled to 16$\times$16 and 120 classes.

Note that the test data points are only used for final evaluation. The data points for validation are used by the NES algorithms and DeepEns + ES baselines during ensemble selection and by DeepEns (RS) for picking the best architecture from the pool to use in the deep ensemble. Note that when considering dataset shift for CIFAR-10, CIFAR-100 and Tiny ImageNet, we also apply two disjoint sets of ``corruptions'' (following the terminology used by Hendrycks \& Dietterich~\citep{hendrycks2018benchmarking}) to the validation and test sets. We never apply any corruption to the training data. More specifically, out of the 19 different corruptions provided by Hendrycks \& Dietterich \citep{hendrycks2018benchmarking}, we randomly apply one from $\{ \texttt{Speckle Noise},$ $\texttt{Gaussian Blur},$ $\texttt{Spatter},$ $\texttt{Saturate} \}$ to each data point in the validation set and one from $\{ \texttt{Gaussian Noise},$ $\texttt{Shot Noise},$ $\texttt{Impulse Noise},$ $\texttt{Defocus Blur},$ $\texttt{Glass Blur},$ $\texttt{Motion Blur},$ $\texttt{Zoom Blur},$ $\texttt{Snow},$ $\texttt{Frost},$ $\texttt{Fog},$ $\texttt{Brightness},$ $\texttt{Contrast},$ $\texttt{Elastic Transform},$ $\texttt{Pixelate},$ $\texttt{JPEG compression} \}$ to each data point in the test set. This choice of validation and test corruptions follows the recommendation of Hendrycks \& Dietterich \citep{hendrycks2018benchmarking}. Also, as mentioned in Section \ref{sec:experiments}, each of these corruptions has 5 severity levels, which yields 5 corresponding severity levels for $\Dvals$ and $\Dtests$.

\vspace{-5pt}
\subsection{Training Routine \& Time}
\label{appsec:hypers}

\vspace{-5pt}
The macro-architecture we use has 16 initial channels and 8 cells (6 normal and 2 reduction) and was trained using a batch size of 100 for 100 epochs for CIFAR-10/100 and 15 epochs for Fashion-MNIST. For Tiny ImageNet, we used 36 initial channels and a batch size of 128 for 100 epochs. Training a single network took roughly 40 minutes and 3 GPU hours\footnote{We used NVIDIA RTX 2080Ti GPUs for training.} for CIFAR-10/100 and Tiny ImageNet, respectively. The networks are optimized using SGD with momentum set to 0.9. We used a learning rate of 0.025 for CIFAR-10/100 and Fashion-MNIST, and 0.1 for Tiny ImageNet. Unlike DARTS, we do not use any data augmentation procedure during training, nor any additional regularization such as ScheduledDropPath~\citep{zoph2018learning} or auxiliary heads, except for the case of Tiny ImageNet, for which we used ScheduledDropPath, gradient clipping and standard data augmentation as default in DARTS. All other hyperparameter settings are exactly as in DARTS~\citep{liu2018darts}.

All results containing error bars are averaged over multiple runs (at least five) with error bars indicating a 95\% confidence interval. We used a budget $K = 400$ for CIFAR-10/100 (corresponding to 267 GPU hours). For Tiny ImageNet on the DARTS search space, unless otherwise stated, we used $K = 200$ (corresponding to 600 GPU hours). Note that only in Figure \ref{fig:deepens_es_ablation_main_paper} and Table \ref{tbl:deepens_es_ablation} we also tried NES algorithms using a higher budget of $K = 400$ on Tiny ImageNet for an equal-cost comparison with DeepEns + ES (RS). For \insixteen on the NAS-Bench-201 search space, we used $K = 1000$ (compute costs are negligible since it is a tabular benchmark). Note that NES algorithms can be parallelized, especially NES-RS which is embarrassingly parallel like deep ensembles, therefore training time can be reduced easily when using multiple GPUs. 

\vspace{-5pt}
\subsection{Implementation Details of NES-RE}\label{sec:app-implementation-nes-re}

\vspace{-5pt}
\paragraph{Parallization.} Running NES-RE on a single GPU requires evaluating hundreds of networks sequentially, which is tedious.
To circumvent this, we distribute the ``while $|\pool| < K$'' loop in Algorithm \ref{alg:nes-re} over multiple GPUs, called worker nodes. We use the parallelism scheme provided by the $\texttt{hpbandster}$~\citep{falkner-icml18a} codebase.\footnote{\url{https://github.com/automl/HpBandSter}} In brief, the master node keeps track of the population and history (lines 1, 4-6, 8 in Algorithm \ref{alg:nes-re}), and it distributes the training of the networks to the individual worker nodes (lines 2, 7 in Algorithm \ref{alg:nes-re}). In our experiments, we always use 20 worker nodes and evolve a population $\popul$ of size $\popsize = 50$ when working over the DARTS search space. Over NAS-Bench-201, we used one worker since it is a tabular NAS benchmark and hence is quick to evaluate on. During iterations of evolution, we use an ensemble size of $\numcand = 10$ to select parent candidates. 

\vspace{-5pt}
\paragraph{Mutations.} We adapt the mutations used in RE to the DARTS search space. As in RE, we first pick a normal or reduction cell at random to mutate and then sample one of the following mutations:

\begin{itemize}
    \item $\texttt{identity}$: no mutation is applied to the cell.
    \item $\texttt{op mutation}$: sample one edge in the cell and replace its operation with another operation sampled from the list of operations described in Appendix \ref{app:arch_search_space_descriptions}.
    \item $\texttt{hidden state mutation}$: sample one intermediate node in the cell, then sample one of its two incoming edges. Replace the input node of that edge with another sampled node, without altering the edge's operation.
\end{itemize}

For example, one possible mutation would be changing the operation on an edge in the cell (as shown in e.g. Figure \ref{fig:ensemble_cells}) from say $5 \times 5$ separable convolution to $3 \times 3$ max pooling. See Real \etal\citep{real2019regularized} for further details and illustrations of these mutations. Note that for NAS-Bench-201, following Dong \& Yang \citep{dong20} we only use $\texttt{op mutation}$. 

\vspace{-5pt}
\paragraph{Adaptation of NES-RE to dataset shifts.} As described in Section \ref{sec:ens-adaptation-shift}, at each iteration of evolution, the validation set used in line 4 of Algorithm \ref{alg:nes-re} is sampled uniformly between $\Dval$ and $\Dvals$ when dealing with dataset shift. In this case, we use shift severity level 5 for $\Dvals$. Once the evolution is complete and the pool $\pool$ has been formed, then for each severity level $s \in \{0, 1, \dots, 5\}$, we apply $\esa$ with $\Dvals$ of severity $s$ to select an ensemble from $\pool$ (line 9 in Algorithm \ref{alg:nes-re}), which is then evaluated on $\Dtests$ of severity $s$.  (Here $s = 0$ corresponds to no shift.) This only applies to CIFAR-10, CIFAR-100 and Tiny ImageNet, as we do not consider dataset shift for Fashion-MNIST and \insixteen. 

\vspace{-5pt}
\subsection{Implementation details of anchored ensembles.}
\label{app:anchored_ensembles}

\vspace{-5pt}
Anchored ensembles \cite{pearce20} are constructed by independently training $M$ base learners, each regularized towards a new initialization sample (called the \textit{anchor point}). Specifically, the $k$-th base learner $f_{\theta_k}$ is trained to minimize the following loss (using the notation in Section \ref{sec:definitions}):
\begin{align*}
    \frac{1}{N}\sum_{i=1}^N \loss(f_{\theta_k}(\bx_i), y_i) + \frac{\lambda}{N} \lVert \Gamma^{1/2} (\theta_k - \accentset{\circ}{\theta}_k ) \rVert_2^2
\end{align*}
where $\loss(f_{\theta_k}(\bx_i), y_i) = - \log \left[ f_{\theta_k}(\bx_i) \right]_{y_i}$ is the cross-entropy loss for the $i$-th datapoint. Moreover, the anchor point $\accentset{\circ}{\theta}_k$ is an independently sampled initialization, and $\lambda$ is the regularization strength. Defining $p$ to be the number of parameters, i.e. $\theta \in \mathbb{R}^p$, and letting $\sigma_i^2$ be the variance of the initialization for the $i$-th parameter $(\theta_{k})_i$, $\Gamma \in \mathbb{R}^{p \times p}$ is the diagonal matrix with $\Gamma_{ii} = 1/(2\sigma_i^2)$.   

Although Pearce \etal\citep{pearce20} do not include a tune-able regularization parameter $\lambda$ (they set $\lambda = 1$), we found anchored ensembles performed poorly without tuning $\lambda$. For CIFAR-10/100, we used $\lambda = 0.4$ and for Tiny ImageNet, we used $\lambda = 0.1$ (tuned by grid search). We followed the recommendation of Pearce \etal\citep{pearce20} in using a different initialization for optimization than the anchor point $\accentset{\circ}{\theta}_k$. Moreover, we turned off weight decay when training anchored ensembles, and we did not apply the regularization to the trainable parameters in batch normalization layers (which are initialized deterministically).

\vspace{-2pt}
\section{Additional Experiments}
\label{app:add_exp}

\vspace{-5pt}
In this section we provide additional results for the experiments conducted in Section~\ref{sec:experiments}. Note that, as with all results shown in Section \ref{sec:experiments}, all evaluations are made on test data unless stated otherwise.

\vspace{-5pt}
\subsection{Supplementary plots complementing Section~\ref{sec:experiments}}
\label{appsubsec:add_exp_darts}

\vspace{-5pt}
\subsubsection{Results on Fashion-MNIST}
\label{appsubsec:fmnist_exp}

\begin{figure}
    \centering
    \includegraphics[width=1.0\linewidth]{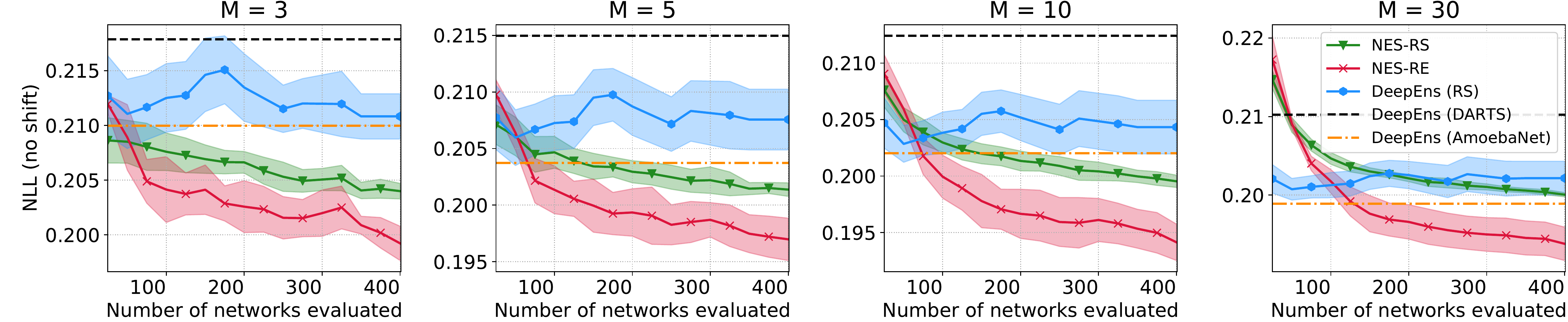}
    \caption{Results on Fashion-MNIST with varying ensembles sizes $M$. Lines show the mean NLL achieved by the ensembles with 95\% confidence intervals.}
    \vspace{-5pt}
    \label{fig:evals_test_loss_fmnist}
\end{figure}

\begin{figure}
    \centering
    \includegraphics[width=1.0\linewidth]{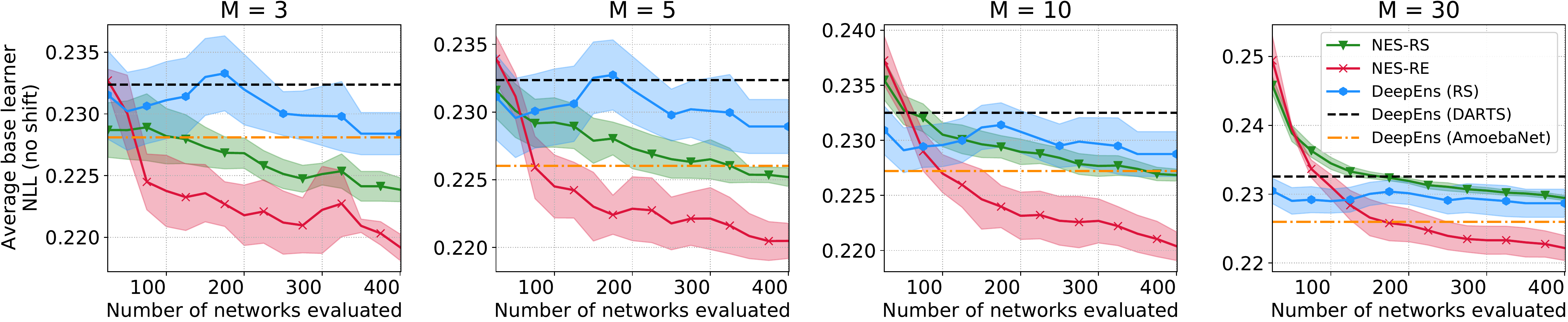}
    \caption{Average base learner loss for NES-RS, NES-RE and DeepEns (RS) on Fashion-MNIST. Lines show the mean NLL and $95\%$ confidence intervals.}
    \vspace{-5pt}
    \label{fig:evals_test_avg_bsl_fmnist}
\end{figure}

\begin{figure}[t!]
    \centering
    \includegraphics[width=1.0\linewidth]{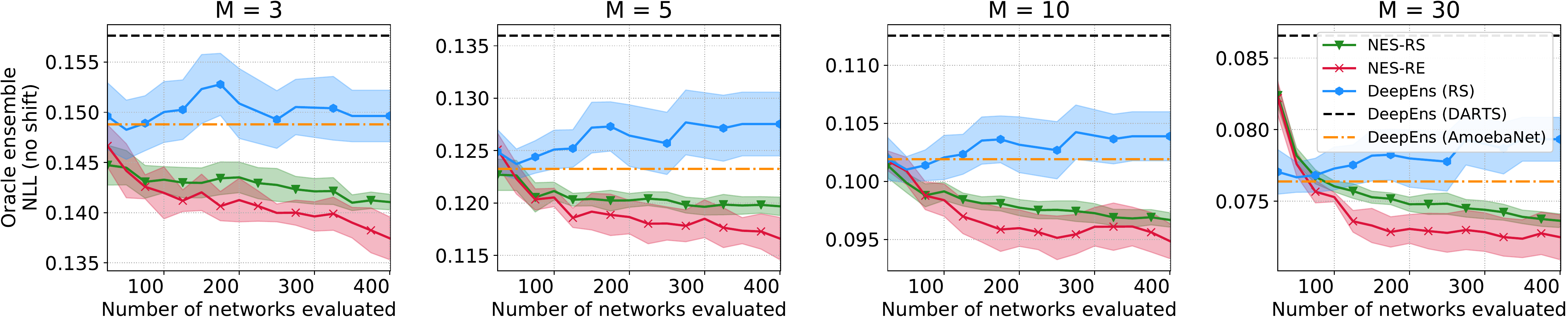}
    \caption{Oracle ensemble loss for NES-RS, NES-RE and DeepEns (RS) on Fashion-MNIST. Lines show the mean NLL and $95\%$ confidence intervals.}
    \vspace{-5pt}
    \label{fig:evals_test_oracle_fmnist}
\end{figure}

As shown in Figure \ref{fig:evals_test_loss_fmnist}, we see a similar trend on Fashion-MNIST as with other datasets: NES ensembles outperform deep ensembles with NES-RE outperforming NES-RS. To understand why NES algorithms outperform deep ensembles on Fashion-MNIST~\citep{fmnist}, we compare the average base learner loss (Figure~\ref{fig:evals_test_avg_bsl_fmnist}) and oracle ensemble loss (Figure~\ref{fig:evals_test_oracle_fmnist}) of NES-RS, NES-RE and DeepEns (RS). Notice that, apart from the case when ensemble size $M = 30$, NES-RS and NES-RE find ensembles with both stronger and more diverse base learners (smaller losses in Figures~\ref{fig:evals_test_avg_bsl_fmnist} and~\ref{fig:evals_test_oracle_fmnist}, respectively). While it is expected that the oracle ensemble loss is smaller for NES-RS and NES-RE compared to DeepEns (RS), it initially appears surprising that DeepEns (RS) has a larger average base learner loss considering that the architecture for the deep ensemble is chosen to minimize the base learner loss. We found that this is due to the loss having a sensitive dependence not only on the architecture but also the initialization of the base learner networks. Therefore, re-training the best architecture by validation loss to build the deep ensemble yields base learners with higher losses due to the use of different random initializations. Fortunately, NES algorithms are not affected by this, since they simply select the ensemble's base learners from the pool without having to re-train anything which allows them to exploit good architectures as well as initializations. Note that, for CIFAR-10-C experiments, this was not the case; base learner losses did not have as sensitive a dependence on the initialization as they did on the architecture.


In Table~\ref{tbl:fmnist_loss_ece_more_sevs}, we compare the classification error and expected calibration error (ECE) of NES algorithms with the deep ensembles baseline for various ensemble sizes on Fashion-MNIST. Similar to the loss, NES algorithms also achieve smaller errors, while ECE remains approximately the same for all methods.

\begin{table}
  \caption{Error and ECE of ensembles on Fashion-MNIST for different ensemble sizes $M$. Best values and all values within $95\%$ confidence interval are bold faced.}\vspace{5pt}
 \label{tbl:fmnist_loss_ece_more_sevs}
 \centering
 \resizebox{\textwidth}{!}{%
 \begin{tabular}{ccccccccccc@{}}
 \toprule
 
   &
   \multicolumn{5}{c}{\textbf{Classification Error} (out of 1)} &
   \multicolumn{5}{c}{\textbf{Expected Calibration Error (ECE)}} \\ \cmidrule(lr){2-6} \cmidrule(lr){7-11}
   \multirow{-2}{*}{$M$} &
   NES-RS &
   NES-RE &
   \begin{tabular}[c]{@{}c@{}}DeepEns\\ (RS)\end{tabular} &
     \begin{tabular}[c]{@{}c@{}}DeepEns\\ (DARTS)\end{tabular} &
   \begin{tabular}[c]{@{}c@{}}DeepEns\\ (AmoebaNet)\end{tabular} &
   NES-RS &
   NES-RE &
   \begin{tabular}[c]{@{}c@{}}DeepEns\\ (RS)\end{tabular}  &
     \begin{tabular}[c]{@{}c@{}}DeepEns\\ (DARTS)\end{tabular} &
   \begin{tabular}[c]{@{}c@{}}DeepEns\\ (AmoebaNet)\end{tabular} \\ \midrule

       3 &  $0.074$\tiny{$\pm 0.001$} &  $\mathbf{0.072}$\tiny{$\pm 0.001$} &  $0.076$\tiny{$\pm 0.001$} &  $0.077$ &  $0.077$ &  
       $0.007$\tiny{$\pm 0.001$} &  $0.007$\tiny{$\pm 0.002$} &  $0.008$\tiny{$\pm 0.001$} &  $\mathbf{0.003}$ &  $0.008$ \\

     5 &   $\mathbf{0.073}$\tiny{$\pm 0.001$} &  $\mathbf{0.071}$\tiny{$\pm 0.002$} &  $0.075$\tiny{$\pm 0.001$} &  $0.077$ &  $0.074$ &                    $\mathbf{0.005}$\tiny{$\pm 0.001$}  &  $\mathbf{0.005}$\tiny{$\pm 0.001$} &                    $\mathbf{0.006}$\tiny{$\pm 0.001$} &  $\mathbf{0.005}$ &  $\mathbf{0.005}$ \\

      10 &  $0.073$\tiny{$\pm 0.001$} &   $\mathbf{0.070}$\tiny{$\pm 0.001$} &  $0.075$\tiny{$\pm 0.001$} &  $0.076$ &  $0.073$ & 
      $\mathbf{0.004}$\tiny{$\pm 0.001$} &  $\mathbf{0.005}$\tiny{$\pm 0.001$} &                    $\mathbf{0.005}$\tiny{$\pm 0.001$}  &  $0.006$ &  $\mathbf{0.005}$ \\

     30 &     $0.073$\tiny{$\pm 0.001$} &   $\mathbf{0.070}$\tiny{$\pm 0.001$} &  $0.074$\tiny{$\pm 0.001$} &  $0.075$ &  $0.073$ &
     $\mathbf{0.004}$\tiny{$\pm 0.001$} &  $\mathbf{0.004}$\tiny{$\pm 0.002$} &  $\mathbf{0.004}$\tiny{$\pm 0.001$} &  $0.008$ &  $\mathbf{0.004}$ 

   \\ \bottomrule
 \end{tabular}%
 }
\end{table}

\vspace{-5pt}
\subsubsection{Entropy on out-of-distribution inputs}

\begin{figure}[t]
    \centering
    \includegraphics[width=.93\linewidth]{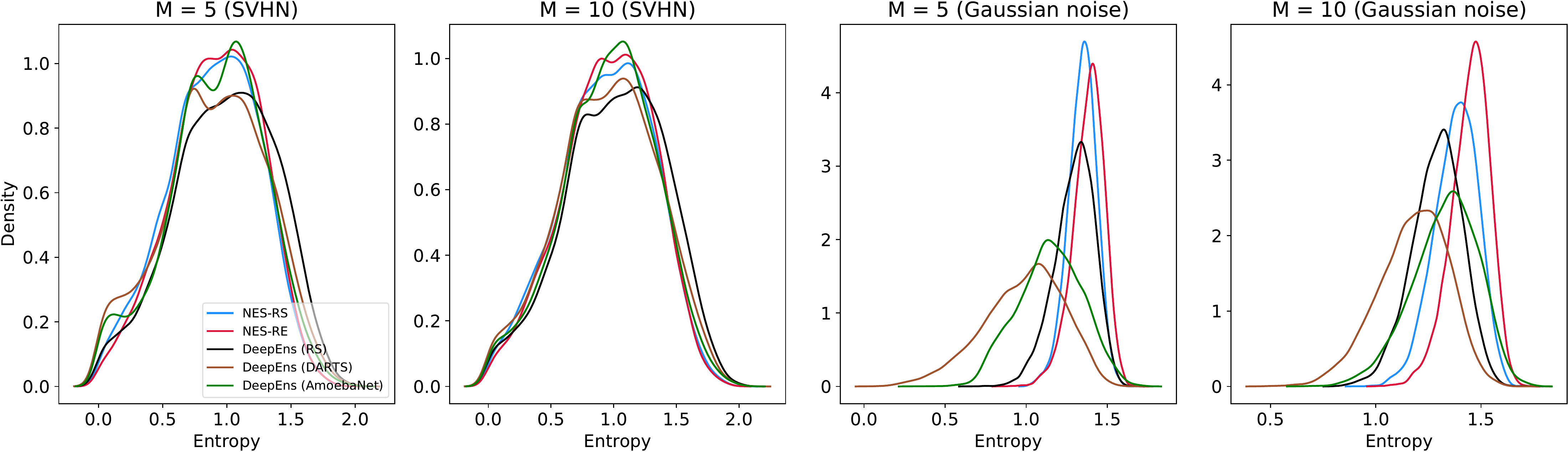}
    \caption{Entropy of predicted probabilities when trained on CIFAR-10 in the DARTS search space.}
    \label{fig:ood-c10}
\end{figure}

\vspace{-5pt}
To assess how well models respond to completely out-of-distribution (OOD) inputs (inputs which do not belong to one of the classes the model can predict), we investigate the entropy of the predicted probability distribution over the classes when the input is OOD. Higher entropy of the predicted probabilities indicates more uncertainty in the model's output. For CIFAR-10 on the DARTS search space, we compare the entropy of the predictions made by NES ensembles with deep ensembles on two types of OOD inputs: images from the SVHN dataset and Gaussian noise. In Figure \ref{fig:ood-c10}, we notice that NES ensembles indicate higher uncertainty when given inputs of Gaussian noise than deep ensembles but behave similarly to deep ensembles for inputs from SVHN.

\vspace{-5pt}
\subsubsection{Additional Results on the DARTS search space}
\label{appsubsec:cifarc_exp}

\begin{figure}[ht]
    \centering
    \captionsetup[subfigure]{justification=centering}
    \begin{subfigure}[t]{0.99\textwidth}
        \centering
        \includegraphics[width=.49\linewidth]{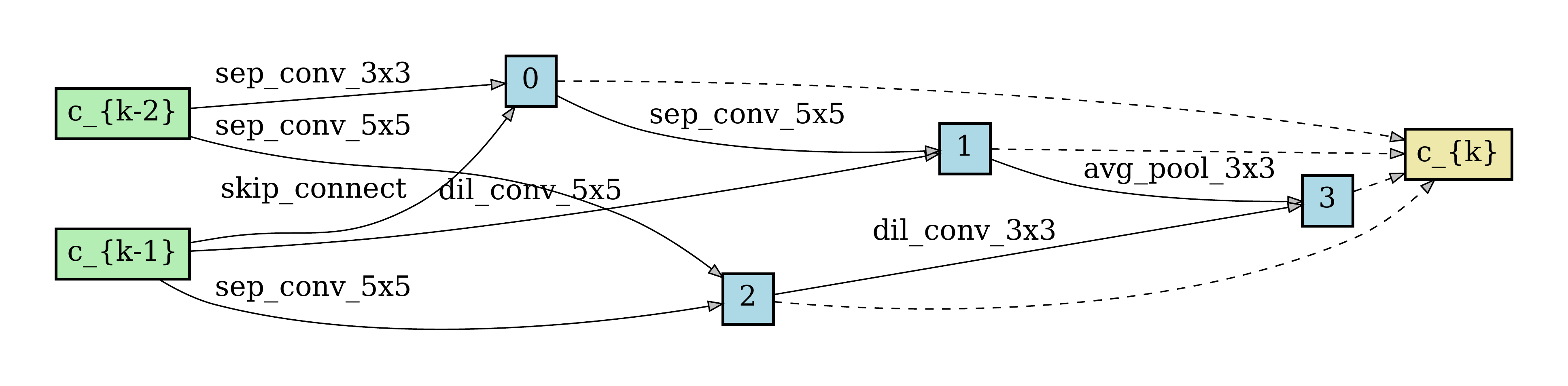}
        \includegraphics[width=.49\linewidth]{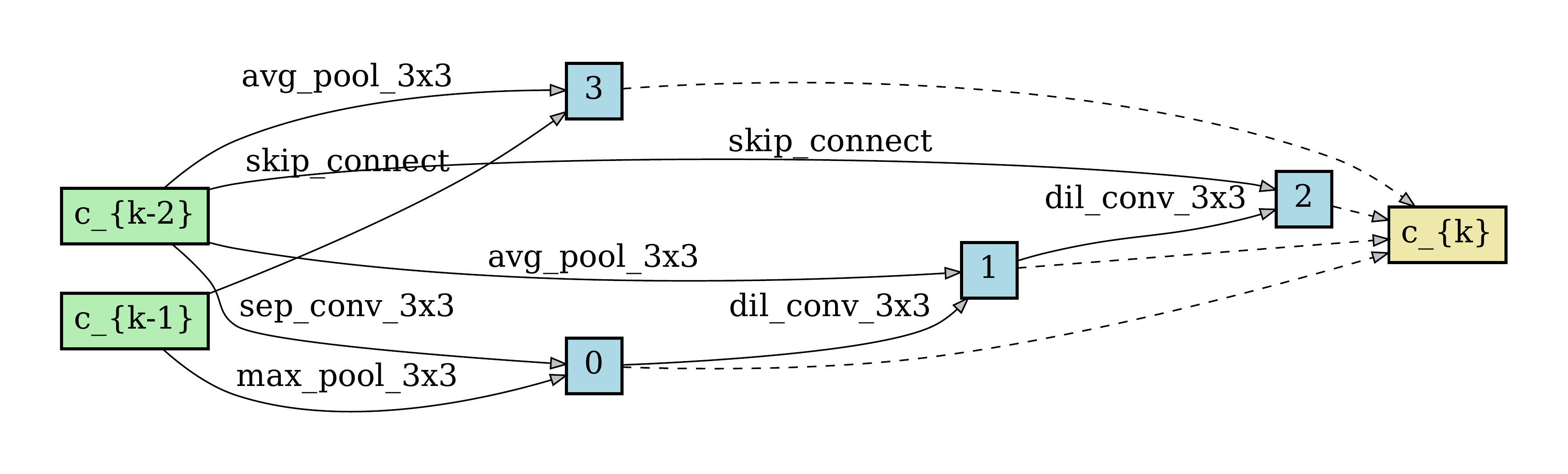}
        \subcaption{First base learner}
    \end{subfigure}\\
    \begin{subfigure}[t]{0.99\textwidth}
        \centering
        \includegraphics[width=.49\linewidth]{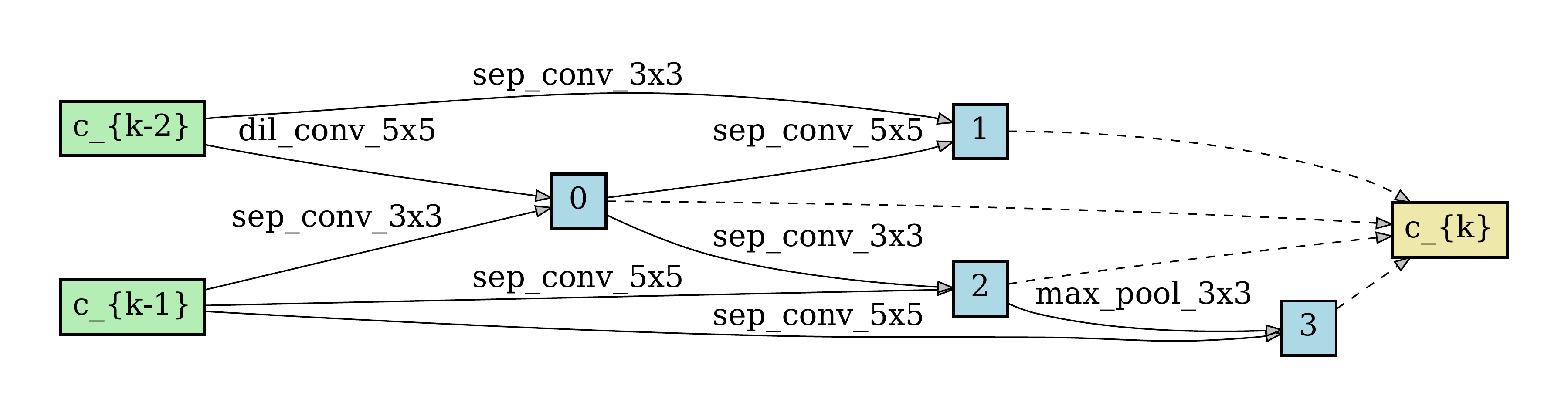}
        \includegraphics[width=.49\linewidth]{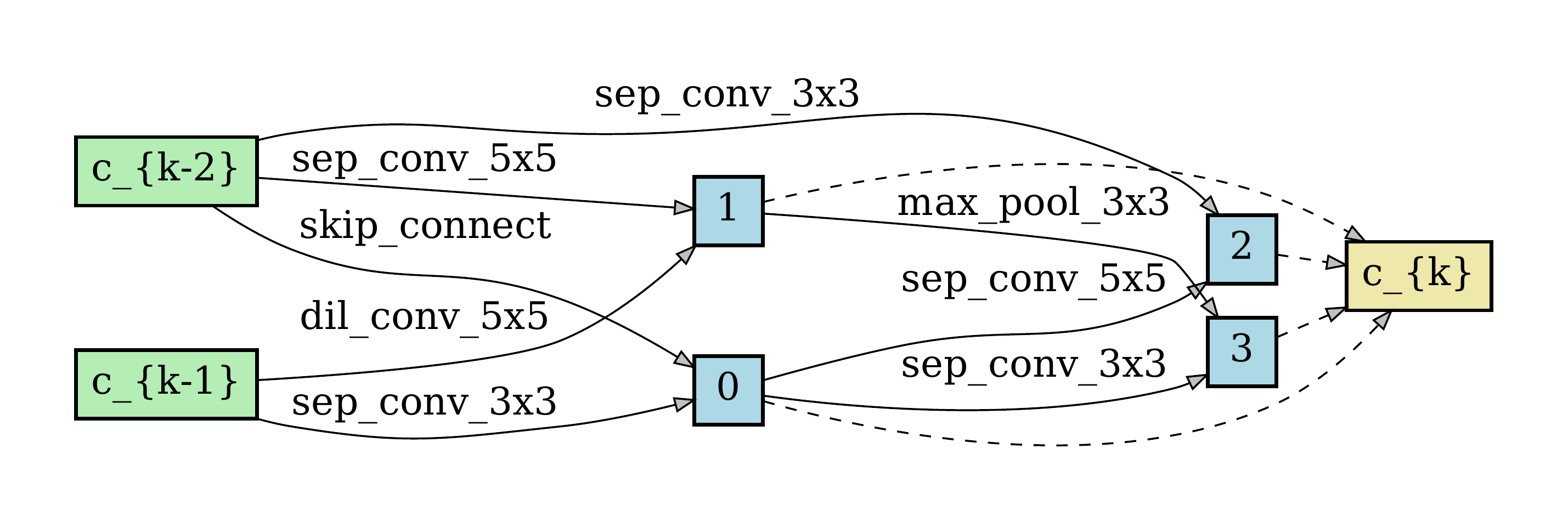}
        \subcaption{Second base learner}
    \end{subfigure}\\
    \begin{subfigure}[t]{0.99\textwidth}
        \centering
        \includegraphics[width=.49\linewidth]{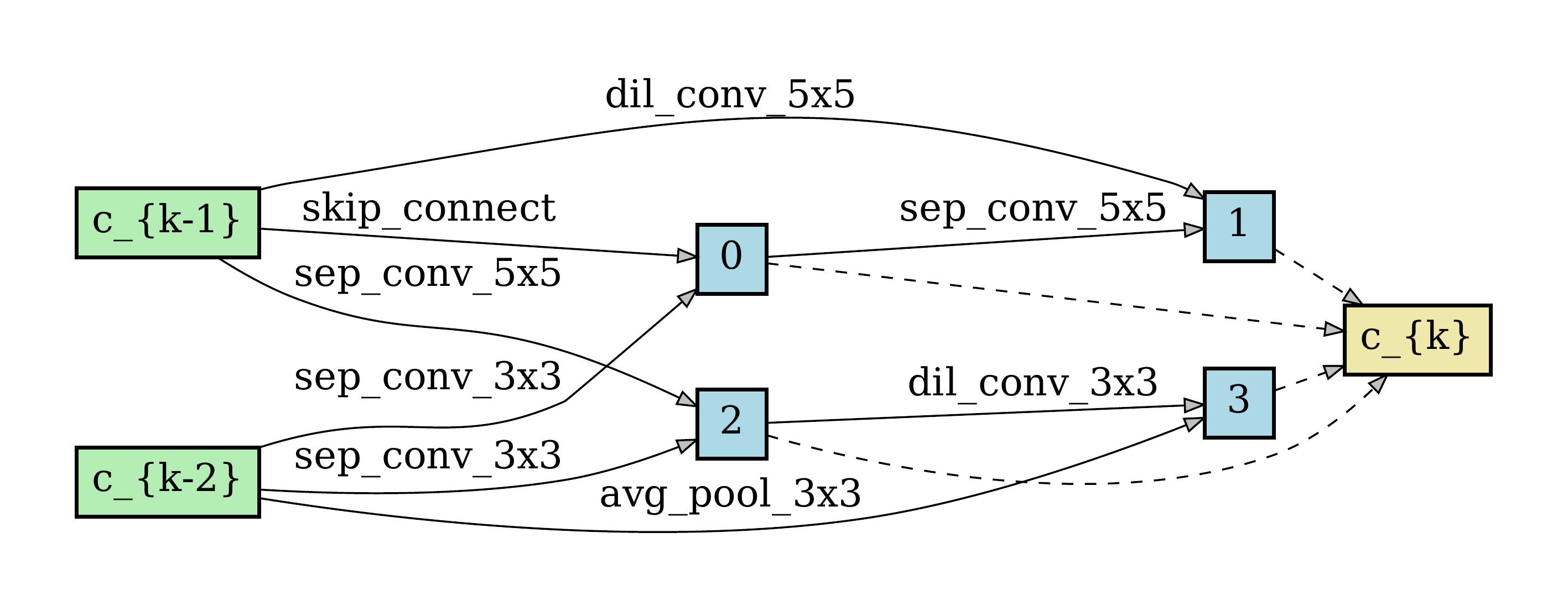}
        \includegraphics[width=.49\linewidth]{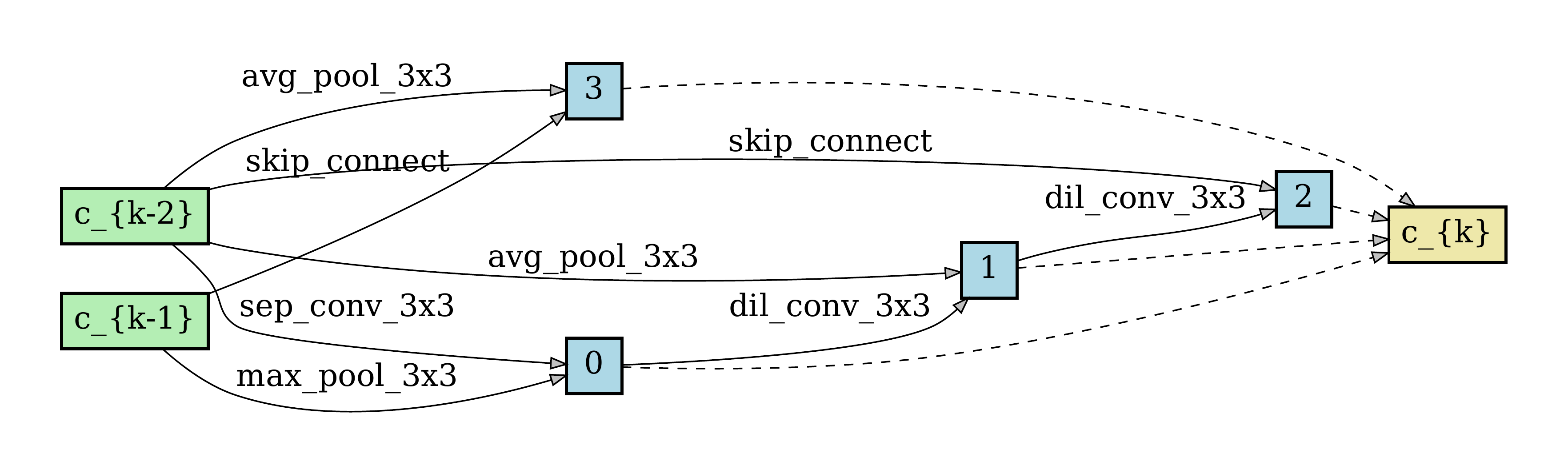}
        \subcaption{Third base learner}
    \end{subfigure}
    
    \caption{Illustration of example cells found by NES-RE for an ensemble of size $M=3$ on Tiny ImageNet over the DARTS search space. In each sub-figure, left is the normal cell and right is the reduction cell. Note that the first and third base learners have the same reduction cell in this instance.} 
    \label{fig:ensemble_cells}
\end{figure}

\vspace{-5pt}
In this section, we provide additional experimental results on CIFAR-10, CIFAR-100 and Tiny ImageNet on the DARTS search space, complementing the results in Section \ref{sec:experiments} as shown in Figures~\ref{fig:test_loss_M_other}-\ref{fig:test_error_budget_other}. We also include examples of architectures chosen by NES-RE in Figure~\ref{fig:ensemble_cells} for Tiny ImageNet, showcasing variation in architectures. 

\paragraph{Results on CIFAR for larger models.}
In additional to the results on CIFAR-10 and CIFAR-100 on the DARTS search space using the settings described in Appendix~\ref{appsec:hypers}, we also train larger models (around 3M parameters) by scaling up the number of stacks cells and initial channels in the network. We run NES and other baselines similarly as done before and plot results in Figure~\ref{fig:test_loss_full_fid} and \ref{fig:test_error_full_fid} for NLL and classification test error with budget $\budget = 90$. As shown, NES algorithms tend to outperform or be competitive with the baselines. Note, more runs are needed including error bars for conclusive results in this case.

\clearpage
\input{text/appendix_plots}
\clearpage

\begin{figure}
    \centering
    \includegraphics[width=.95\linewidth]{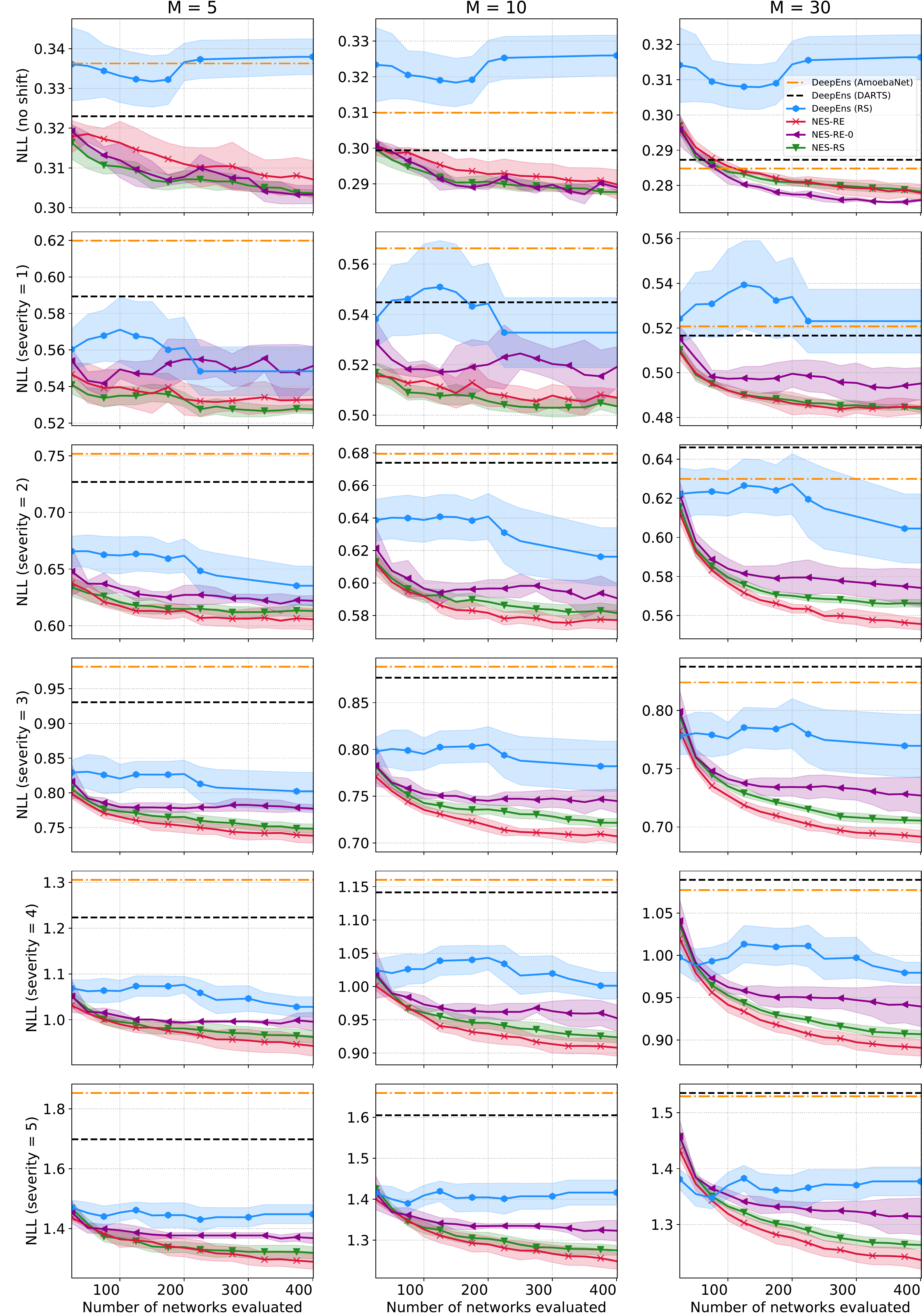}
    \caption{Results on CIFAR-10~\citep{hendrycks2018benchmarking} with varying ensembles sizes $M$ and shift severity. Lines show the mean NLL achieved by the ensembles with 95\% confidence intervals. See Appendix \ref{appsubsec:cifarc_exp} for the definition of NES-RE-0.}%
    \label{fig:evals_test_loss_full}
\end{figure}

\clearpage
\newpage

\vspace{-5pt}
\subsubsection{Additional Results on the NAS-Bench-201 search space}
\label{appsubsec:nb201_exp}

In this section, we provide additional experimental results on CIFAR-10, CIFAR-100 and ImageNet-16-120 on the NAS-Bench-201 search space, complementing the results in Section \ref{sec:experiments} as shown in Figures~\ref{fig:nb201_c10_budget}-\ref{fig:nb201_error}.

\vspace{-5pt}
\subsection{Ablation study: NES-RE optimizing only on $\Dval$}\label{app:nes_re_0}

We also include a variant of NES-RE, called NES-RE-0, in Figure \ref{fig:evals_test_loss_full}. NES-RE and NES-RE-0 are the same, except that NES-RE-0 uses the validation set $\Dval$ without any shift during iterations of evolution, as in line 4 of Algorithm \ref{alg:nes-re}. Following the discussion in Appendix \ref{sec:app-implementation-nes-re}, recall that this is unlike NES-RE, where we sample the validation set to be either $\Dval$ or $\Dvals$ at each iteration of evolution. Therefore, NES-RE-0 evolves the population without taking into account dataset shift, with $\Dvals$ only being used for the post-hoc ensemble selection step in line 9 of Algorithm \ref{alg:nes-re}.

As shown in the Figure \ref{fig:evals_test_loss_full}, NES-RE-0 shows a minor improvement over NES-RE in terms of loss for ensemble size $M = 30$ in the absence of dataset shift. This is in line with expectations, because  evolution in NES-RE-0 focuses on finding base learners which form strong ensembles for in-distribution data. On the other hand, when there is dataset shift, the performance of NES-RE-0 ensembles degrades, yielding higher loss and error than both NES-RS and NES-RE. Nonetheless, NES-RE-0 still manages to outperform the DeepEns baselines consistently. We draw two conclusions on the basis of these results: (1) NES-RE-0 can be a competitive option in the absence of dataset shift. (2) Sampling the validation set, as done in NES-RE, to be $\Dval$ or $\Dvals$ in line 4 of Algorithm \ref{alg:nes-re} plays an important role is returning a final pool $\pool$ of base learners from which $\esa$ can select ensembles robust to dataset shift.

\vspace{-5pt}
\subsection{What if deep ensembles use ensemble selection over initializations?}\label{app:deepens_esa_ablation}

\begin{figure} 
    \begin{minipage}{.33\linewidth}
        \includegraphics[width=\linewidth]{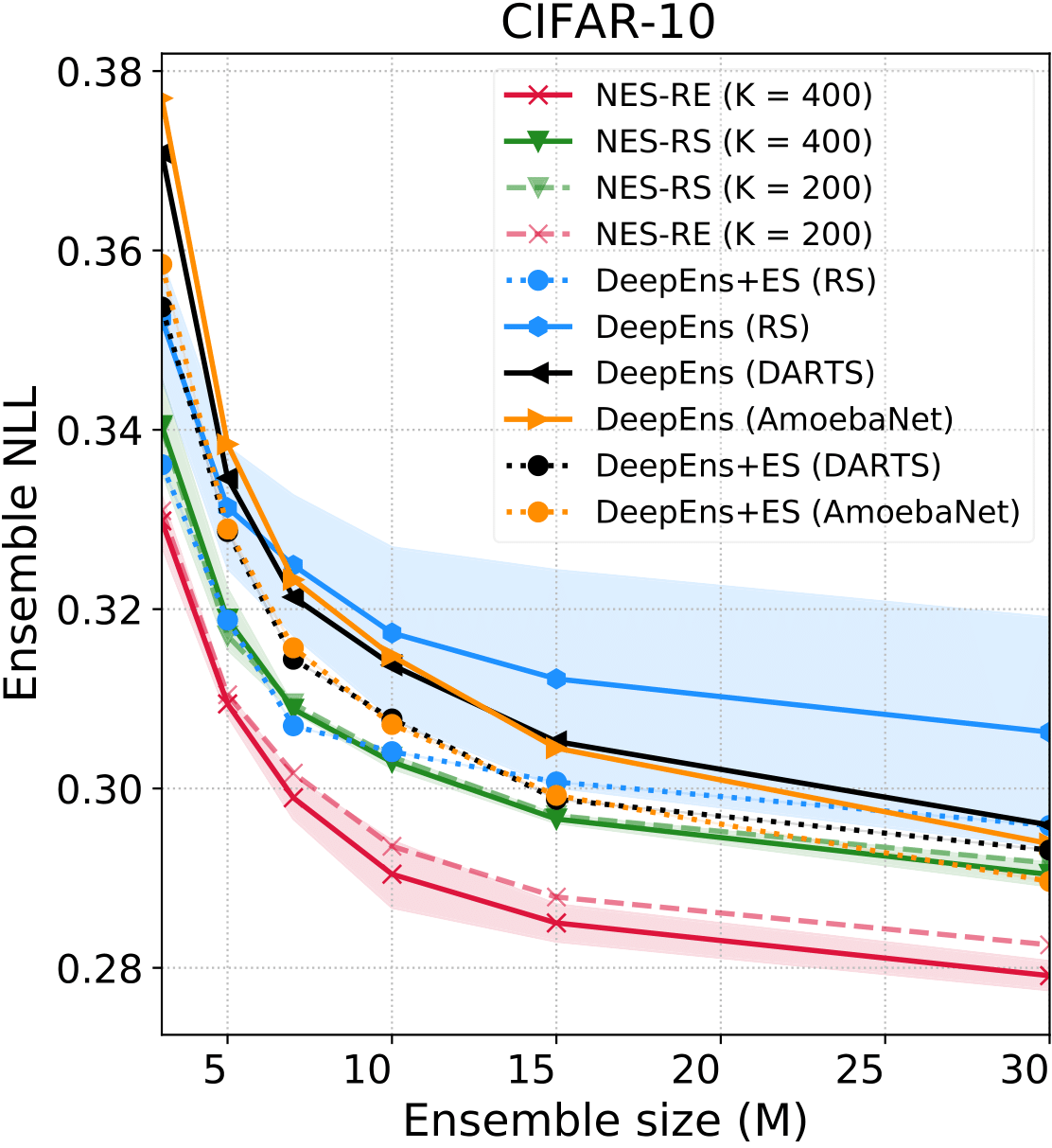}
    \end{minipage}\hfill
    \begin{minipage}{.66\linewidth}
        \resizebox{\linewidth}{!}{%
        \setlength{\tabcolsep}{1.8pt}
            \begin{tabular}{@{}lcccc@{}}    
            & Tiny ImageNet \\ 
            \toprule
            \multicolumn{1}{c}{\multirow{2}{*}{\textbf{Method}}} &
            \multicolumn{2}{c}{No dataset shift} &
            \multicolumn{2}{c}{Dataset shift: severity 5} \\ \cmidrule(l){2-5} 
            \multicolumn{1}{c}{} & \textbf{NLL} & \textbf{Error (\%)} & \textbf{NLL} & \textbf{Error (\%)} \\ \midrule
            DeepEns (DARTS) &
            $1.59$ &
            $39.08$ &
             $3.75$ &
              $71.58$ \\
            DeepEns + ES (DARTS) &
            $1.57$ &
            $38.68$ &
            $3.68$ &
  $70.90$ \\
DeepEns (AmoebaNet) &
  $1.55$ &
  $38.46$ &
  $3.76$ &
  $71.72$ \\
DeepEns + ES (AmoebaNet) &
  $1.54$ &
  $38.12$ &
  $3.70$ &
  $71.68$ \\
DeepEns (RS) &
  $1.51$\tiny{$\pm 0.00 $} &
  $37.46$\tiny{$\pm 0.27 $} &
  $3.60$\tiny{$\pm 0.03 $} &
  $70.48$\tiny{$\pm 0.38 $} \\
DeepEns + ES (RS) &
  $1.50$ &
  $36.98$ &
  $3.55$ &
  $\mathbf{ 70.10 }$ \\
NES-RS ($K=200$) &
  $1.51$\tiny{$\pm 0.01 $} &
  $37.42$\tiny{$\pm 0.21 $} &
  $3.53$\tiny{$\pm 0.01 $} &
  $\mathbf{ 69.93 }$\tiny{$\pm 0.23 $} \\
NES-RS ($K=400$) & $1.48$\tiny{$\pm 0.01 $}            & $37.45$\tiny{$\pm 0.25 $}            & $\mathbf{ 3.51 }$\tiny{$\pm 0.01 $} & $\mathbf{ 70.15 }$\tiny{$\pm 0.34 $} \\
NES-RE ($K=200$) &
  $1.48$\tiny{$\pm 0.01 $} &
  $36.98$\tiny{$\pm 0.57 $} &
  $3.55$\tiny{$\pm 0.02 $} &
  $70.22$\tiny{$\pm 0.13 $} \\
NES-RE ($K=400$) & $\mathbf{ 1.44 }$\tiny{$\pm 0.02 $} & $\mathbf{ 36.55 }$\tiny{$\pm 0.17 $} & $3.54$\tiny{$\pm 0.02 $}            & $\mathbf{ 69.98 }$\tiny{$\pm 0.24 $} \\ \bottomrule
        \end{tabular}
        } \vspace{2mm}
    \end{minipage}
    \caption{Comparison of NES to deep ensembles with ensemble selection over initializations over the DARTS search space. \textsc{Left}: NLL vs ensemble size on CIFAR-10 (similar to ablation in Figure \ref{fig:deepens_es_ablation_main_paper}). \textsc{Right}: Test NLL and classification error NES vs. deep ensembles on Tiny ImageNet with ensemble size $M = 10$.} 
    \label{tbl:deepens_es_ablation}
\end{figure}

We provide additional experimental results in this section for comparing NES to deep ensembles with ensemble selection building on those shown in Figure \ref{fig:deepens_es_ablation_main_paper} in Section \ref{sec:experiments}. The table in Figure \ref{tbl:deepens_es_ablation}-right compares the methods with respect to classification error as well as loss at different dataset shift severities. In the absence of any dataset shift, we find that NES-RE outperforms all baselines with respect to both metrics, specially at the higher budget of $K = 400$ and NES-RS performs competitively. At shift severity 5, NES algorithms also outperform the baselines, with NES-RS performing slightly better than NES-RE. The plot in Figure \ref{tbl:deepens_es_ablation}-left shows similar results on CIFAR-10 over the DARTS search space for the same experiment as the one conducted on Tiny ImageNet and shown in Figure \ref{fig:deepens_es_ablation_main_paper}.

\vspace{-5pt}
\subsection{Comparing NES to ensembles with other varying hyperparameters}\label{app:hyperdeepens}

Since varying the architecture in an ensemble improves predictive performance and uncertainty estimation as demonstrated in Section \ref{sec:experiments}, it is natural to ask what other hyperparameters should be varied in an ensemble. It is also unclear which hyperparameters might be more important than others. Note that concurrent work by Wenzel \etal\citep{wenzel2020hyperparameter} has shown that varying hyperparameters such as $L_2$ regularization strength, dropout rate and label smoothing parameter also improves upon deep ensembles. While these questions lie outside the scope of our work and are left for future work, we conduct preliminary experiments to address them.

\begin{figure*}
    \centering
    \captionsetup[subfigure]{justification=centering}
    \begin{subfigure}[t]{0.32\textwidth}
        \centering
        \includegraphics[width=.49\linewidth]{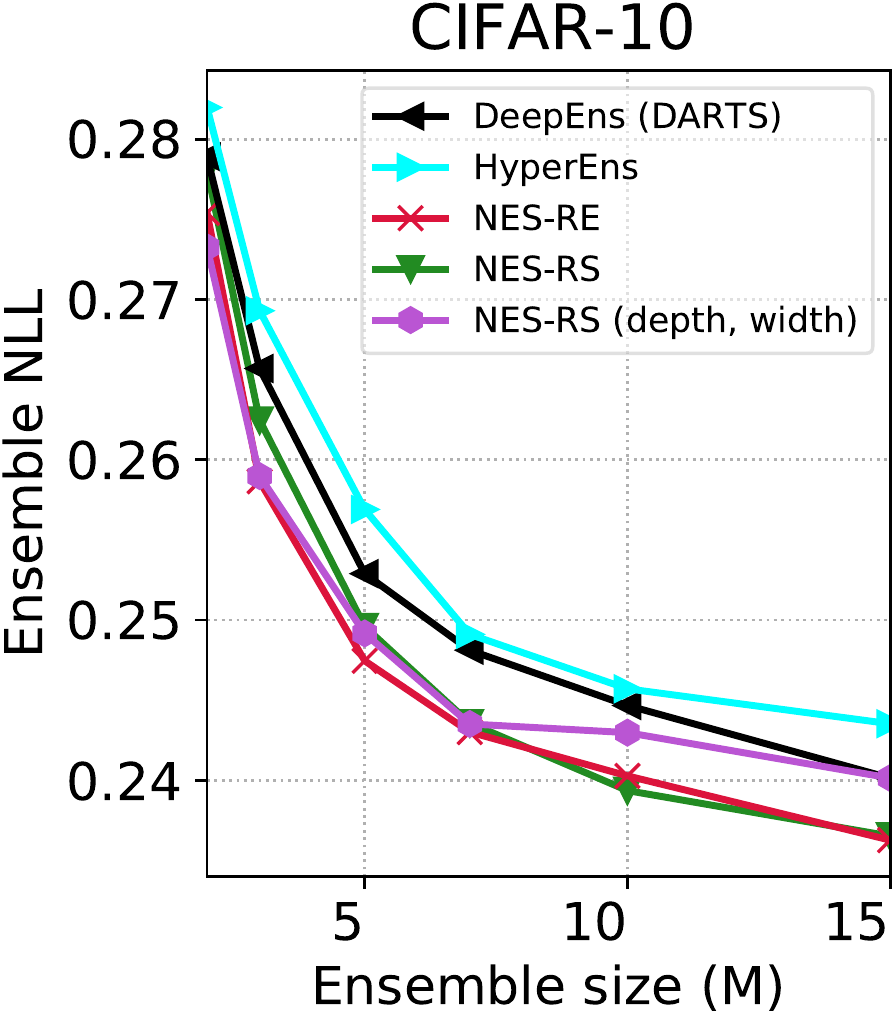}
        \includegraphics[width=.49\linewidth]{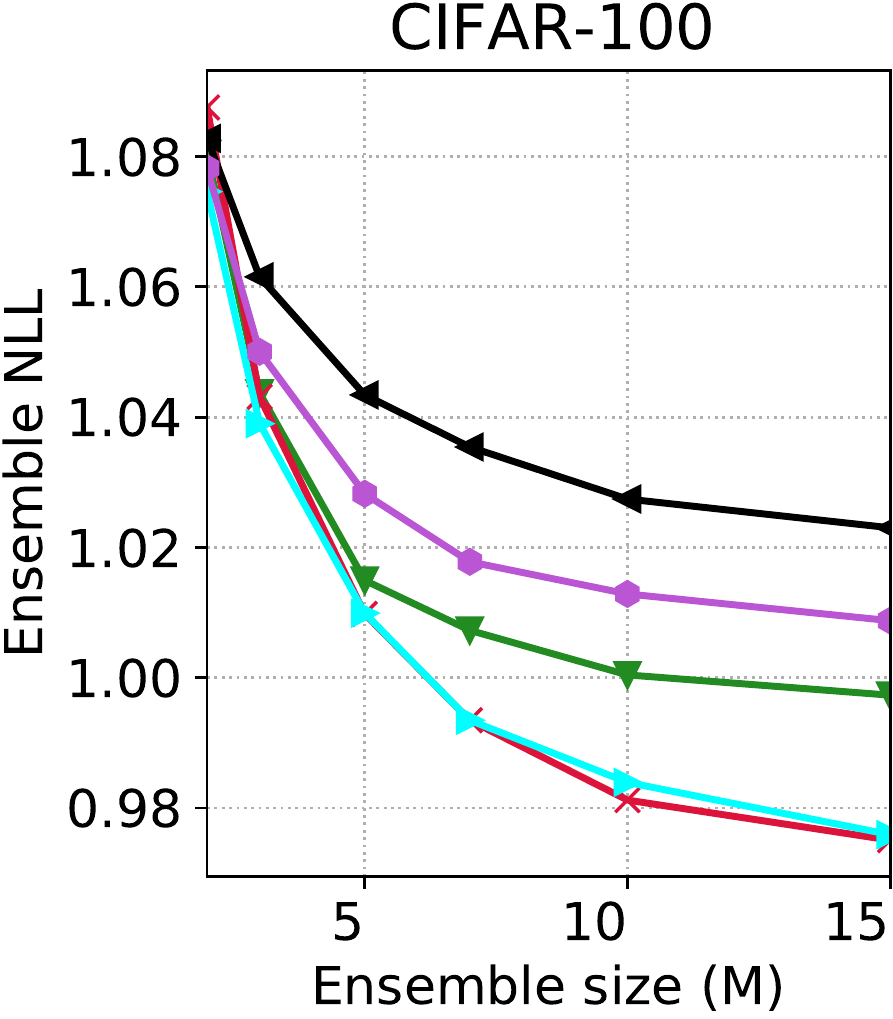}
        \subcaption{No data shift}
        \label{fig:test_loss_hyper_0}
    \end{subfigure}
    ~\hspace{.03cm}
    \begin{subfigure}[t]{0.32\textwidth}
        \centering
        \includegraphics[width=.49\linewidth]{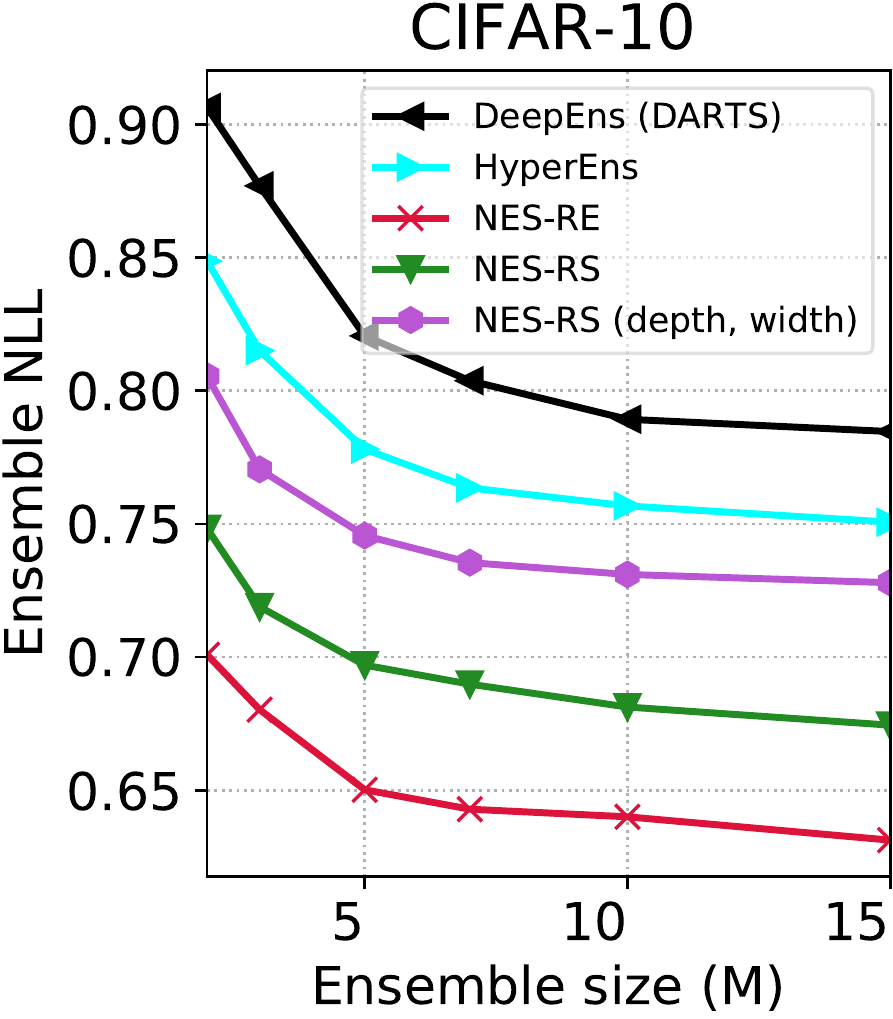}
        \includegraphics[width=.49\linewidth]{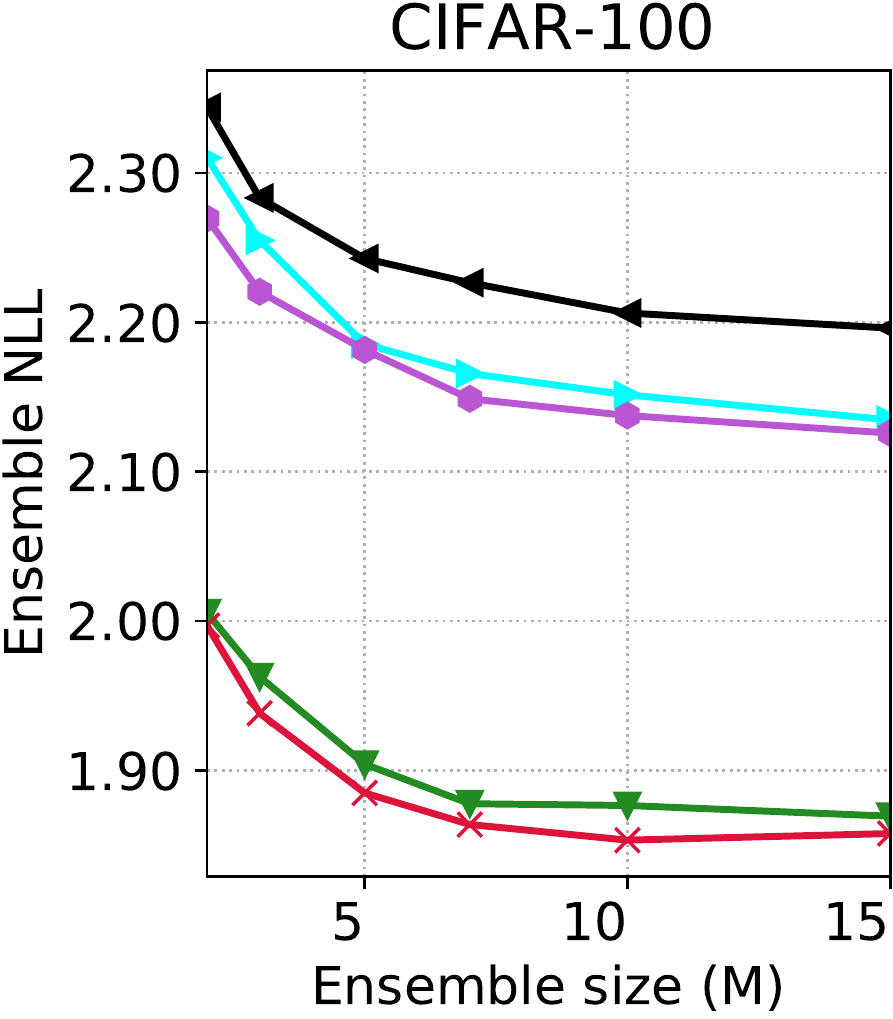}
        \subcaption{Data shift (severity 3)}
        \label{fig:test_loss_hyper_3}
    \end{subfigure}
    ~\hspace{.03cm}
    \begin{subfigure}[t]{0.32\textwidth}
        \centering
        \includegraphics[width=.49\linewidth]{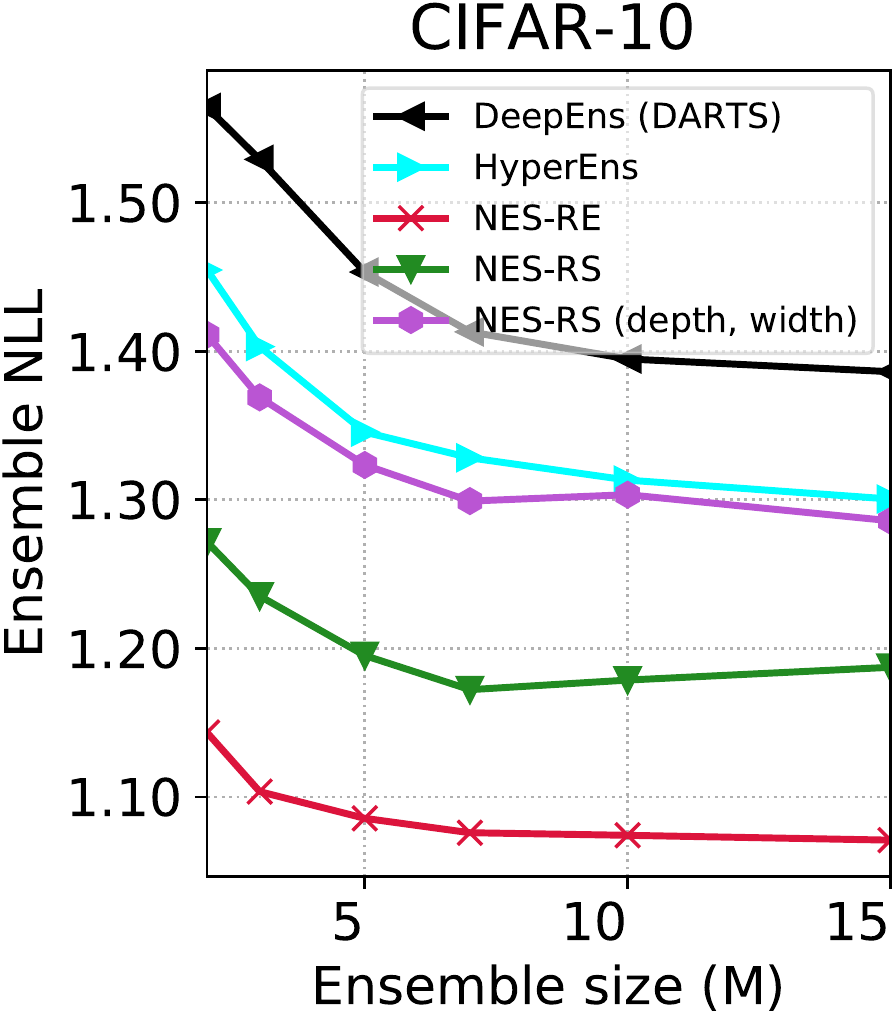}
        \includegraphics[width=.49\linewidth]{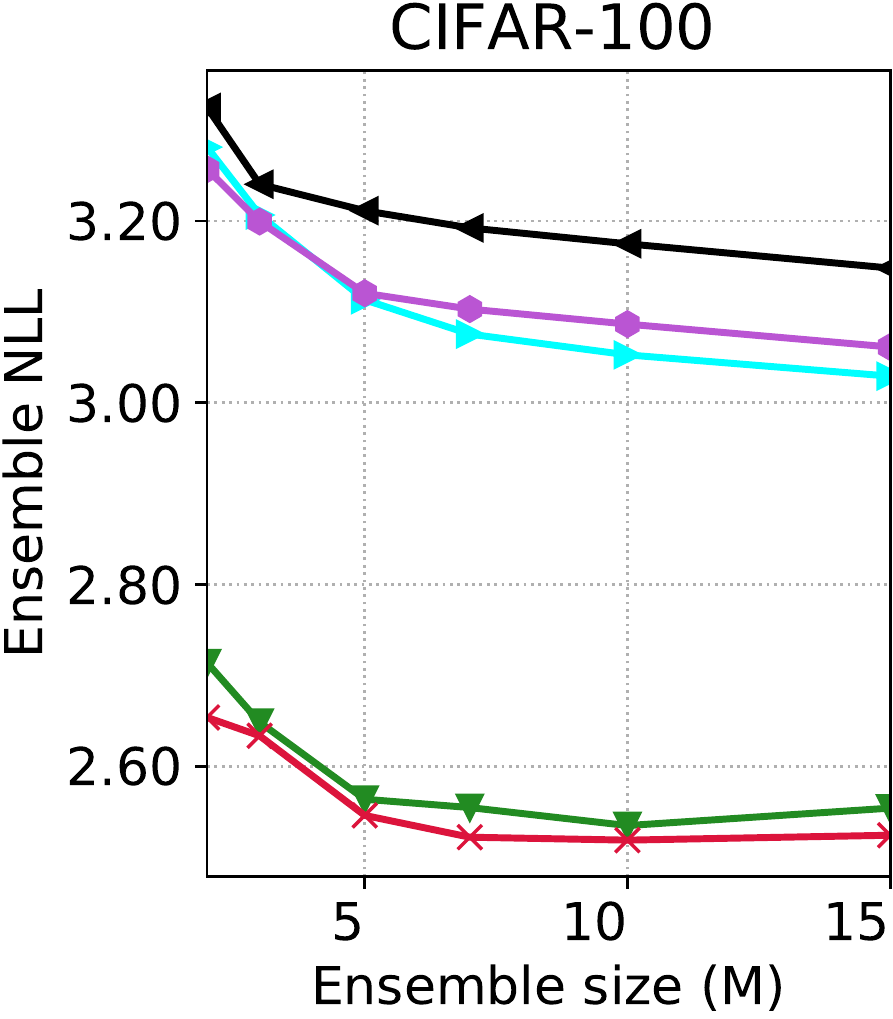}
        \subcaption{Data shift (severity 5)}
        \label{fig:test_loss_hyper_5}
    \end{subfigure}
    
    \caption{Plots show NLL vs. ensemble sizes comparing NES to the baselines introduced in Appendix \ref{app:hyperdeepens} on CIFAR-10 and CIFAR-100 with and without respective dataset shifts~\citep{hendrycks2018benchmarking}.}
    \label{fig:test_loss_hyper}
\end{figure*}

\begin{wrapfigure}[33]{R}{.43\textwidth} 
\vspace{-18pt}
    \centering
    \begin{subfigure}[t]{\linewidth}
        \centering
        \includegraphics[width=.49\linewidth]{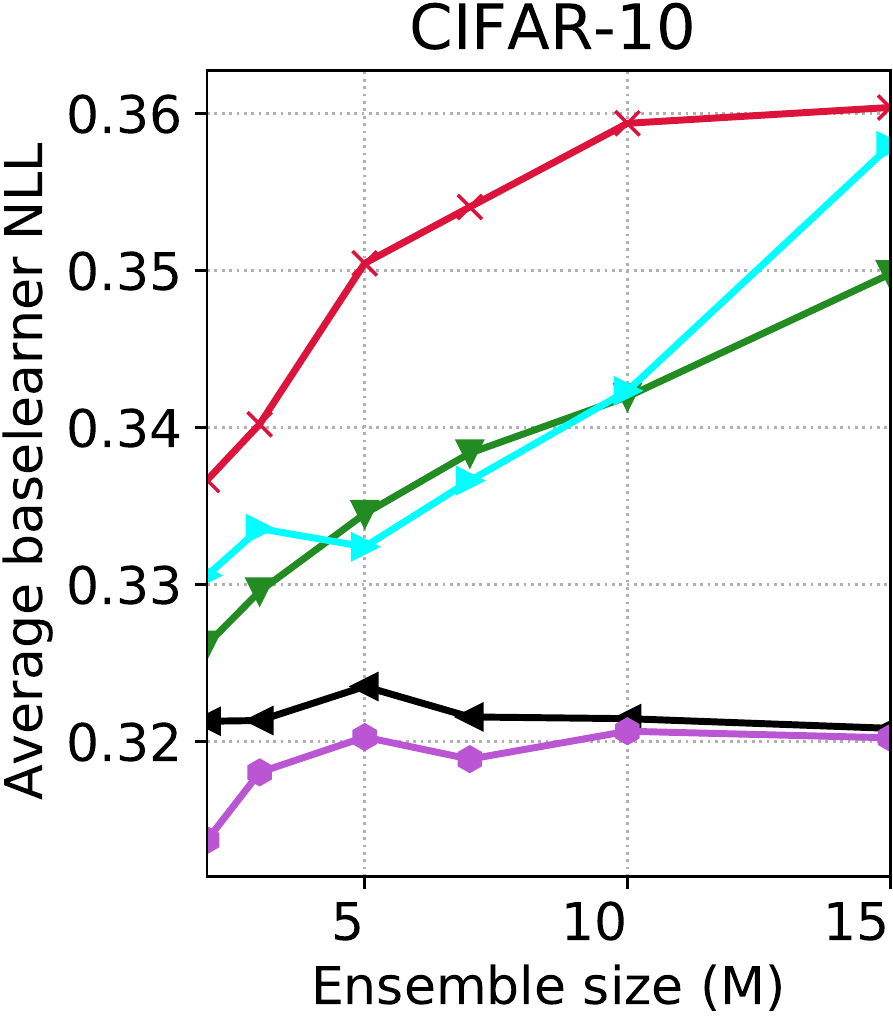}
        \includegraphics[width=.49\linewidth]{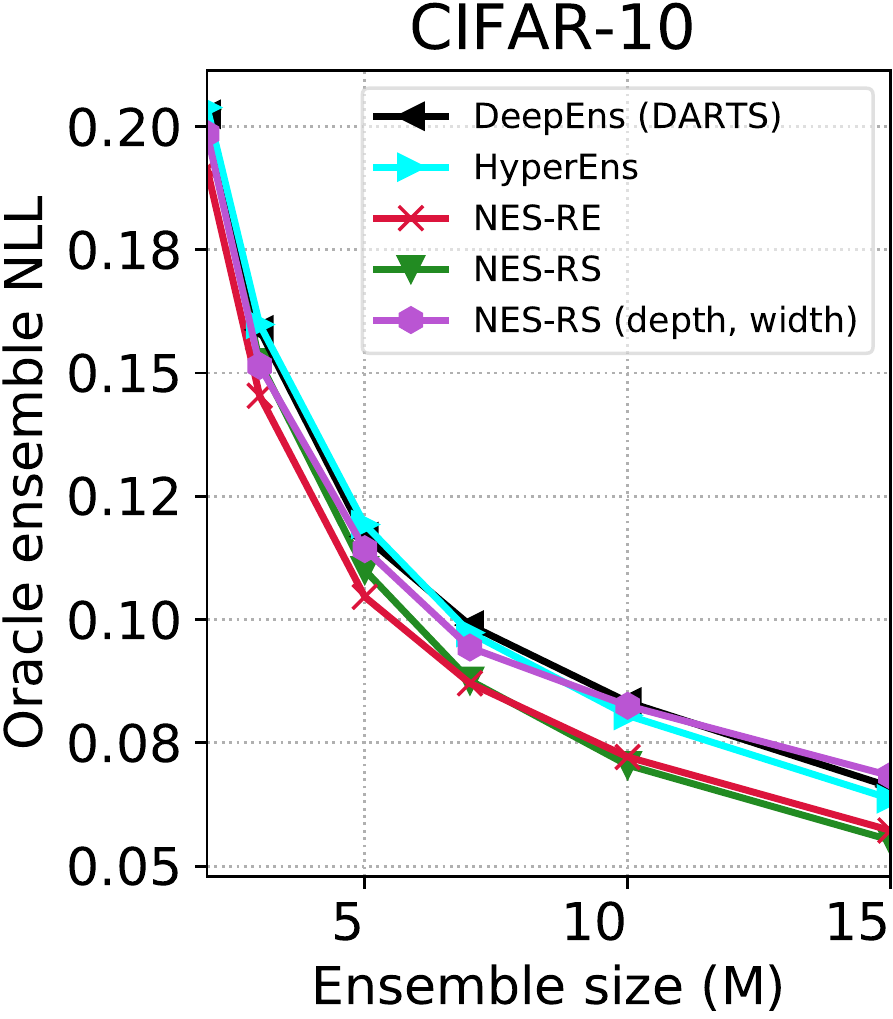}\\
        \includegraphics[width=.49\linewidth]{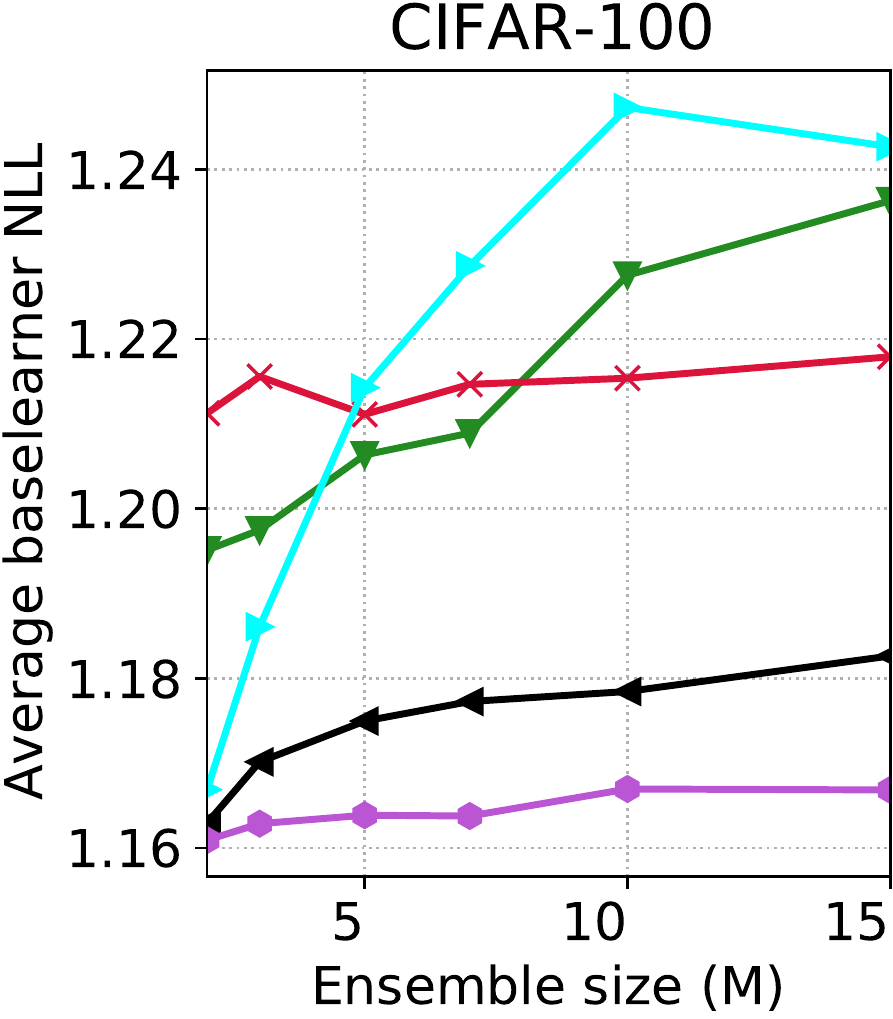}
        \includegraphics[width=.49\linewidth]{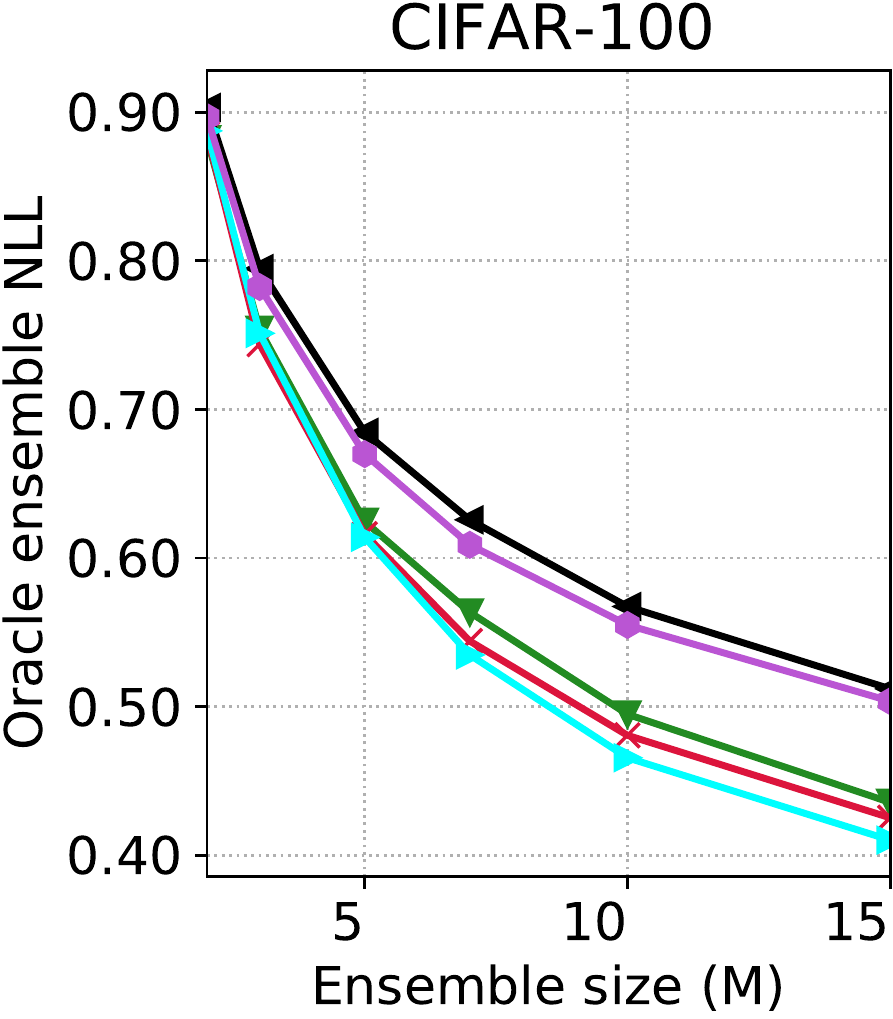}
        \subcaption{Average base learner loss and oracle ensemble loss for NES and the baselines introduced in Appendix \ref{app:hyperdeepens} on CIFAR-10 and CIFAR-100. Recall that small oracle ensemble loss generally corresponds to higher diversity.}
        \vspace{3pt}
    \label{fig:avg_oracle_M_hyper}
    \end{subfigure}
    \begin{subfigure}[t]{\linewidth}
    \centering
\resizebox{0.99\linewidth}{!}{%
\setlength{\tabcolsep}{1.8pt}
\begin{tabular}{@{}ccccccc@{}}
\toprule
\multirow{2}{*}{\textbf{Dataset}} &
  \multirow{2}{*}{\textbf{\begin{tabular}[c]{@{}c@{}}Shift\\ Severity\end{tabular}}} &
  \multicolumn{5}{c}{\textbf{DARTS search space}} \\ \cmidrule(l){3-7} 
 &
  &
  \begin{tabular}[c]{@{}c@{}}DeepEns\\ (DARTS)\end{tabular} &
  HyperEns &
  \begin{tabular}[c]{@{}c@{}}NES-RS\\ (depth, width)\end{tabular} &
  NES-RS &
  NES-RE \\ \midrule
\multirow{3}{*}{C10}  & 0 & $8.2$  & $8.1$           & $8.0$  & $8.0$  & $\mathbf{7.7}$  \\
                      & 3 & $25.9$ & $25.0$          & $24.1$ & $22.5$ & $\mathbf{21.5}$ \\
                      & 5 & $43.3$ & $40.8$          & $42.1$ & $38.1$ & $\mathbf{34.9}$ \\ \midrule
\multirow{3}{*}{C100} & 0 & $28.8$ & $\mathbf{28.1}$ & $28.8$ & $28.4$ & $28.4$          \\
                      & 3 & $54.0$ & $52.7$          & $53.1$ & $48.9$ & $\mathbf{48.5}$ \\
                      & 5 & $68.4$ & $67.2$          & $67.9$ & $61.3$ & $\mathbf{60.7}$ \\ \bottomrule
\end{tabular}
}
        \subcaption{Classification errors comparing NES to the baselines introduced in Appendix \ref{app:hyperdeepens} for different shift severities and $M = 10$. Best values are bold faced.}
    \label{tbl:error_hyper}
    \end{subfigure}
    \caption{See Appendix \ref{app:hyperdeepens} for details.}
\end{wrapfigure}

In this section, we consider two additional baselines working over the DARTS search space on CIFAR-10/100:
\begin{enumerate}[leftmargin=*]
    \item \textbf{HyperEns:} Optimize a fixed architecture, train $K$ random initializations of it \textit{where the learning rate and $L_2$ regularization strength are also sampled randomly} and select the final ensemble of size $M$ from the pool using $\esa$. This is similar to \texttt{hyper ens} from Wenzel \etal\citep{wenzel2020hyperparameter}. 
    \item \textbf{NES-RS (depth, width):} As described in Appendix \ref{app:arch_search_space_descriptions}, NES navigates a complex (non-Euclidean) search space of architectures by varying the cell, which involves changing both the DAG structure of the cell and the operations at each edge of the DAG. We consider a baseline in which we keep the cell fixed (the optimized DARTS cell) and only vary the width and depth of the overall architecture. More specifically, we vary the number of \textit{initial channels} $\in \{12, 14, 16, 18, 20 \}$ (width) and the number of \textit{layers} $\in \{ 5, 8, 11\}$ (depth). We apply NES-RS over this substantially simpler search space of architectures as usual: train $K$ randomly sampled architectures (i.e. sampling only depth and width) to form a pool and select the ensemble from it.
\end{enumerate}

The results shown in Figures \ref{fig:test_loss_hyper} and Table \ref{tbl:error_hyper} compare the two baselines above to DeepEns (DARTS), NES-RS and NES-RE.\footnote{Note that runs of DeepEns (DARTS), NES-RE and NES-RS differ slightly in this section relative to Section \ref{sec:experiments}, as we tune the learning rate and $L_2$ regularization strength for each dataset instead of using the defaults used in Liu \etal\citep{liu2018darts}. This yields a fair comparison: HyperEns varies the learning rate and $L_2$ regularization while using a fixed, optimized architecture (DARTS), whereas NES varies the architecture while using fixed, optimized learning rate and $L_2$ regularization strength.} As shown in Figure \ref{fig:test_loss_hyper}, NES-RE tends to outperform the baselines, though is at par with HyperEns on CIFAR-100 without dataset shift (Figure \ref{fig:test_loss_hyper_0}). Under the presence of dataset shift (Figures \ref{fig:test_loss_hyper_3} and \ref{fig:test_loss_hyper_5}), both NES algorithms substantially outperform all baselines. Note that both HyperEns and NES-RS (depth, width) follow the same protocol as NES-RS and NES-RE: ensemble selection uses a shifted validation dataset when evaluating on a shifted test dataset. In terms of classification error, the observations are similar as shown in Table \ref{tbl:error_hyper}. Lastly, we view the diversity of the ensembles from the perspective of oracle ensemble loss in Figure \ref{fig:avg_oracle_M_hyper}. As in Section \ref{sec:experiments}, results here also suggest that NES agorithms tend to find more diverse ensembles despite having higher average base learner loss.

\subsection{Sensitivity to the size of the validation dataset $\Dval$ and overfitting during ensemble selection}\label{app:val-data-sensitivity-overfitting}

\begin{wrapfigure}[30]{R}{.43\textwidth} 
\vspace{-17pt}
    \centering
    \begin{subfigure}[t]{\linewidth}
        \centering
        \includegraphics[width=.49\linewidth]{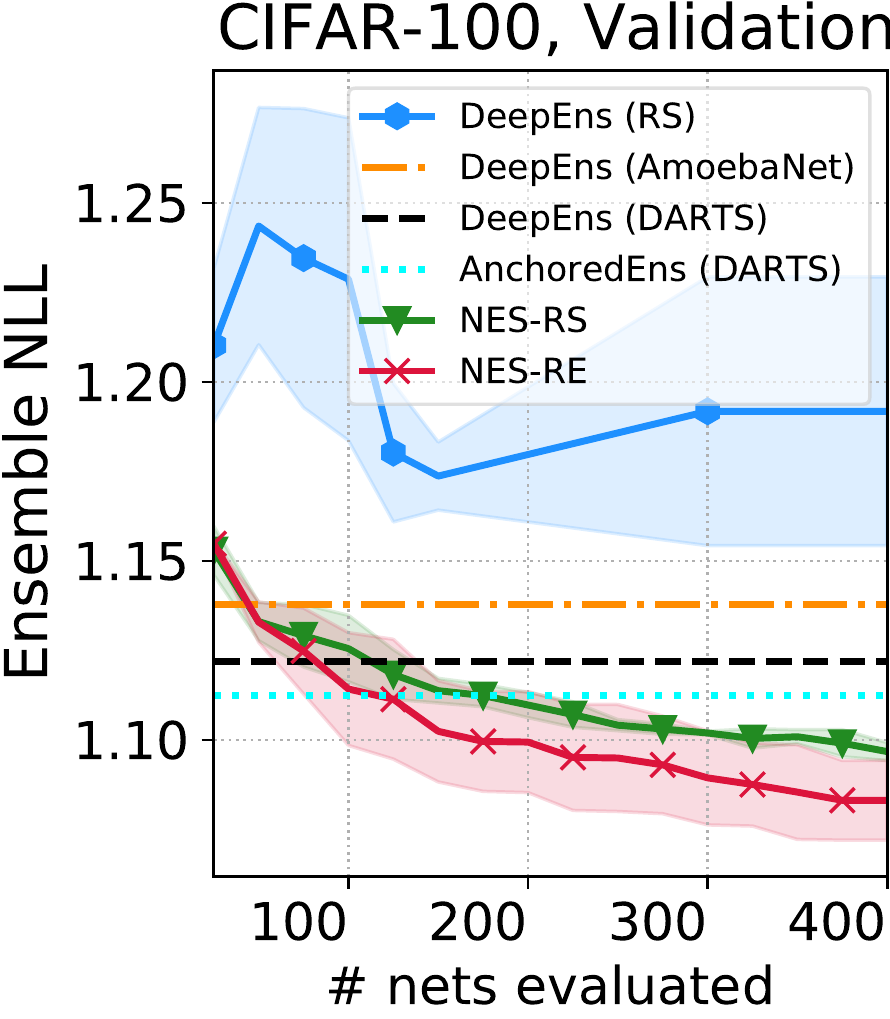}
        \includegraphics[width=.49\linewidth]{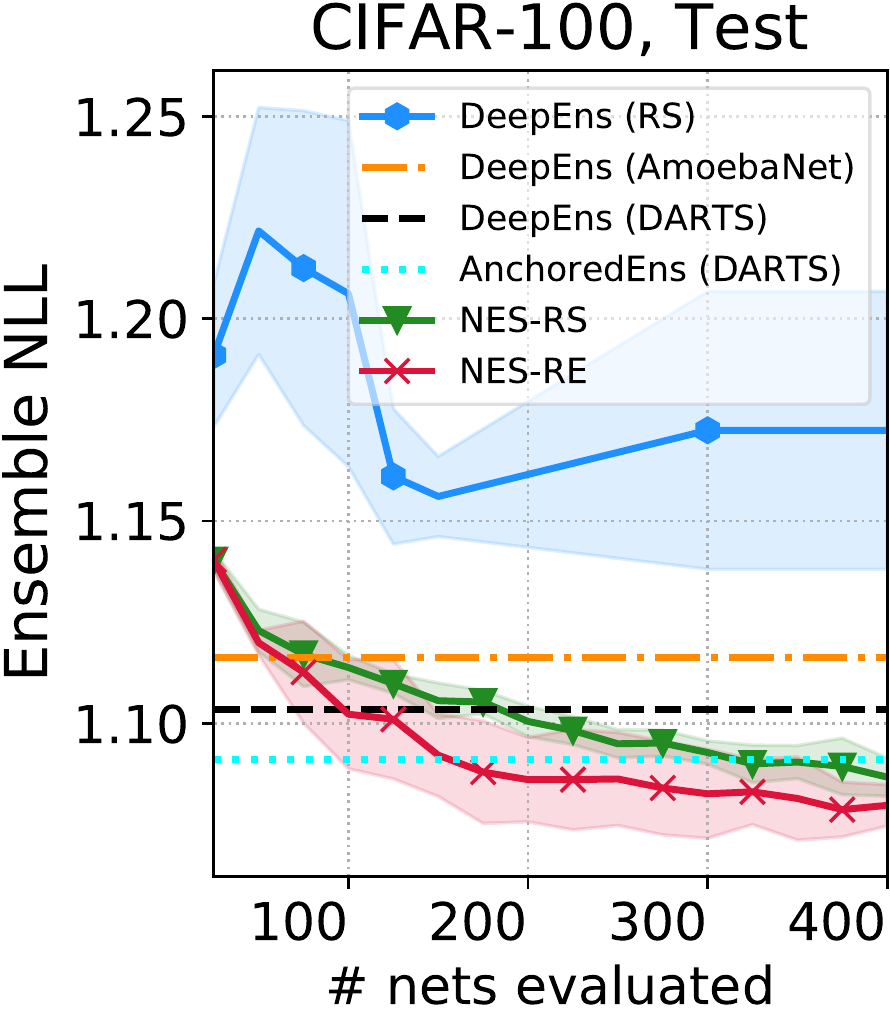}
        \subcaption{Ensemble loss over the validation and test datasets over time (i.e. with increasing budget $\budget$). These curves being similar indicates no overfitting to the validation dataset during ensemble selection.}
        \vspace{3pt}
        \label{fig:val-size-overfitting}
    \end{subfigure}
    ~\hspace{.05cm}
    \begin{subfigure}[t]{\linewidth}
        \centering
        \includegraphics[width=.49\linewidth]{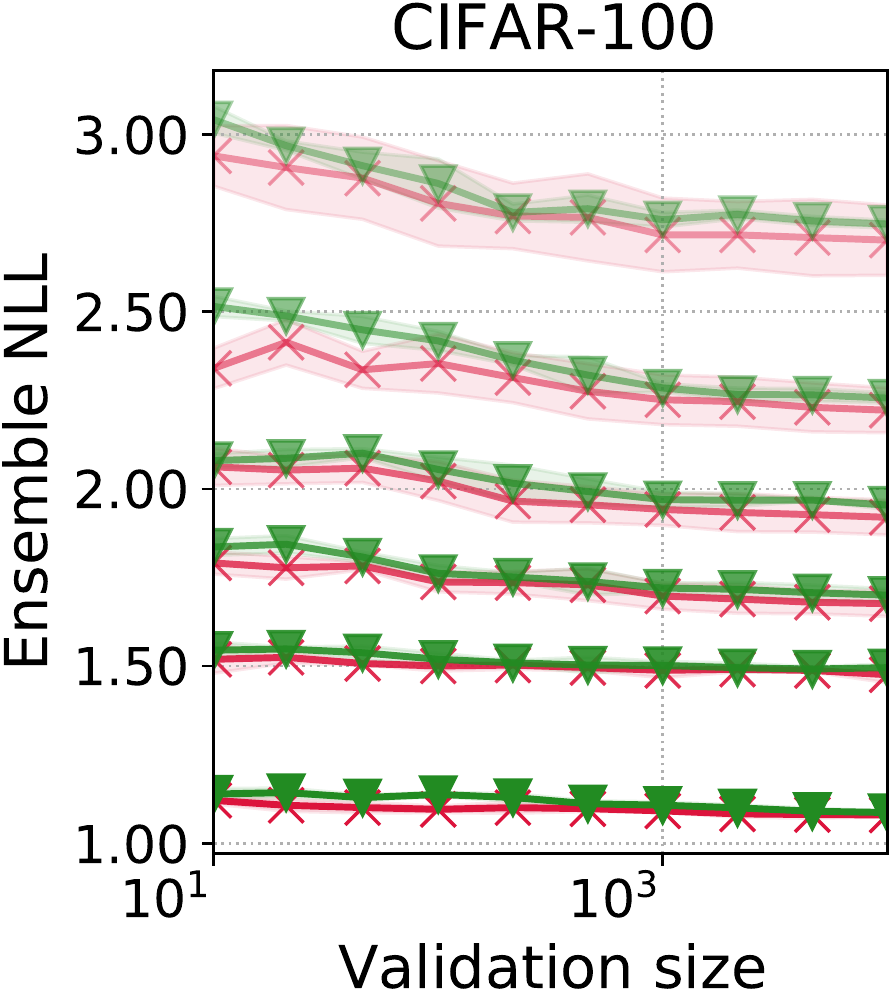}
        \includegraphics[width=.49\linewidth]{figures/validation_data_study/tiny_metric_loss_M_10.pdf}
        \subcaption{Test performance of NES algorithms with varying validation data sizes. Each curve corresponds to one particular dataset shift severity (0-5 with 0 being no shift), with more transparent curves corresponding to higher shift severities. The loss stays relatively constant even with 10$\times$ fewer validation data samples, indicating NES is not very sensitive to validation dataset size.}
        \label{fig:val-size-sensitivity}
    \end{subfigure}
    \caption{See Appendix \ref{app:val-data-sensitivity-overfitting} for details.}
\end{wrapfigure}

In this section, we consider how sensitive the performance of NES is to changes in the size of the validation dataset $\Dval$, and we discuss why overfitting to $\Dval$ is not a concern in our experiments. Recall that $\Dval$ is used by NES algorithms during ensemble selection from the pool of base learners as outlined in Algorithm \ref{alg:esa}, and we use $\Dval$ or $\Dvals$ depending on whether dataset shift is expected at test time as described in Section \ref{sec:ens-adaptation-shift}. Over the DARTS search space for CIFAR-100 and Tiny ImageNet, we measure the test loss of NES as a function of different validation dataset sizes. Specifically, we measure test loss of the ensembles selected using validation datasets of different sizes (with as few as 10 validation samples). Figure \ref{fig:val-size-sensitivity} shows this for different levels of dataset shift severity. The results indicate that NES is insensitive to the validation set size, achieving almost the same loss when using 10$\times$ fewer validation data. 

NES also did not overfit to the validation data in our experiments. This can be seen in Figure \ref{fig:val-size-overfitting} for CIFAR-100 over the DARTS search space, which shows that both the test and validation losses decrease over time (i.e. as the pool size $\budget$ increases). This finding is consistent across datasets and search spaces in our experiments. We also remark that overfitting is unexpected, because ensemble selection ``fits a small number of parameters'' since it only selects which base learners are added to the ensemble.

\subsection{Re-training the selected architectures on $\Dtrain + \Dval$}
\label{app:dtrain_and_dval}

\begin{wrapfigure}[12]{R}{.25\textwidth}
\vspace{-16pt}
    \centering
    \includegraphics[width=.99\linewidth]{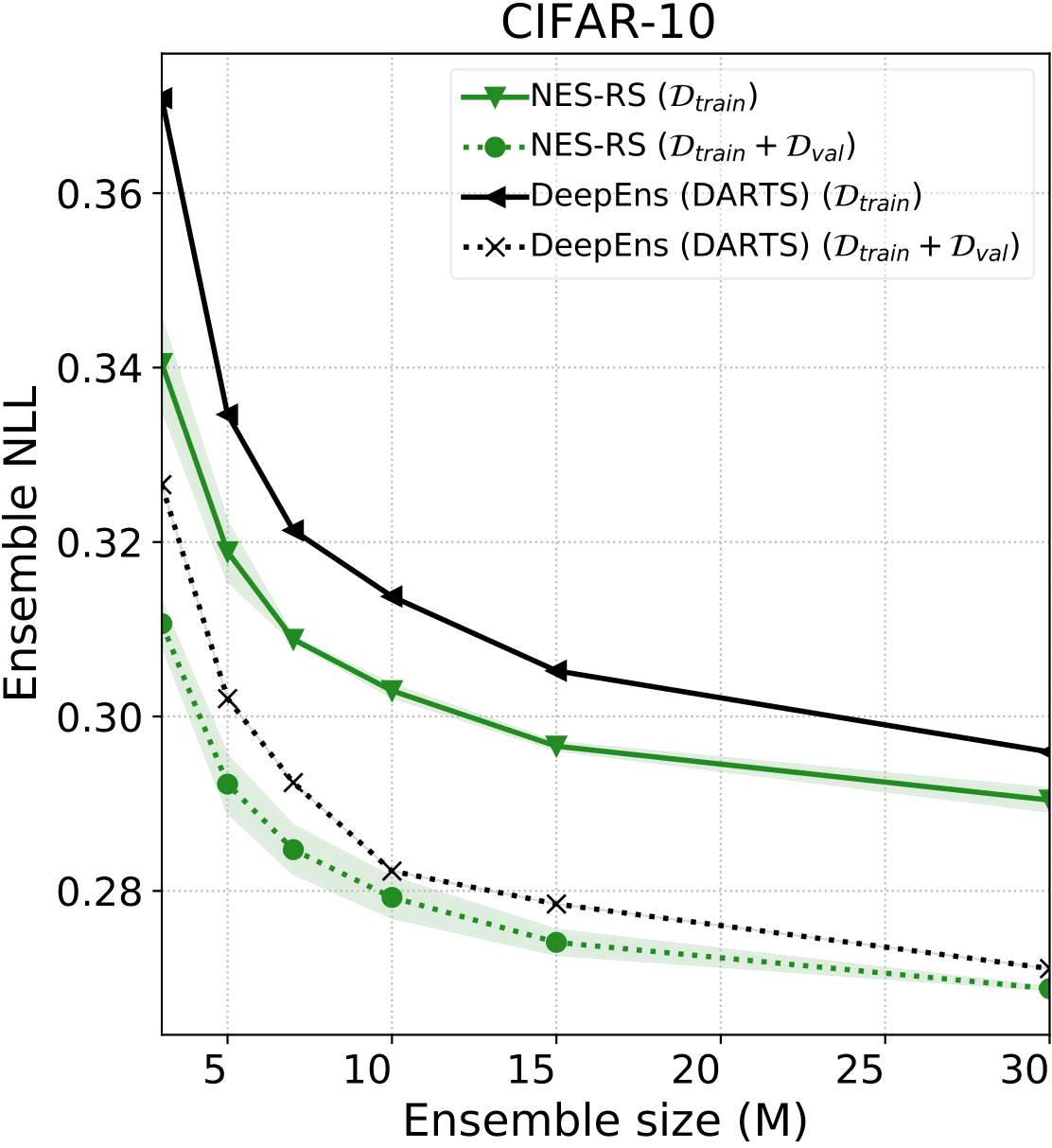}
    \caption{Re-training the selection architectures on $\Dtrain + \Dval$.}
    \label{fig:train_and_val}
\end{wrapfigure}

In practice, one might choose to re-trained the selected architectures from NES on the $\Dtrain + \Dval$ to make maximal use of the data available. We re-trained the ensembles constructed by NES and the best performing deep ensemble baseline, DeepEns (DARTS), on $\Dtrain + \Dval$, finding that both ensembles improve since the base learners improve due to more training data (see Figure~\ref{fig:train_and_val}). Note that, as shown in Appendix~\ref{app:val-data-sensitivity-overfitting}, the performance of ensemble selection (which uses $\Dval$) is relatively insensitive to the size of $\Dval$. Therefore, one way to bypass the additional cost of having to re-train the ensembles on $\Dtrain + \Dval$ is to simply pick a very small $\Dval$, such that the performance of the models when trained on $\Dtrain + \Dval$ is approximately the same as when trained on $\Dtrain$.

\subsection{Averaging logits vs. averaging probabilities in an ensemble}
\label{app:logits_vs_probabilities}

\begin{figure}
    \centering
    \includegraphics[width=.99\linewidth]{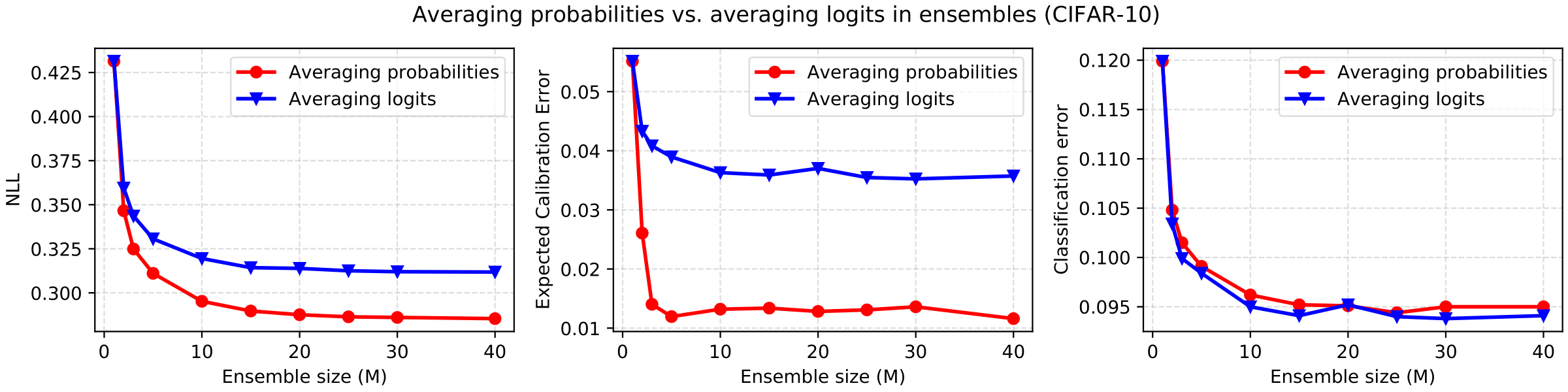}
    \caption{See Appendix \ref{app:logits_vs_probabilities} for details.}
    \label{fig:logits_vs_probs}
\end{figure}

In Figure~\ref{fig:logits_vs_probs}, we explore whether ensembles should be constructed by averaging probabilities (i.e. post-softmax) as done in our work or by averaging the logits (i.e. pre-softmax). We find that while classification error of the resulting ensembles is similar, ensembles with averaged probabilities perform notably better in terms of uncertainty estimation (NLL) and are better calibrated as well (ECE).

\subsection{Comparison of ensemble selection algorithms}
\label{app:esa_comparison}

In Figure \ref{fig:esa_comparison}, we compare the performance of NES for different choices of the ensemble selection algorithm (ESA) used in stage 2 (ensemble selection from pool). In addition to our default $\esa$ without replacement, we experiment with the following seven choices:

\begin{itemize}[leftmargin=*]
    \item $\topm$: selects the top M (ensemble size) models by validation performance.
    \item $\quickgreedy$: Starting with the best network by validation performance, add the next best network to the ensemble only if it improves validation performance, iterating until the ensemble size is M or all models have been considered (returning an ensemble of size at most M).
    \item $\esa$ with replacement: $\esa$ but with replacement which allows repetitions of base learners.
    \item $\stacking$ (Weighted): Linearly combines all base learners in the pool with learned stacking weights (which are positive and sum to 1). Then keeps only the base learners with the M largest stacking weights and weights them using the renormalized learned stacking weights.
    \item $\stacking$ (Unweighted): Linearly combines all base learners in the pool with learned stacking weights (which are positive and sum to 1). Then keeps only the base learners with the M largest stacking weights and combines them by a simple, unweighted average.
    \item $\esa$ with Bayesian model averaging by likelihood: Selects the base learners in the ensemble using $\esa$ and then takes an average weighted by the (normalized) validation likelihoods of the base learners.
    \item $\esa$ with Bayesian model averaging by accuracy: Selects the base learners in the ensemble using $\esa$ and then takes an average weighted by the (normalized) validation accuracies of the base learners.
\end{itemize}

\begin{figure}
    \centering
    \begin{subfigure}[t]{0.48\textwidth}
        \centering
        \includegraphics[width=.49\linewidth]{figures/rebuttal/nes-rs-es-error-bar-1.pdf}
        \includegraphics[width=.49\linewidth]{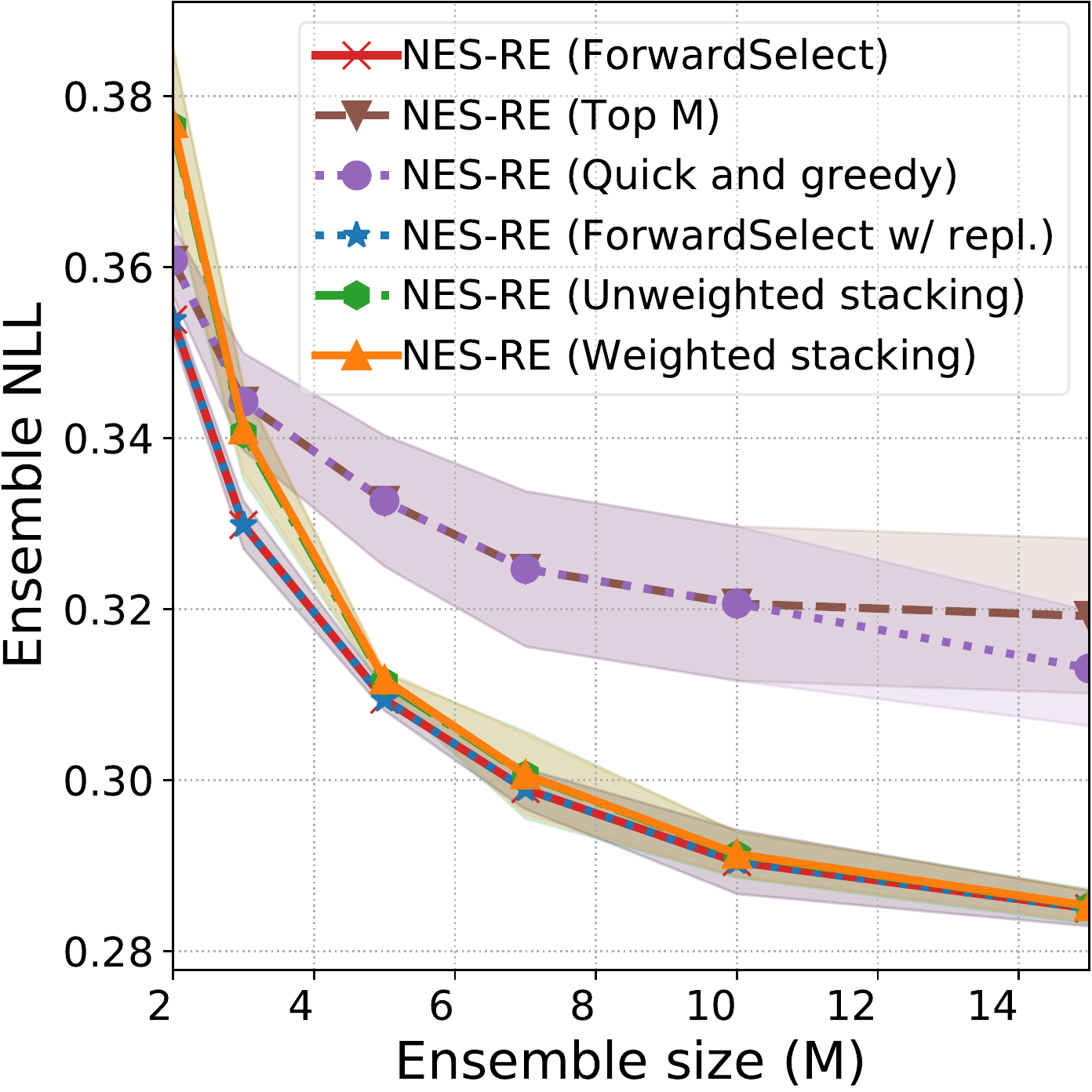}
        \subcaption{\textsc{Left:} NES-RS. \textsc{Right:} NES-RE.}
        \label{fig:other_esa}
    \end{subfigure}
    ~\hspace{.05cm}
    \begin{subfigure}[t]{0.48\textwidth}
        \centering
        \includegraphics[width=.49\linewidth]{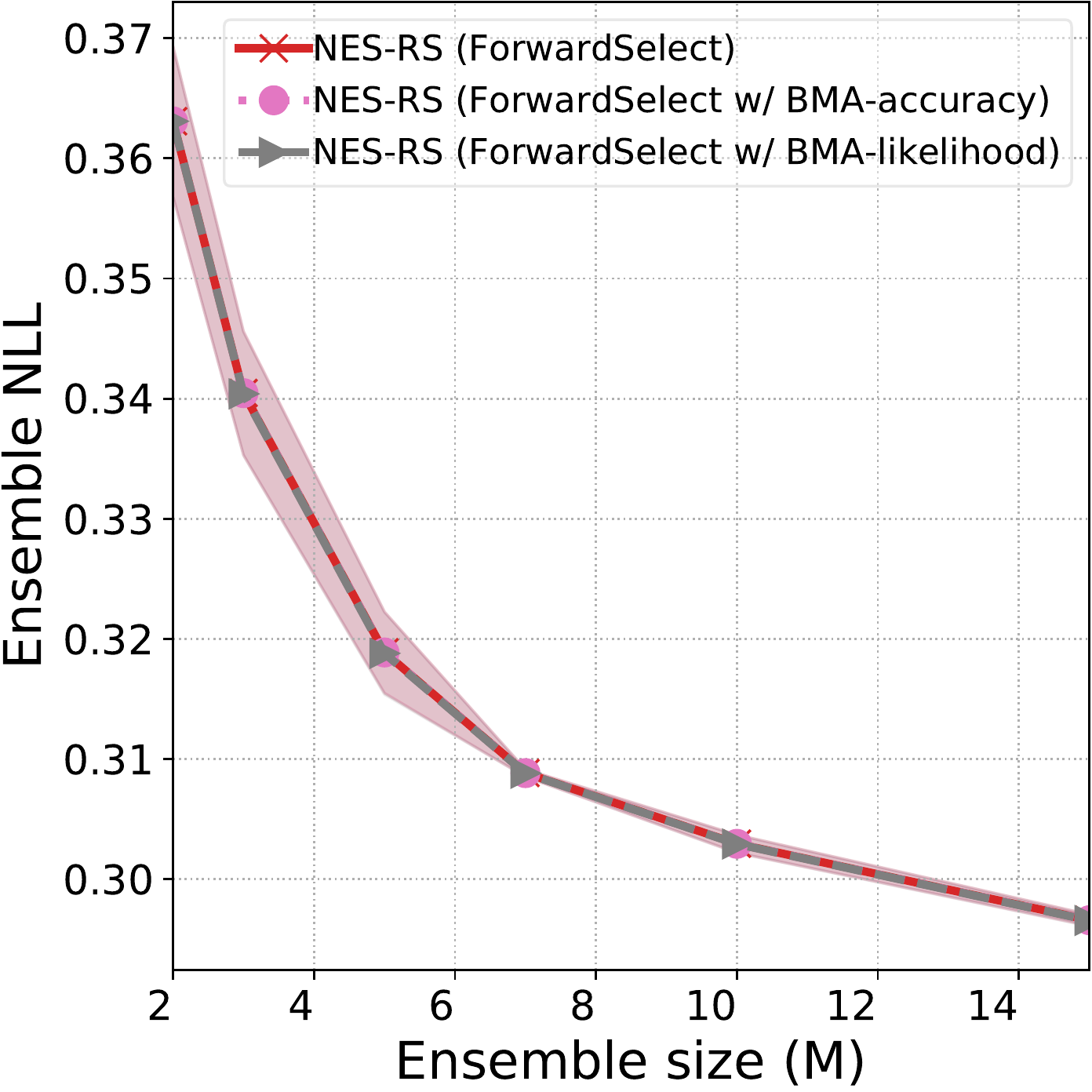}
        \includegraphics[width=.49\linewidth]{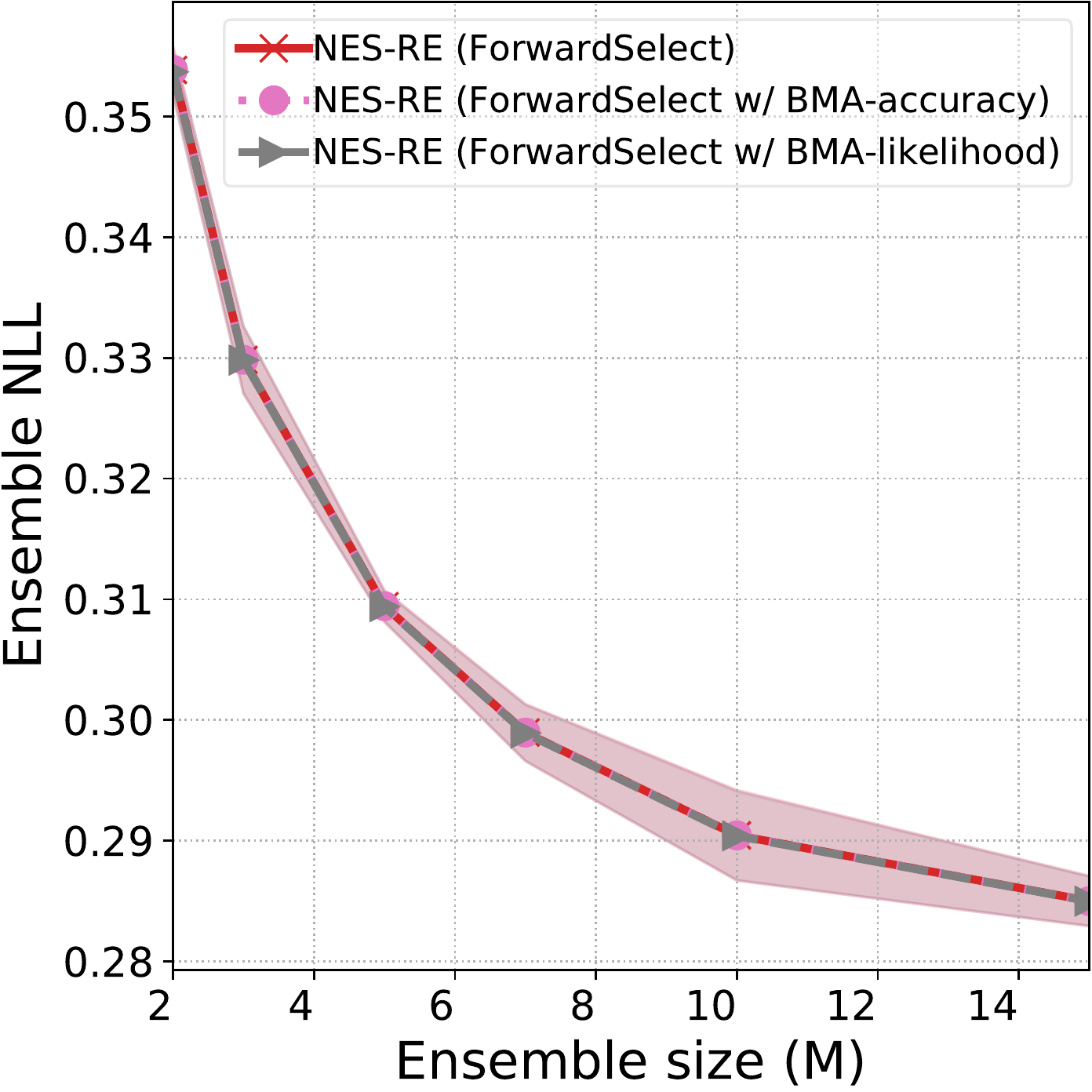}
        \subcaption{\textsc{Left:} NES-RS. \textsc{Right:} NES-RE.}
        \label{fig:bma}
    \end{subfigure}
    \caption{$\esa$ compared to other ensemble selection algorithms. Experiments were conducted on CIFAR-10 over the DARTS search space. See Appendix \ref{app:esa_comparison} for details.}
    \label{fig:other_esa_and_bma}
\end{figure}

\begin{wrapfigure}[15]{R}{.6\textwidth}
    \vspace{-4mm}
    \centering
    \includegraphics[width=.49\linewidth]{figures/rebuttal/esa-diversity-nes-rs.pdf}
    \includegraphics[width=.49\linewidth]{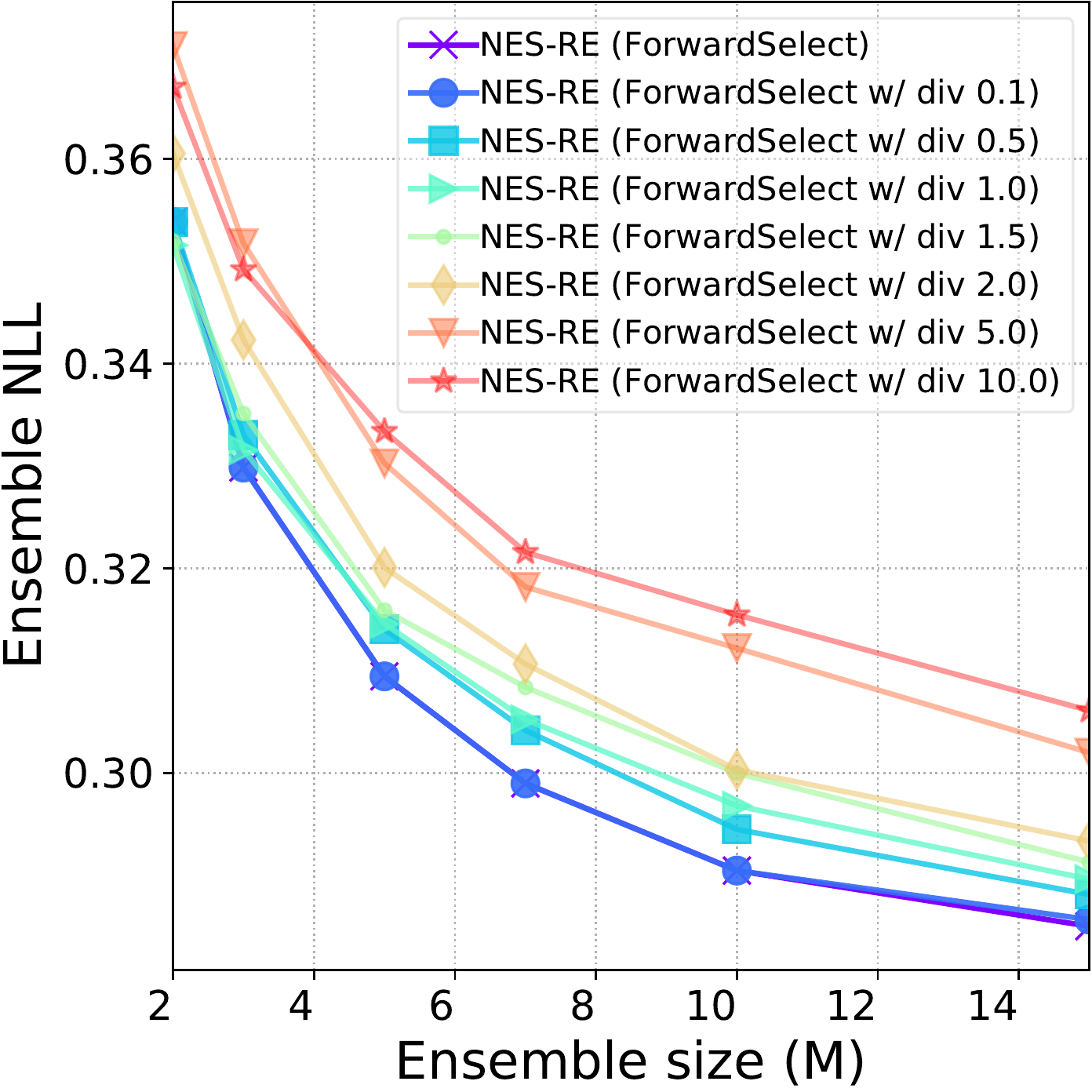}
    \caption{Ensemble selection with explicit diversity regularization. Experiments were conducted on CIFAR-10 over the DARTS search space.}
    \label{fig:esa_diversity}
\end{wrapfigure}

In Figure~\ref{fig:other_esa_and_bma} we provide additional results including the one in Figure \ref{fig:esa_comparison}. From the plots we can observe that:

\begin{itemize}[leftmargin=*]
    \item $\esa$ performs better than or at par with all ESAs considered here.
    \item $\stacking$ tends to perform competitively but still worse than $\esa$, and whether weighted averaging is used or not has a very minor impact on performance, since the learned weights tend to be close to uniform.
    \item Bayesian model averaging (BMA) also has a very minor impact (almost invisible in the plots in Figure~\ref{fig:bma}), because the likelihoods and accuracies of individual base learners are very similar (a consequence of multi-modal loss landscapes in neural networks with different models achieving similar losses) hence the BMA weights are close to uniform. The NLL achieved by the BMA ensemble only differed at the 4th or 5th decimal places compared to unweighted ensembles with the same base learners.
\end{itemize}

Lastly, we assess the impact of \textit{explicitly} regularizing for diversity during ensemble selection as follows: we use $\esa$ as before but instead use it to minimize the objective ``validation loss - diversity strength $\times$ diversity'' where diversity is defined as the average (across base learners and data points) $L_2$ distance between a base learner's predicted class probabilities and the ensemble's predicted class probabilities. The plots below show the results for different choices of the ``diversity strength'' hyperparameter. In summary, if appropriately tuned, $\esa$ with diversity performs slightly better than usual $\esa$ for NES-RS, though for NES-RE, the diversity term seems to harm performance as shown in Figure \ref{fig:esa_diversity}-right. 

\subsection{Generating the architecture pool using weight-sharing NAS algorithms}
\label{app:one_shot_nes}

\begin{wrapfigure}[18]{R}{.6\textwidth}
    \vspace{-4mm}
    \centering
    \includegraphics[width=.49\linewidth]{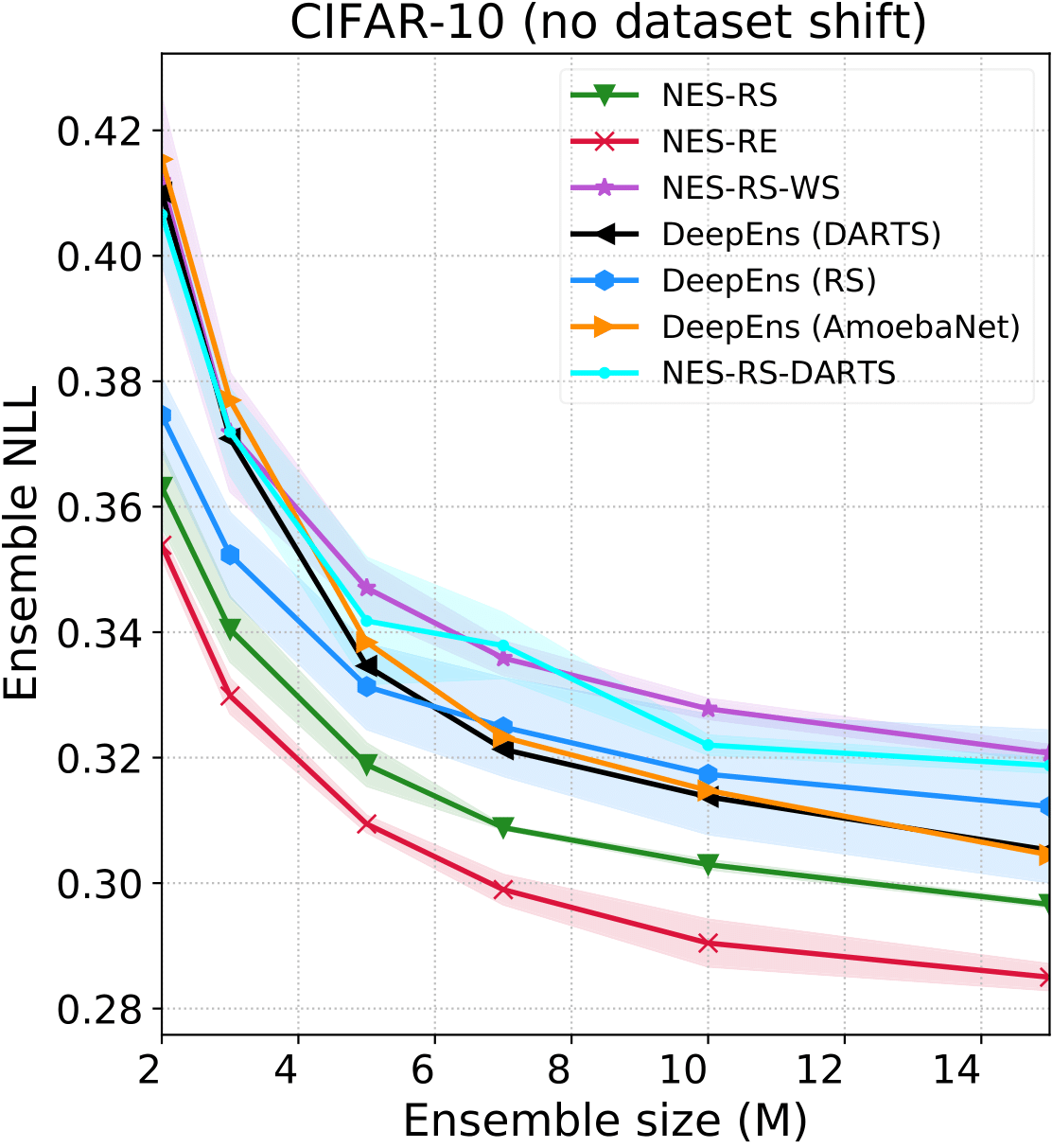}
    \includegraphics[width=.49\linewidth]{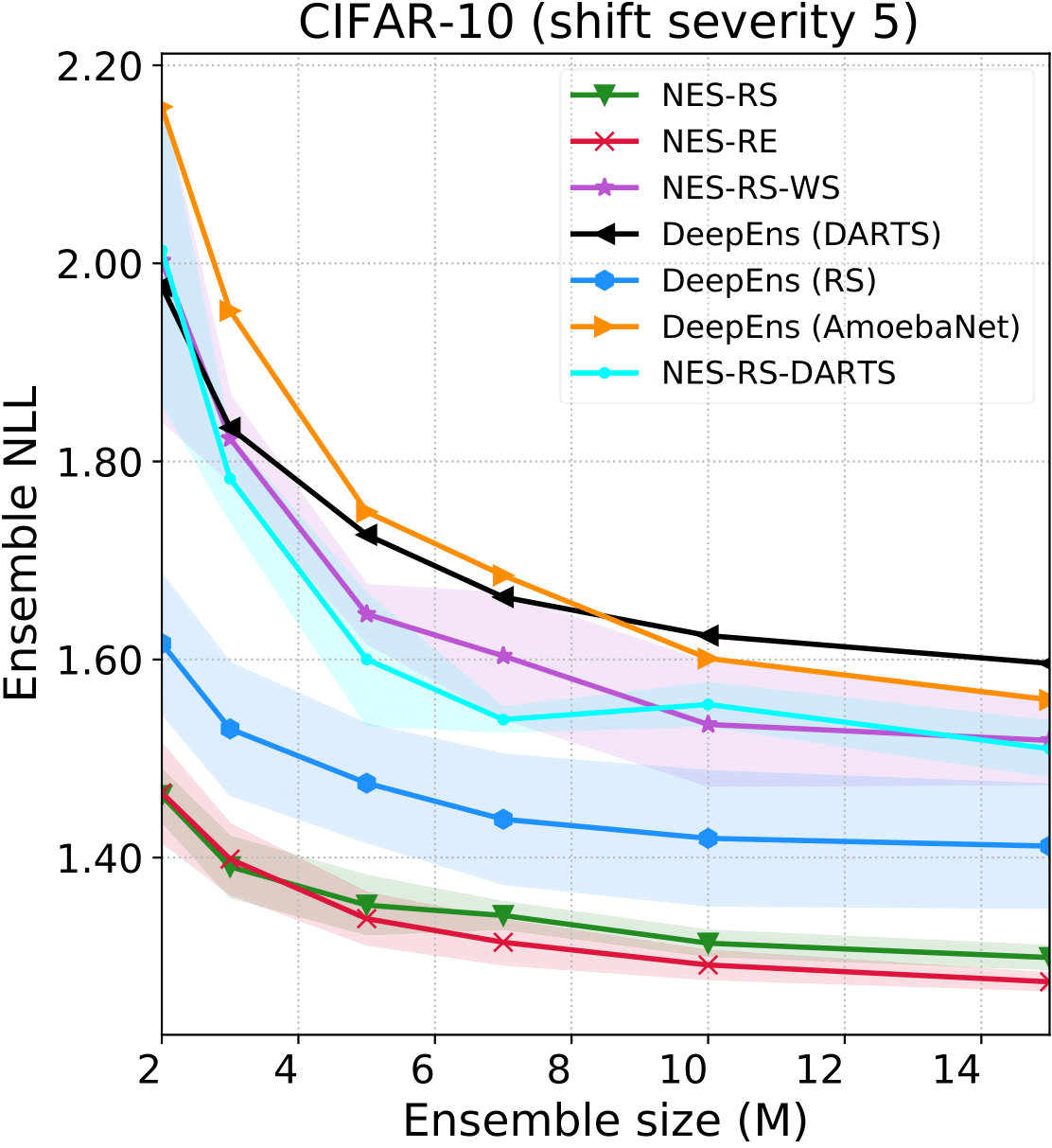}
    \caption{Comparison of NES-RS-WS and NES-RS-DARTS with the other baselines and NES-RS and NES-RE. The plots show the NLL vs. ensemble sizes on CIFAR-10 with and without dataset shifts over the DARTS search space. Mean NLL shown with 95 confidence intervals.}
    \label{fig:one_shot_nes}
\end{wrapfigure}

In order to accelerate the NES search phase, we generated the pool using the weight sharing schemes proposed by Random Search with Weight Sharing~\citep{li-uai19a} and DARTS~\citep{liu2018darts}. Specifically, we trained one-shot weight-sharing models using each of these two algorithms, then we sampled architectures from the weight-shared models uniformly at random to build the pool. Next, we ran $\esa$ as usual to select ensembles from these pools. Finally, we retrained the selected base learners from scratch using the same training pipeline as NES and the deep ensemble baselines. In Figure~\ref{fig:one_shot_nes}, we refer to the results of these methods as NES-RS-WS and NES-RS-DARTS, respectively. As we can see, their performance is worse than NES-RS and NES-RE, even though the computation cost is reduced significantly. This is likely due to the low correlation between the performance of the architectures when evaluated using the shared weights and the performance when re-trained in isolation. Prior work has shown this is caused by weight interference and co-adaptation which occurs during the weight-sharing model training~\citep{Yu2020Evaluating, zela2020nasbenchshot}. Nonetheless, we believe this is a promising avenue for future research requiring more development, as it substantially reduces the cost of exploring the search space.

\clearpage
\section{NAS Best Practice Checklist}\label{app:nas_best_practice_checklist}
Our experimental setup resembles that of NAS papers, therefore, to foster reproducibility, we now describe how we addressed the individual points of the NAS best practice checklist by Lindauer \& Hutter~\citep{lindauer2019best}.

\begin{enumerate}

\item \textbf{Best Practices for Releasing Code}\\[0.2cm]
For all experiments you report: 
\begin{enumerate}
  \item Did you release code for the training pipeline used to evaluate the final architectures?
    \answerYes{Code for our experiments, including NES and baselines, is open-source.}
  \item Did you release code for the search space?
  \answerYes{}
  \item Did you release the hyperparameters used for the final evaluation pipeline, as well as random seeds?
  \answerYes{}
  \item Did you release code for your NAS method?
  \answerYes{}
  \item Did you release hyperparameters for your NAS method, as well as random seeds?
  \answerYes{}   
\end{enumerate}

\item \textbf{Best practices for comparing NAS methods}
\begin{enumerate}
  \item For all NAS methods you compare, did you use exactly the same NAS benchmark, including the same dataset (with the same training-test split), search space and code for training the architectures and hyperparameters for that code?
    \answerYes{All base learners follow the same training routine and use architectures from the same search space.}
  \item Did you control for confounding factors (different hardware, versions of DL libraries, different runtimes for the different methods)?
    \answerYes{}
	\item Did you run ablation studies?
    \answerYes{See Section~\ref{sec:analysis_and_ablations} and Appendices \ref{app:nes_re_0}, \ref{app:deepens_esa_ablation}, \ref{app:hyperdeepens}, \ref{app:val-data-sensitivity-overfitting}, \ref{app:dtrain_and_dval}, \ref{app:logits_vs_probabilities}, \ref{app:esa_comparison}, \ref{app:one_shot_nes}.}
	\item Did you use the same evaluation protocol for the methods being compared?
    \answerYes{}
	\item Did you compare performance over time?
    \answerYes{As shown in all figures with $\budget$ (which is proportional to training time) on the $x$-axis.}
	\item Did you compare to random search?
    \answerYes{}
	\item Did you perform multiple runs of your experiments and report seeds?
    \answerYes{We ran NES with multiple seeds and averaged over them in Section \ref{sec:experiments}.}
	\item Did you use tabular or surrogate benchmarks for in-depth evaluations?
    \answerYes{We performed experiments over the tabular NAS-Bench-201 search space, including an in-depth evaluation of NES by comparing it to the deep ensemble of the best architecture in the search space.} 

\end{enumerate}

\item \textbf{Best practices for reporting important details}
\begin{enumerate}
  \item Did you report how you tuned hyperparameters, and what time and resources
this required?
    \answerYes{We relied on default hyperparameter choices from Liu \etal\citep{liu2018darts} as described in Appendix \ref{app:exp_details}.}
  \item Did you report the time for the entire end-to-end NAS method
(rather than, e.g., only for the search phase)?
    \answerYes{See Figure \ref{fig:deepens_es_ablation_main_paper} and Appendix \ref{appsec:hypers}.}
  \item Did you report all the details of your experimental setup?
    \answerYes{} 

\end{enumerate}

\end{enumerate}

%% file: text/appendix_plots.tex
\begin{figure}
    \centering
    \captionsetup[subfigure]{justification=centering}
    \begin{subfigure}[t]{0.49\textwidth}
        \centering
        \includegraphics[width=.31\linewidth]{figures/c10/ens_size/test/evals/metric_loss_0.pdf}
        \includegraphics[width=.31\linewidth]{figures/c100/ens_size/test/evals/metric_loss_sev_0.pdf}
        \includegraphics[width=.31\linewidth]{figures/tiny/ens_size/test/evals/metric_loss_sev_0.pdf}
        \subcaption{No data shift}
        \label{fig:test_loss_M_shift0_replica}
    \end{subfigure}%
    \begin{subfigure}[t]{0.49\textwidth}
        \centering
        \includegraphics[width=.31\linewidth]{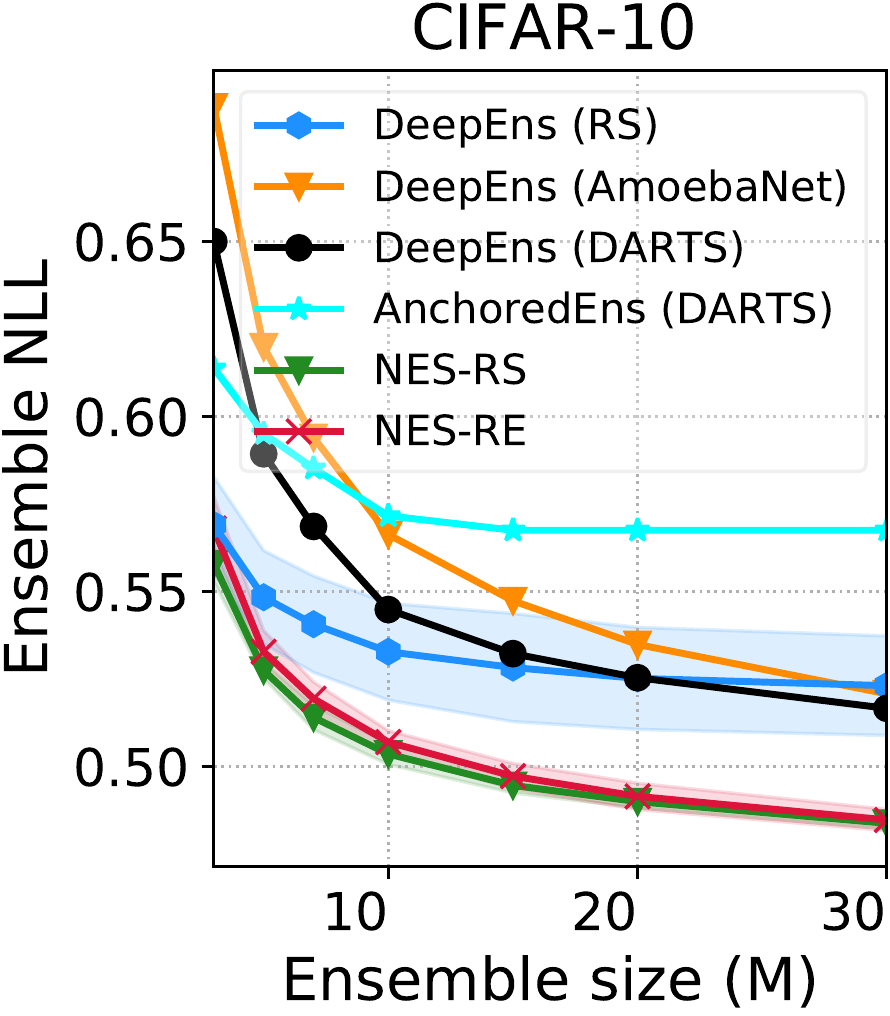}
        \includegraphics[width=.31\linewidth]{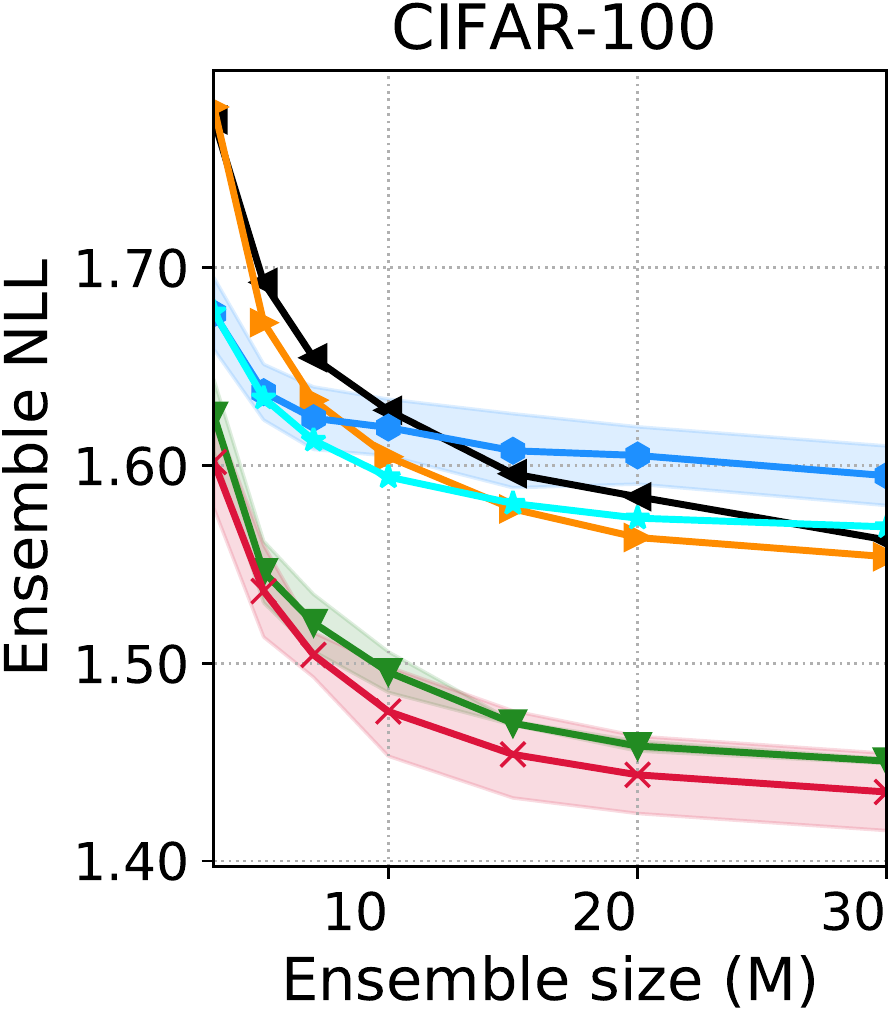}
        \includegraphics[width=.31\linewidth]{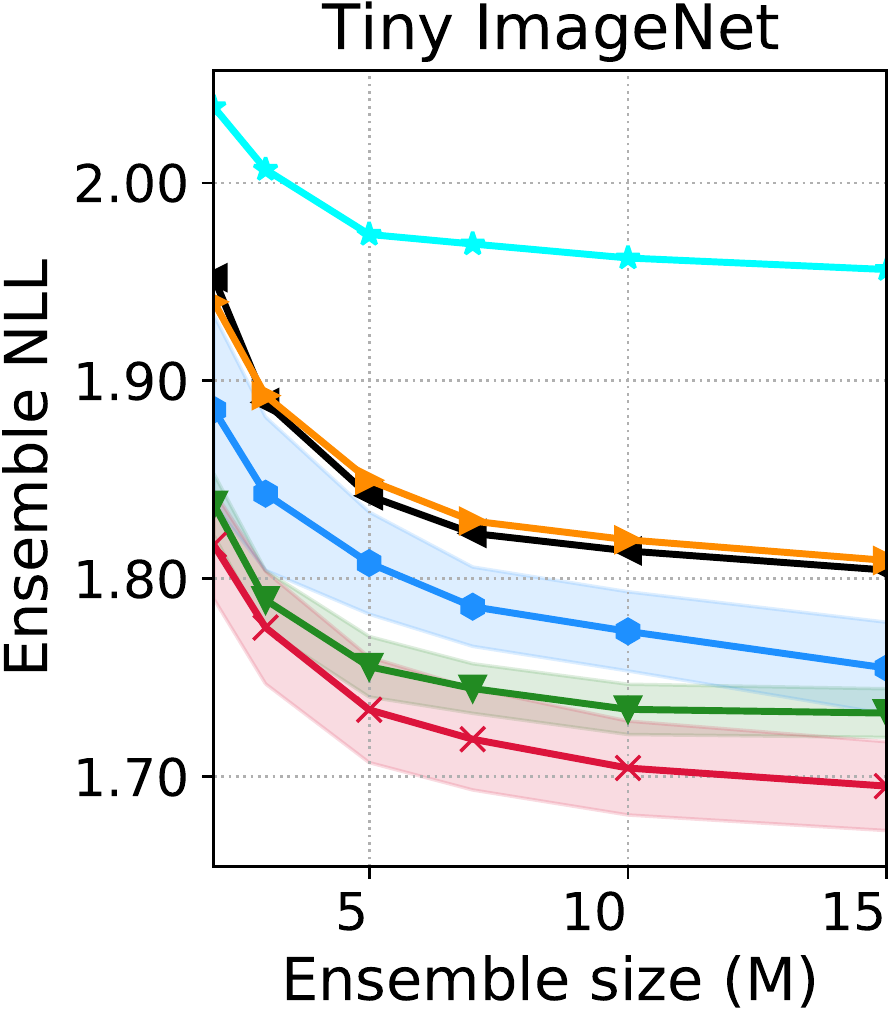}
        \subcaption{Data shift (severity 1)}
        \label{fig:test_loss_M_shift1}
    \end{subfigure}\\%
    \begin{subfigure}[t]{0.49\textwidth}
        \centering
        \includegraphics[width=.31\linewidth]{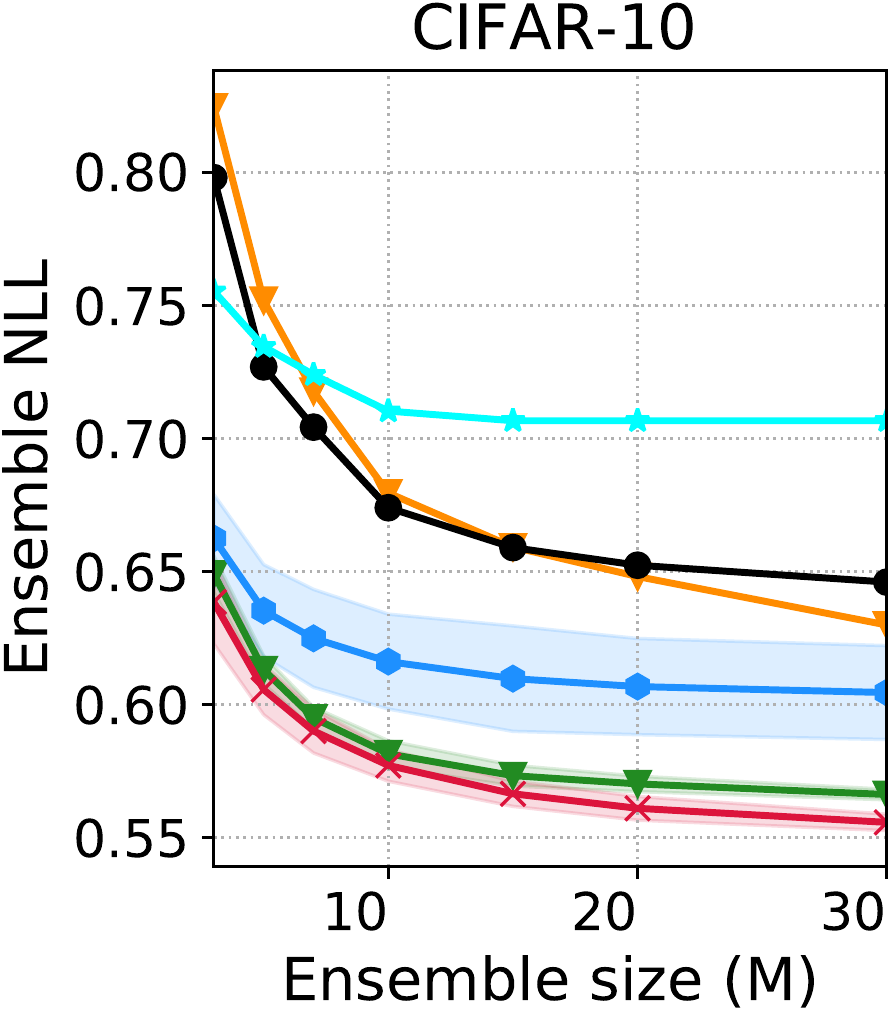}
        \includegraphics[width=.31\linewidth]{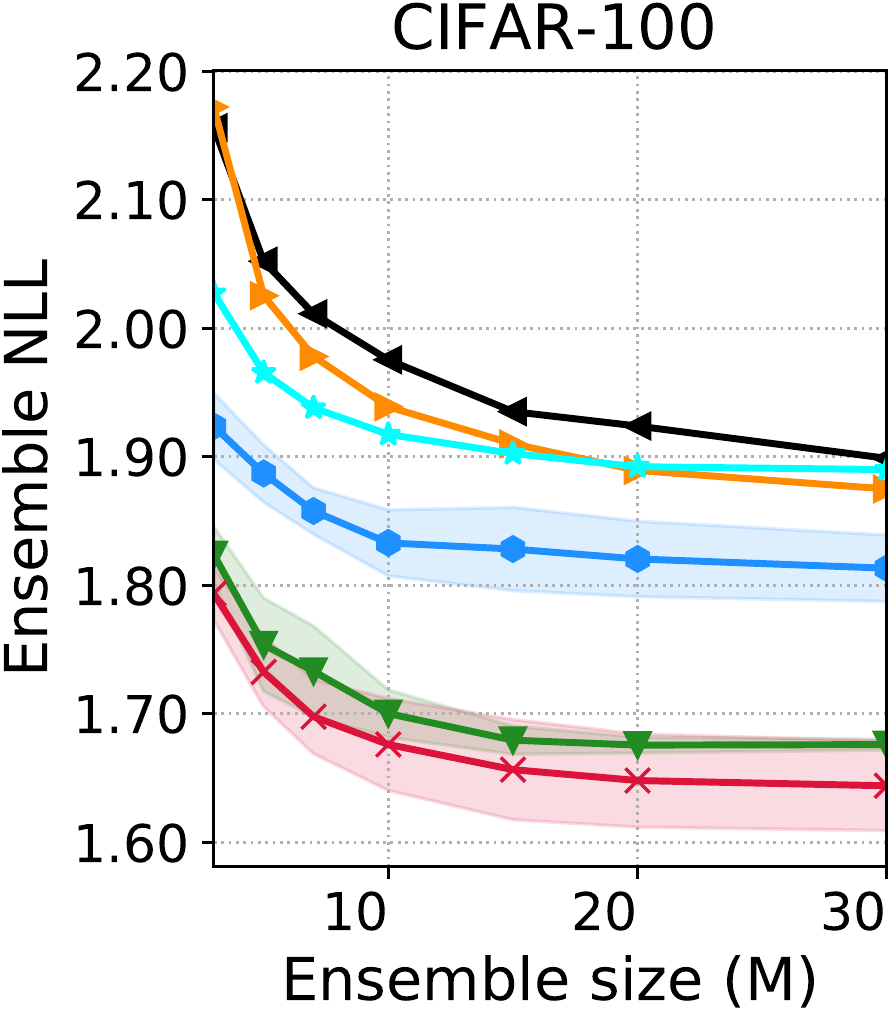}
        \includegraphics[width=.31\linewidth]{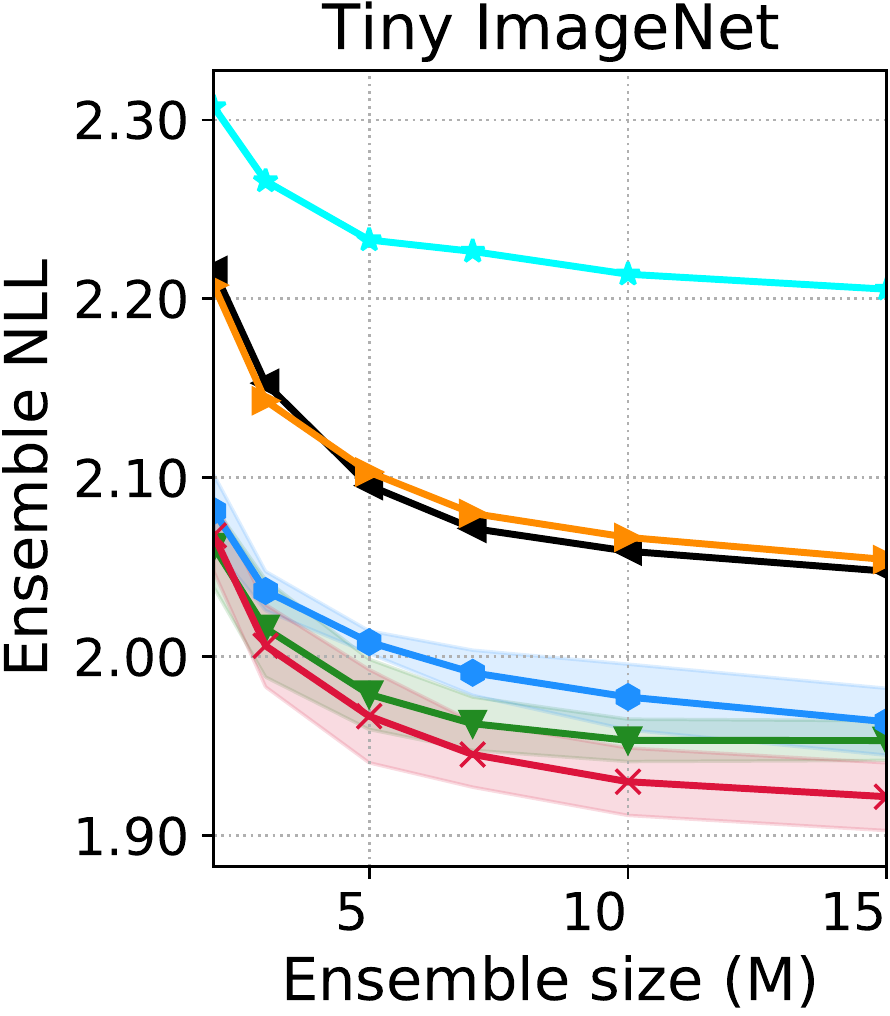}
        \subcaption{Data shift (severity 2)}
        \label{fig:test_loss_M_shift2}
    \end{subfigure}
    \begin{subfigure}[t]{0.49\textwidth}
        \centering
        \includegraphics[width=.31\linewidth]{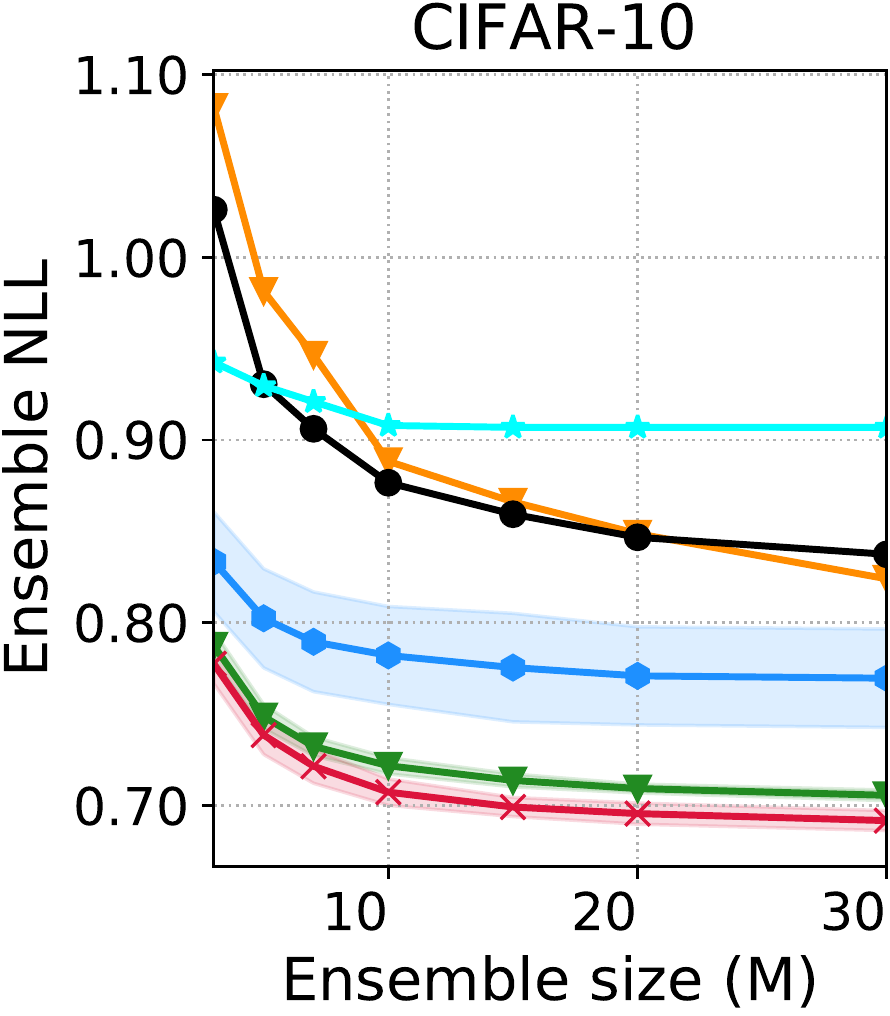}
        \includegraphics[width=.31\linewidth]{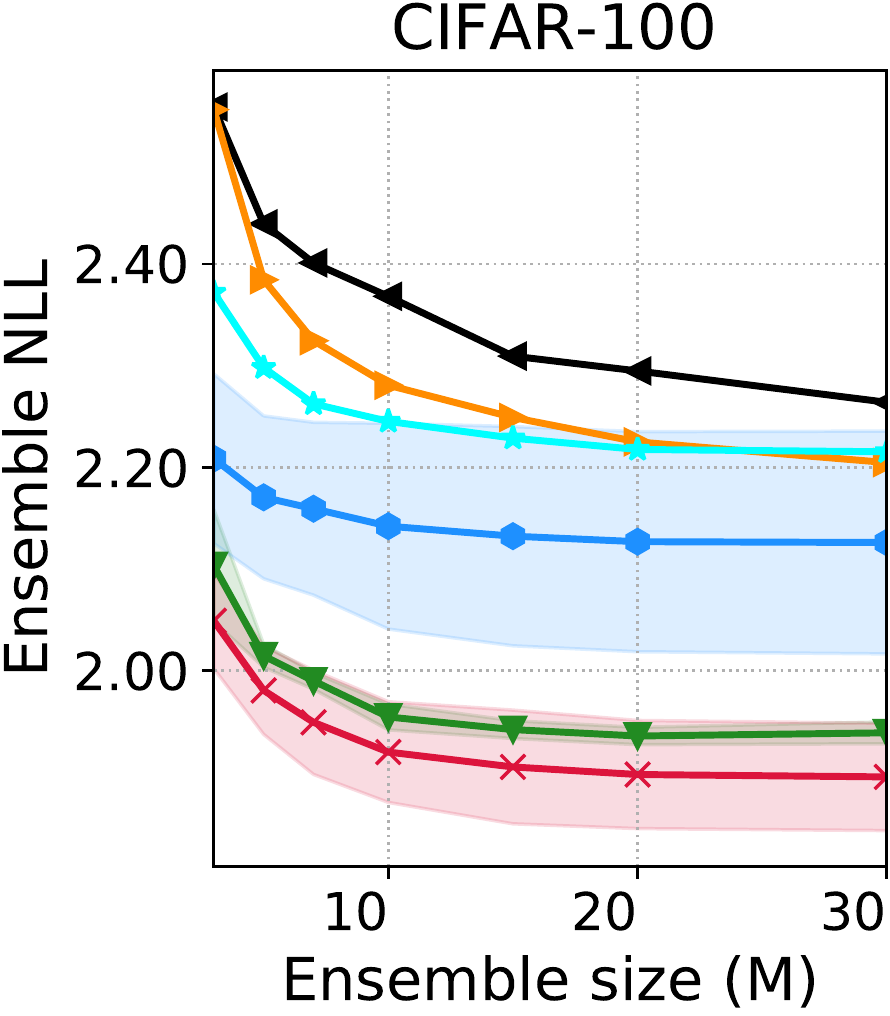}
        \includegraphics[width=.31\linewidth]{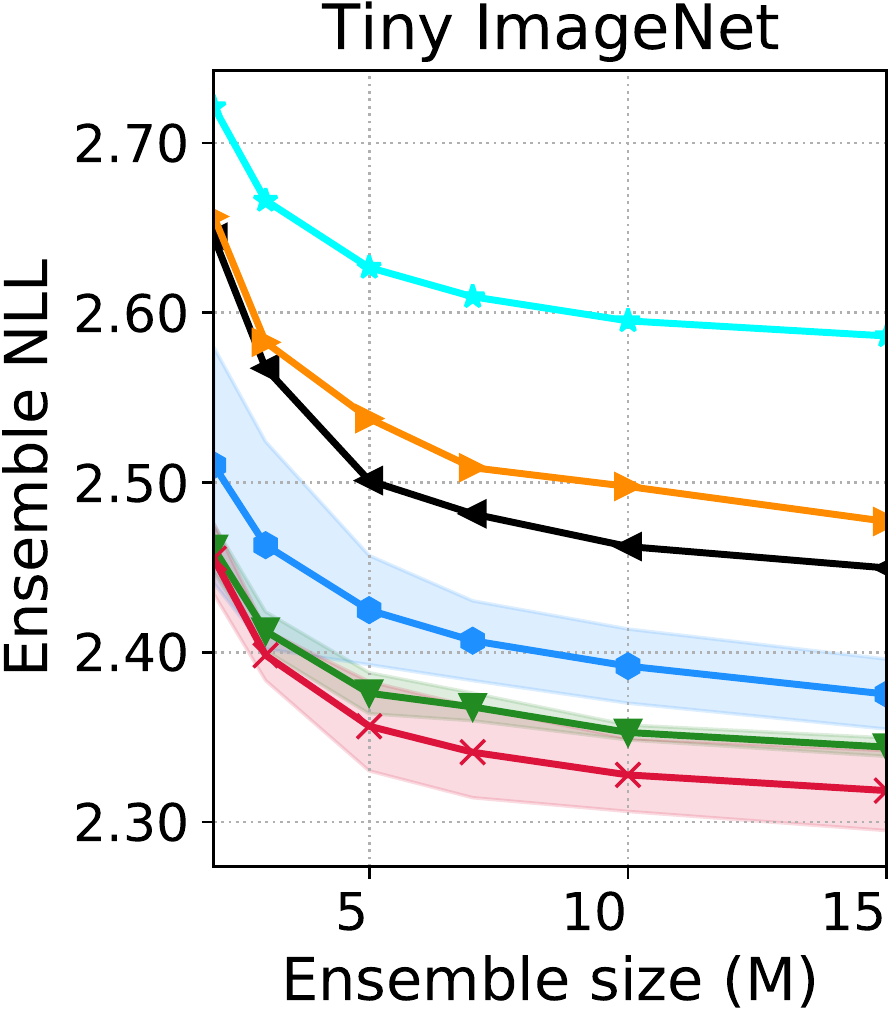}
        \subcaption{Data shift (severity 3)}
        \label{fig:test_loss_M_shift3}
    \end{subfigure}\\%
    \begin{subfigure}[t]{0.49\textwidth}
        \centering
        \includegraphics[width=.31\linewidth]{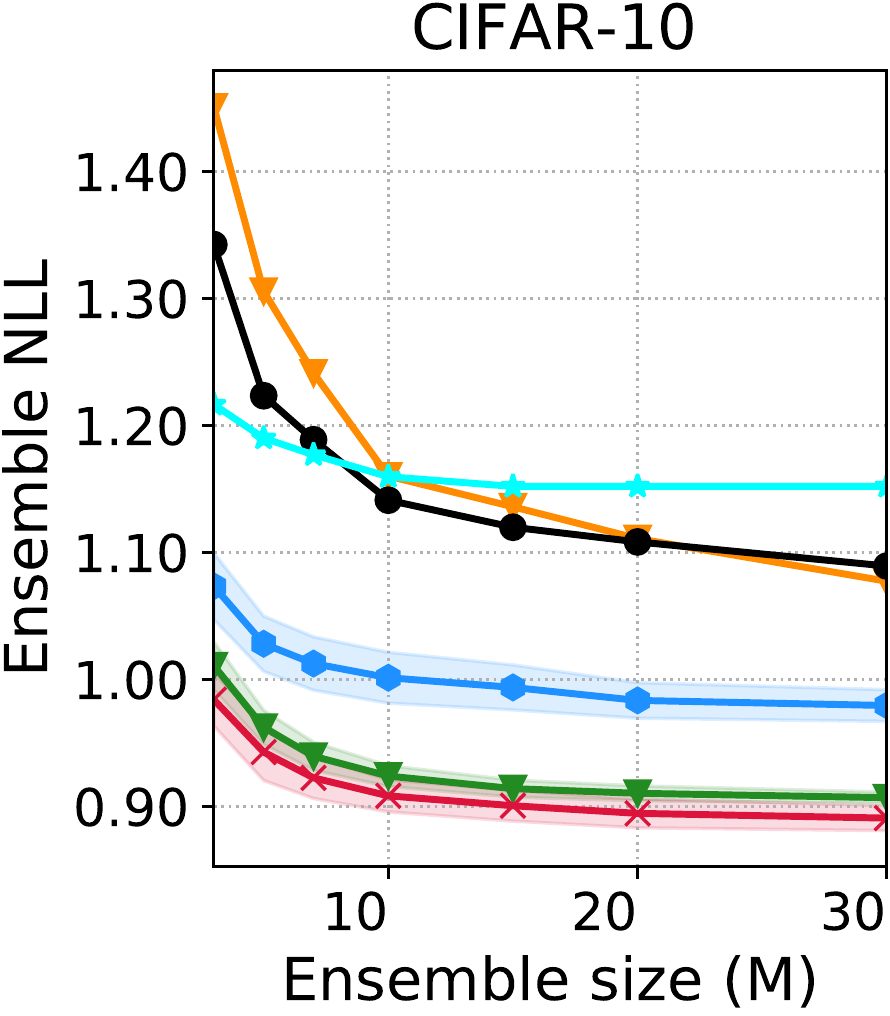}
        \includegraphics[width=.31\linewidth]{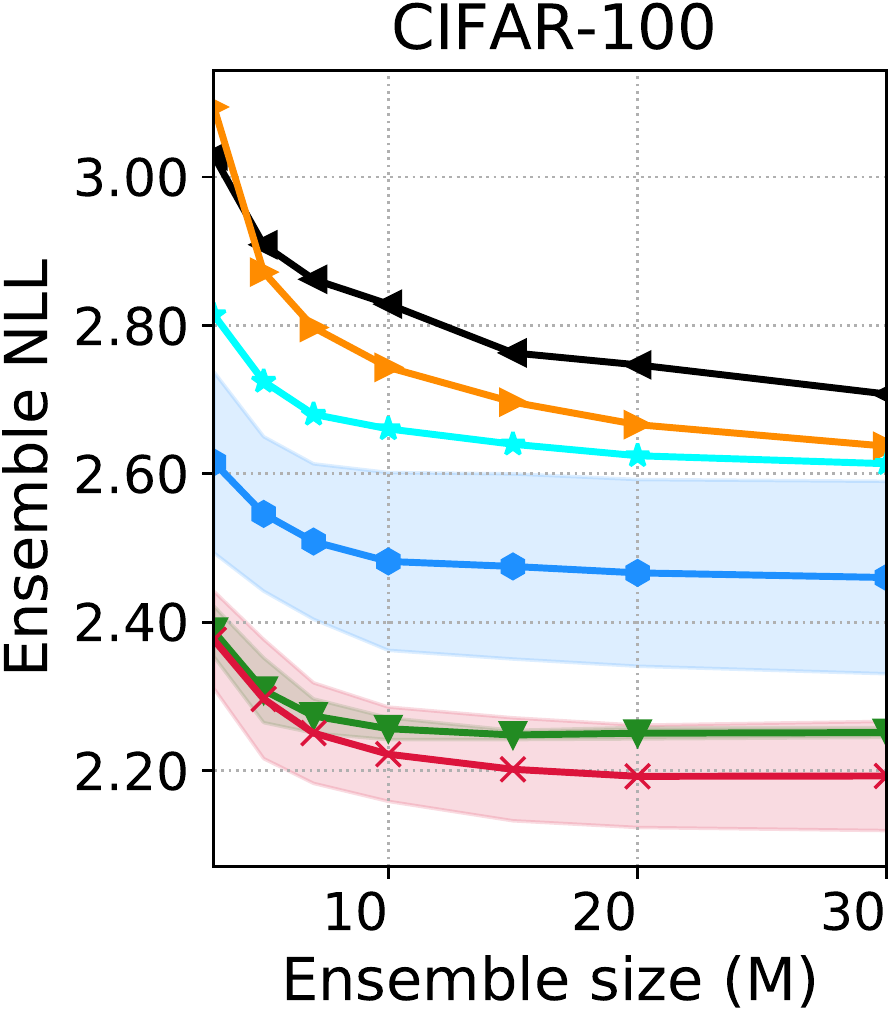}
        \includegraphics[width=.31\linewidth]{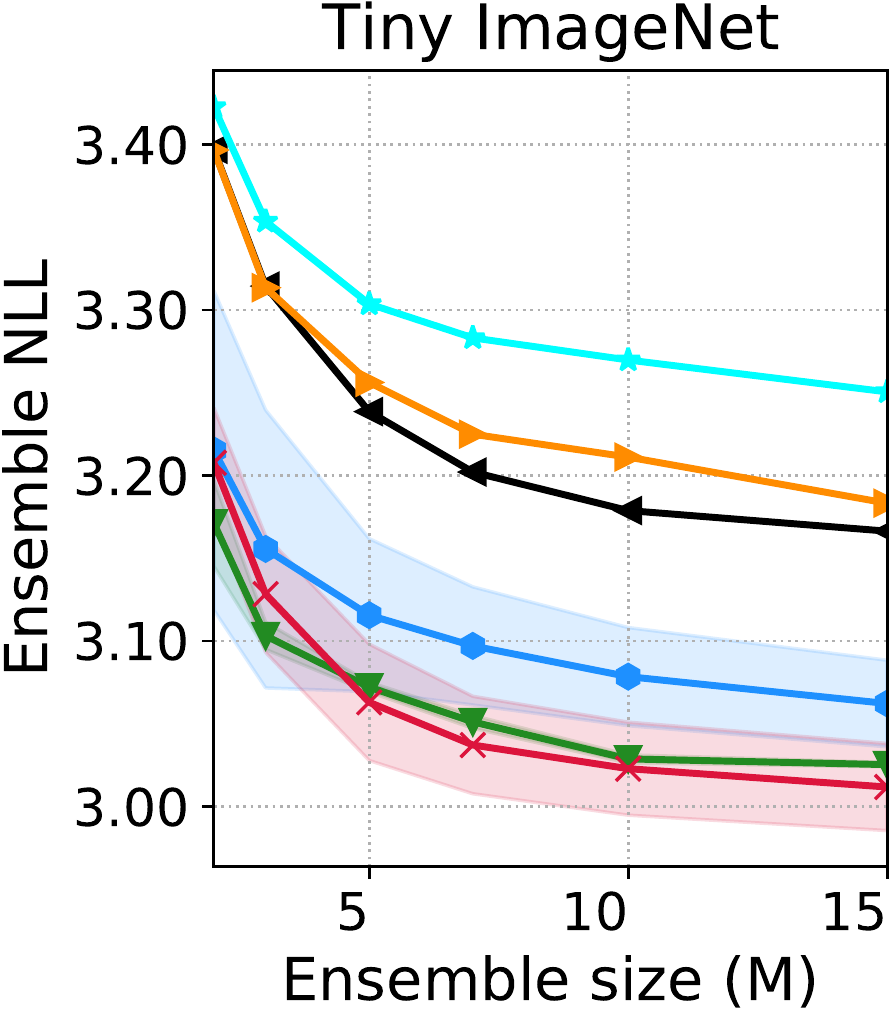}
        \subcaption{Data shift (severity 4)}
        \label{fig:test_loss_M_shift4}
    \end{subfigure}
        \begin{subfigure}[t]{0.49\textwidth}
        \centering
        \includegraphics[width=.31\linewidth]{figures/c10/ens_size/test/evals/metric_loss_5.pdf}
        \includegraphics[width=.31\linewidth]{figures/c100/ens_size/test/evals/metric_loss_sev_5.pdf}
        \includegraphics[width=.31\linewidth]{figures/tiny/ens_size/test/evals/metric_loss_sev_5.pdf}
        \subcaption{Data shift (severity 5)}
        \label{fig:test_loss_M_shift5_replica}
    \end{subfigure}%
    
    \caption{NLL vs. ensemble sizes on CIFAR-10, CIFAR-100 and Tiny ImageNet with varying dataset shifts~\citep{hendrycks2018benchmarking} over DARTS space.}
    \label{fig:test_loss_M_other}
\end{figure}

\begin{figure*}
    \centering
    \captionsetup[subfigure]{justification=centering}
    \begin{subfigure}[t]{0.49\textwidth}
        \centering
        \includegraphics[width=.3\linewidth]{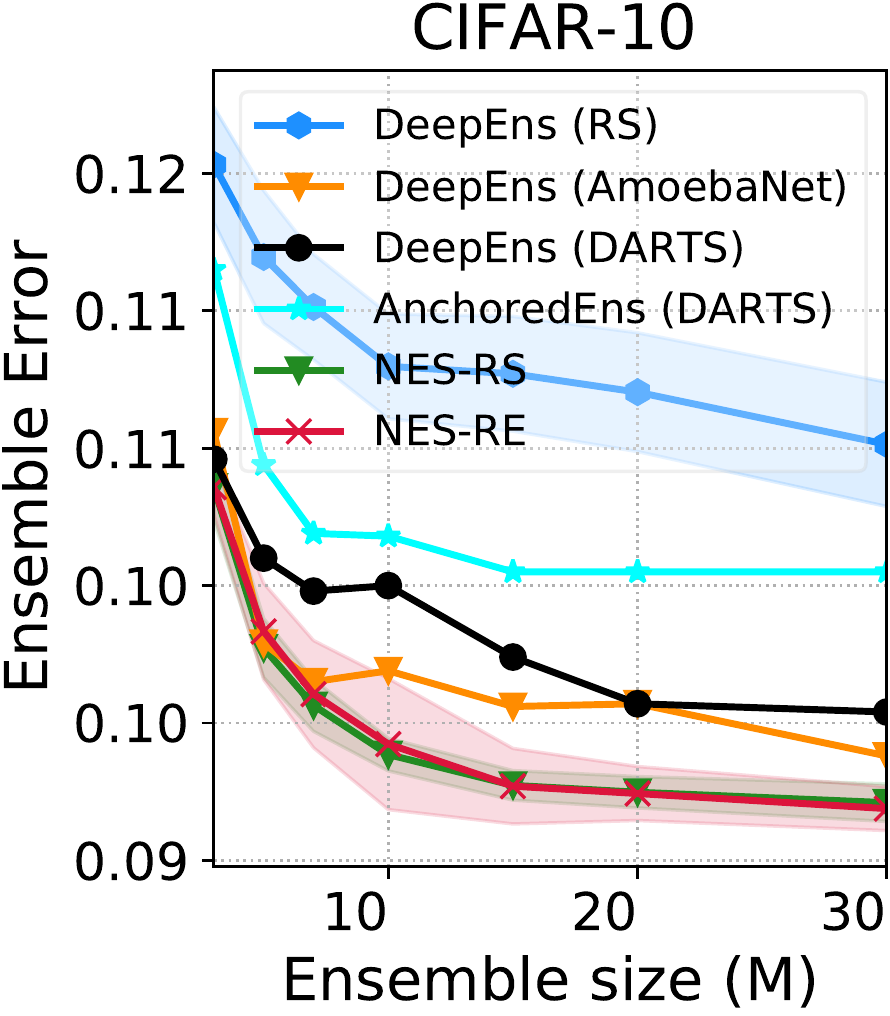}
        \includegraphics[width=.3\linewidth]{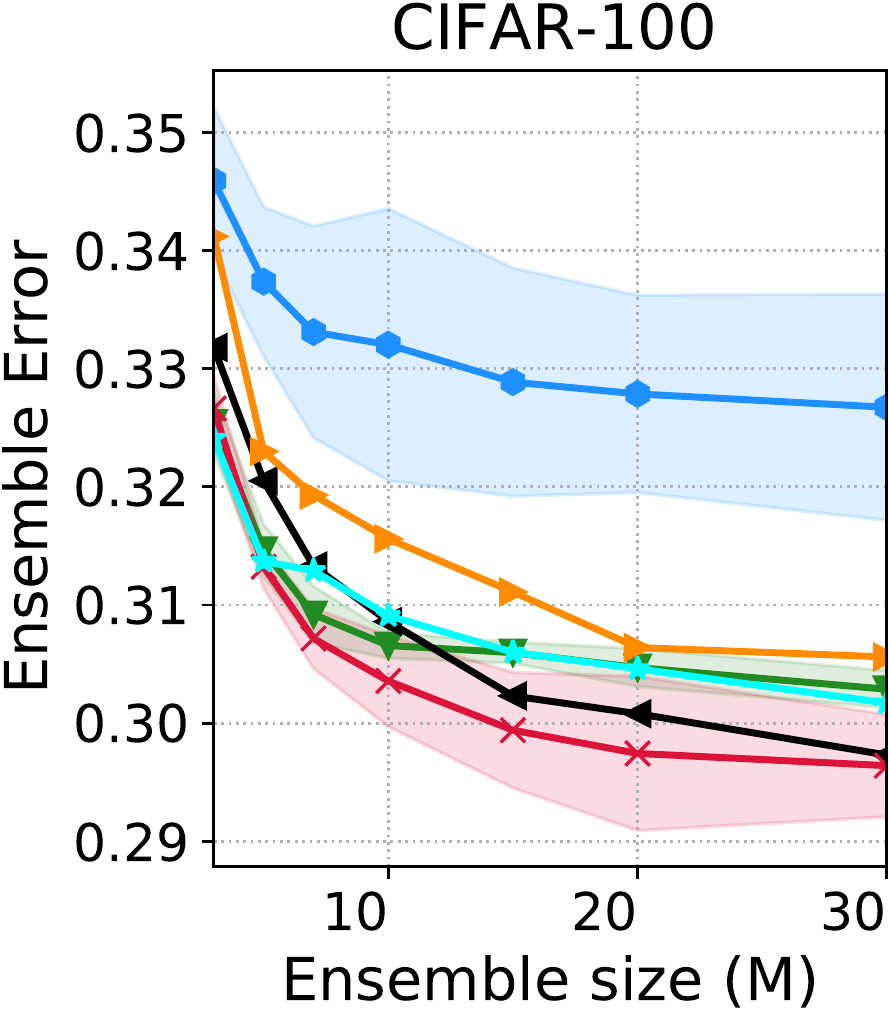}
        \includegraphics[width=.3\linewidth]{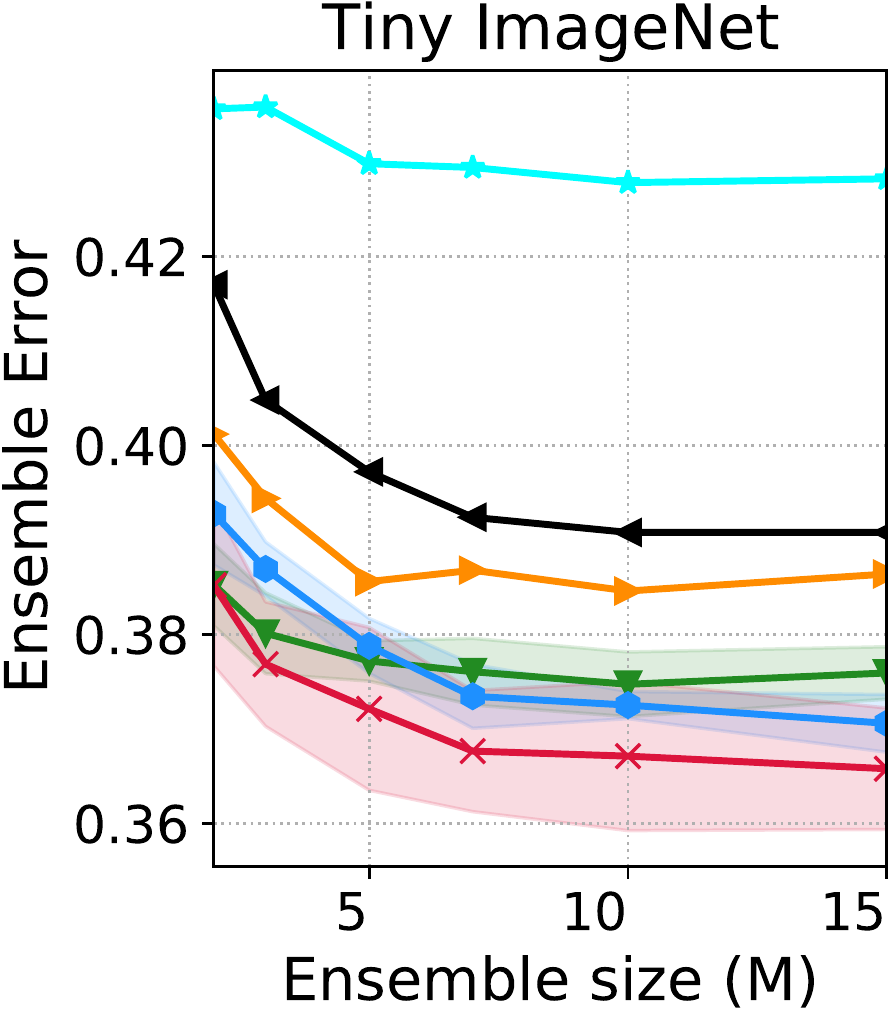}
        \subcaption{No data shift}
        \label{fig:test_error_M_shift0}
    \end{subfigure}
    \begin{subfigure}[t]{0.49\textwidth}
        \centering
        \includegraphics[width=.3\linewidth]{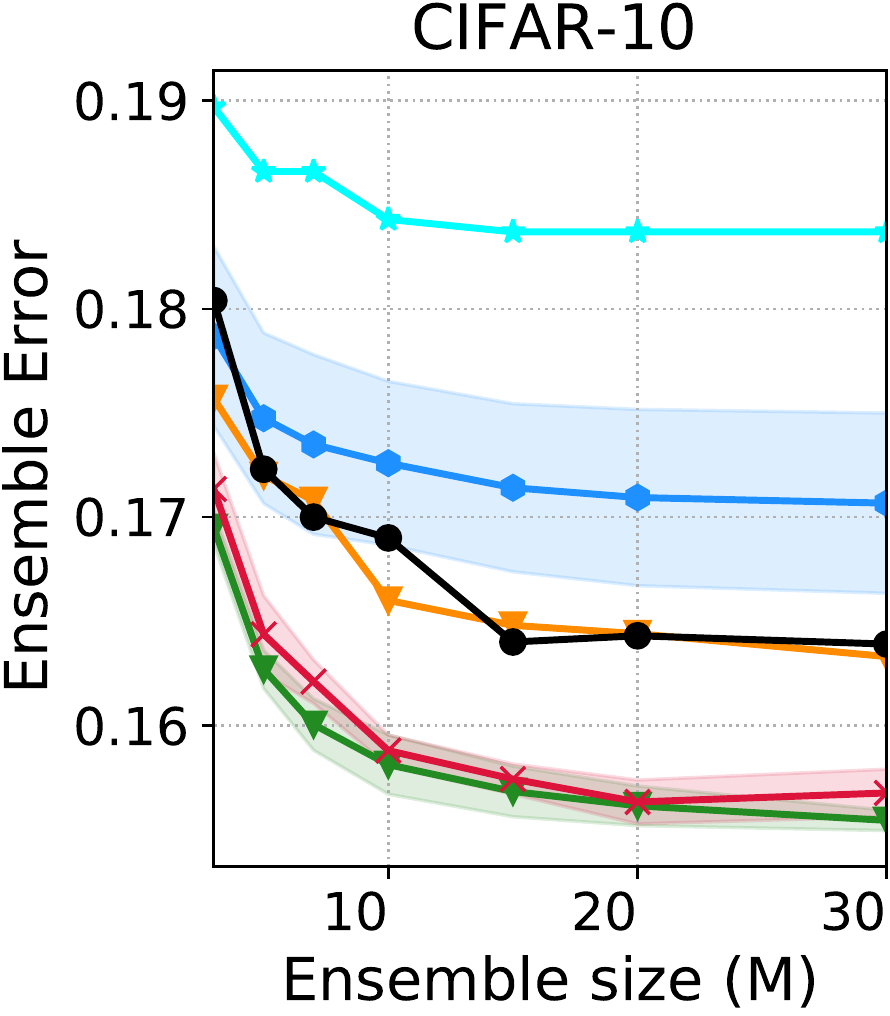}
        \includegraphics[width=.3\linewidth]{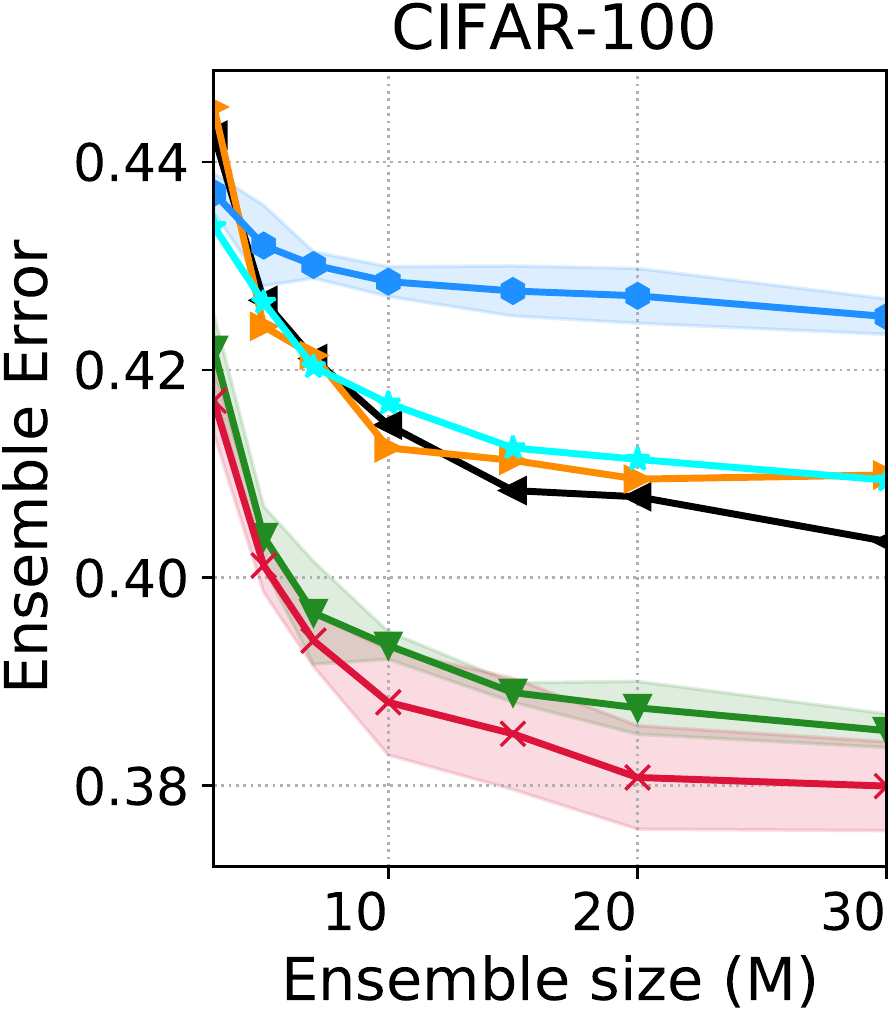}
        \includegraphics[width=.3\linewidth]{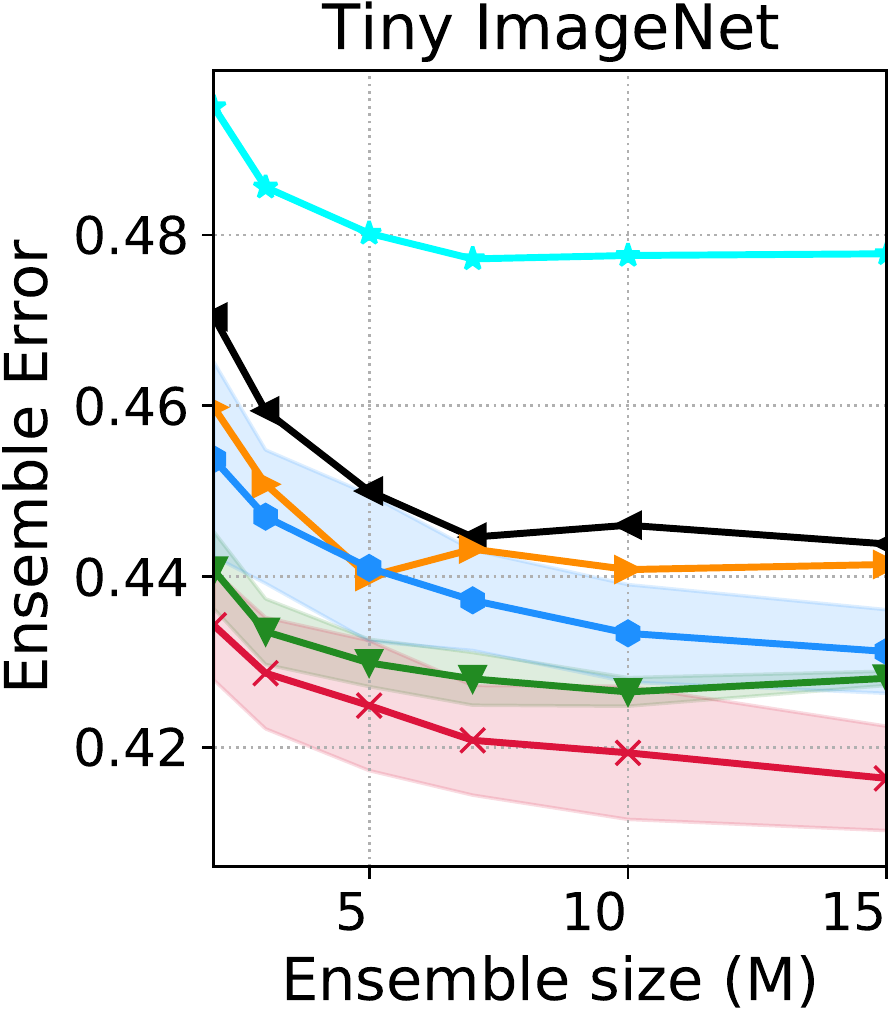}
        \subcaption{Data shift (severity 1)}
        \label{fig:test_error_M_shift1}
    \end{subfigure}\\
    \begin{subfigure}[t]{0.49\textwidth}
        \centering
        \includegraphics[width=.3\linewidth]{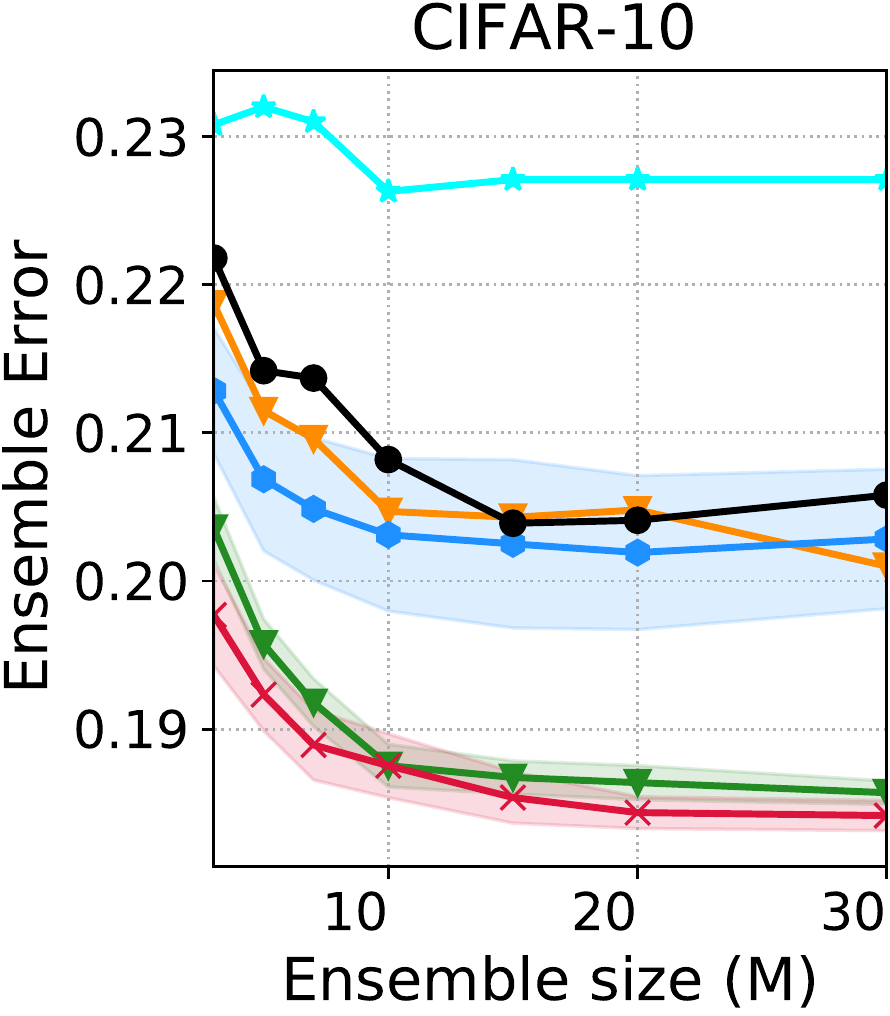}
        \includegraphics[width=.3\linewidth]{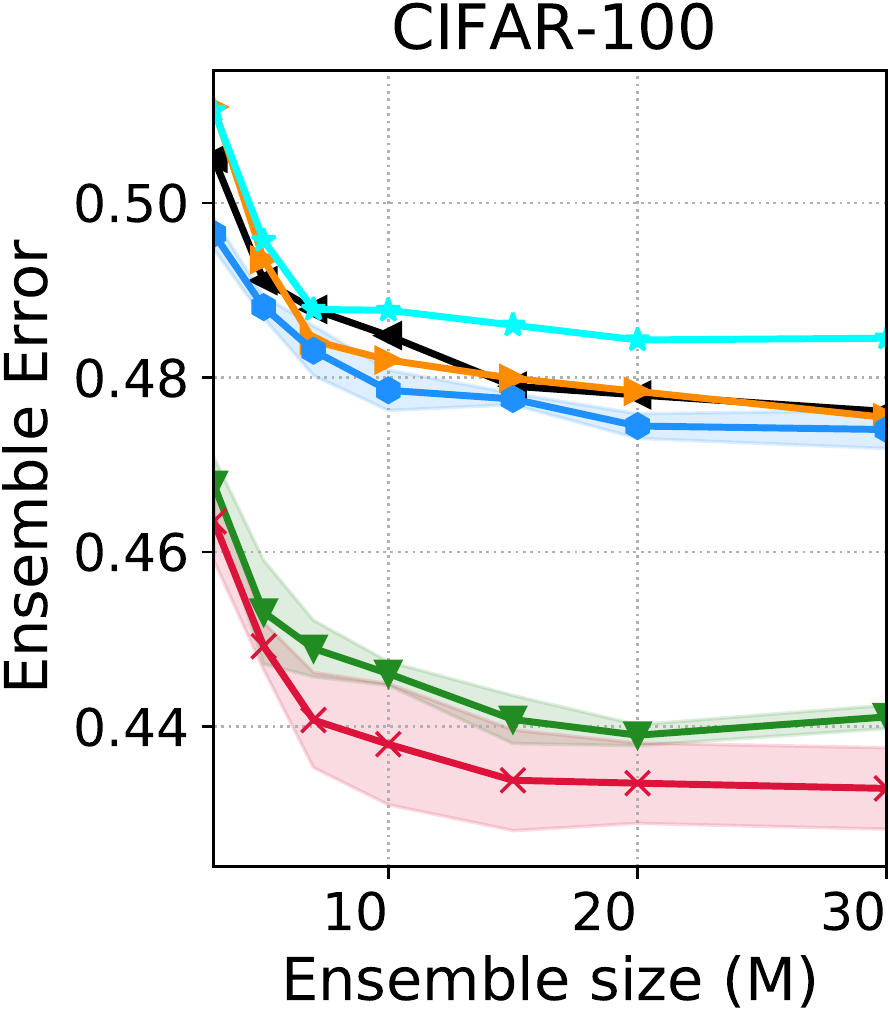}
        \includegraphics[width=.3\linewidth]{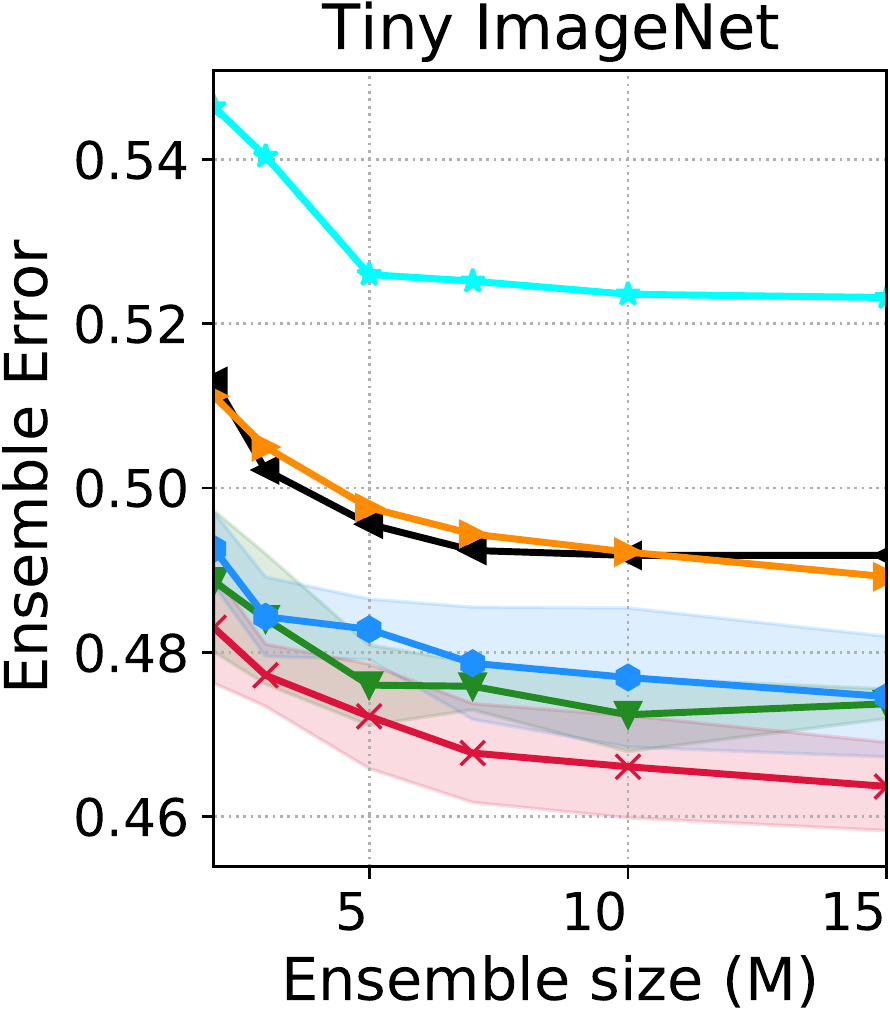}
        \subcaption{Data shift (severity 2)}
        \label{fig:test_error_M_shift2}
    \end{subfigure}
    \begin{subfigure}[t]{0.49\textwidth}
        \centering
        \includegraphics[width=.3\linewidth]{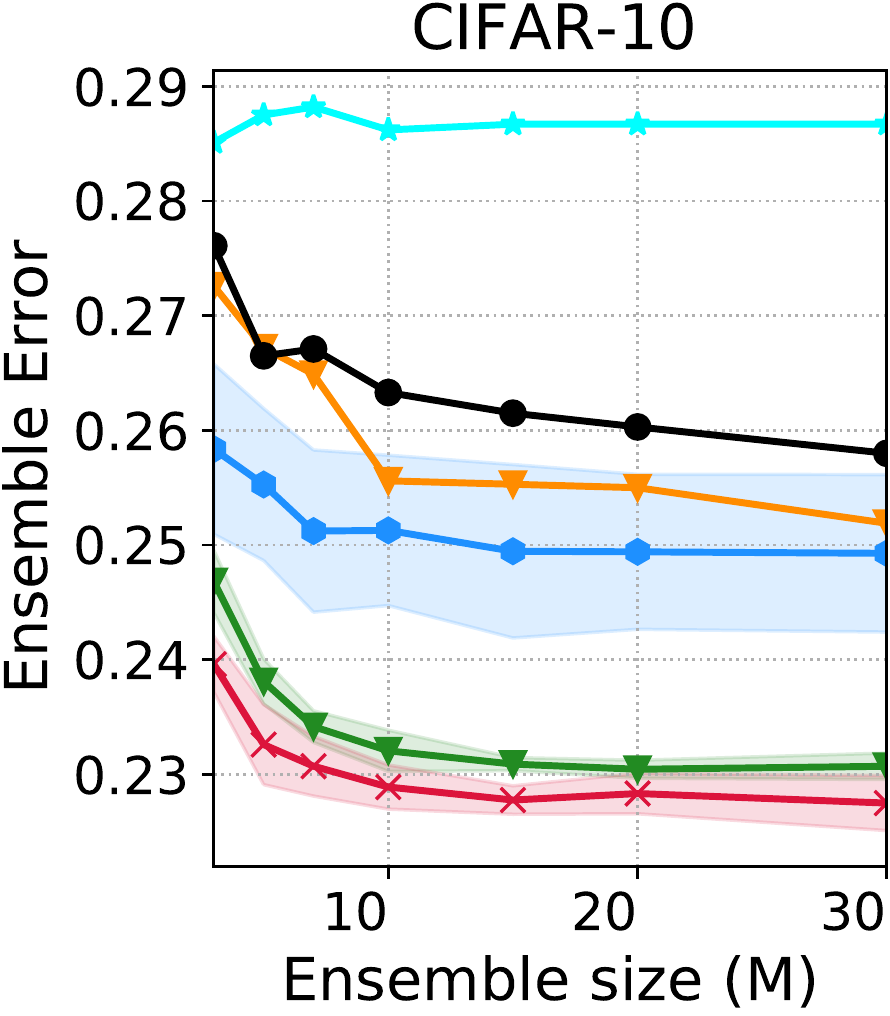}
        \includegraphics[width=.3\linewidth]{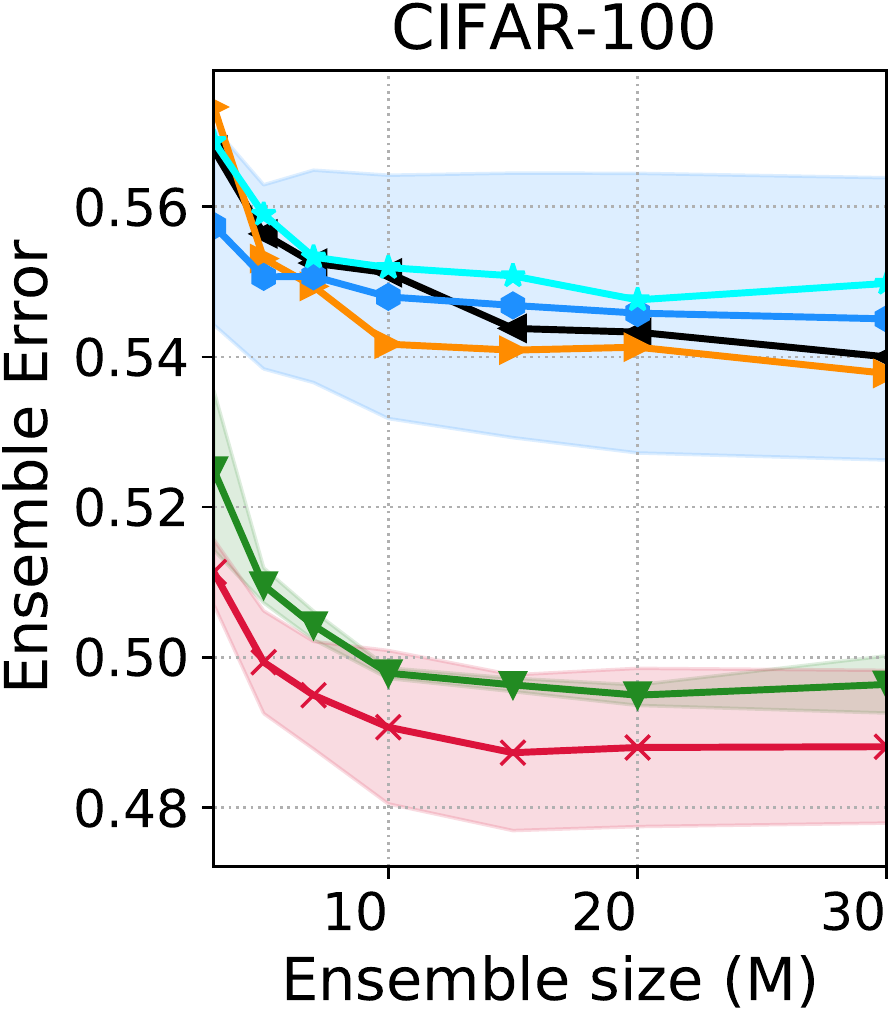}
        \includegraphics[width=.3\linewidth]{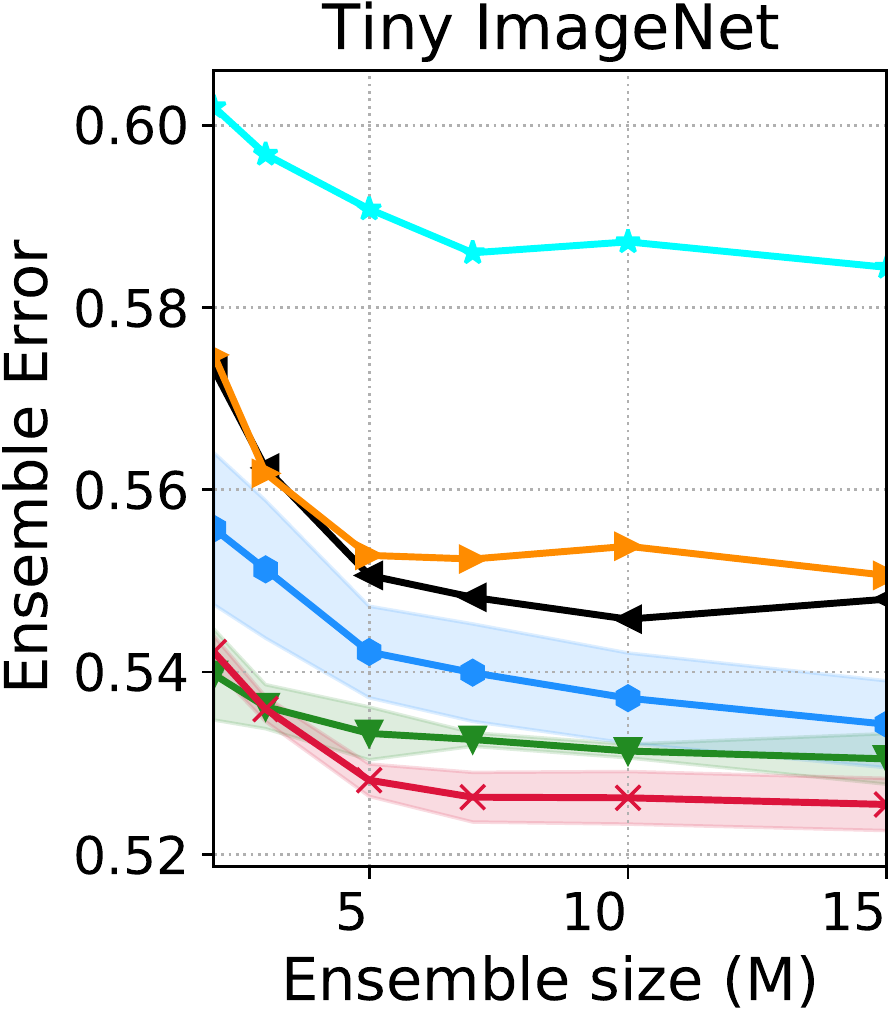}
        \subcaption{Data shift (severity 3)}
        \label{fig:test_error_M_shift3}
    \end{subfigure}\\ 
    \begin{subfigure}[t]{0.49\textwidth}
        \centering
        \includegraphics[width=.3\linewidth]{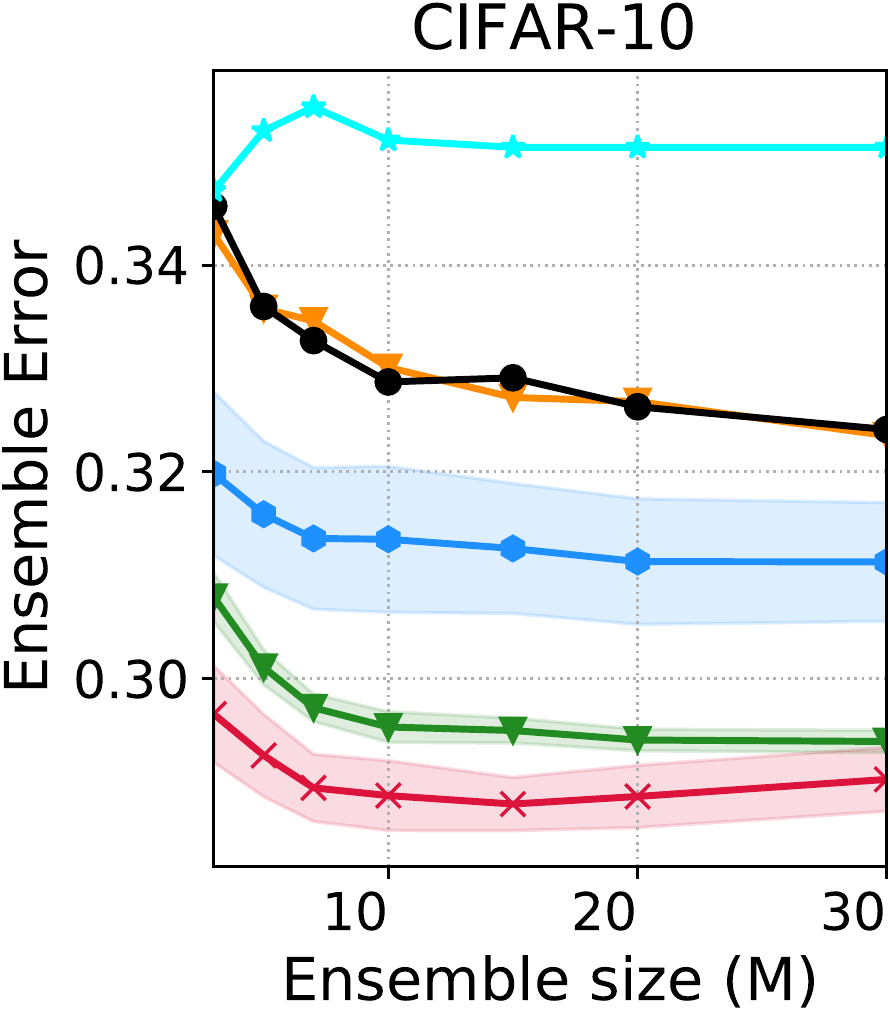}
        \includegraphics[width=.3\linewidth]{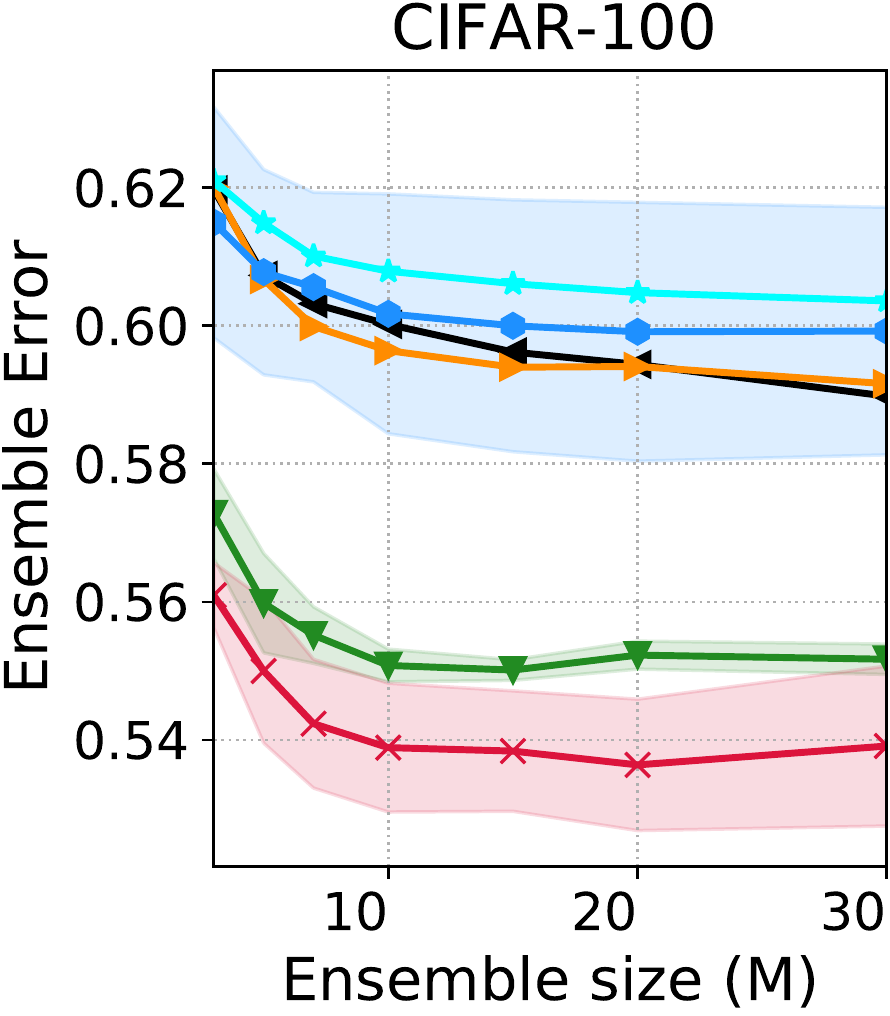}
        \includegraphics[width=.3\linewidth]{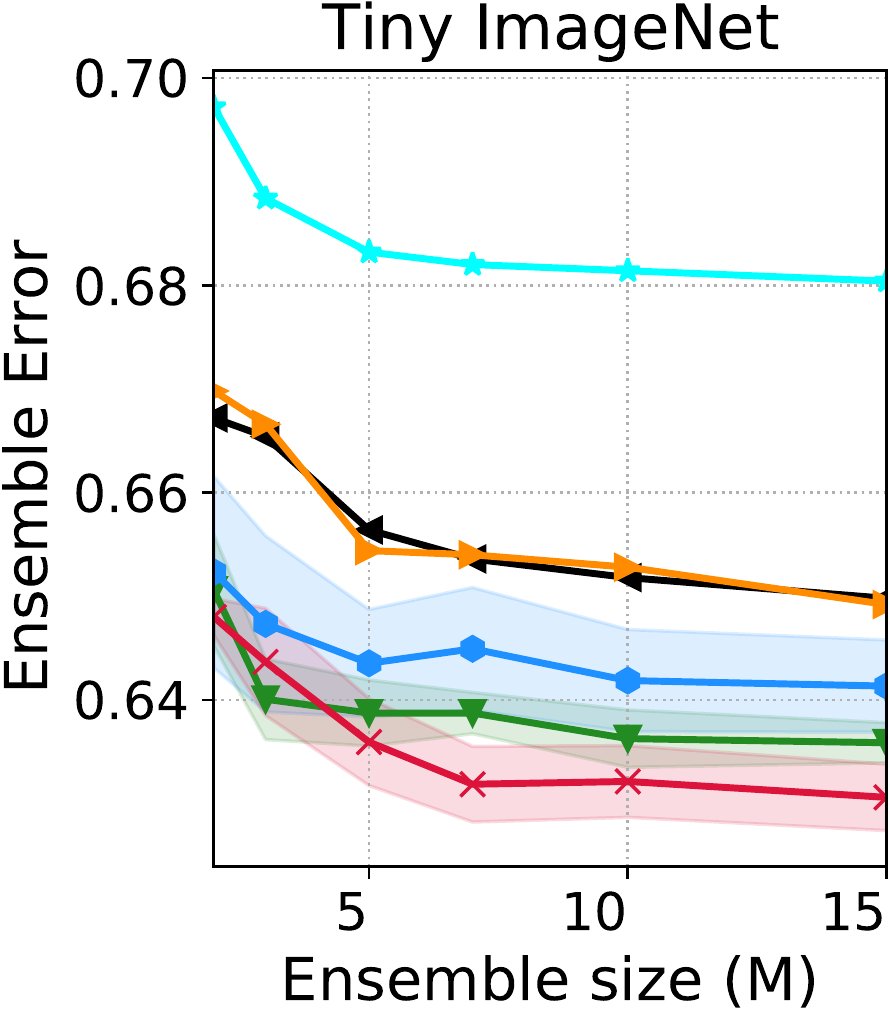}
        \subcaption{Data shift (severity 4)}
        \label{fig:test_error_M_shift4}
    \end{subfigure}
    \begin{subfigure}[t]{0.49\textwidth}
        \centering
        \includegraphics[width=.3\linewidth]{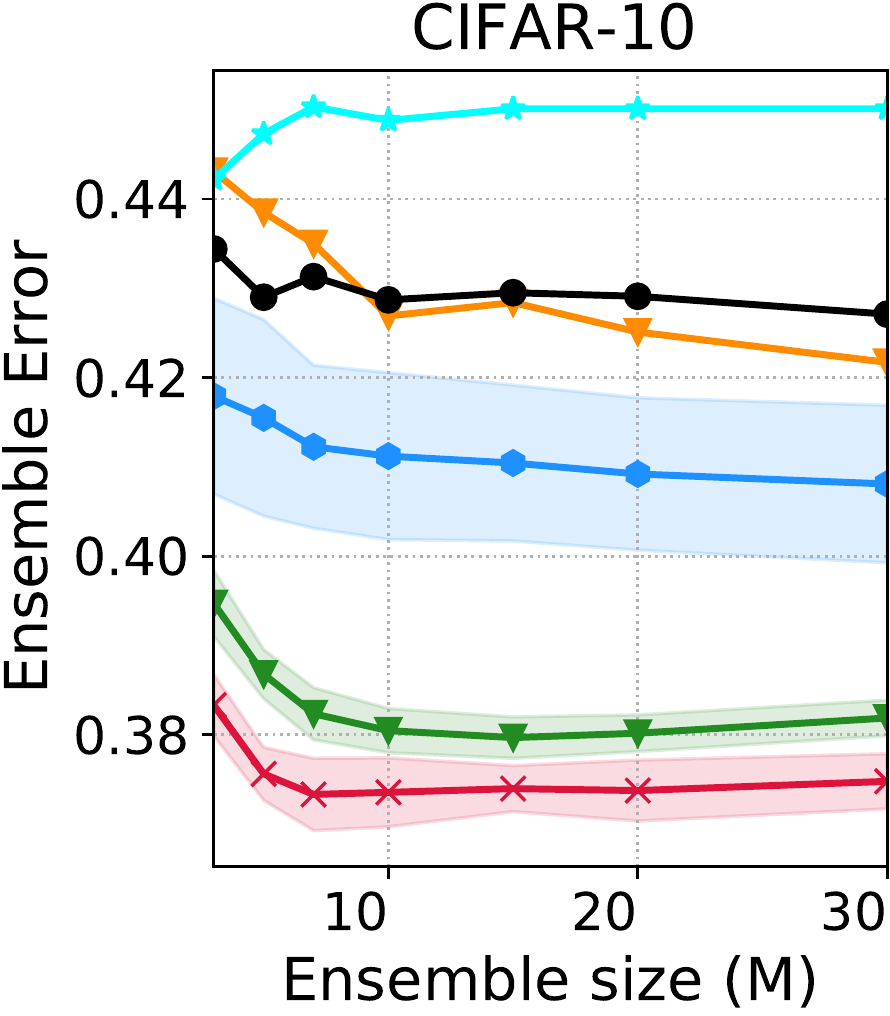}
        \includegraphics[width=.3\linewidth]{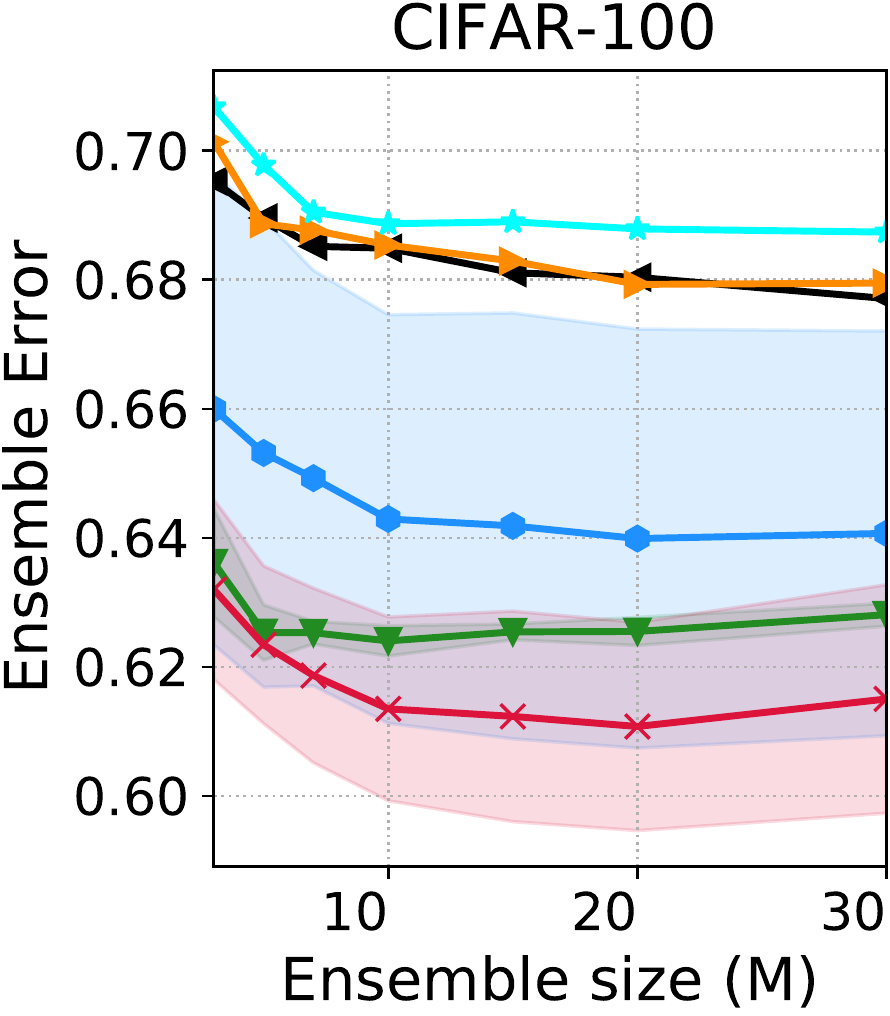}
        \includegraphics[width=.3\linewidth]{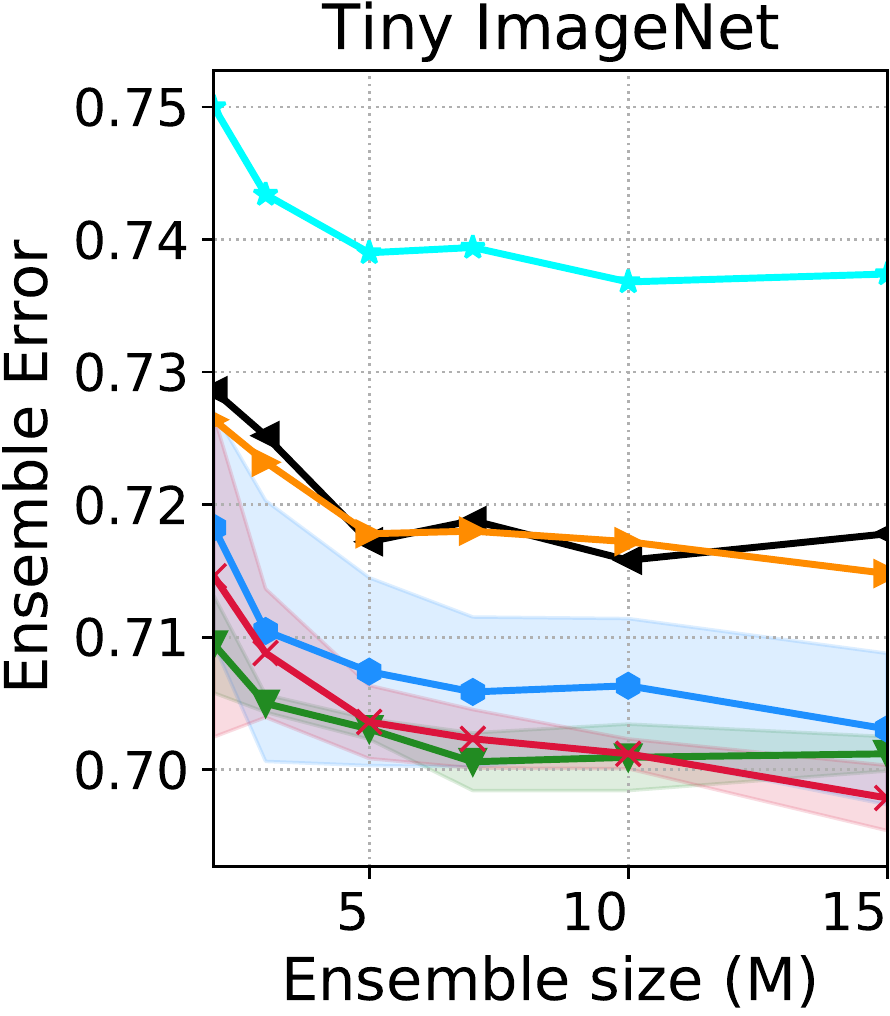}
        \subcaption{Data shift (severity 5)}
        \label{fig:test_error_M_shift5}
    \end{subfigure}
    
    \caption{Classification error rate (between 0-1) vs. ensemble size on DARTS search space.}
    \label{fig:test_error_M_other}
\end{figure*}

\begin{figure*}
    \centering
    \captionsetup[subfigure]{justification=centering}
    \begin{subfigure}[t]{0.31\textwidth}
        \centering
        \includegraphics[width=0.48\linewidth]{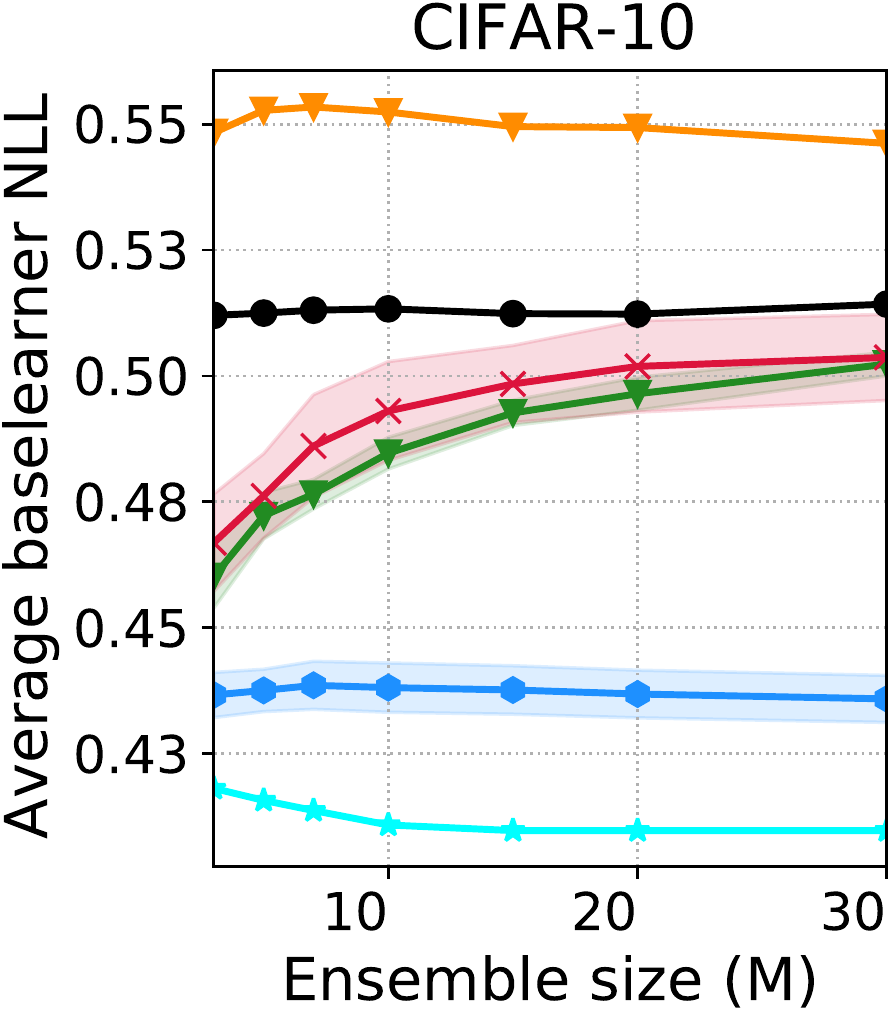}
        \includegraphics[width=0.48\linewidth]{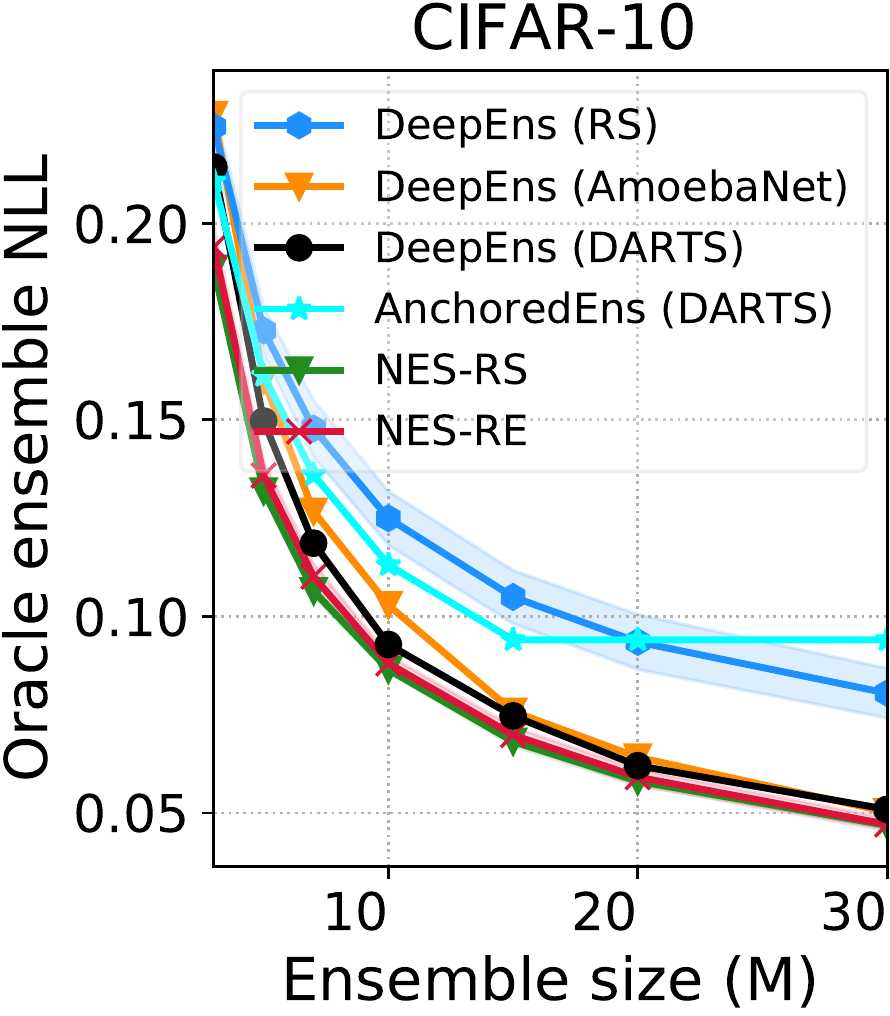}
        \subcaption{No data shift}
        \label{fig:test_avg_oracle_M_shift0-c10}
    \end{subfigure}%
    ~\hspace{.1cm}
    \begin{subfigure}[t]{0.31\textwidth}
        \centering
        \includegraphics[width=0.48\linewidth]{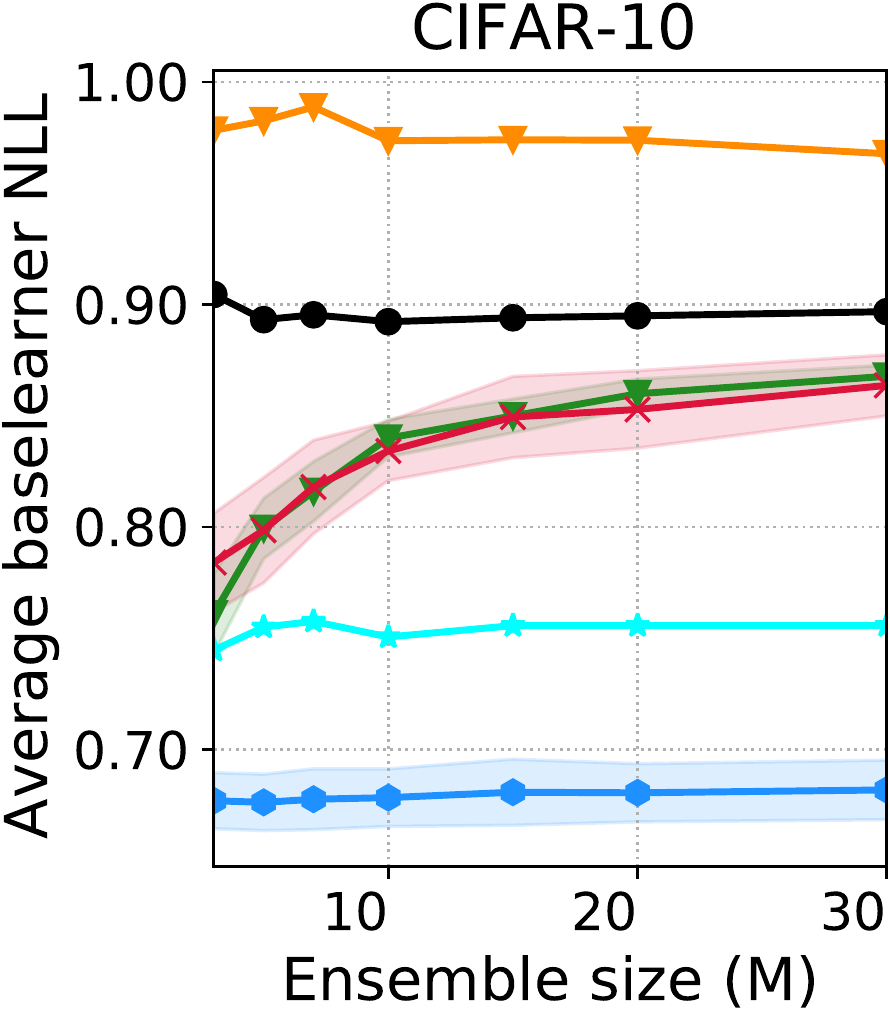}
        \includegraphics[width=0.48\linewidth]{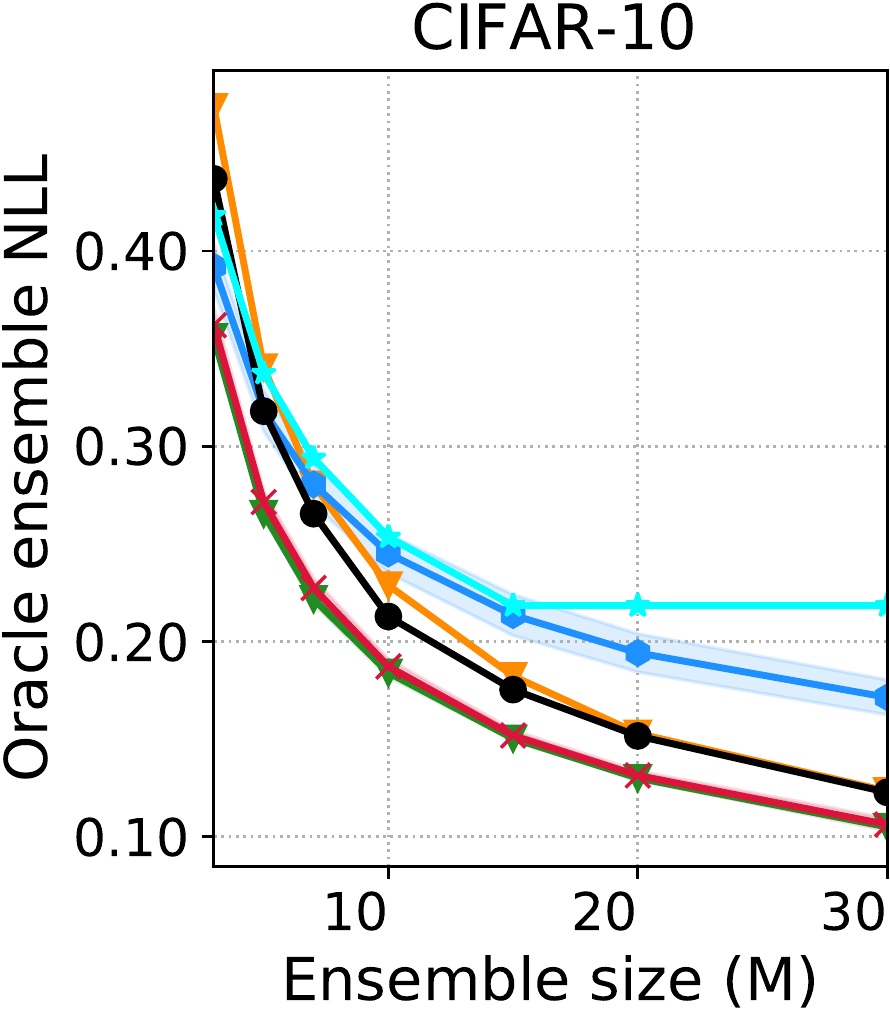}
        \subcaption{Data shift (severity 1)}
        \label{fig:test_avg_oracle_M_shift1}
    \end{subfigure}
    ~\hspace{.1cm}
    \begin{subfigure}[t]{0.31\textwidth}
        \centering
        \includegraphics[width=0.48\linewidth]{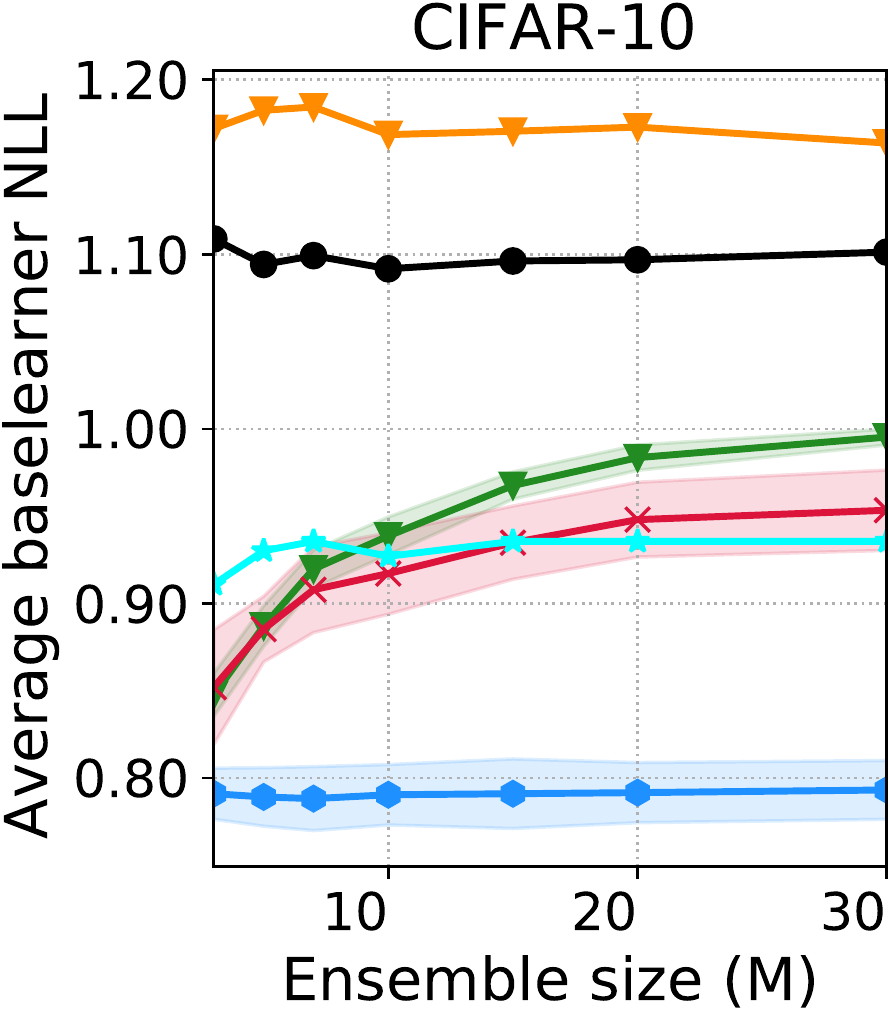}
        \includegraphics[width=0.48\linewidth]{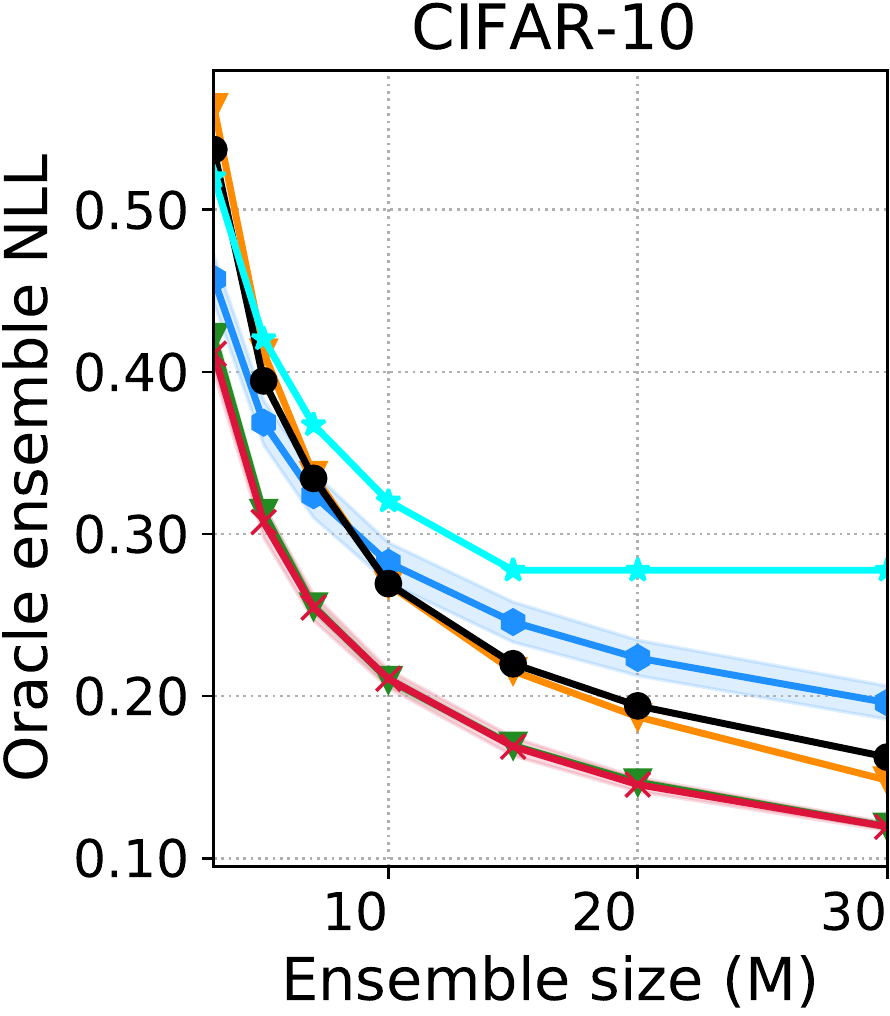}
        \subcaption{Data shift (severity 2)}
        \label{fig:test_avg_oracle_M_shift2}
    \end{subfigure} \\%
    \begin{subfigure}[t]{0.31\textwidth}
        \centering
        \includegraphics[width=.48\linewidth]{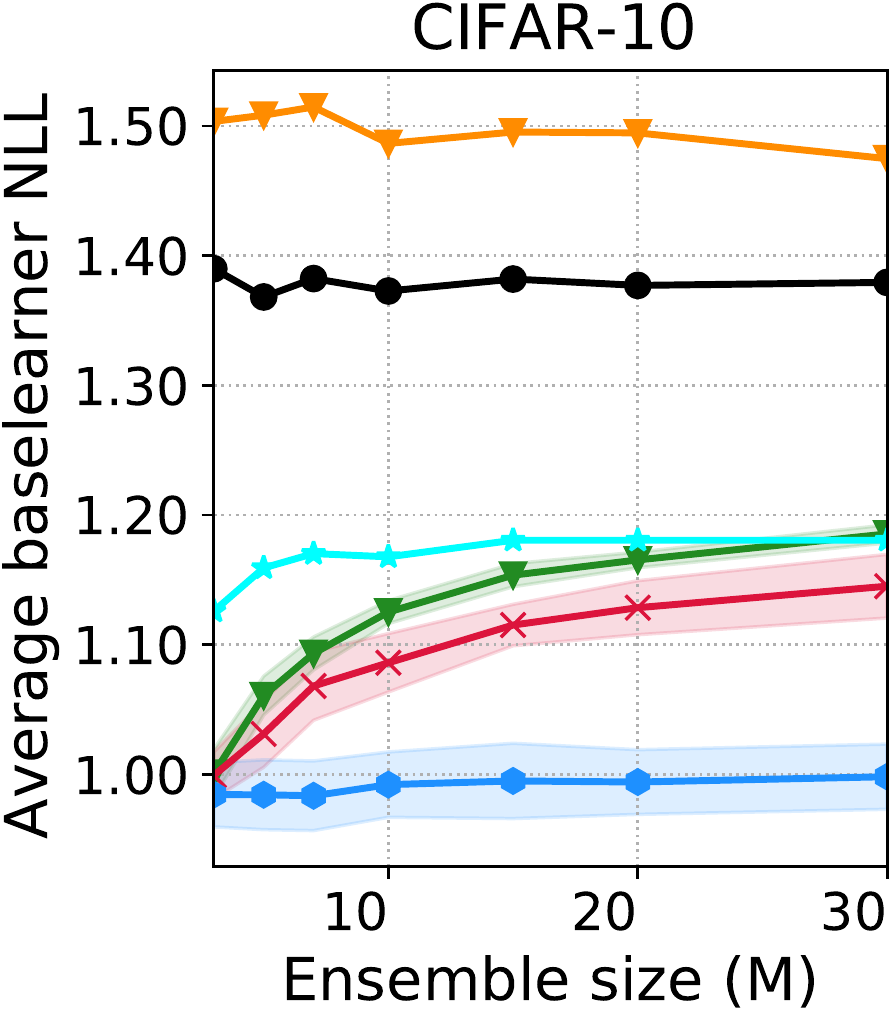}
        \includegraphics[width=.48\linewidth]{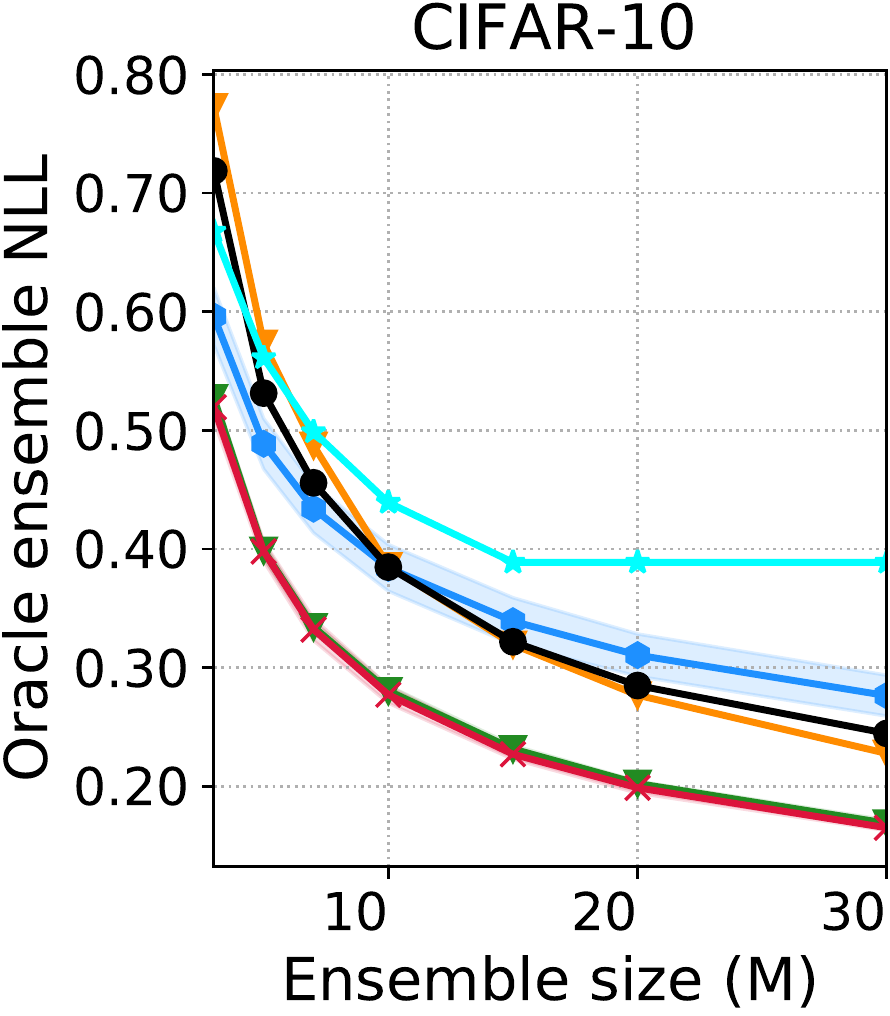}
        \subcaption{Data shift (severity 3)}
        \label{fig:test_avg_oracle_M_shift3}
    \end{subfigure}
    ~\hspace{.1cm}
    \begin{subfigure}[t]{0.31\textwidth}
        \centering
        \includegraphics[width=0.48\linewidth]{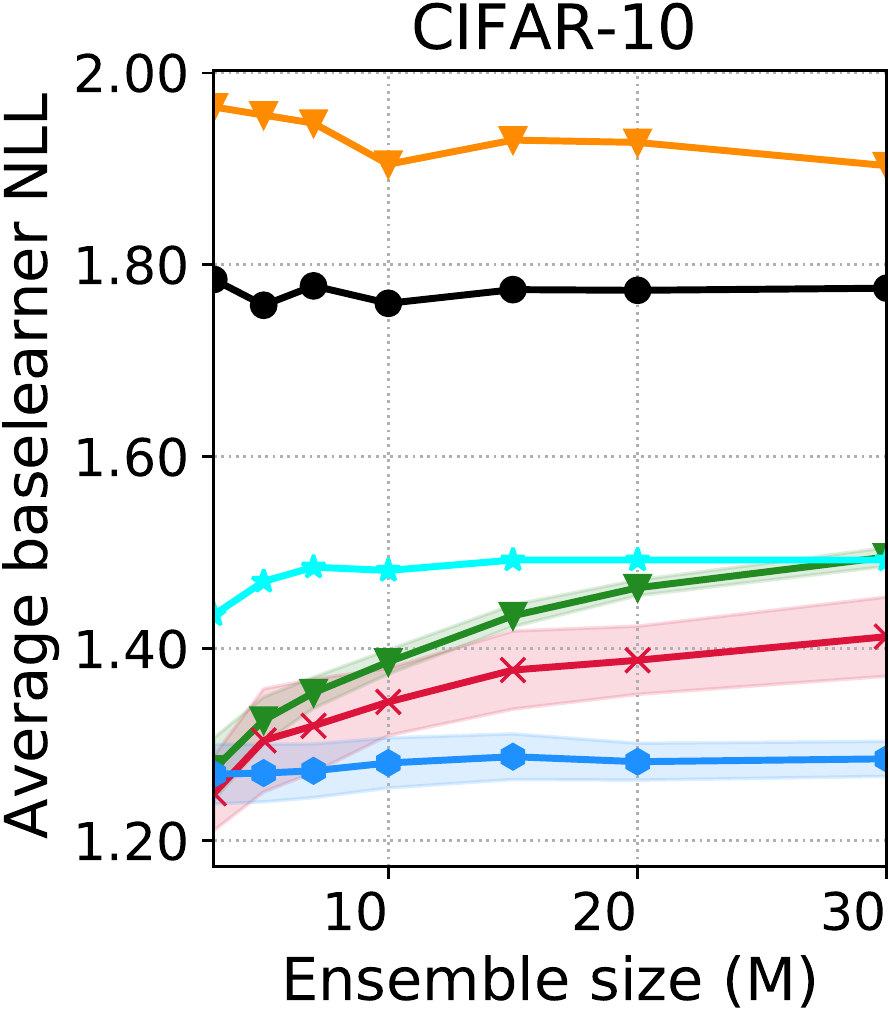}
        \includegraphics[width=0.48\linewidth]{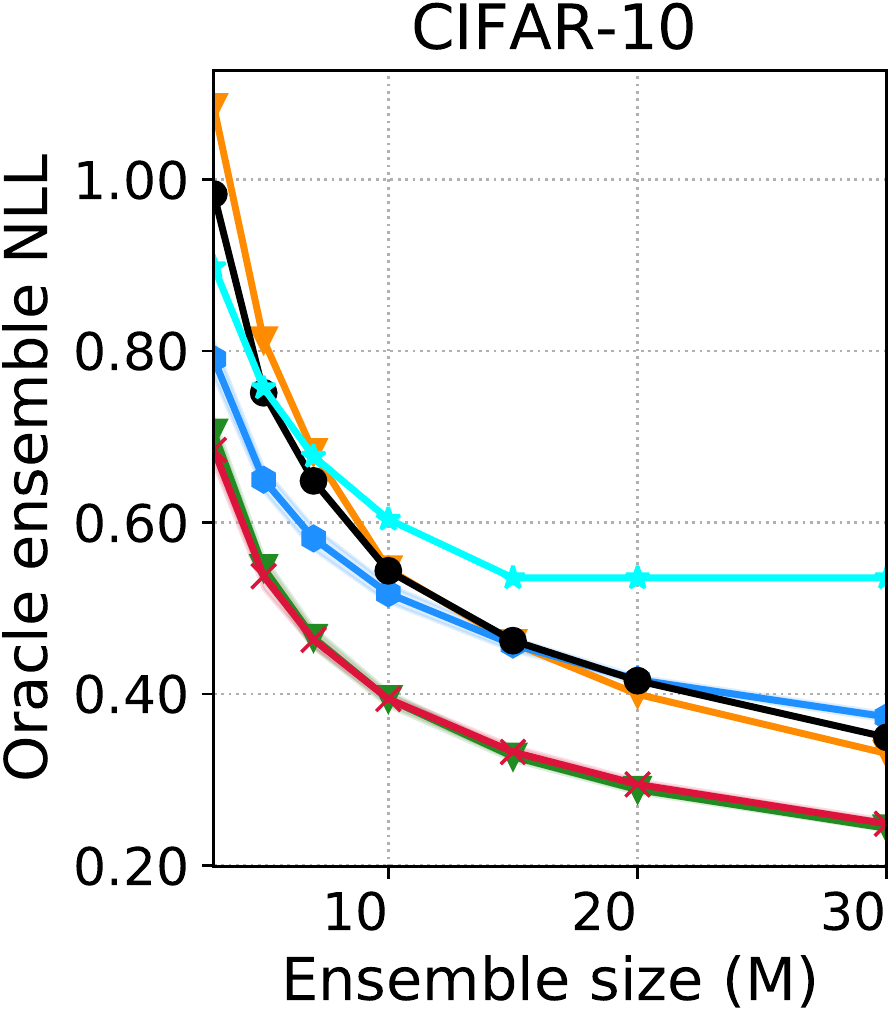}
        \subcaption{Data shift (severity 4)}
        \label{fig:test_avg_oracle_M_shift4}
    \end{subfigure}%
    ~\hspace{.1cm}
    \begin{subfigure}[t]{0.31\textwidth}
        \centering
        \includegraphics[width=0.48\linewidth]{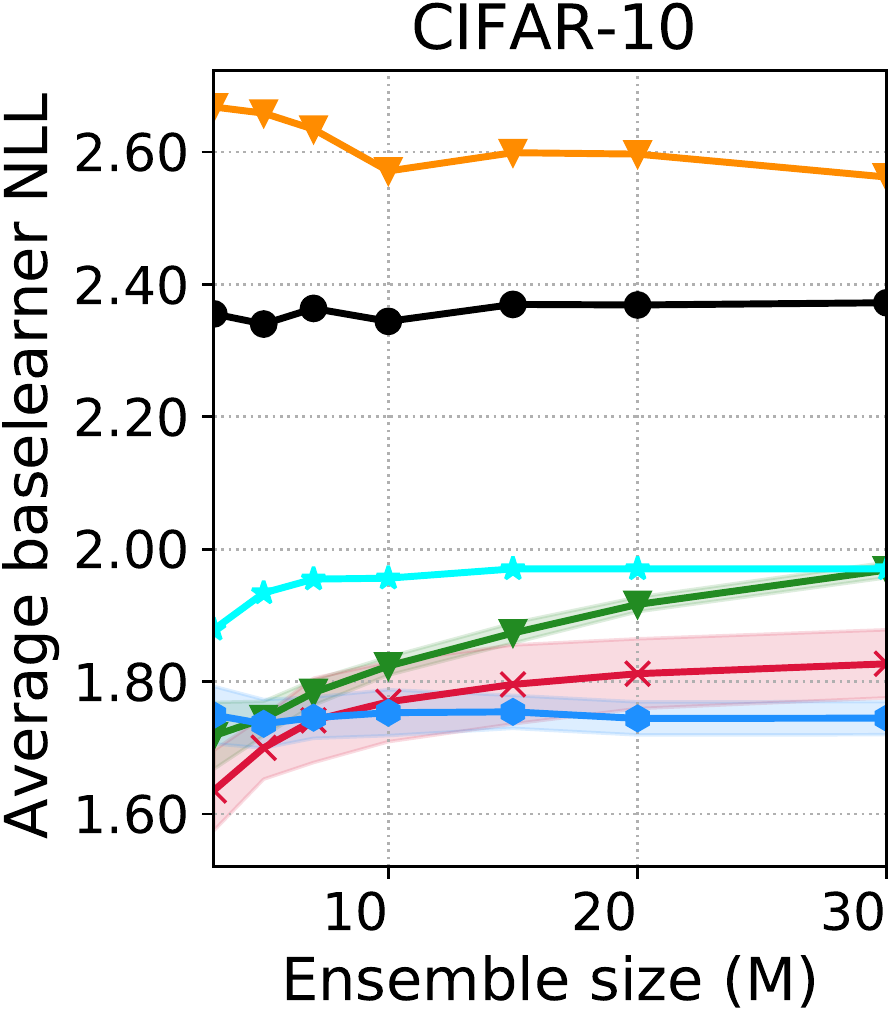}
        \includegraphics[width=0.48\linewidth]{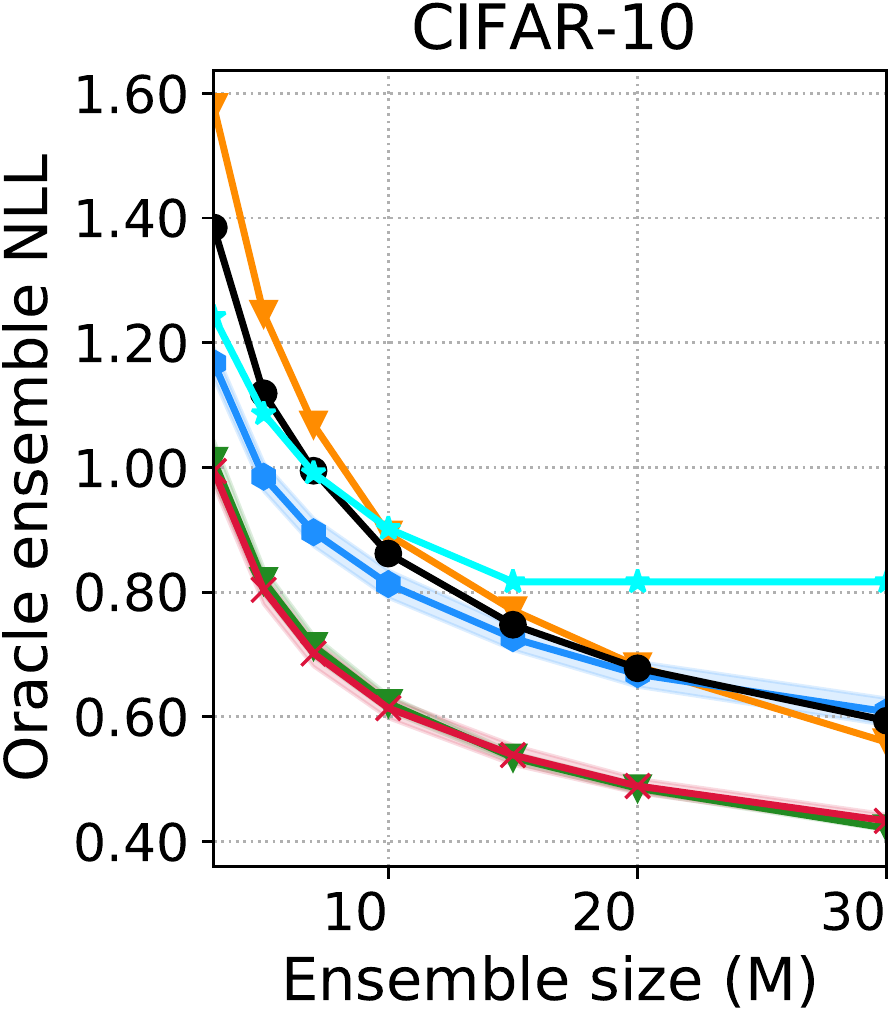}
        \subcaption{Data shift (severity 5)}
        \label{fig:test_avg_oracle_M_shift5}
    \end{subfigure}
    
    \caption{Average base learner and oracle ensemble NLL across ensemble sizes and shift severities on CIFAR-10 over DARTS search space.}
    \label{fig:test_avg_oracle_M_other-c10}
\end{figure*}

\vspace{3cm}

\begin{figure*}
    \centering
    \captionsetup[subfigure]{justification=centering}
    \begin{subfigure}[t]{0.31\textwidth}
        \centering
        \includegraphics[width=0.48\linewidth]{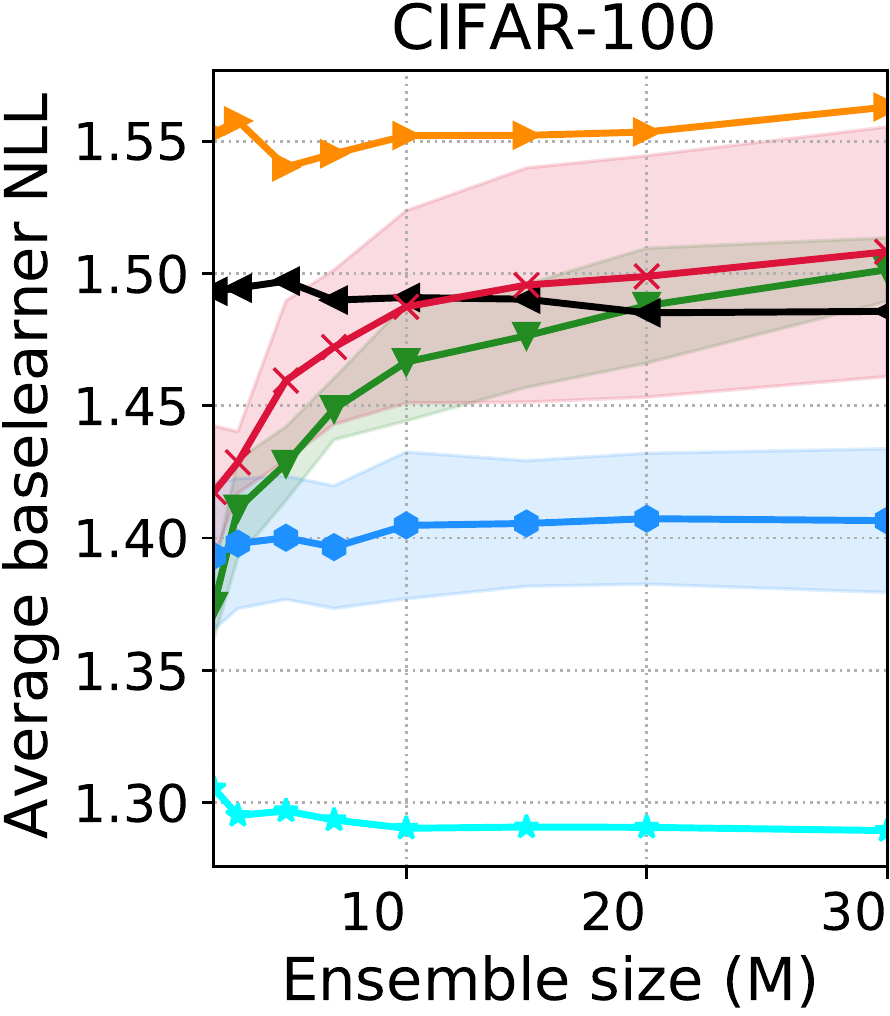}
        \includegraphics[width=0.48\linewidth]{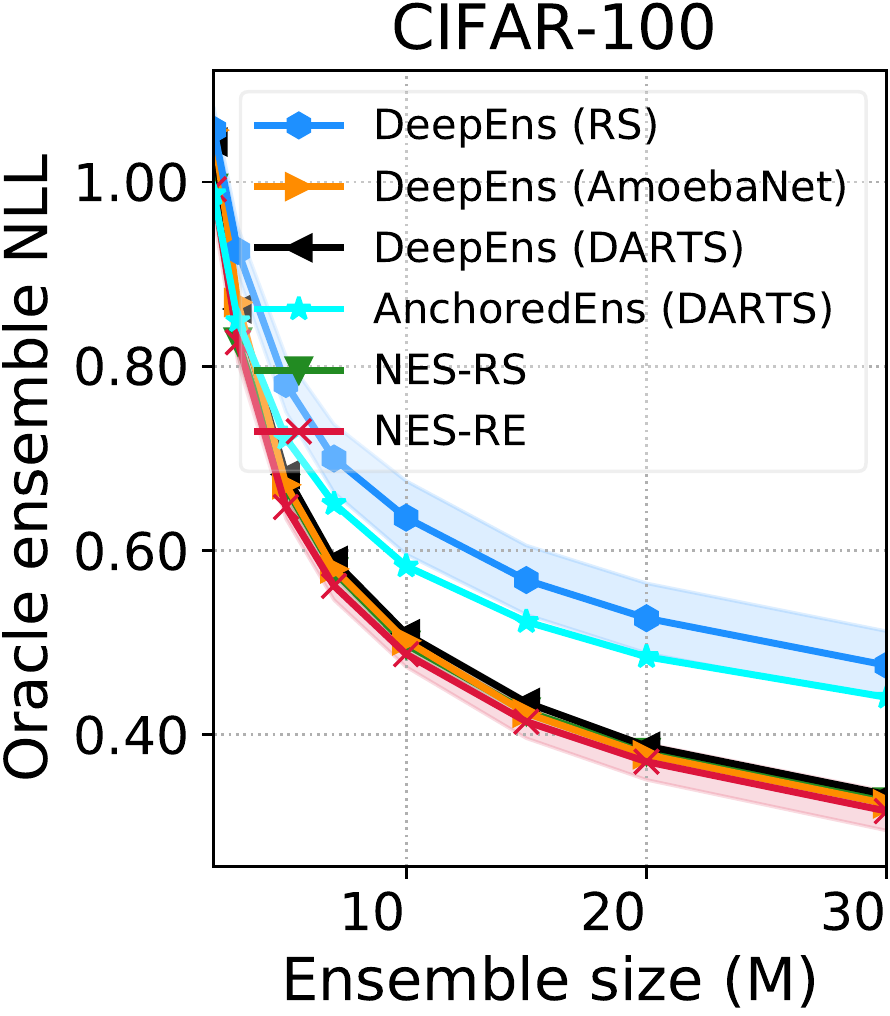}
        \subcaption{No data shift}
    \end{subfigure}%
    ~\hspace{.1cm}
    \begin{subfigure}[t]{0.31\textwidth}
        \centering
        \includegraphics[width=0.48\linewidth]{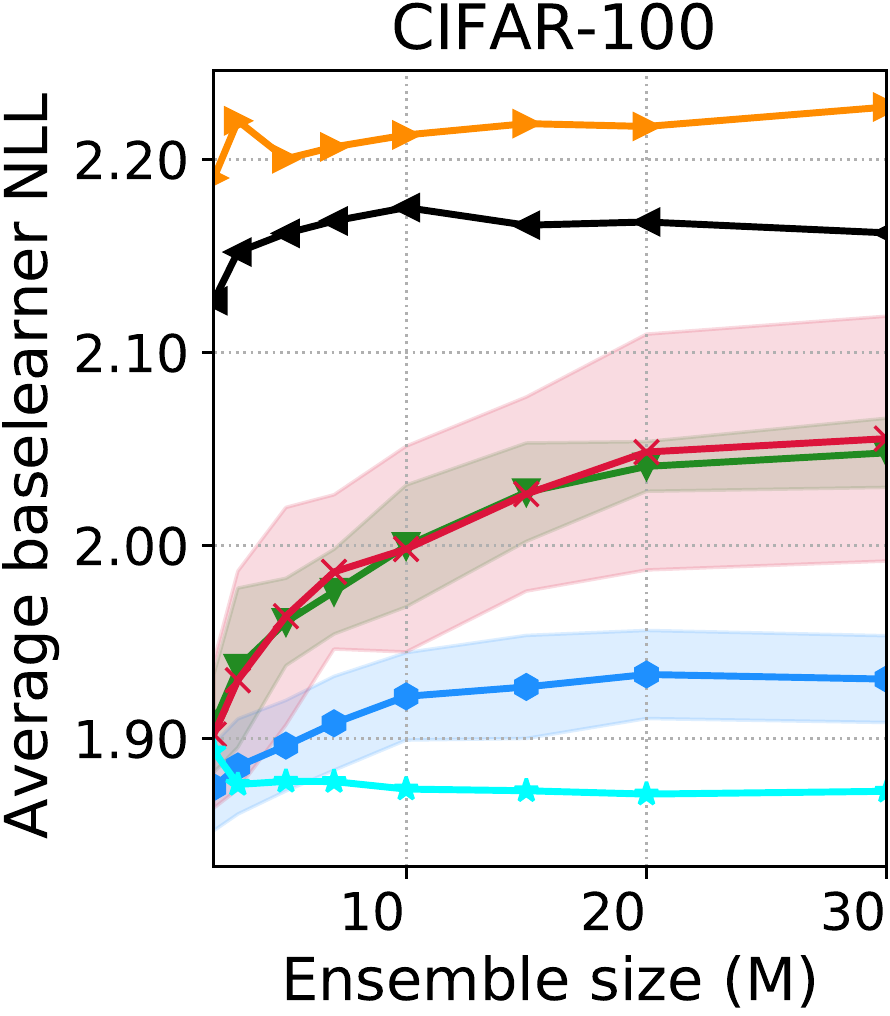}
        \includegraphics[width=0.48\linewidth]{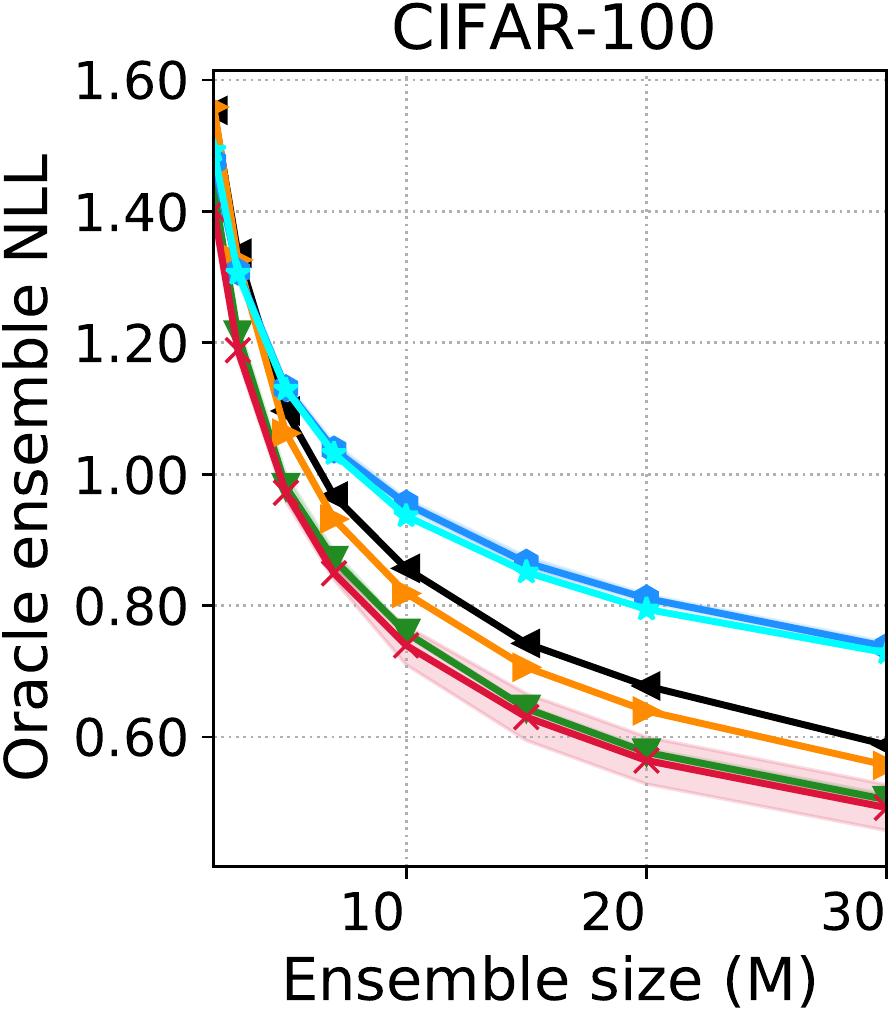}
        \subcaption{Data shift (severity 1)}
    \end{subfigure}
    ~\hspace{.1cm}
    \begin{subfigure}[t]{0.31\textwidth}
        \centering
        \includegraphics[width=0.48\linewidth]{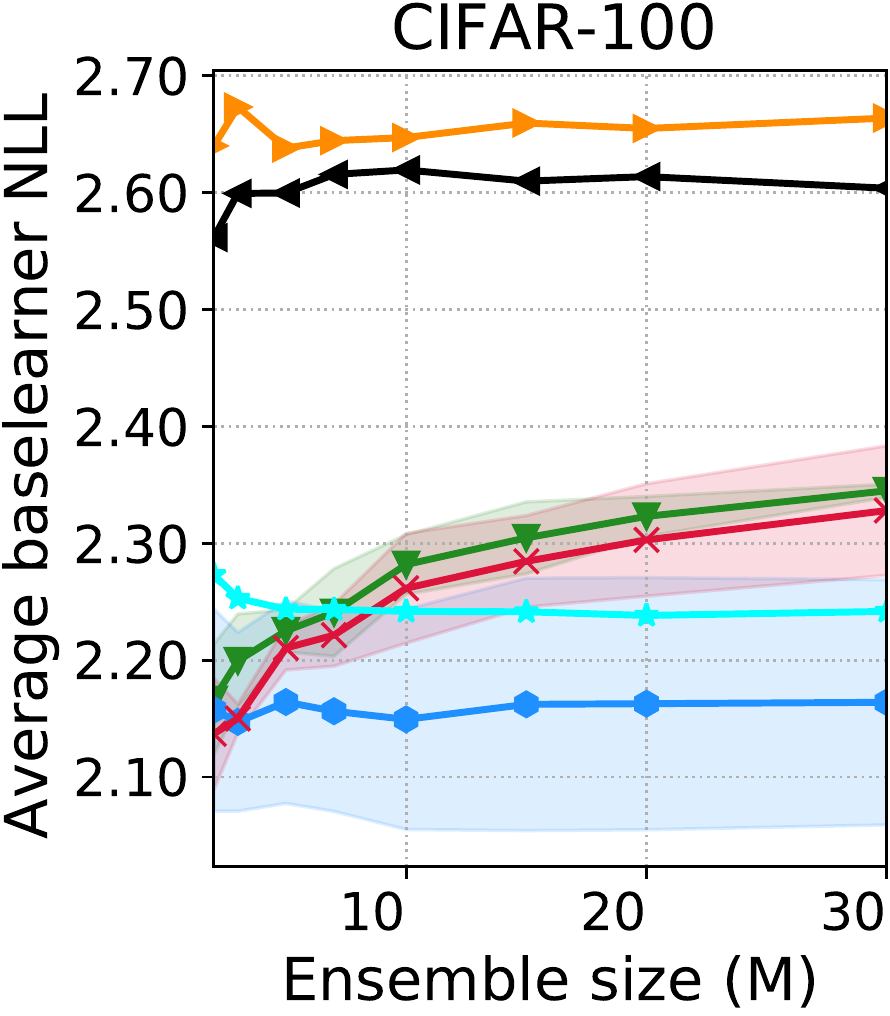}
        \includegraphics[width=0.48\linewidth]{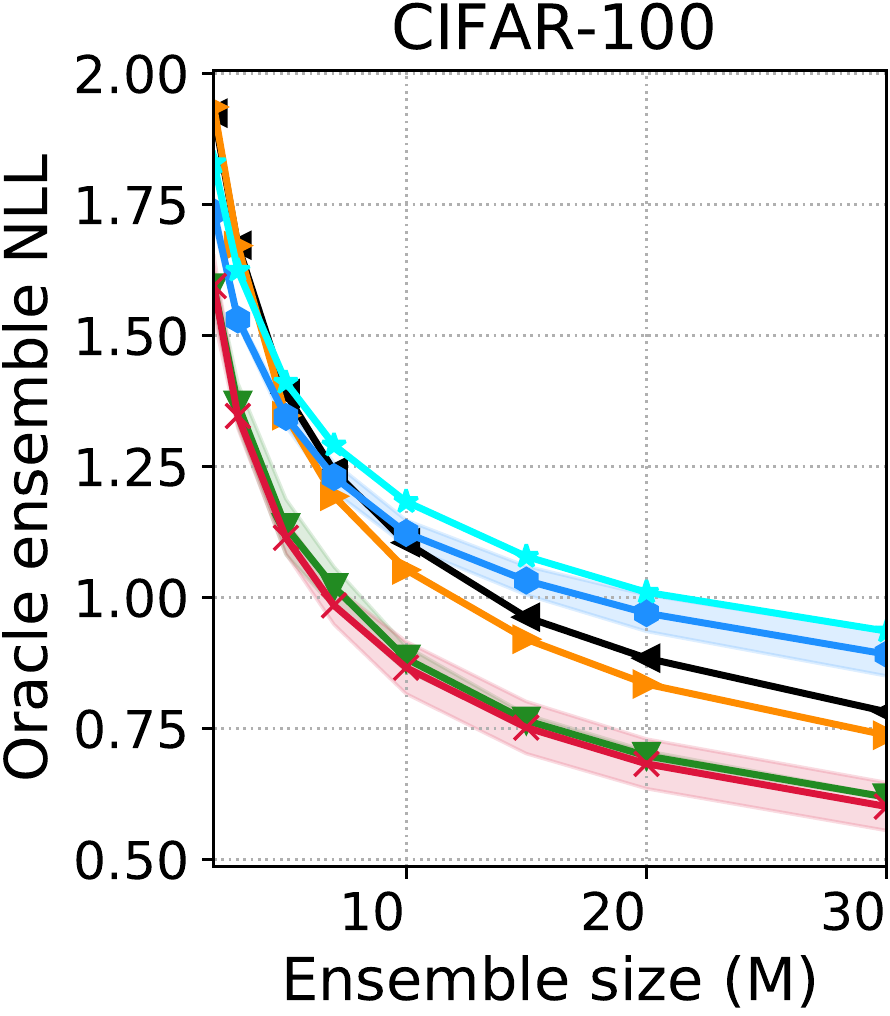}
        \subcaption{Data shift (severity 2)}
    \end{subfigure}\\%
    \begin{subfigure}[t]{0.31\textwidth}
        \centering
        \includegraphics[width=.48\linewidth]{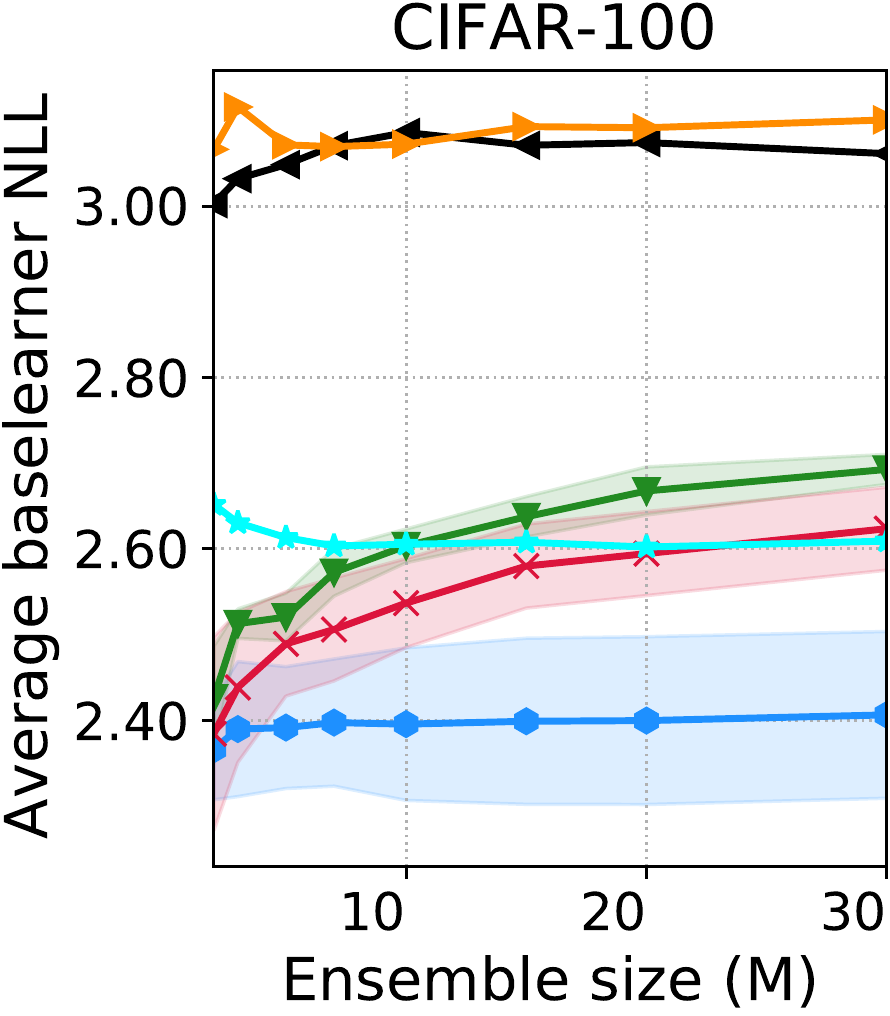}
        \includegraphics[width=.48\linewidth]{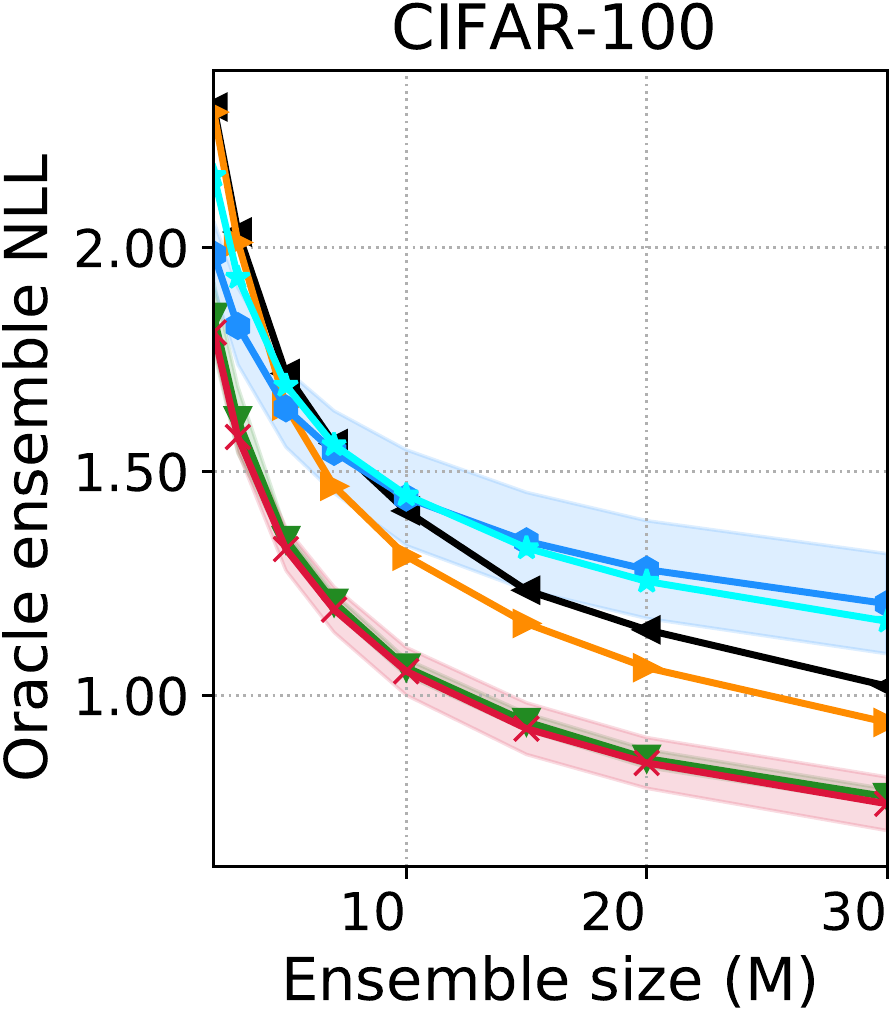}
        \subcaption{Data shift (severity 3)}
    \end{subfigure}
    ~\hspace{.1cm}
    \begin{subfigure}[t]{0.31\textwidth}
        \centering
        \includegraphics[width=0.48\linewidth]{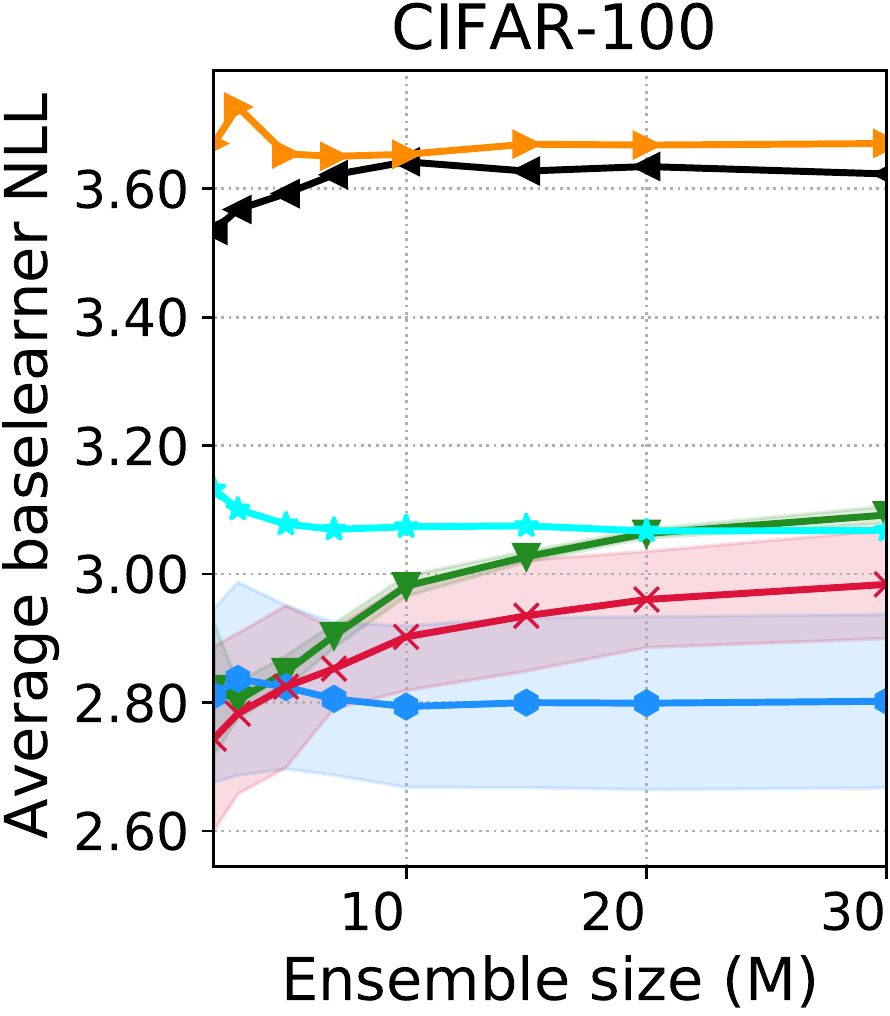}
        \includegraphics[width=0.48\linewidth]{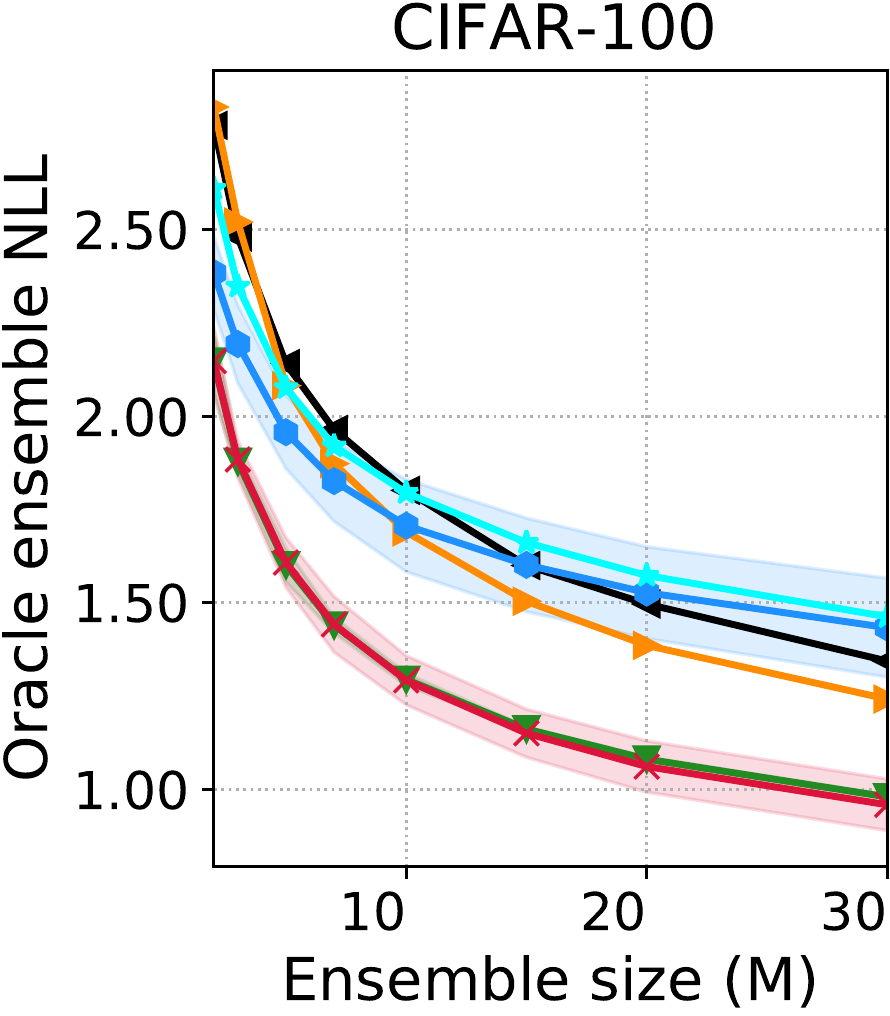}
        \subcaption{Data shift (severity 4)}
    \end{subfigure}%
    ~\hspace{.1cm}
    \begin{subfigure}[t]{0.31\textwidth}
        \centering
        \includegraphics[width=0.48\linewidth]{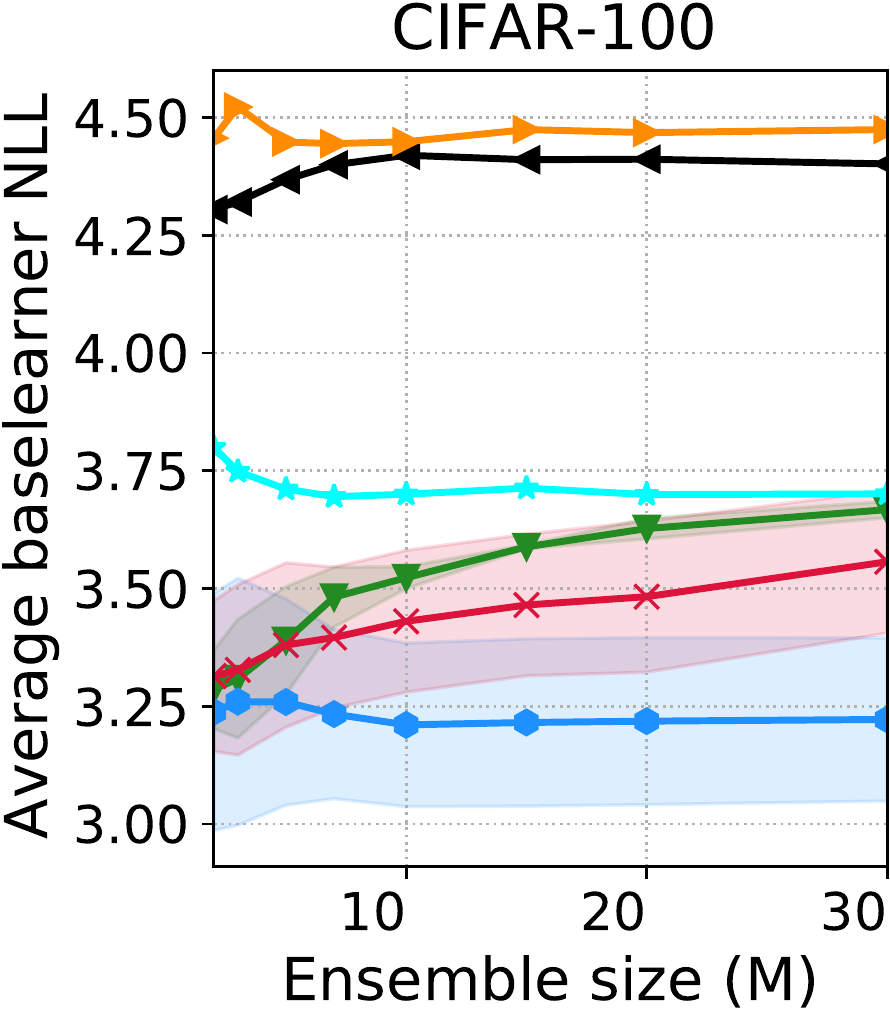}
        \includegraphics[width=0.48\linewidth]{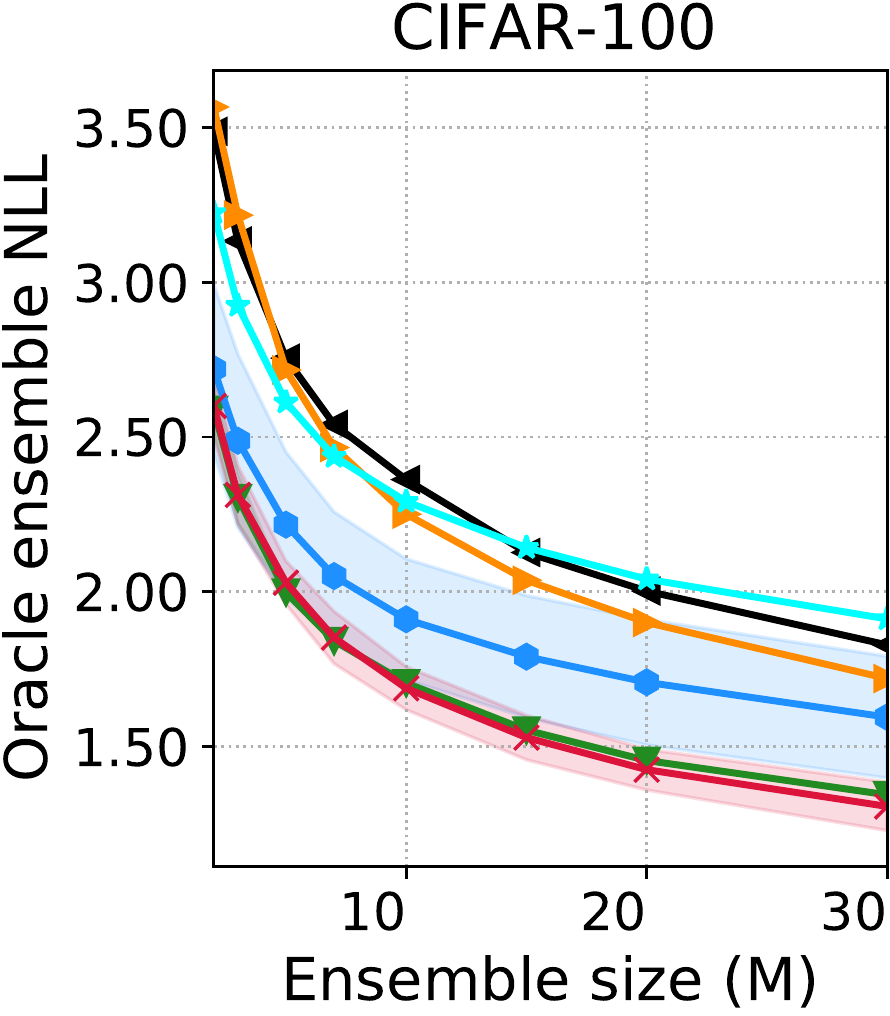}
        \subcaption{Data shift (severity 5)}
    \end{subfigure}
    
    \caption{Average base learner and oracle ensemble NLL across ensemble sizes and shift severities on CIFAR-100 over DARTS search space.}
    \label{fig:test_avg_oracle_M_other-c100}
\end{figure*}

\begin{figure*}
    \centering
    \captionsetup[subfigure]{justification=centering}
    \begin{subfigure}[t]{0.31\textwidth}
        \centering
        \includegraphics[width=0.48\linewidth]{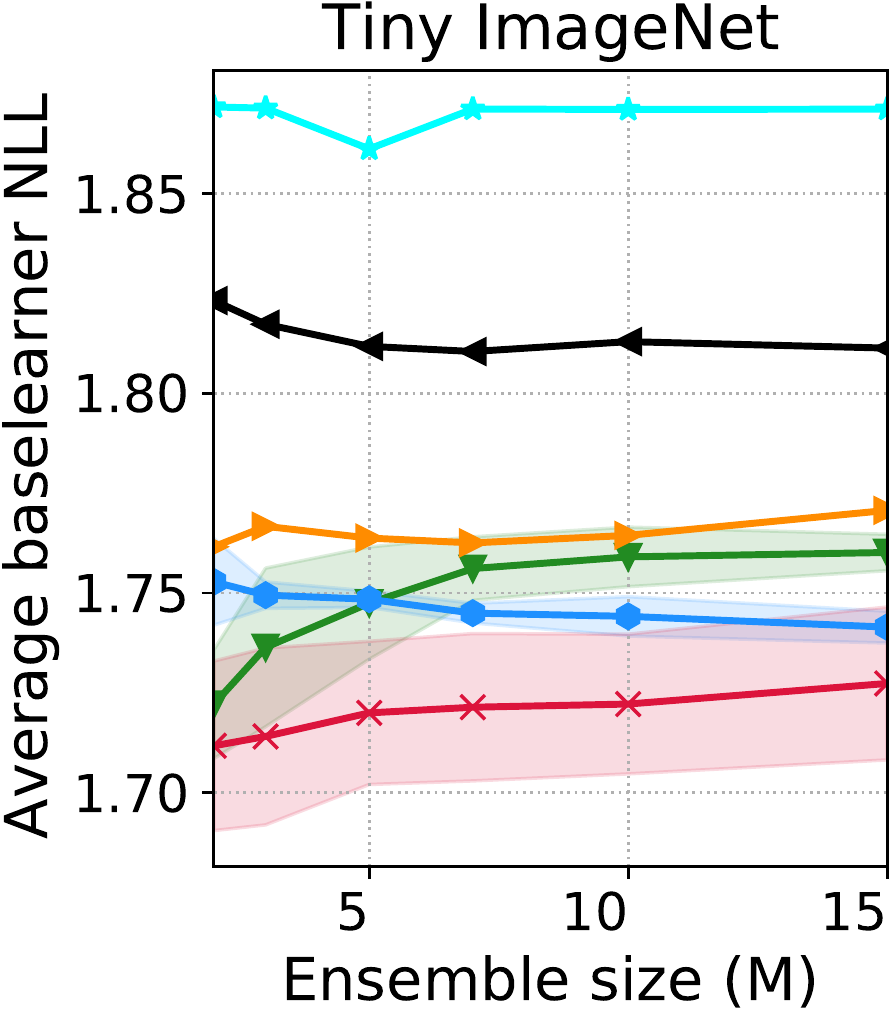}
        \includegraphics[width=0.48\linewidth]{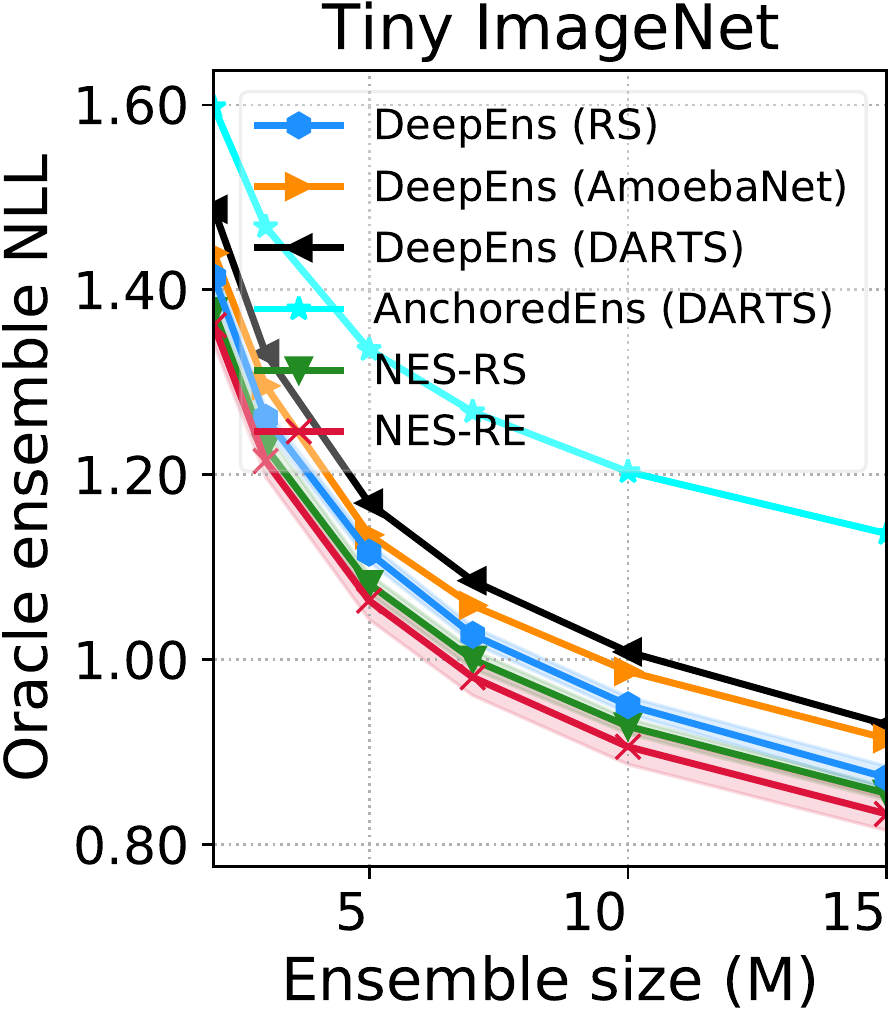}
        \subcaption{No data shift}
    \end{subfigure}%
    ~\hspace{.1cm}
    \begin{subfigure}[t]{0.31\textwidth}
        \centering
        \includegraphics[width=0.48\linewidth]{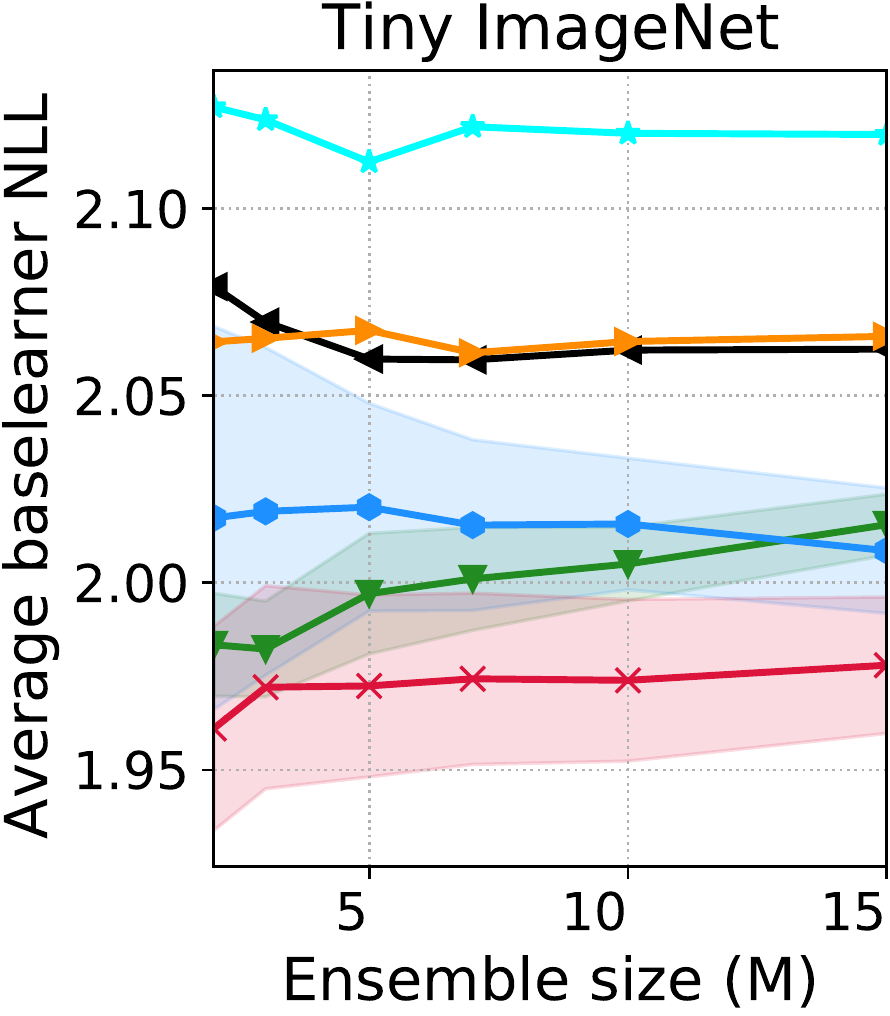}
        \includegraphics[width=0.48\linewidth]{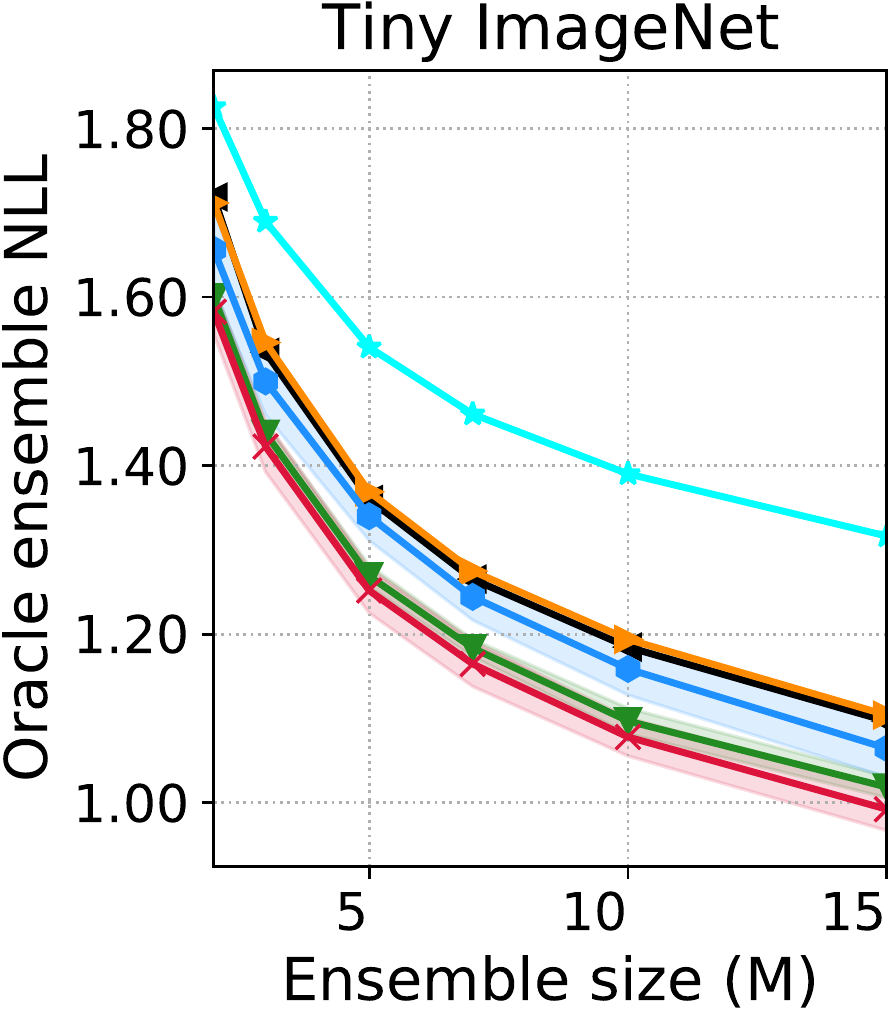}
        \subcaption{Data shift (severity 1)}
    \end{subfigure}
    ~\hspace{.1cm}
    \begin{subfigure}[t]{0.31\textwidth}
        \centering
        \includegraphics[width=0.48\linewidth]{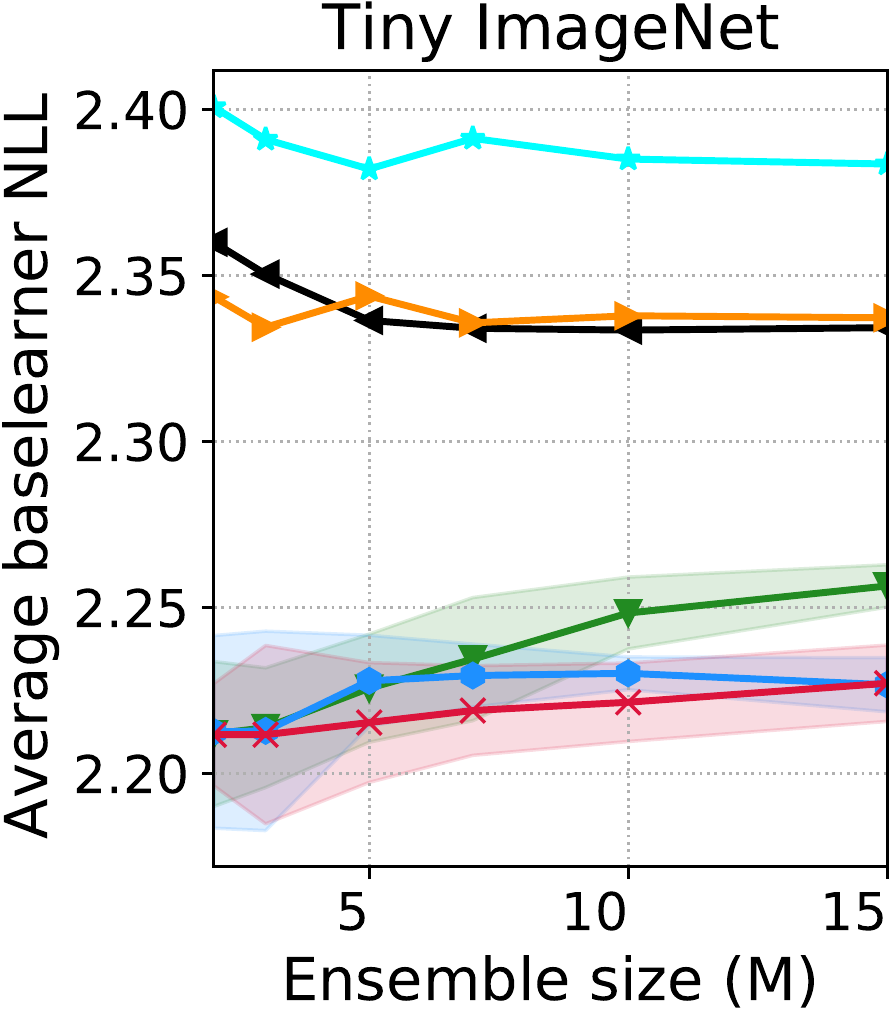}
        \includegraphics[width=0.48\linewidth]{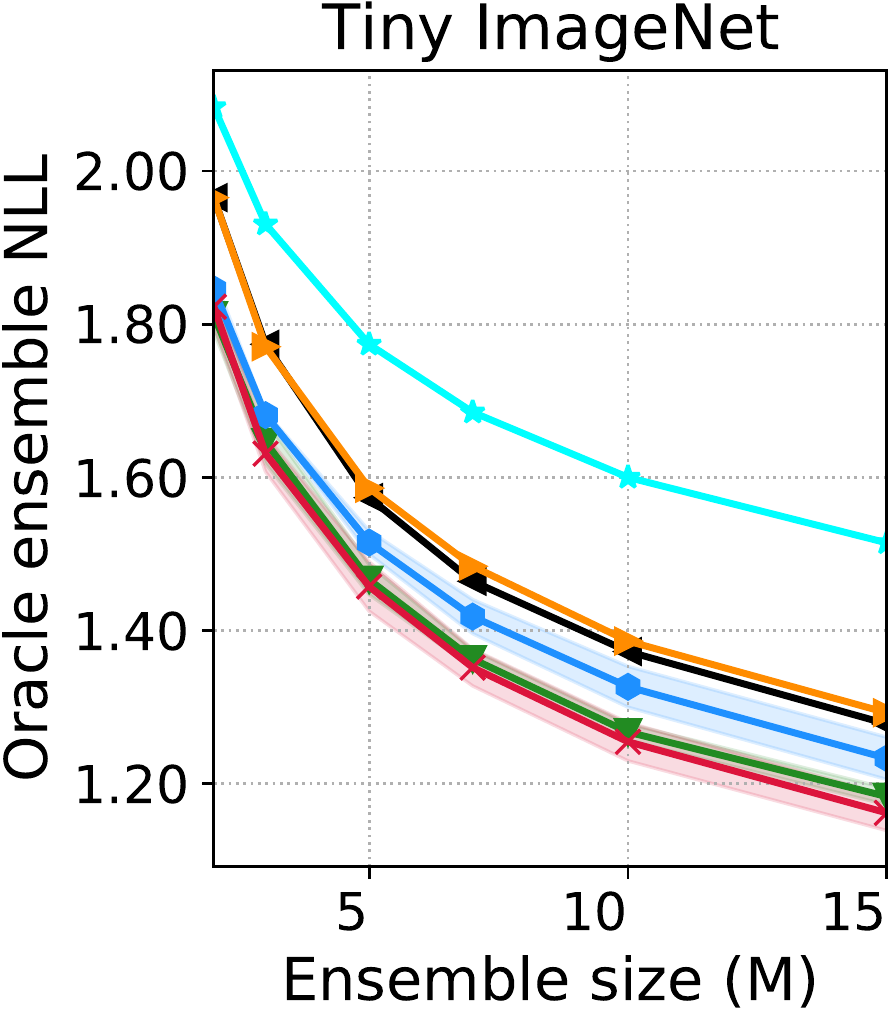}
        \subcaption{Data shift (severity 2)}
    \end{subfigure}\\ %
    \begin{subfigure}[t]{0.31\textwidth}
        \centering
        \includegraphics[width=.48\linewidth]{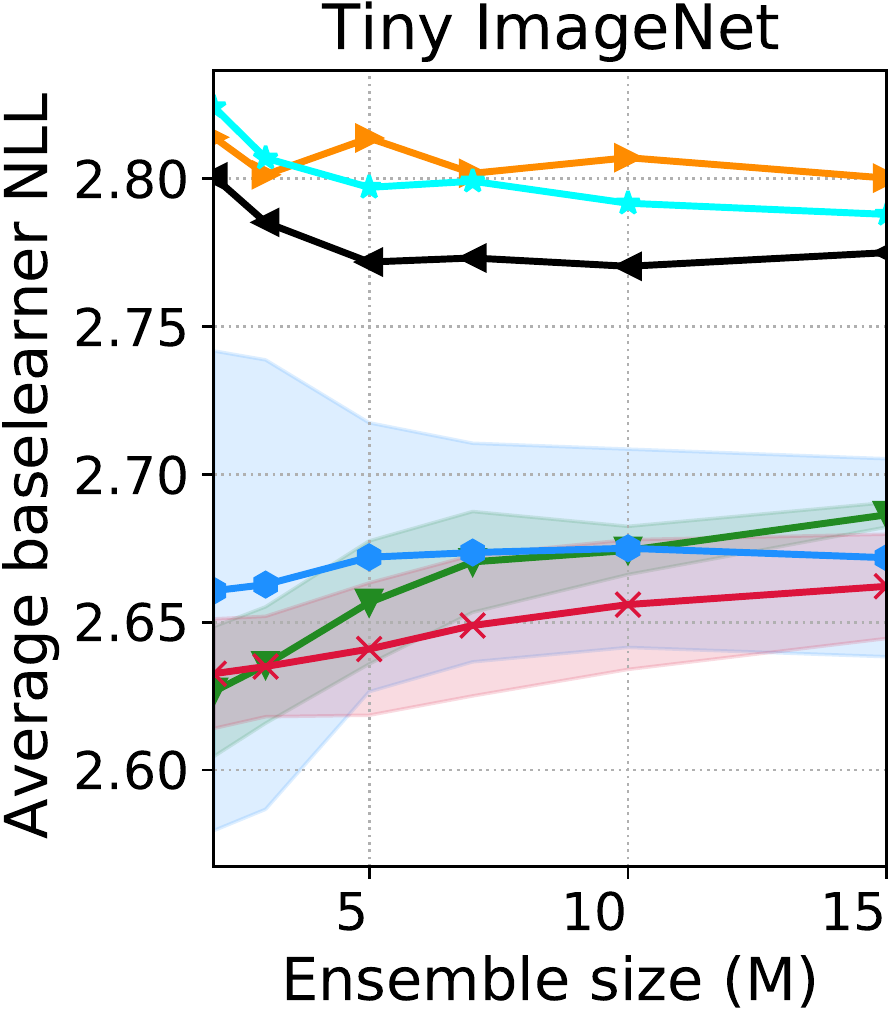}
        \includegraphics[width=.48\linewidth]{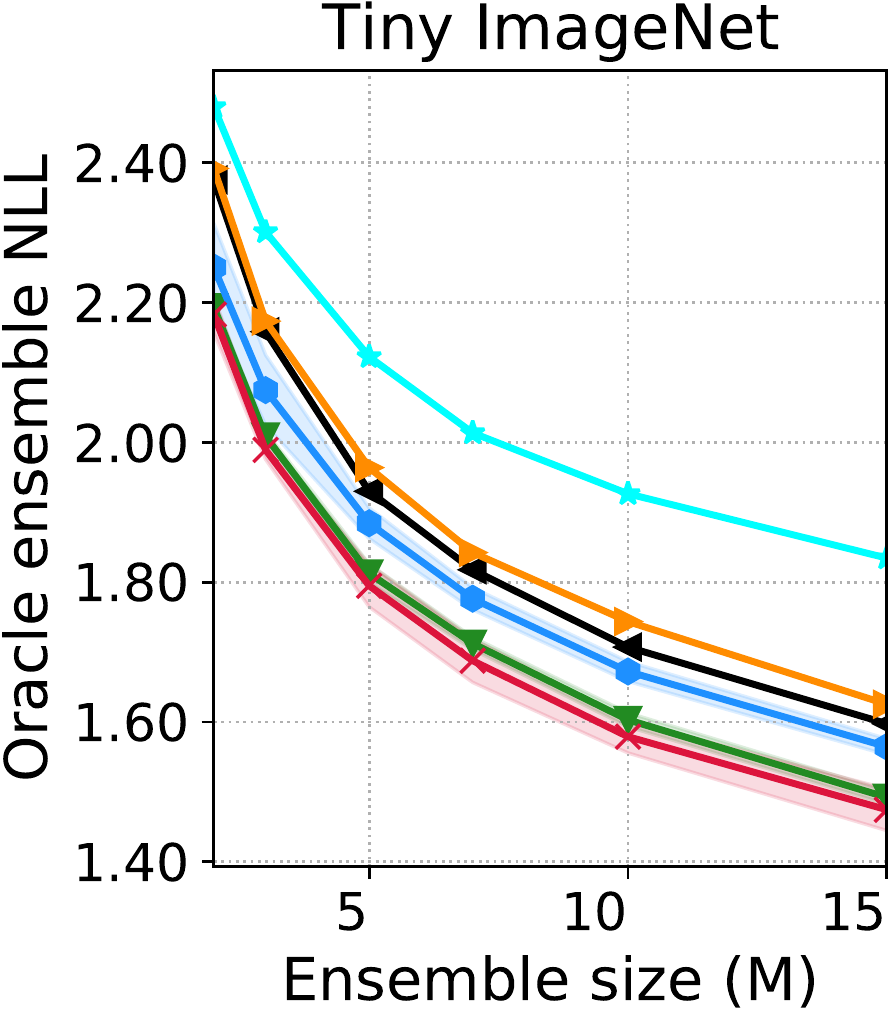}
        \subcaption{Data shift (severity 3)}
    \end{subfigure}
    ~\hspace{.1cm}
    \begin{subfigure}[t]{0.31\textwidth}
        \centering
        \includegraphics[width=0.48\linewidth]{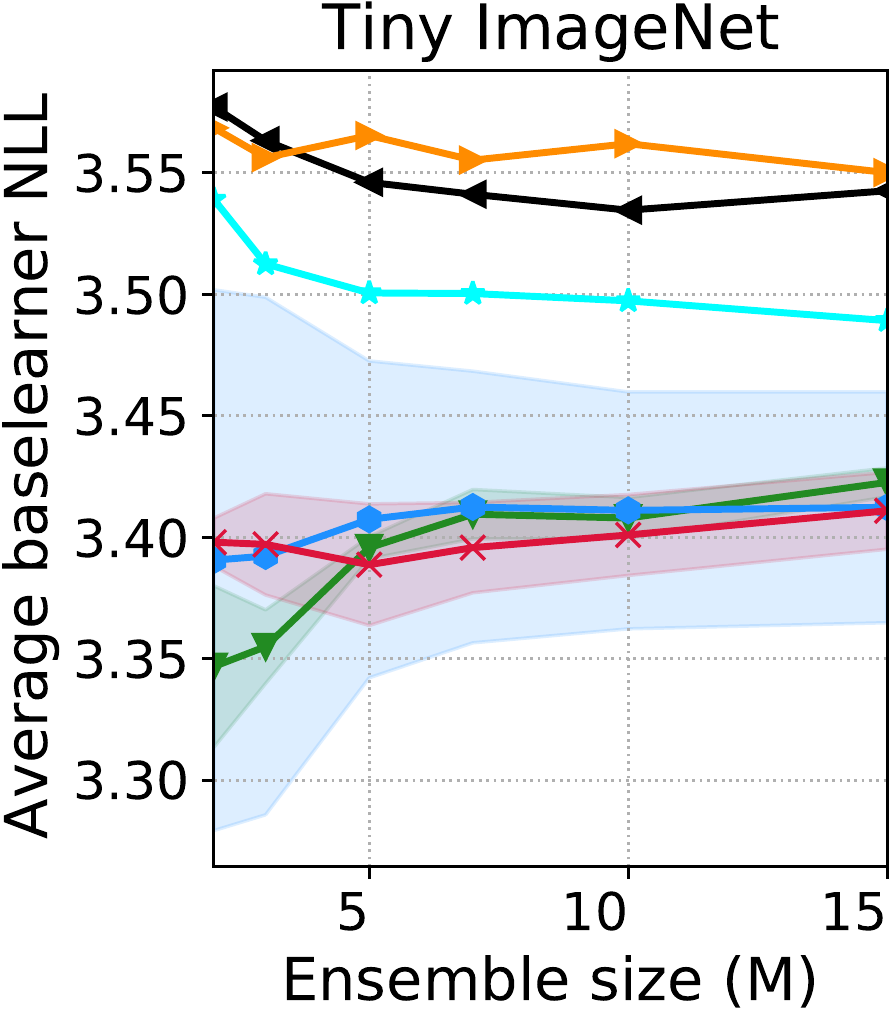}
        \includegraphics[width=0.48\linewidth]{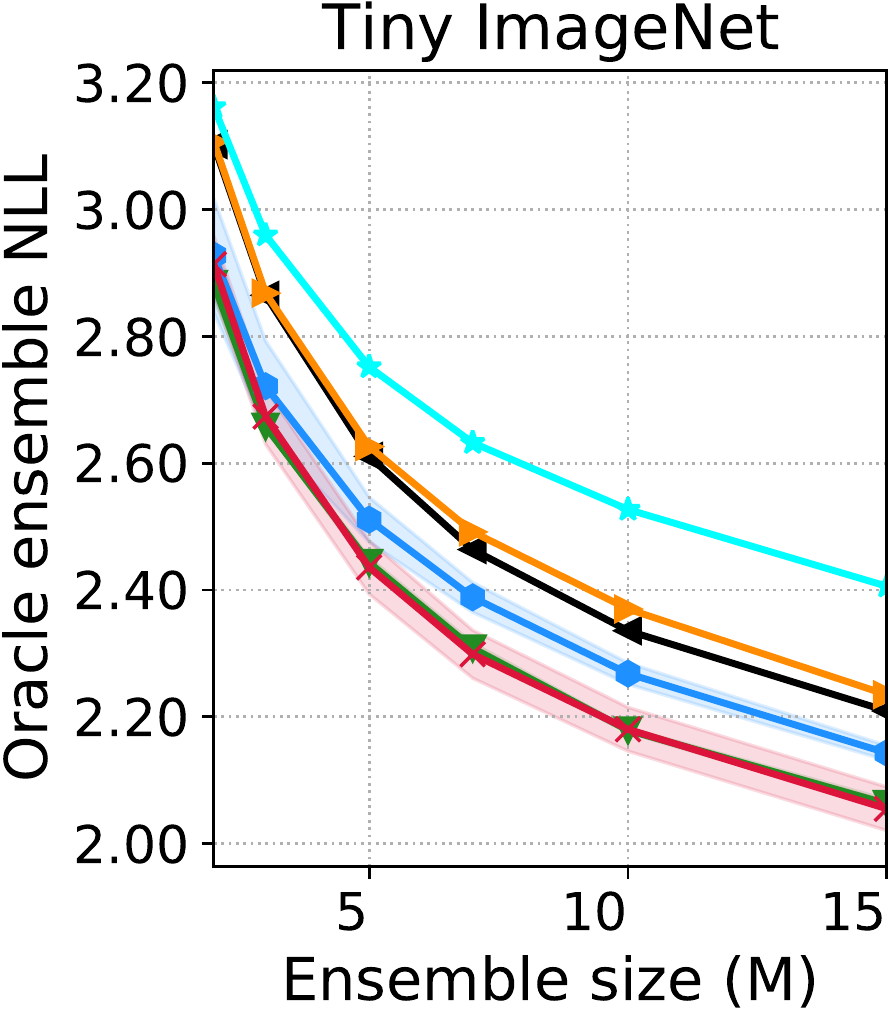}
        \subcaption{Data shift (severity 4)}
    \end{subfigure}%
    ~\hspace{.1cm}
    \begin{subfigure}[t]{0.31\textwidth}
        \centering
        \includegraphics[width=0.48\linewidth]{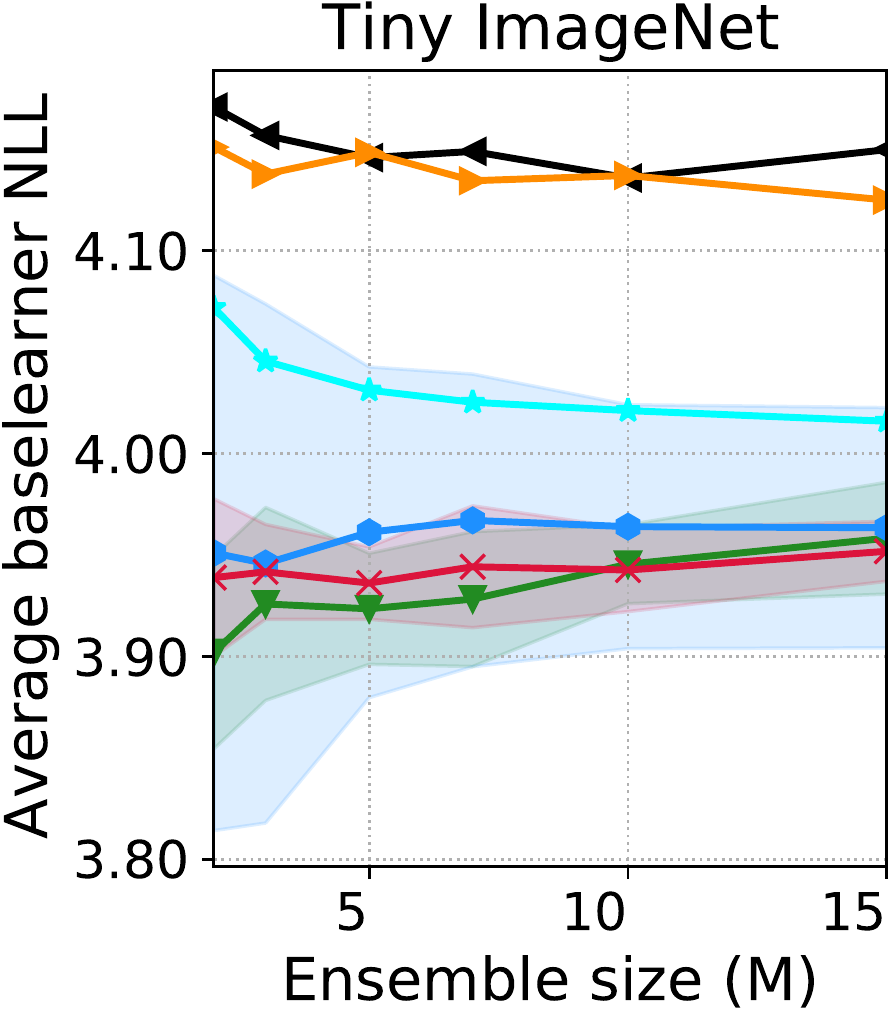}
        \includegraphics[width=0.48\linewidth]{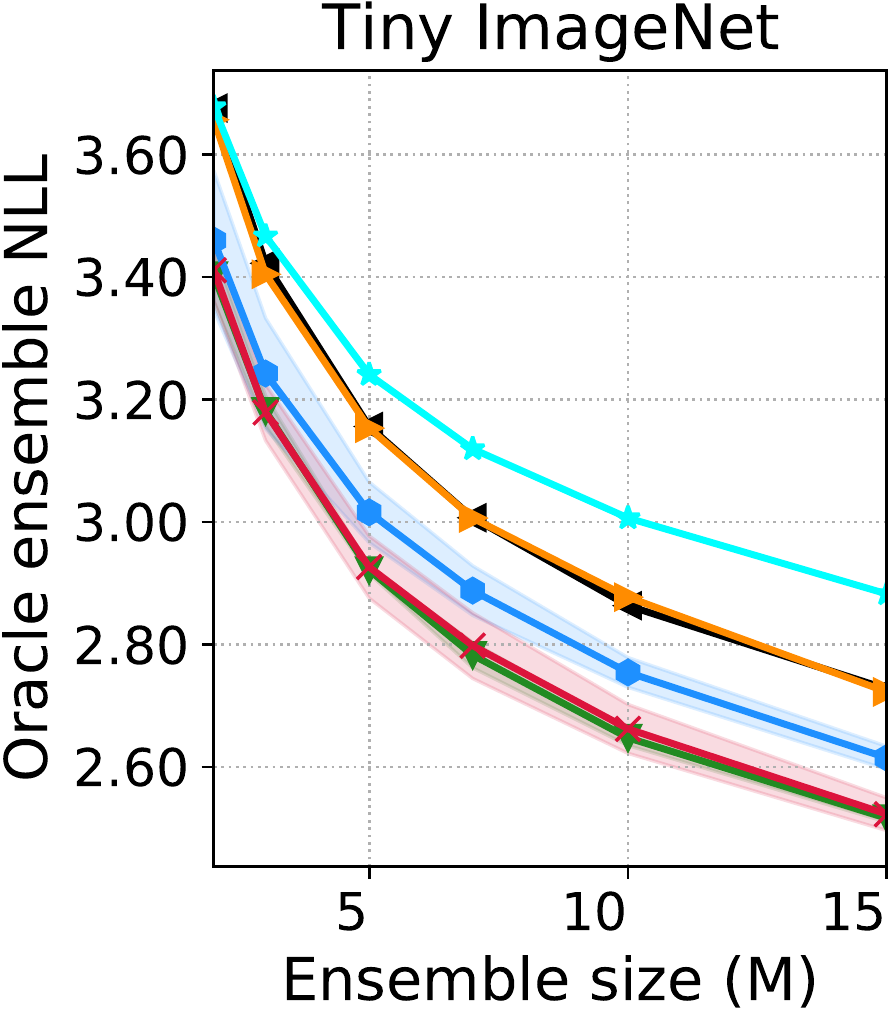}
        \subcaption{Data shift (severity 5)}
    \end{subfigure}
    
    \caption{Average base learner and oracle ensemble NLL across ensemble sizes and shift severities on Tiny ImageNet over DARTS search space.}
    \label{fig:test_avg_oracle_M_other-tiny}
\end{figure*}

\begin{figure*}
    \centering
    \captionsetup[subfigure]{justification=centering}
    \begin{subfigure}[t]{0.49\textwidth}
        \centering
        \includegraphics[width=.30\linewidth]{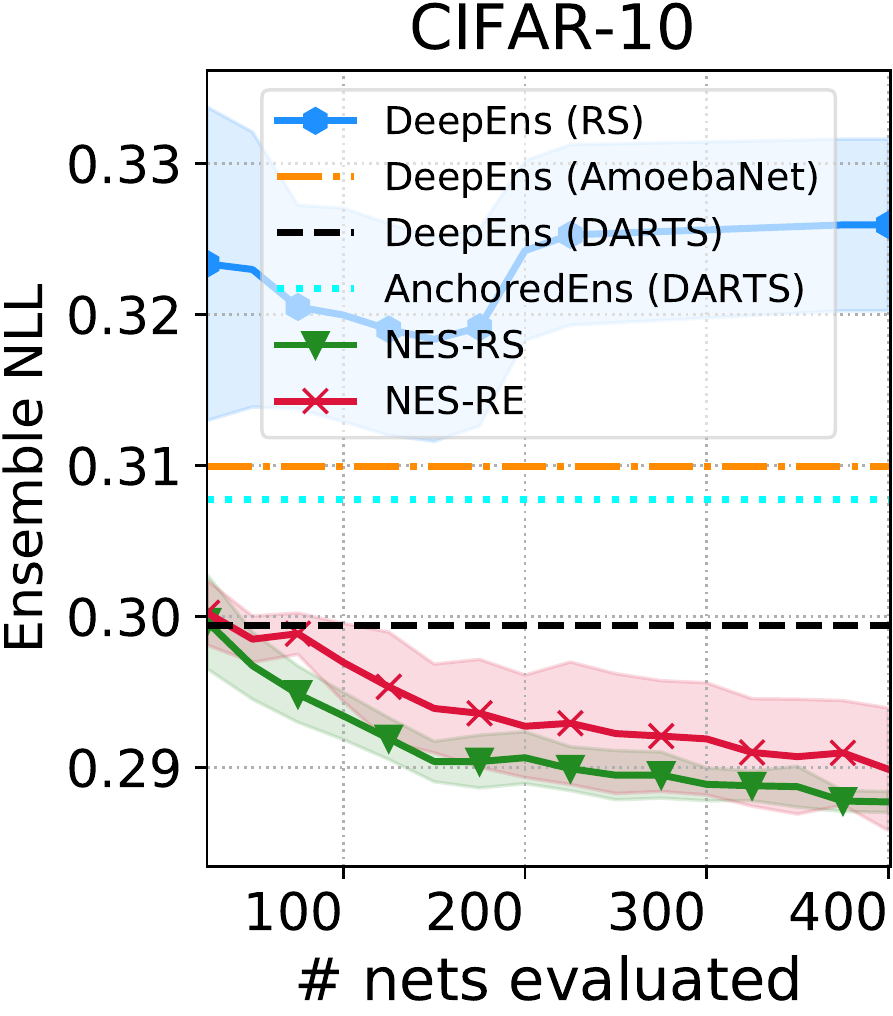}
        \includegraphics[width=.30\linewidth]{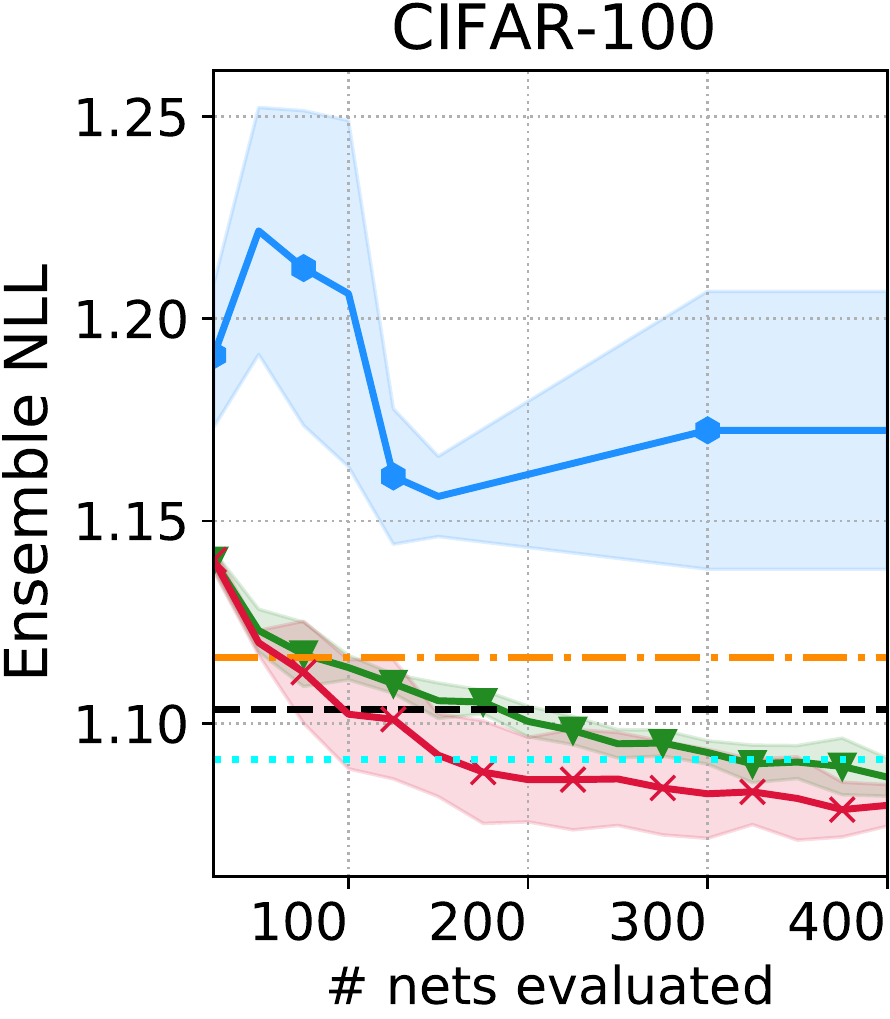}
        \includegraphics[width=.30\linewidth]{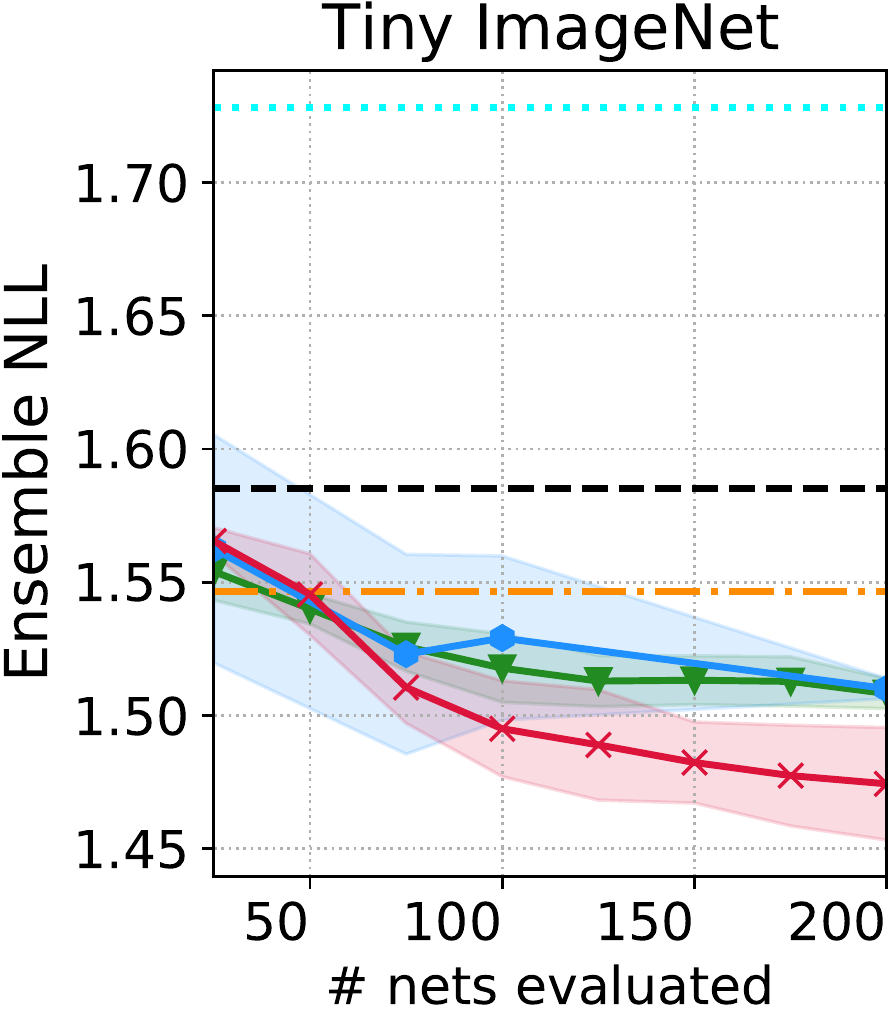}
        \subcaption{No data shift}
    \end{subfigure}%
    \begin{subfigure}[t]{0.49\textwidth}
        \centering
        \includegraphics[width=.30\linewidth]{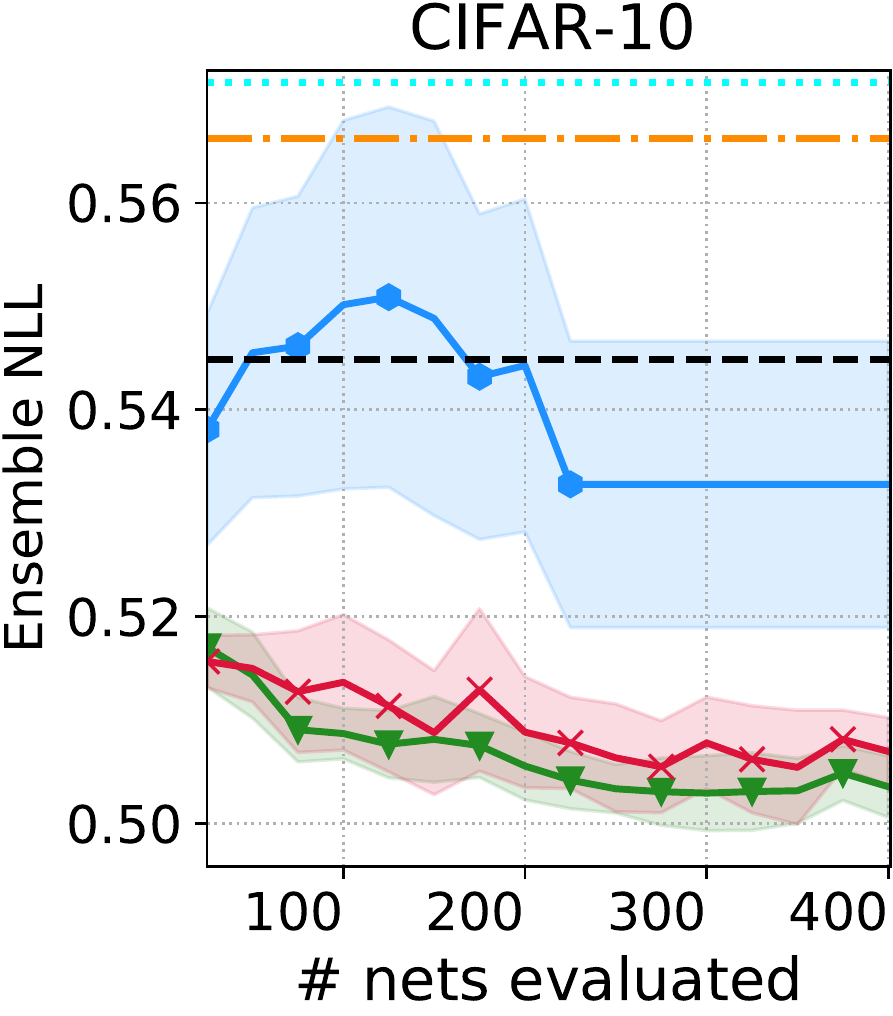}
        \includegraphics[width=.30\linewidth]{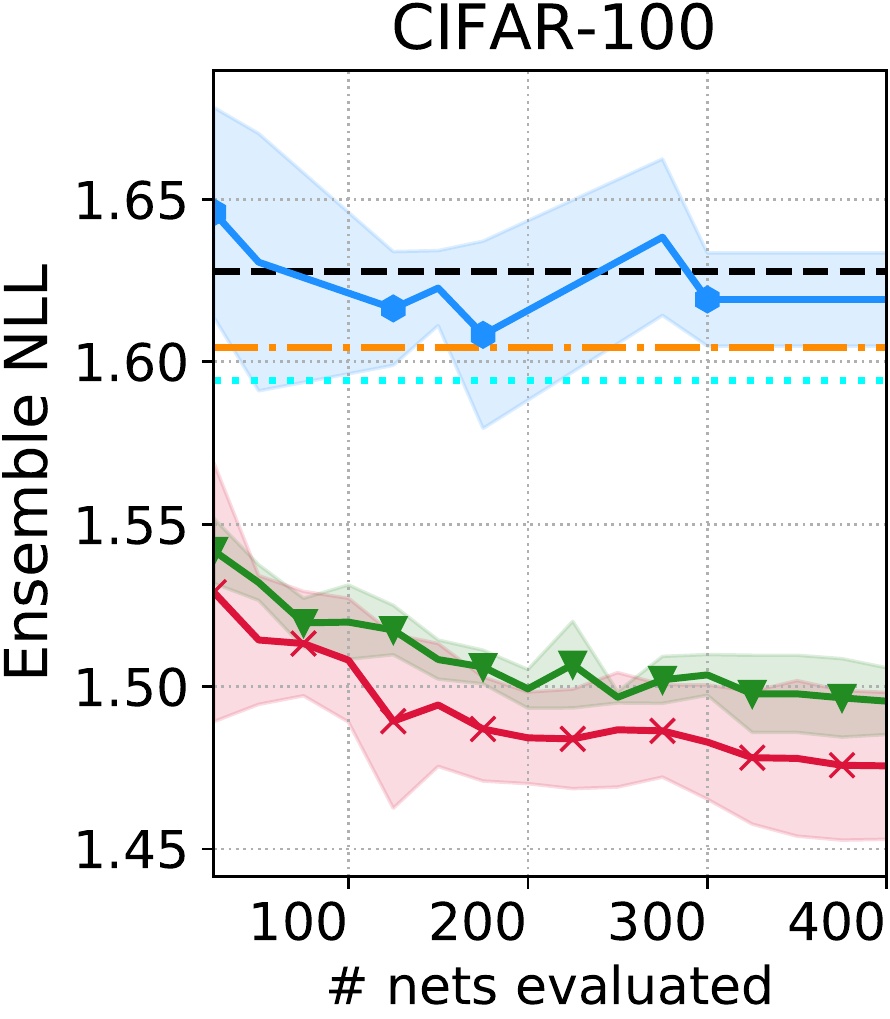}
        \includegraphics[width=.30\linewidth]{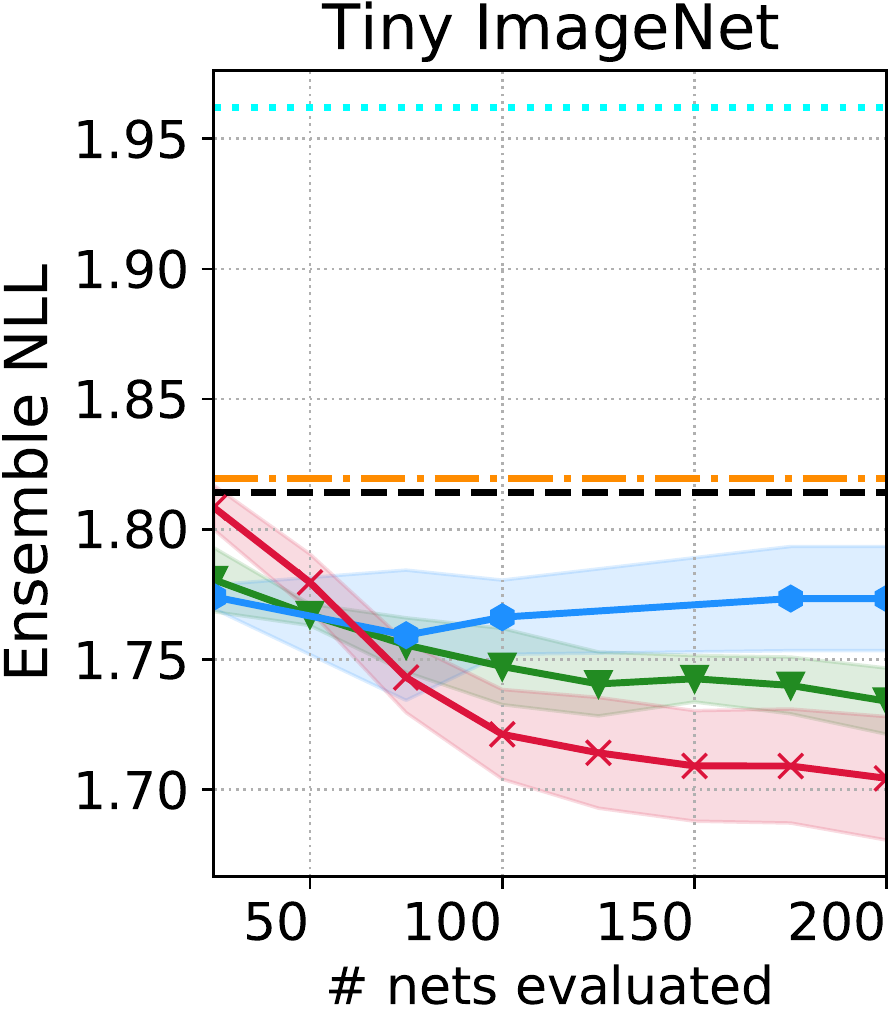}
        \subcaption{Data shift (severity 1)}
    \end{subfigure}\\ %
    \begin{subfigure}[t]{0.49\textwidth}
        \centering
        \includegraphics[width=.30\linewidth]{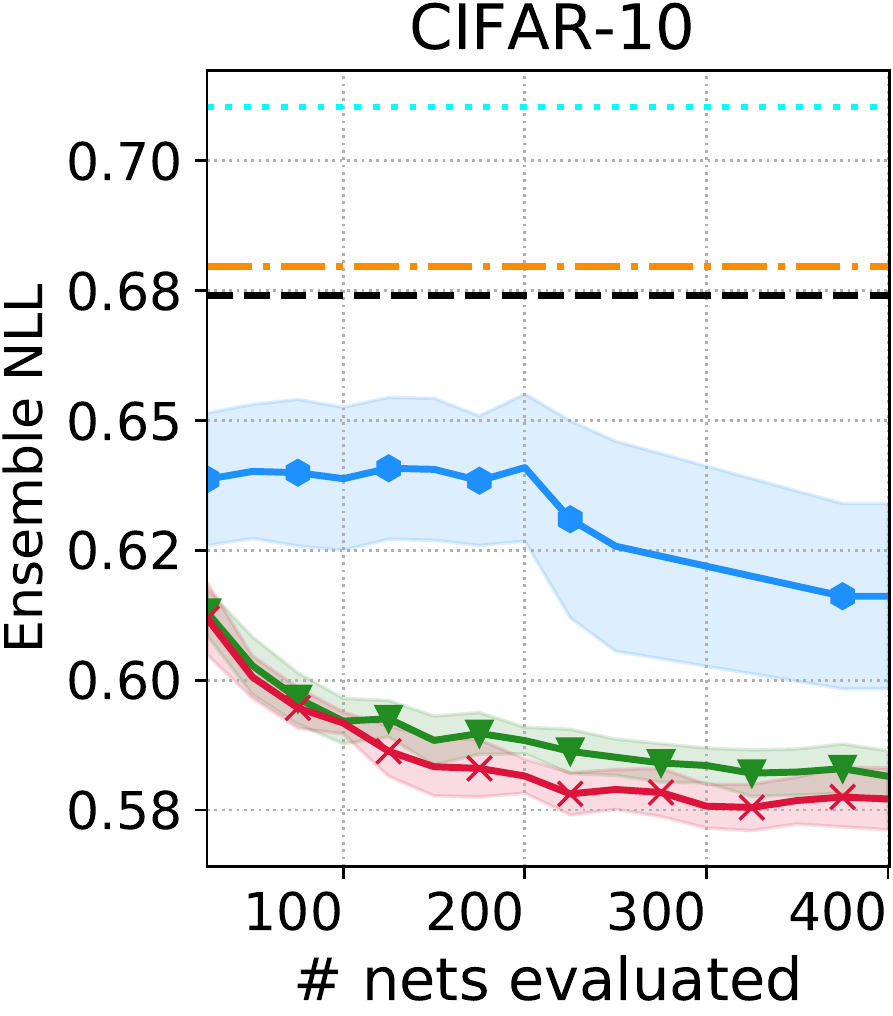}
        \includegraphics[width=.30\linewidth]{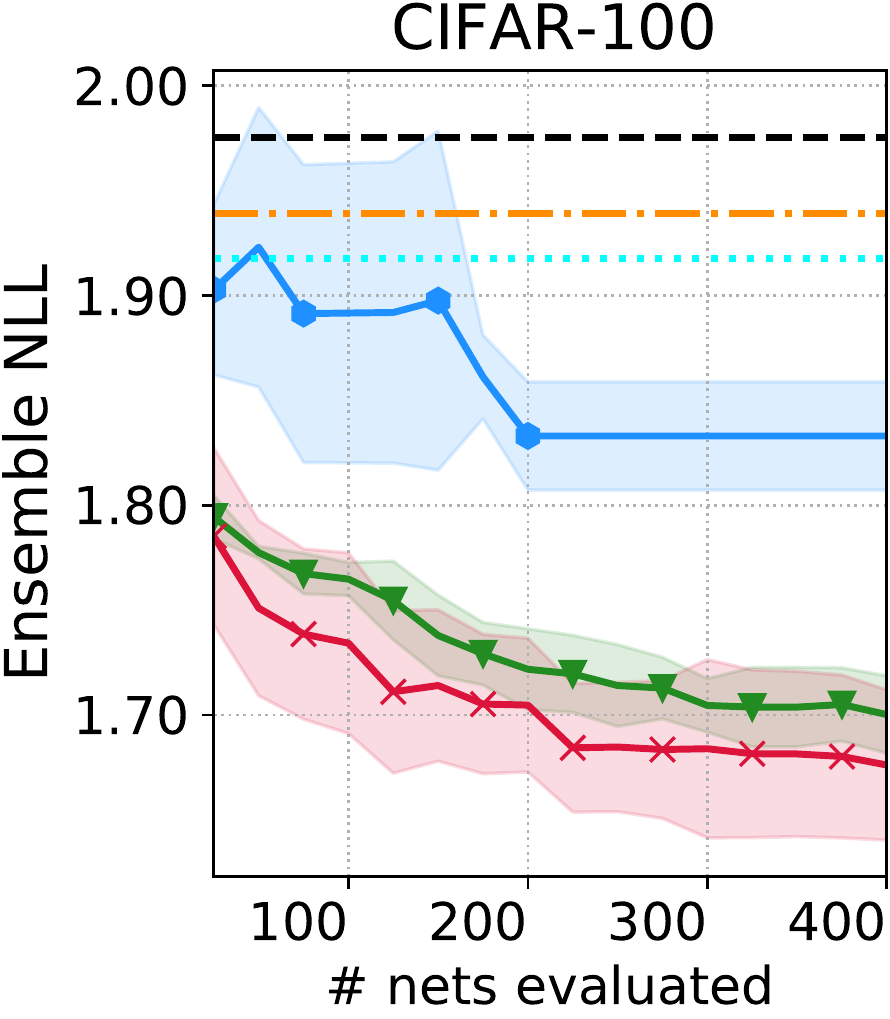}
        \includegraphics[width=.30\linewidth]{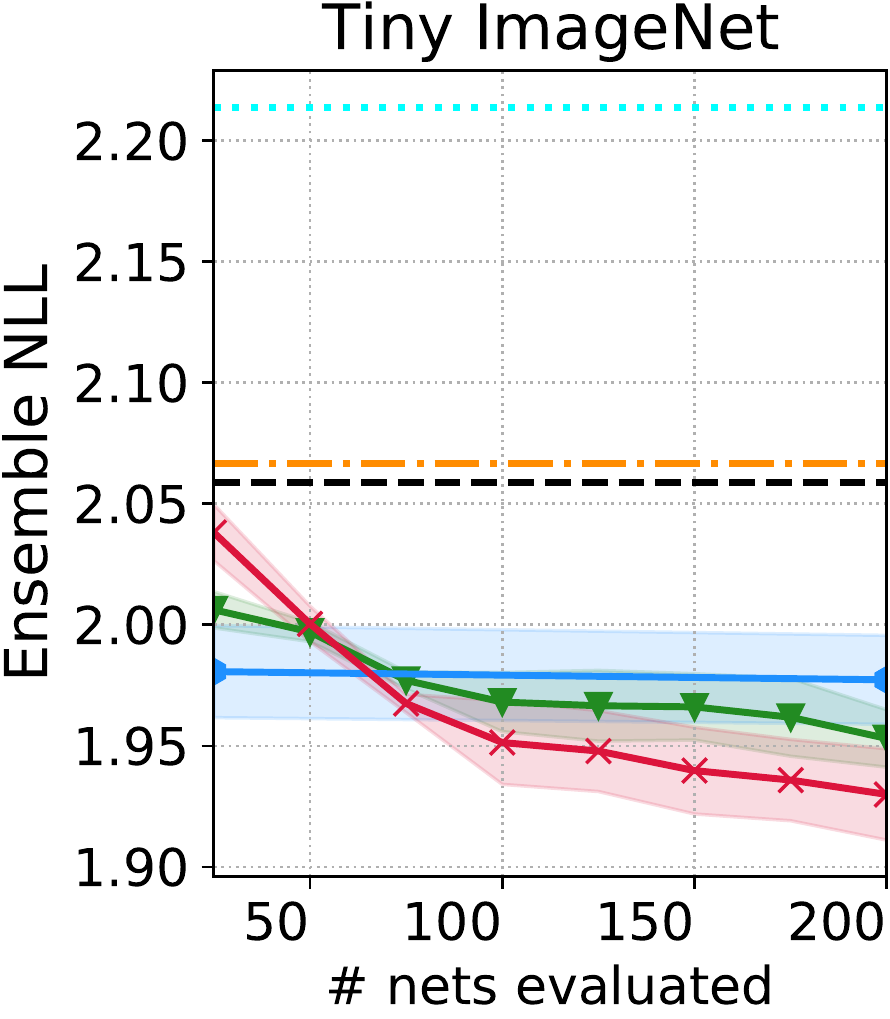}
        \subcaption{Data shift (severity 2)}
    \end{subfigure}%
    \begin{subfigure}[t]{0.49\textwidth}
        \centering
        \includegraphics[width=.30\linewidth]{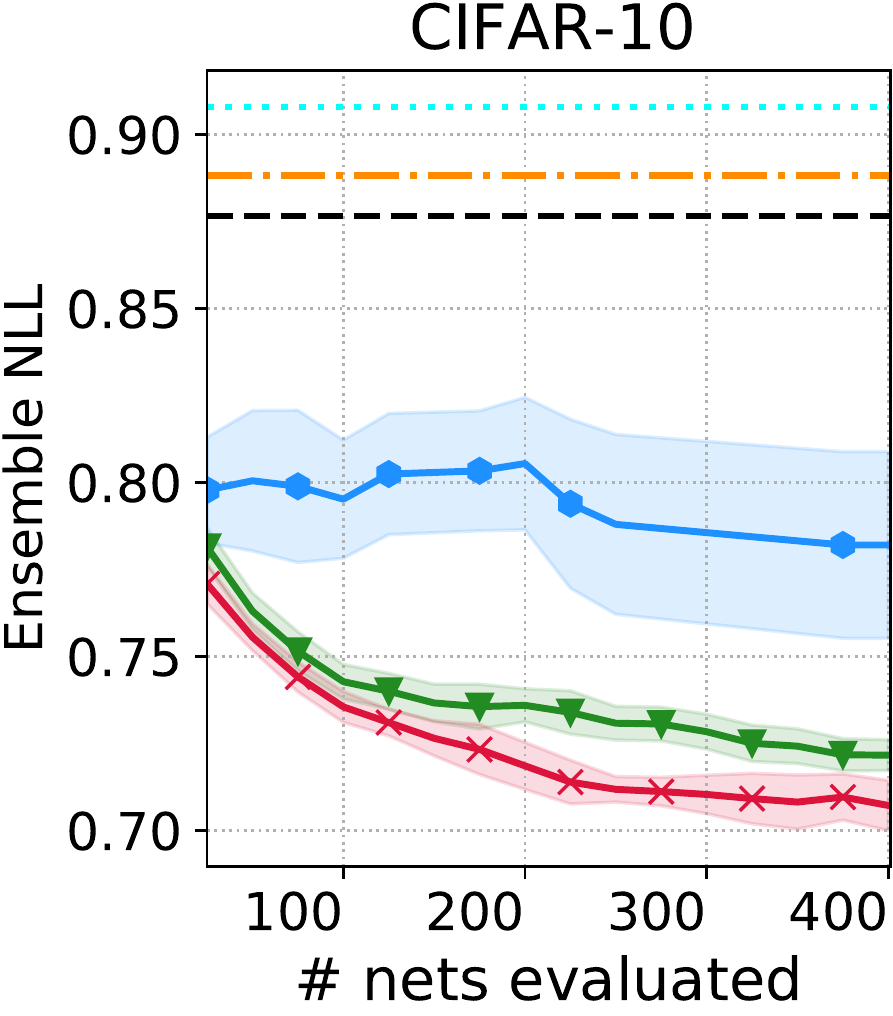}
        \includegraphics[width=.30\linewidth]{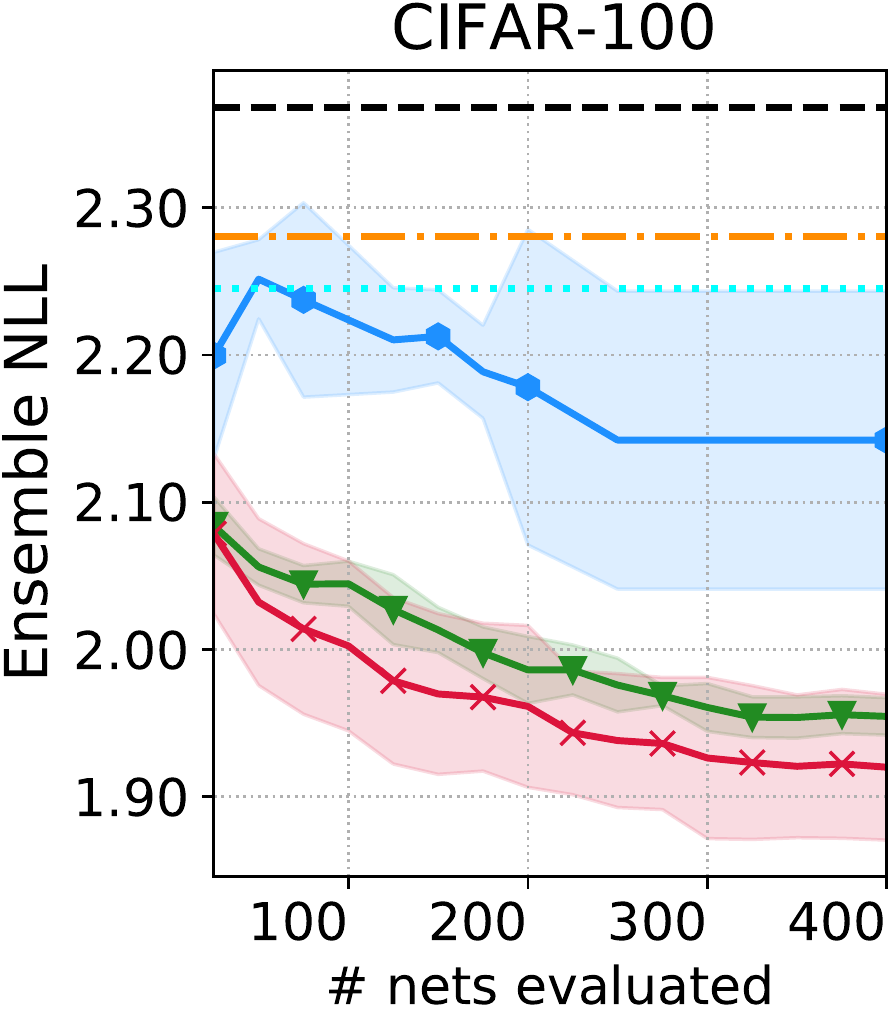}
        \includegraphics[width=.30\linewidth]{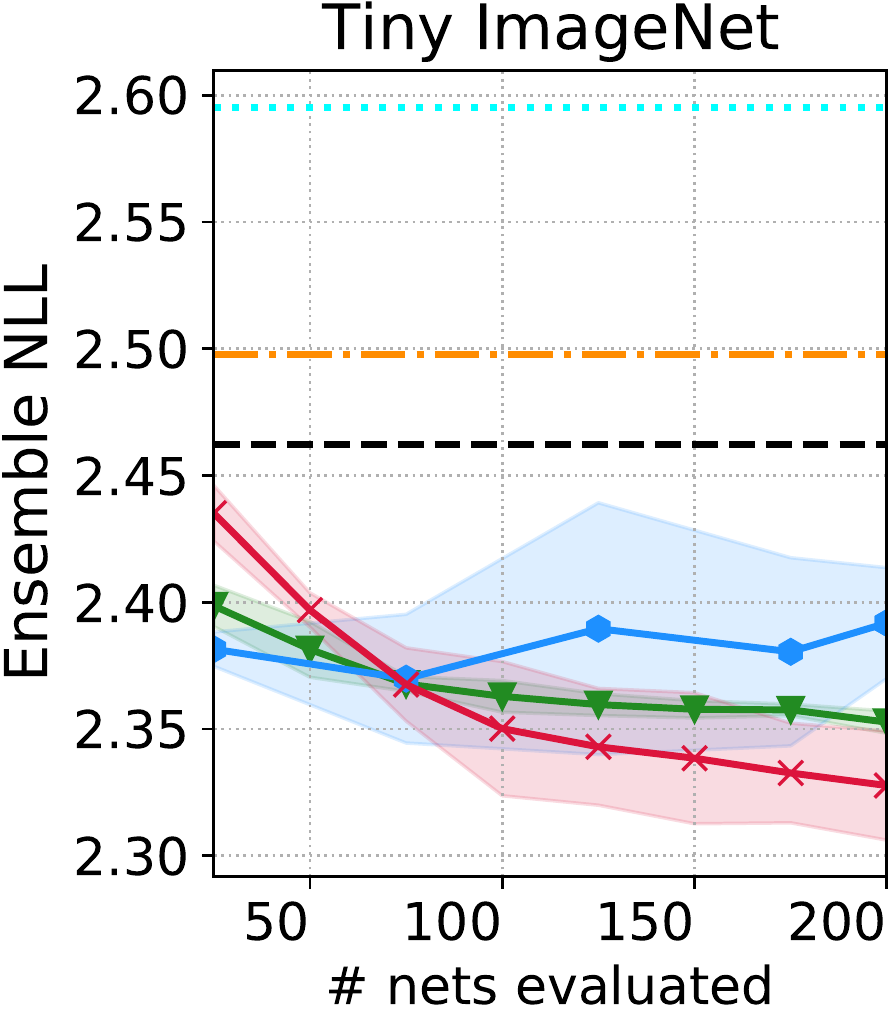}
        \subcaption{Data shift (severity 3)}
    \end{subfigure}\\ %
    \begin{subfigure}[t]{0.49\textwidth}
        \centering
        \includegraphics[width=.30\linewidth]{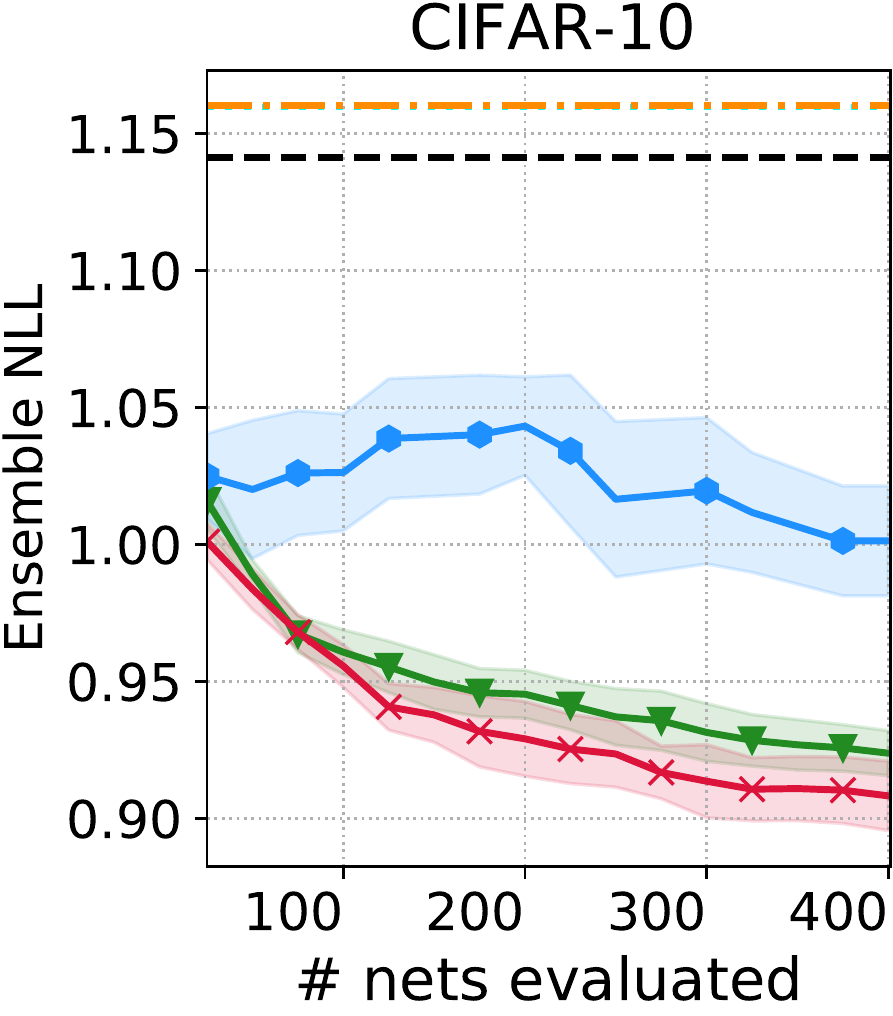}
        \includegraphics[width=.30\linewidth]{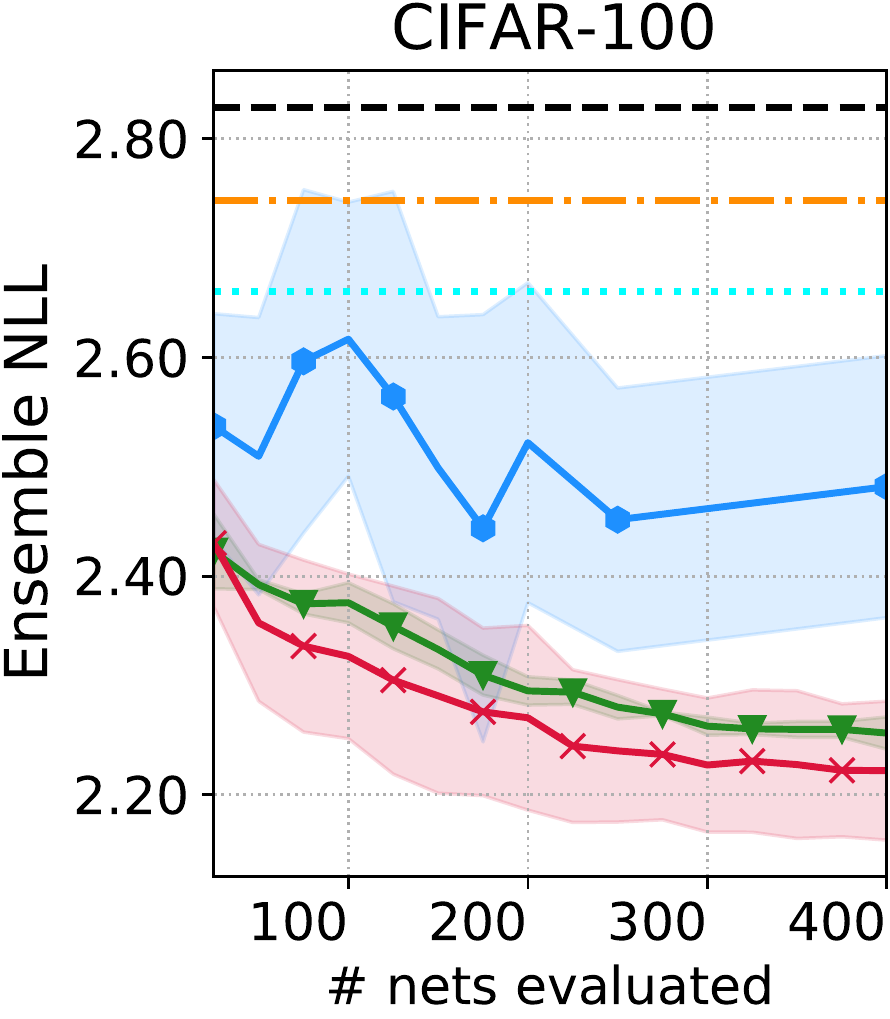}
        \includegraphics[width=.30\linewidth]{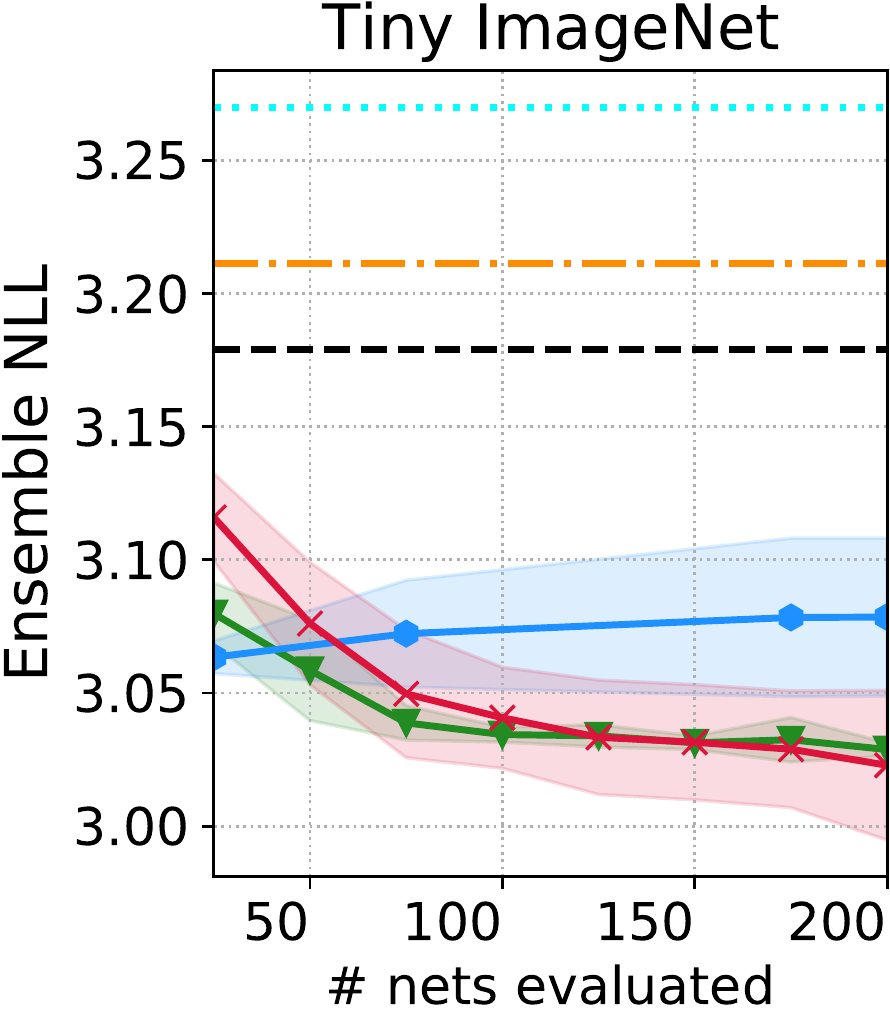}
        \subcaption{Data shift (severity 4)}
    \end{subfigure}%
    \begin{subfigure}[t]{0.49\textwidth}
        \centering
        \includegraphics[width=.30\linewidth]{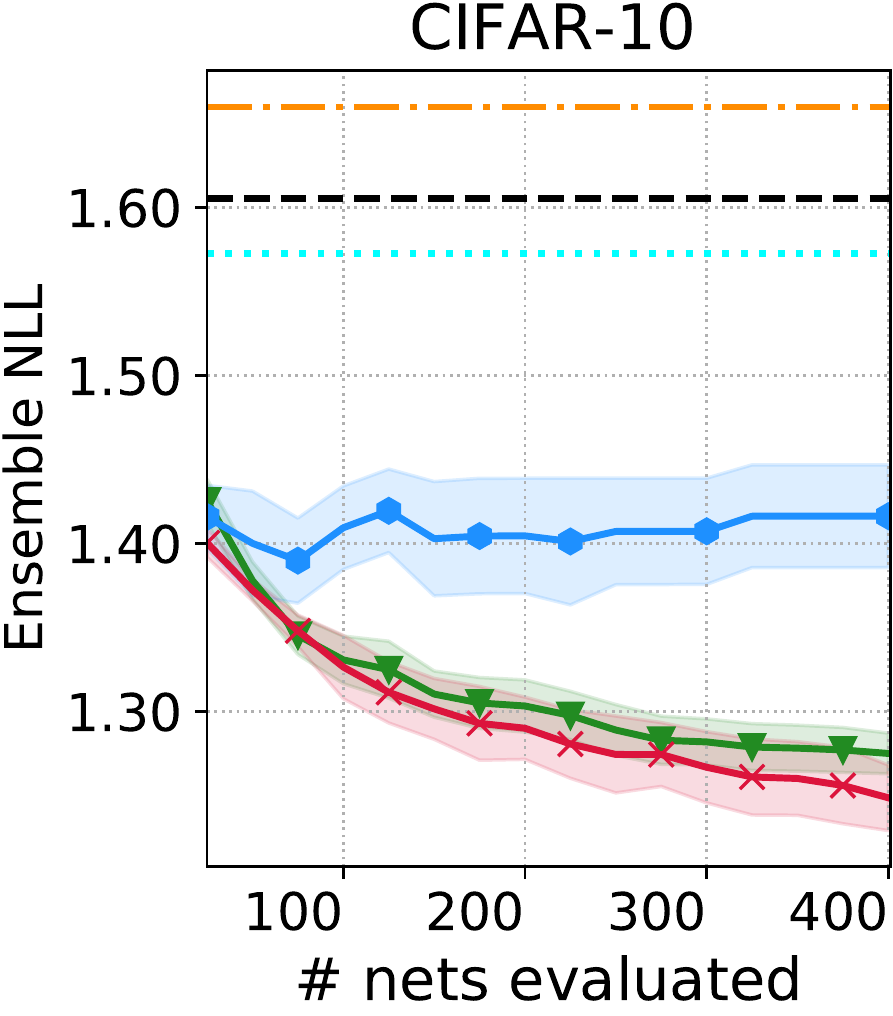}
        \includegraphics[width=.30\linewidth]{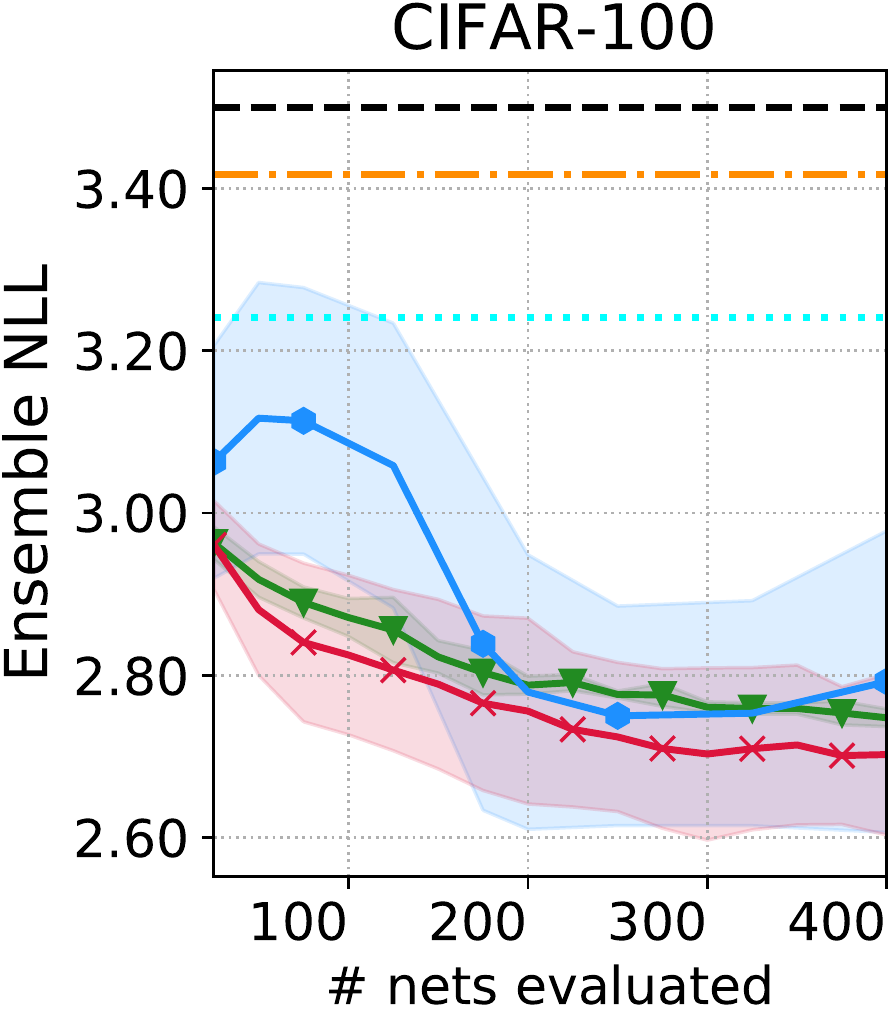}
        \includegraphics[width=.30\linewidth]{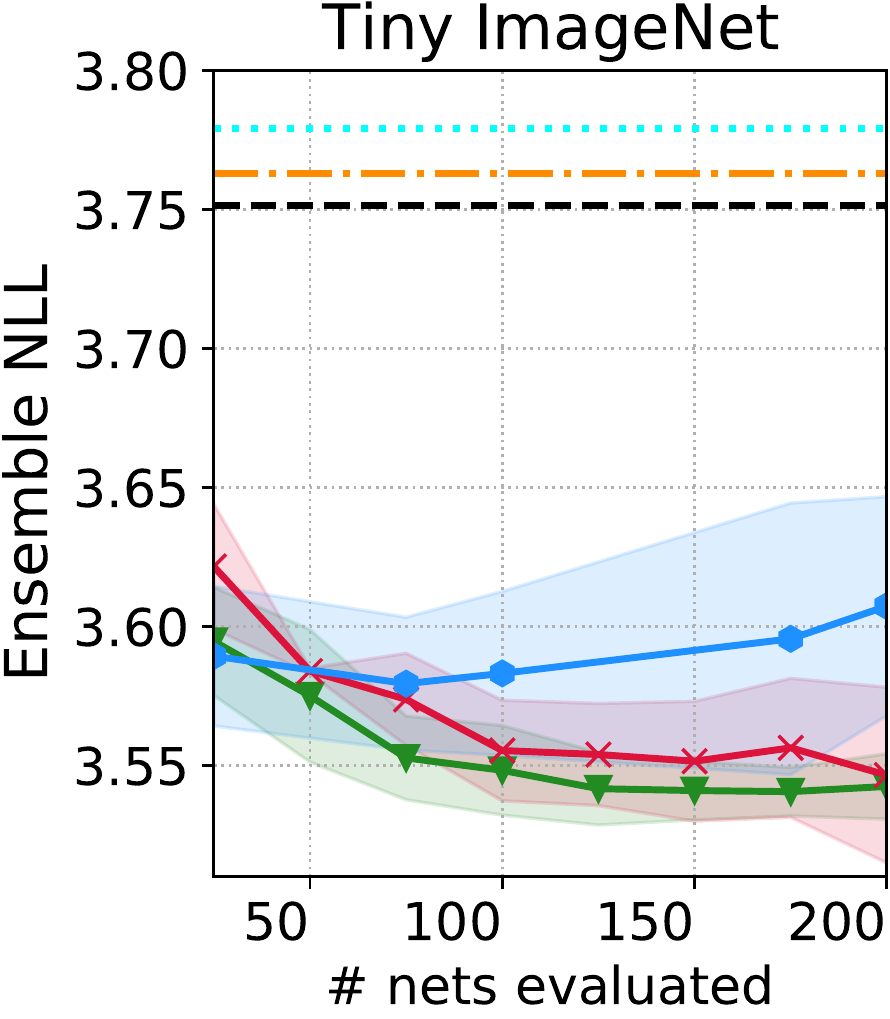}
        \subcaption{Data shift (severity 5)}
    \end{subfigure}
    
    \caption{Ensemble NLL vs. budget $\budget$. Ensemble size fixed at $M = 10$.}
    \label{fig:test_loss_budget_other}
\end{figure*}

\begin{figure*}
    \centering
    \captionsetup[subfigure]{justification=centering}
    \begin{subfigure}[t]{0.49\textwidth}
        \centering
        \includegraphics[width=.29\linewidth]{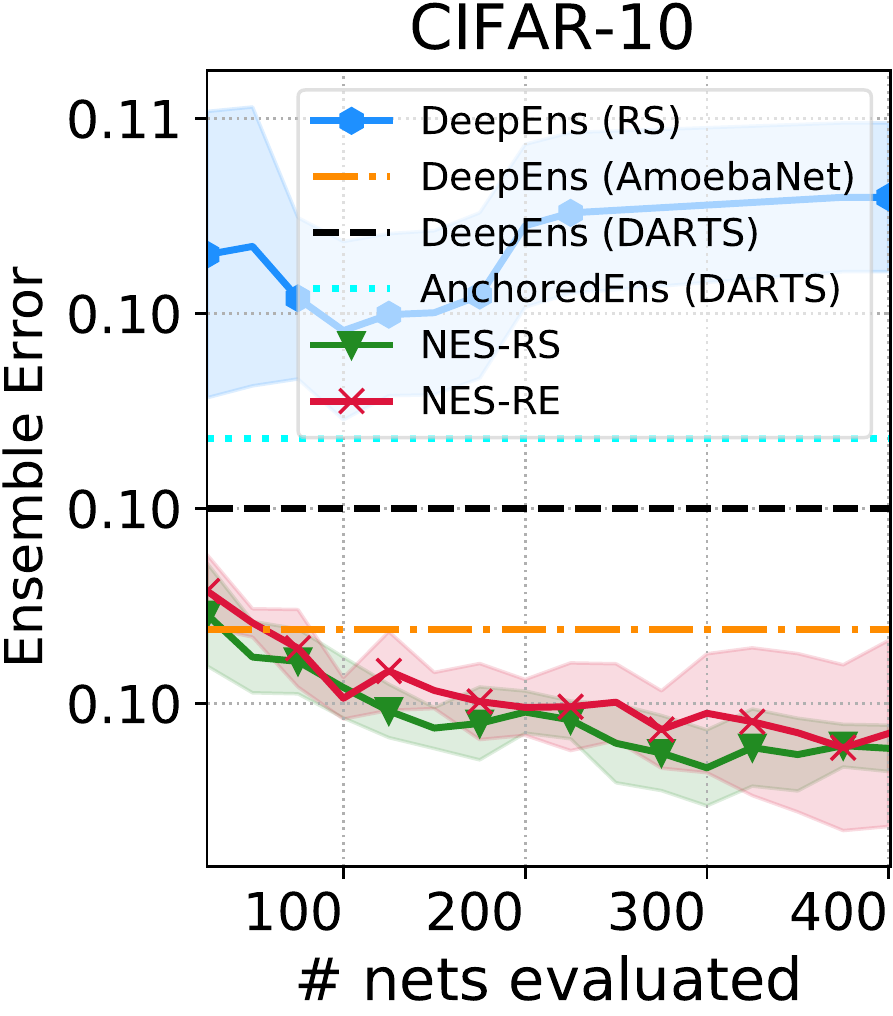}
        \includegraphics[width=.29\linewidth]{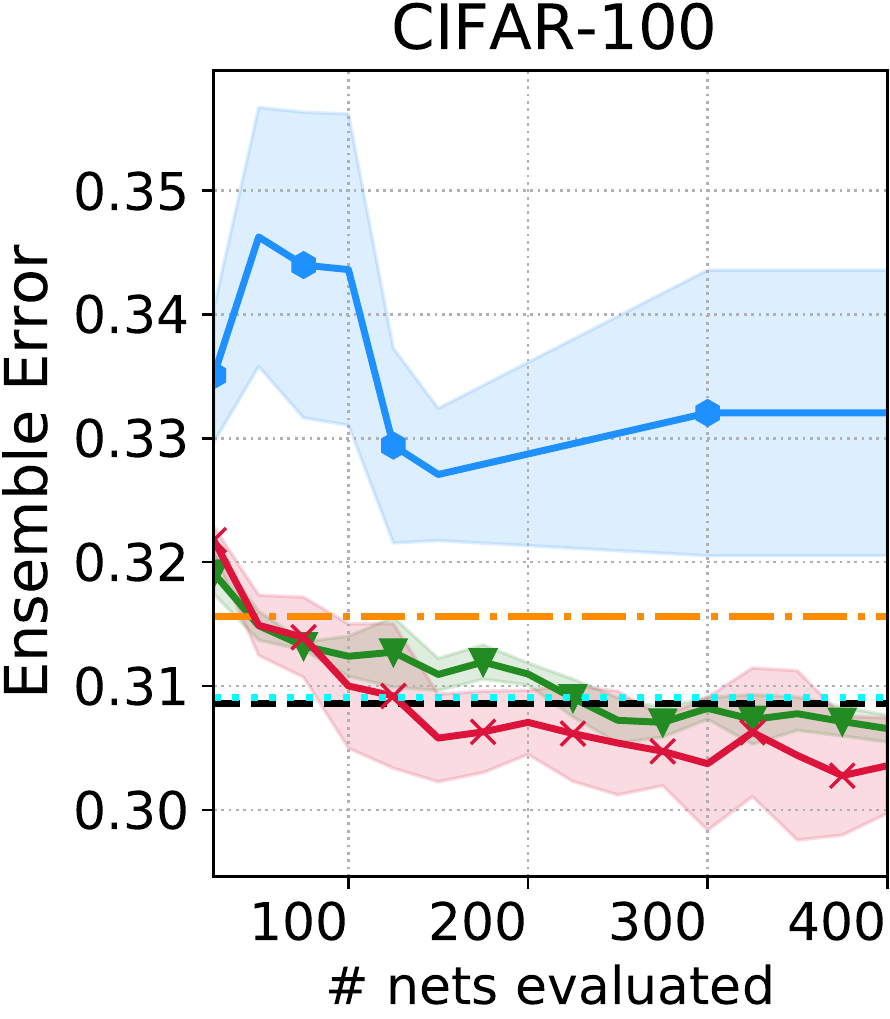}
        \includegraphics[width=.29\linewidth]{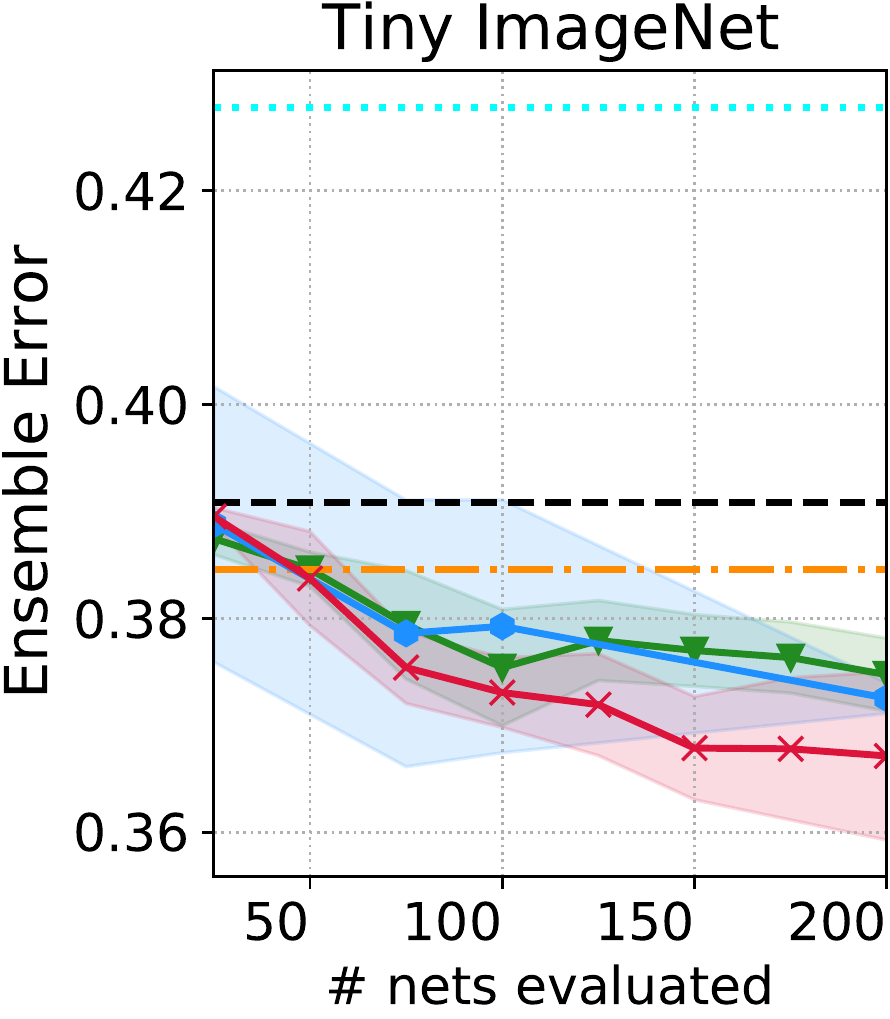}
        \subcaption{No data shift}
    \end{subfigure}%
    \begin{subfigure}[t]{0.49\textwidth}
        \centering
        \includegraphics[width=.29\linewidth]{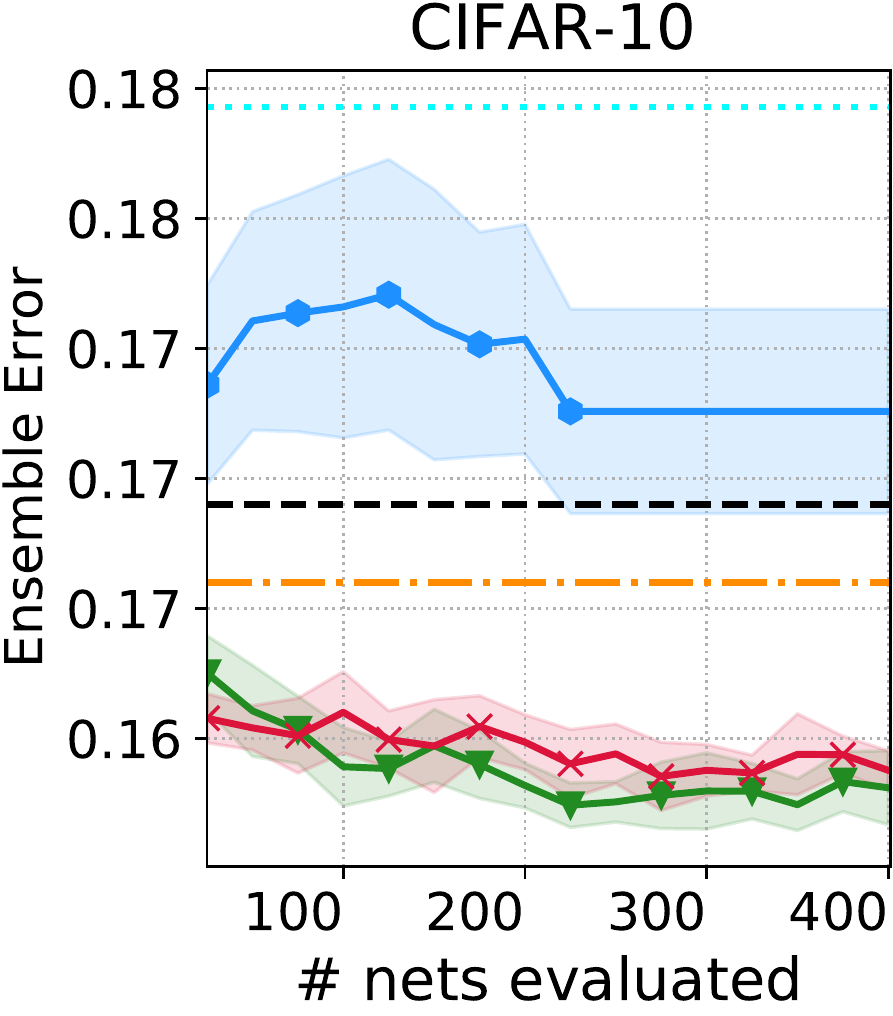}
        \includegraphics[width=.29\linewidth]{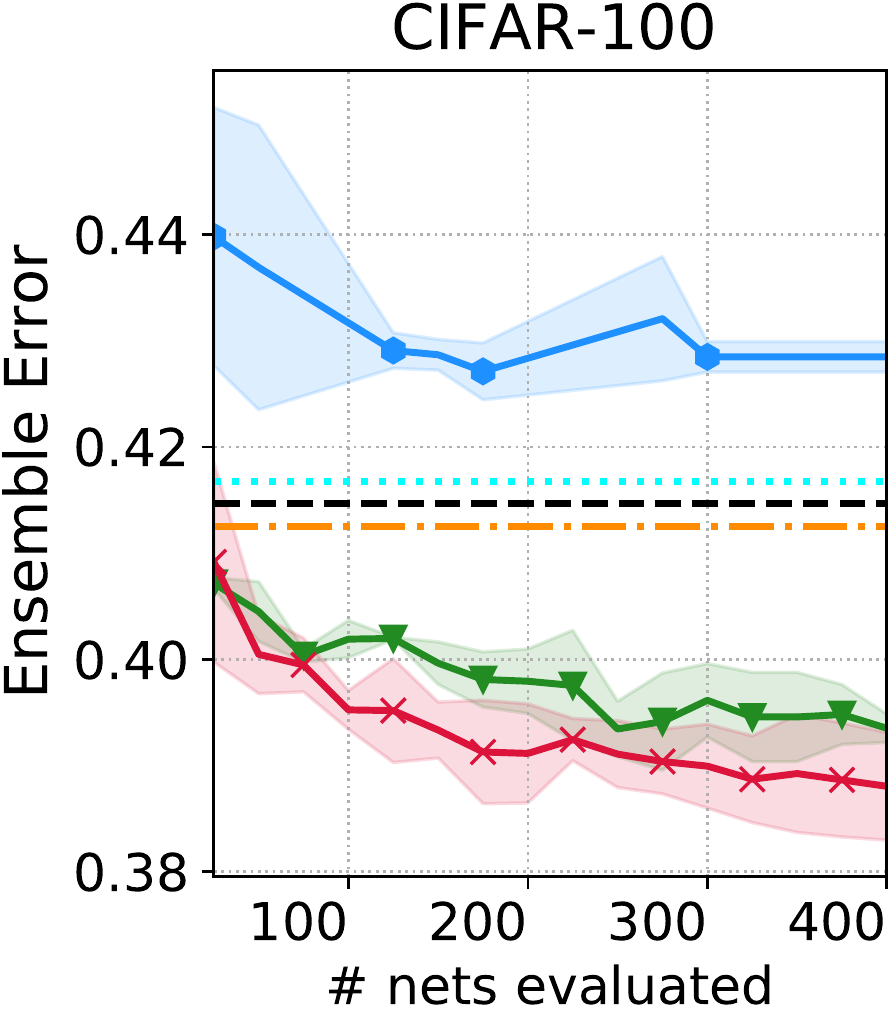}
        \includegraphics[width=.29\linewidth]{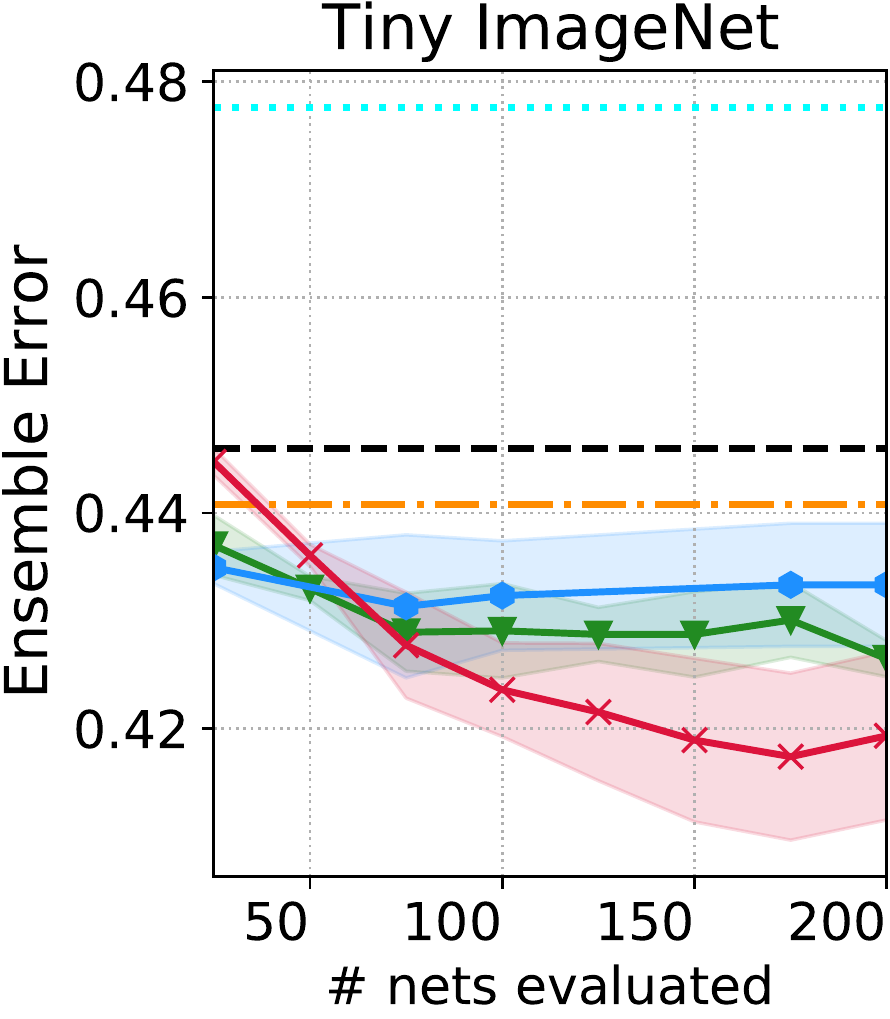}
        \subcaption{Data shift (severity 1)}
    \end{subfigure}\\ %
    \begin{subfigure}[t]{0.49\textwidth}
        \centering
        \includegraphics[width=.29\linewidth]{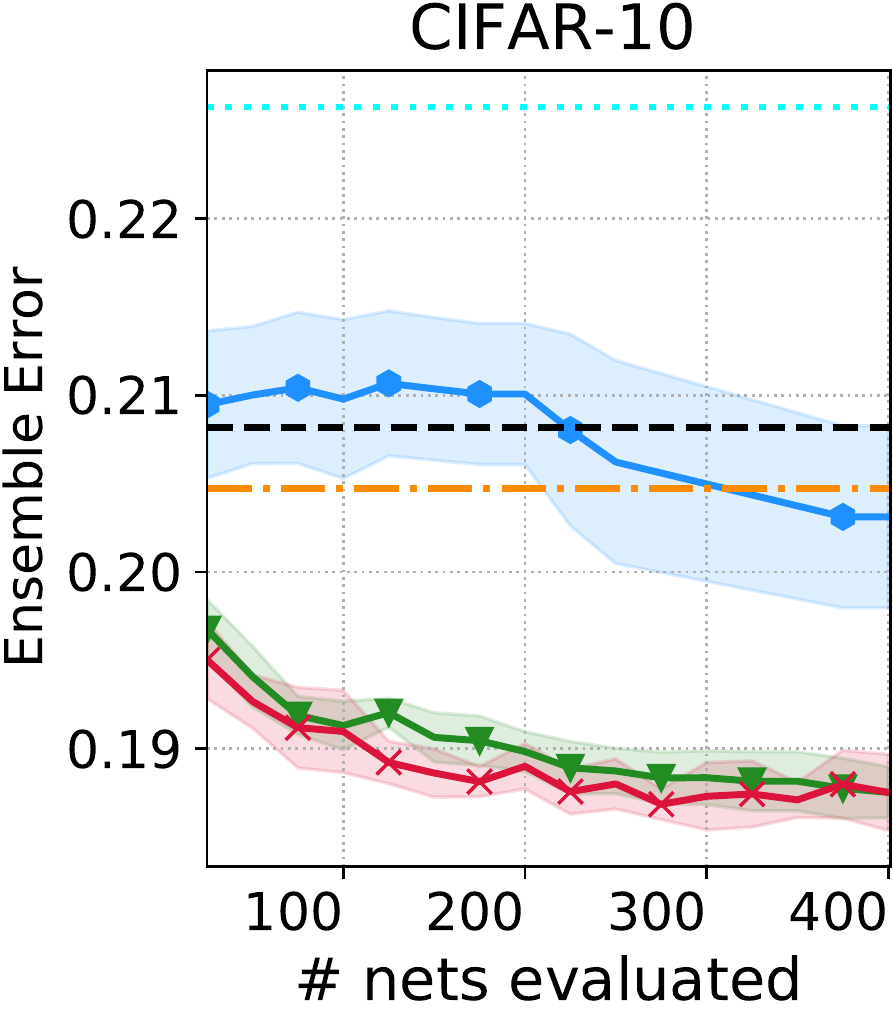}
        \includegraphics[width=.29\linewidth]{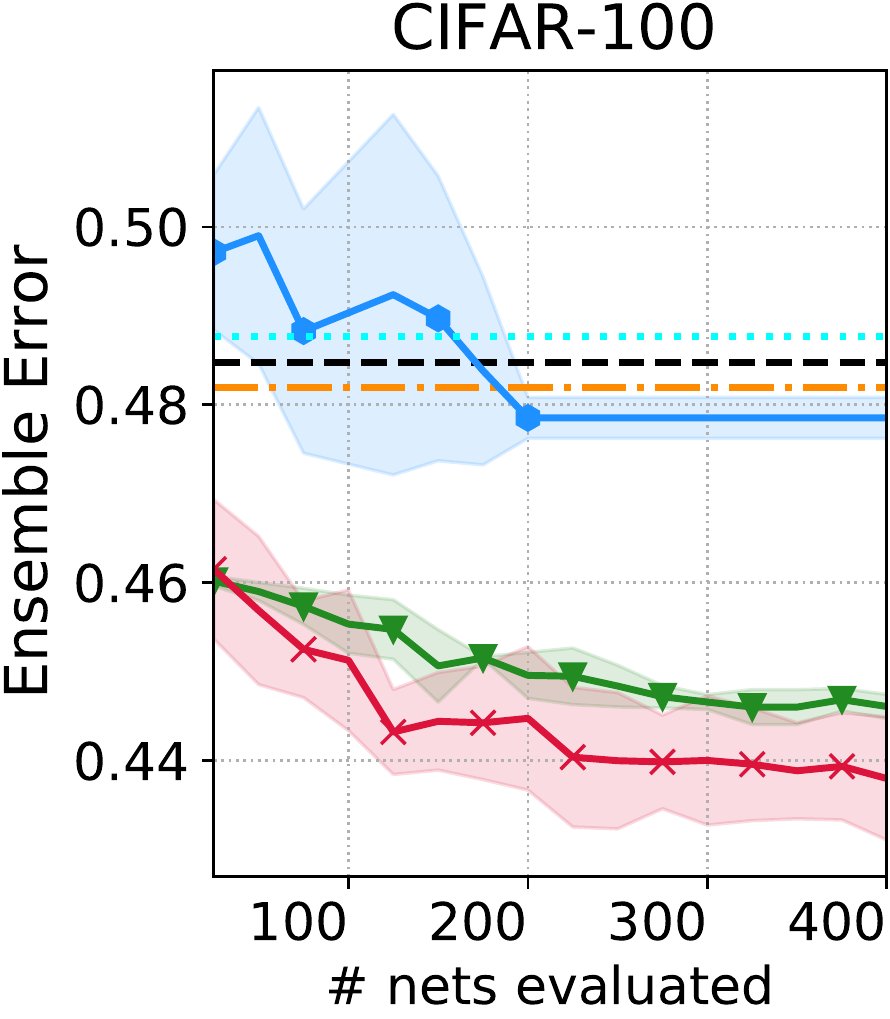}
        \includegraphics[width=.29\linewidth]{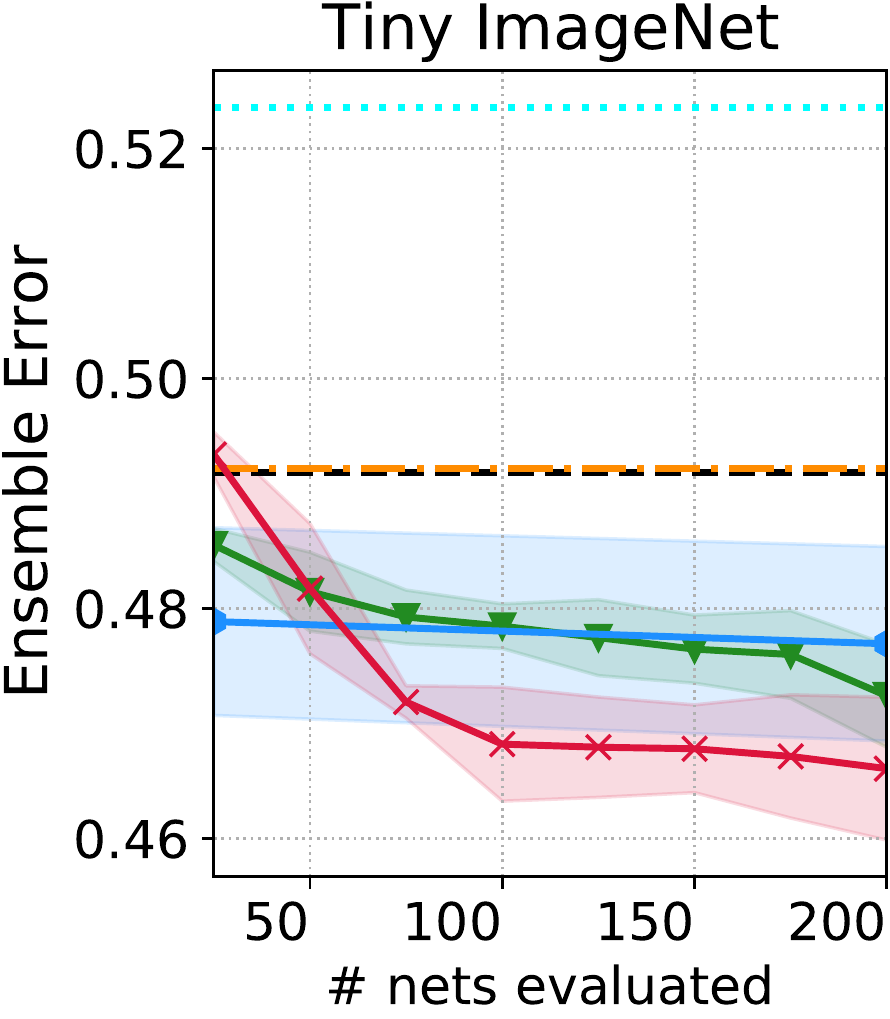}
        \subcaption{Data shift (severity 2)}
    \end{subfigure}%
    \begin{subfigure}[t]{0.49\textwidth}
        \centering
        \includegraphics[width=.29\linewidth]{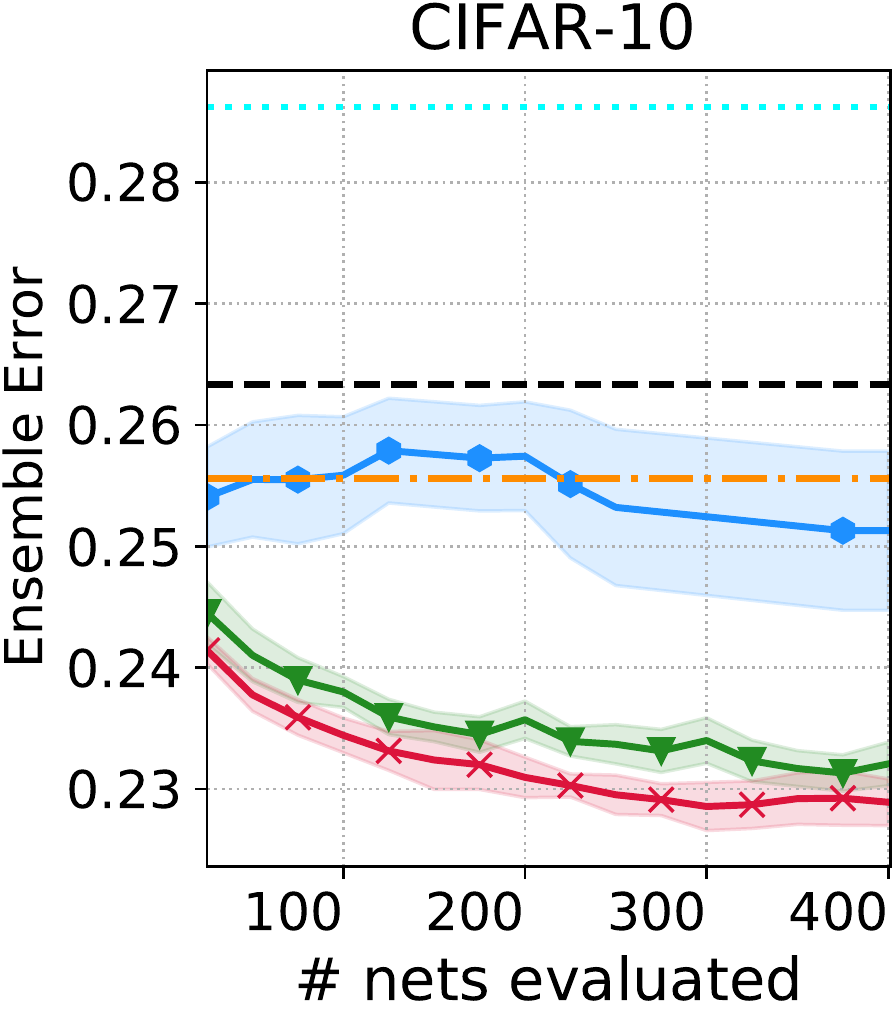}
        \includegraphics[width=.29\linewidth]{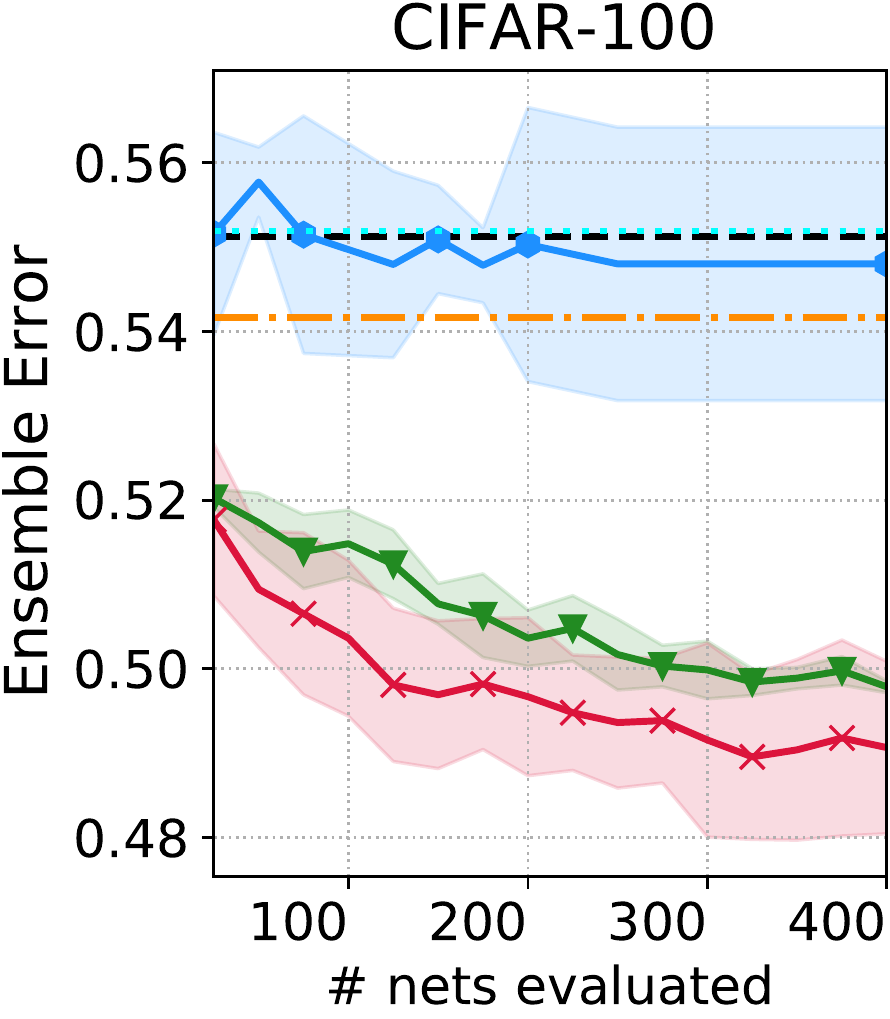}
        \includegraphics[width=.29\linewidth]{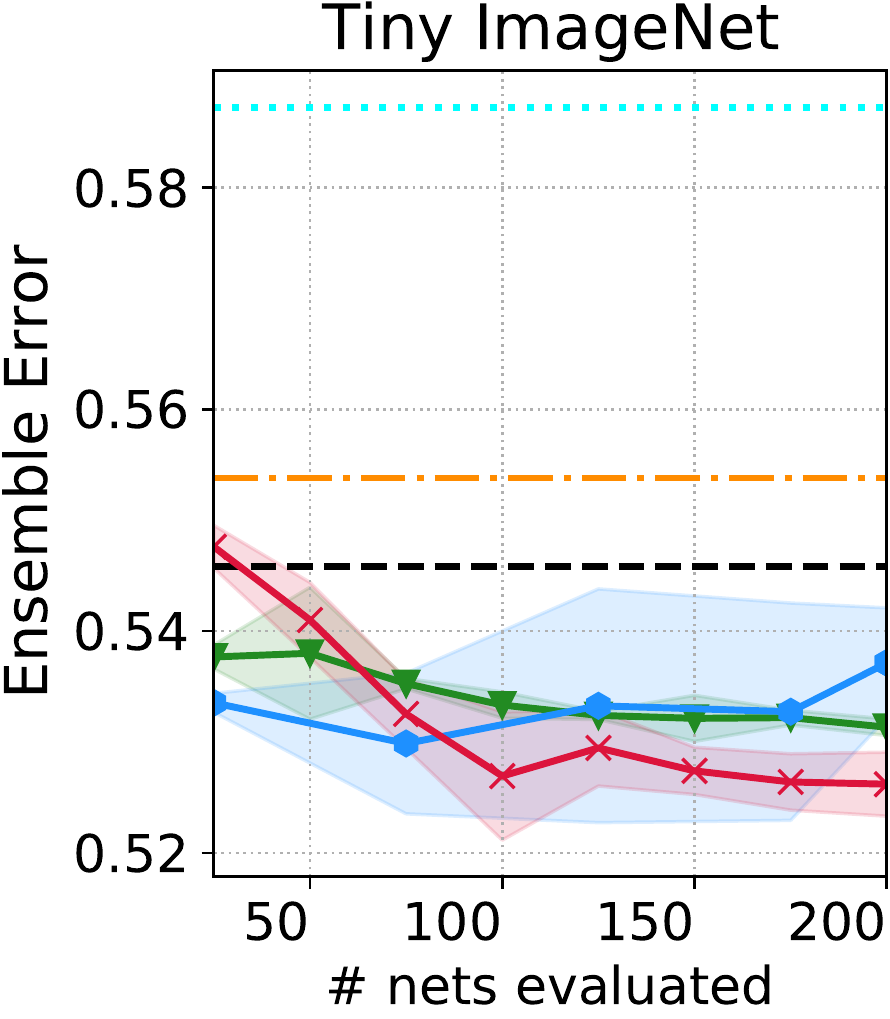}
        \subcaption{Data shift (severity 3)}
    \end{subfigure}\\ %
    \begin{subfigure}[t]{0.49\textwidth}
        \centering
        \includegraphics[width=.29\linewidth]{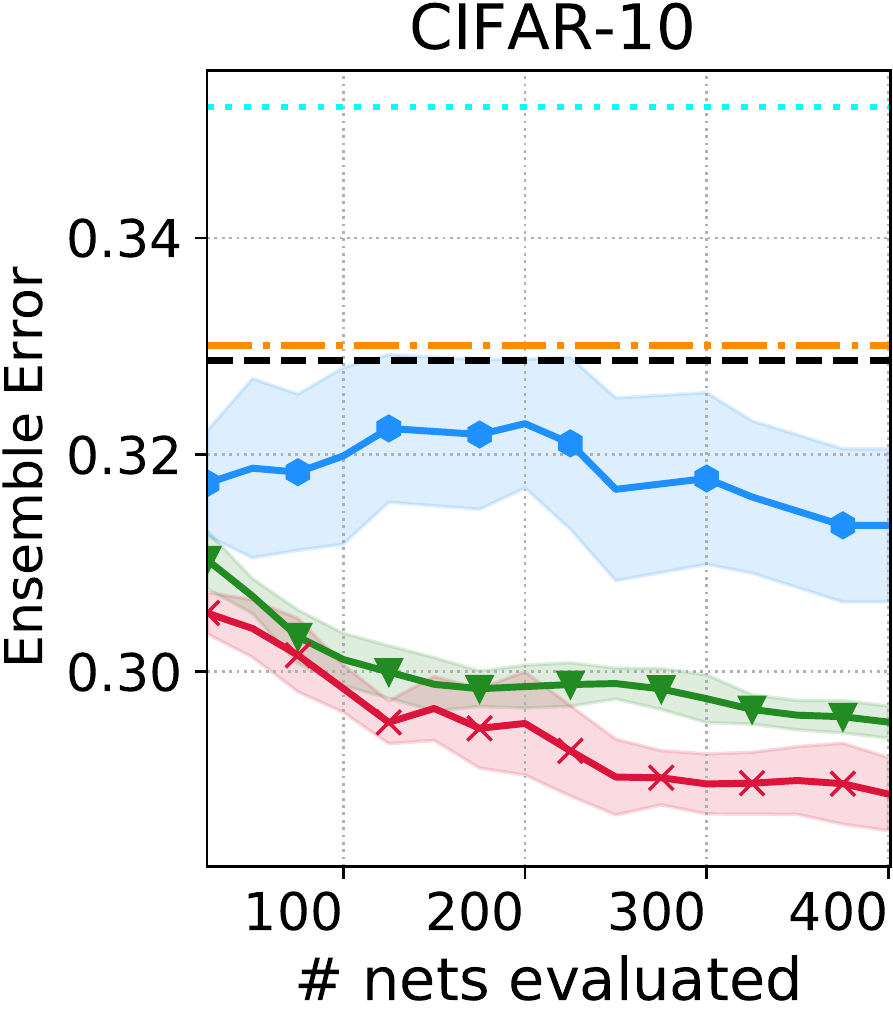}
        \includegraphics[width=.29\linewidth]{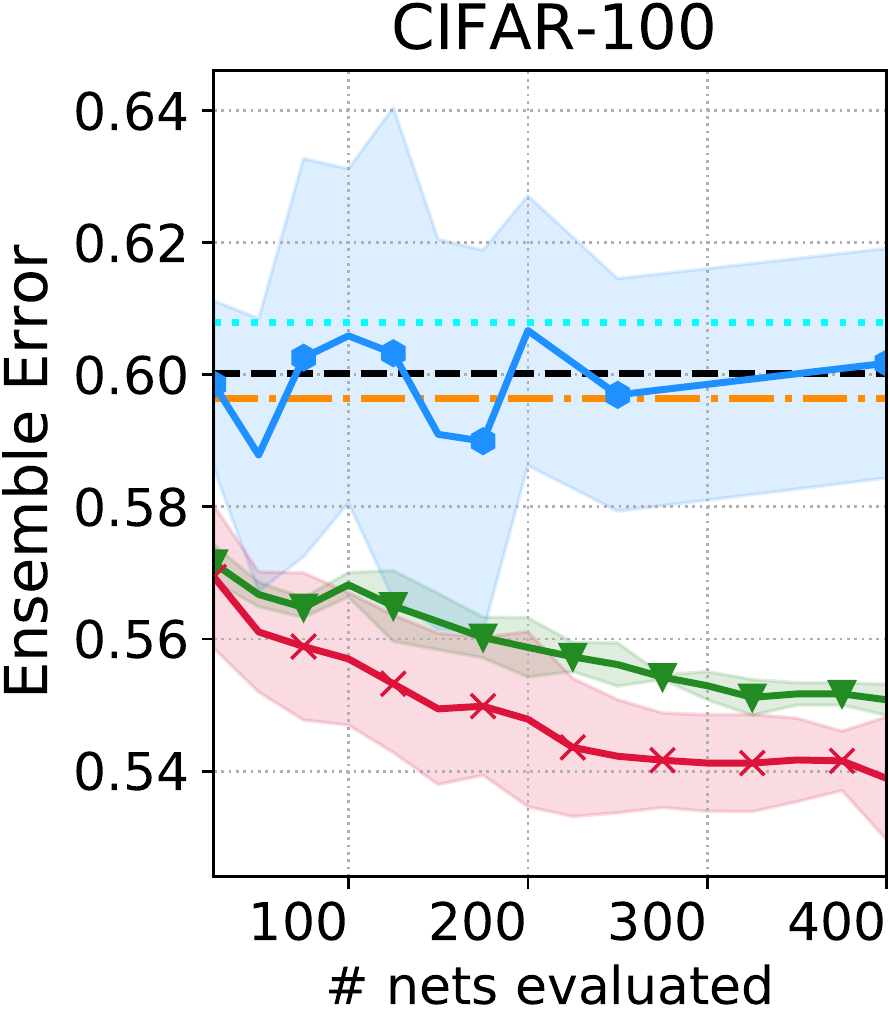}
        \includegraphics[width=.29\linewidth]{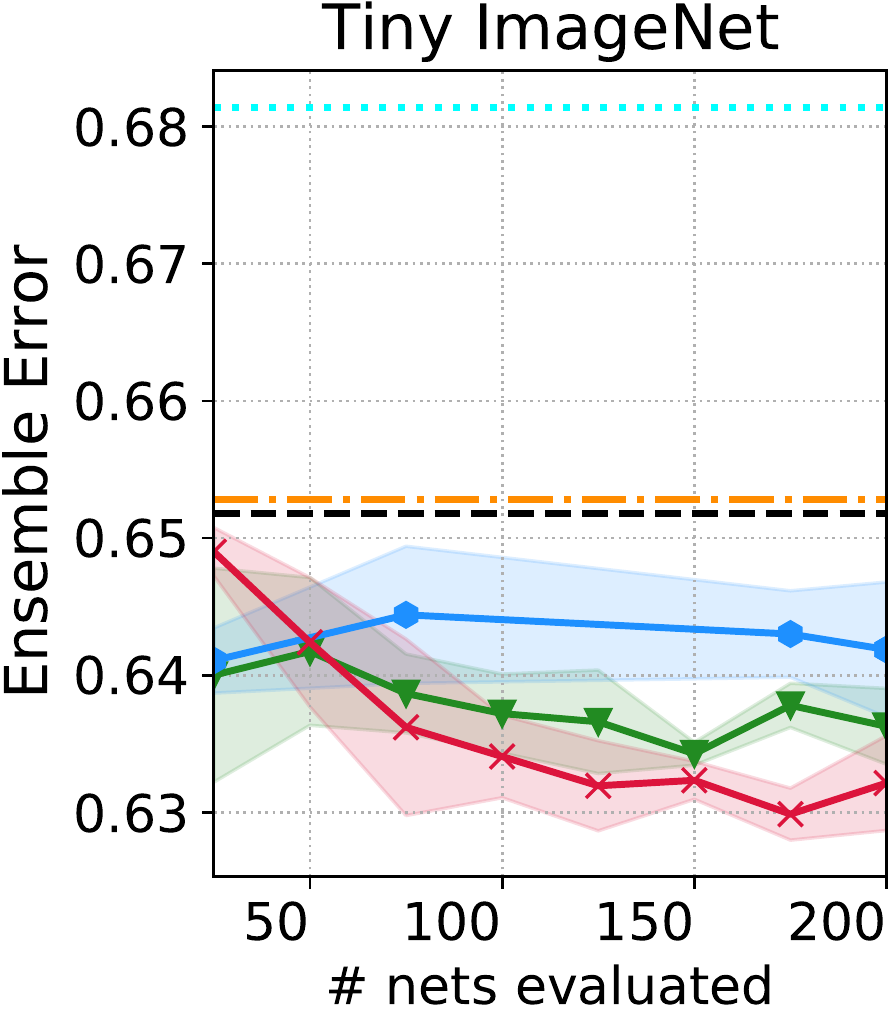}
        \subcaption{Data shift (severity 4)}
    \end{subfigure}%
    \begin{subfigure}[t]{0.49\textwidth}
        \centering
        \includegraphics[width=.29\linewidth]{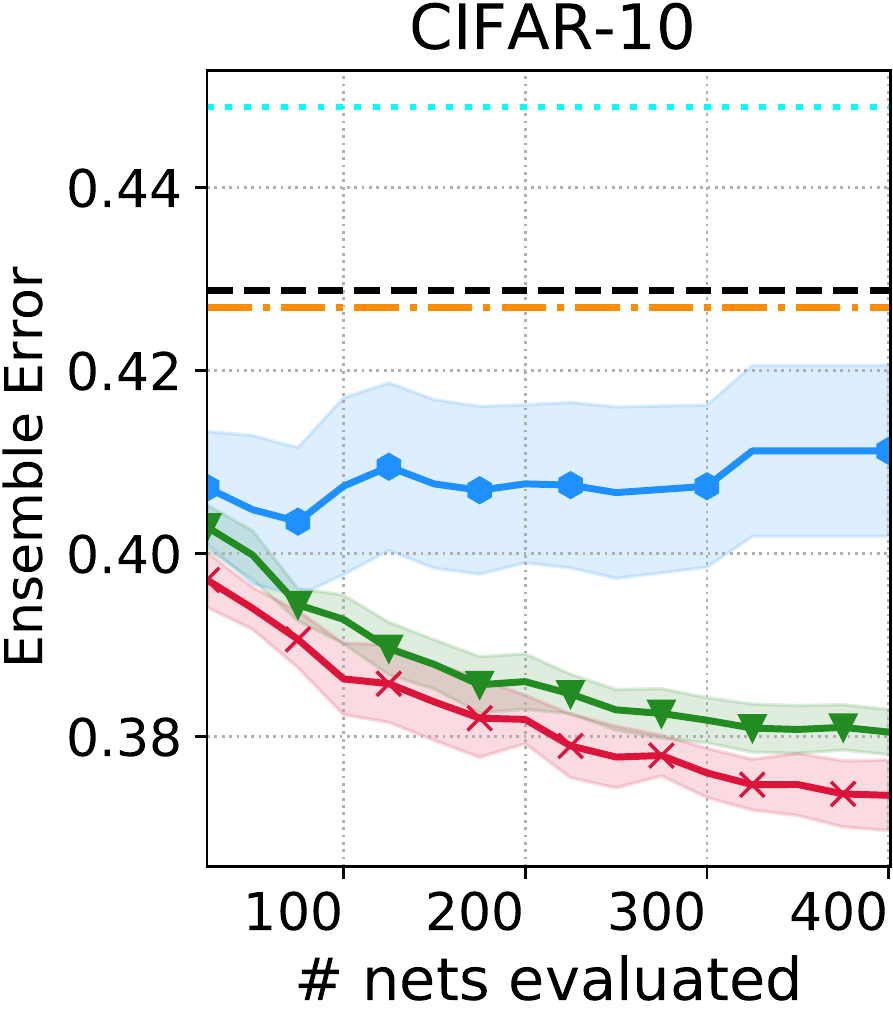}
        \includegraphics[width=.29\linewidth]{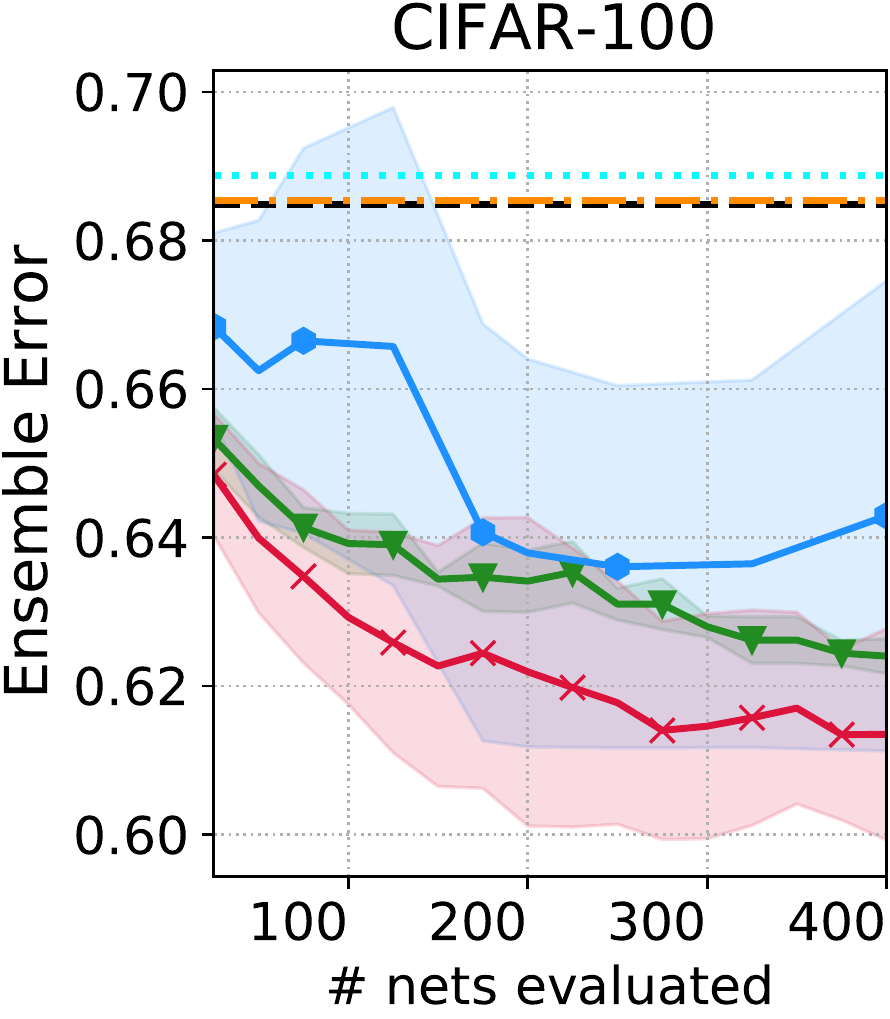}
        \includegraphics[width=.29\linewidth]{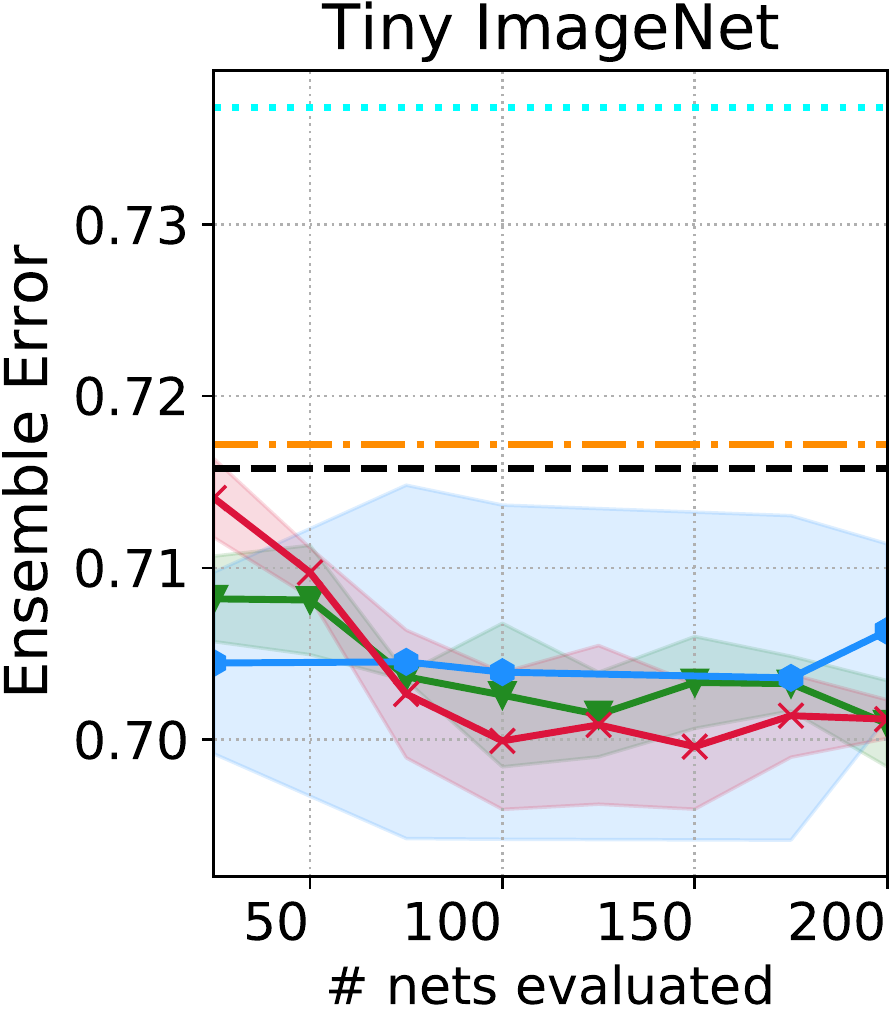}
        \subcaption{Data shift (severity 5)}
    \end{subfigure}
    
    \caption{Ensemble error vs. budget $\budget$. Ensemble size fixed at $M = 10$.}
    \label{fig:test_error_budget_other}
\end{figure*}

\begin{figure*}
    \centering
    \captionsetup[subfigure]{justification=centering}
    \begin{subfigure}[t]{0.49\textwidth}
        \centering
        \includegraphics[width=.29\linewidth]{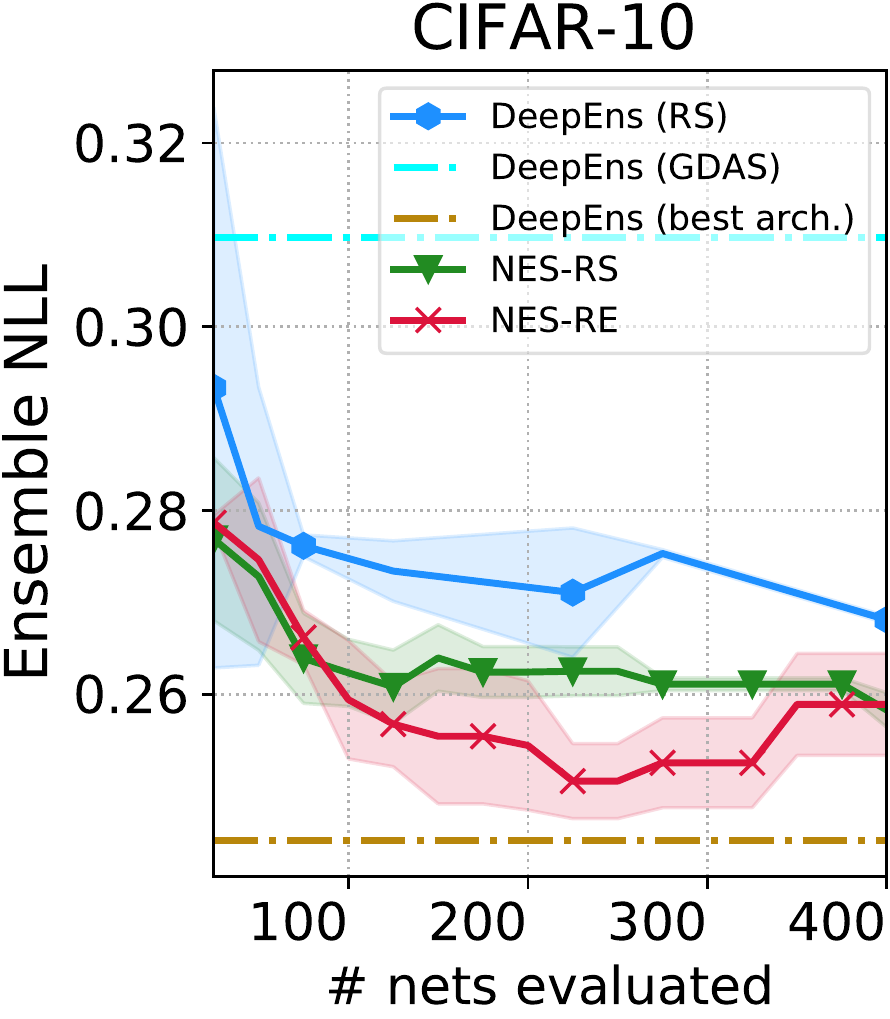}
        \includegraphics[width=.29\linewidth]{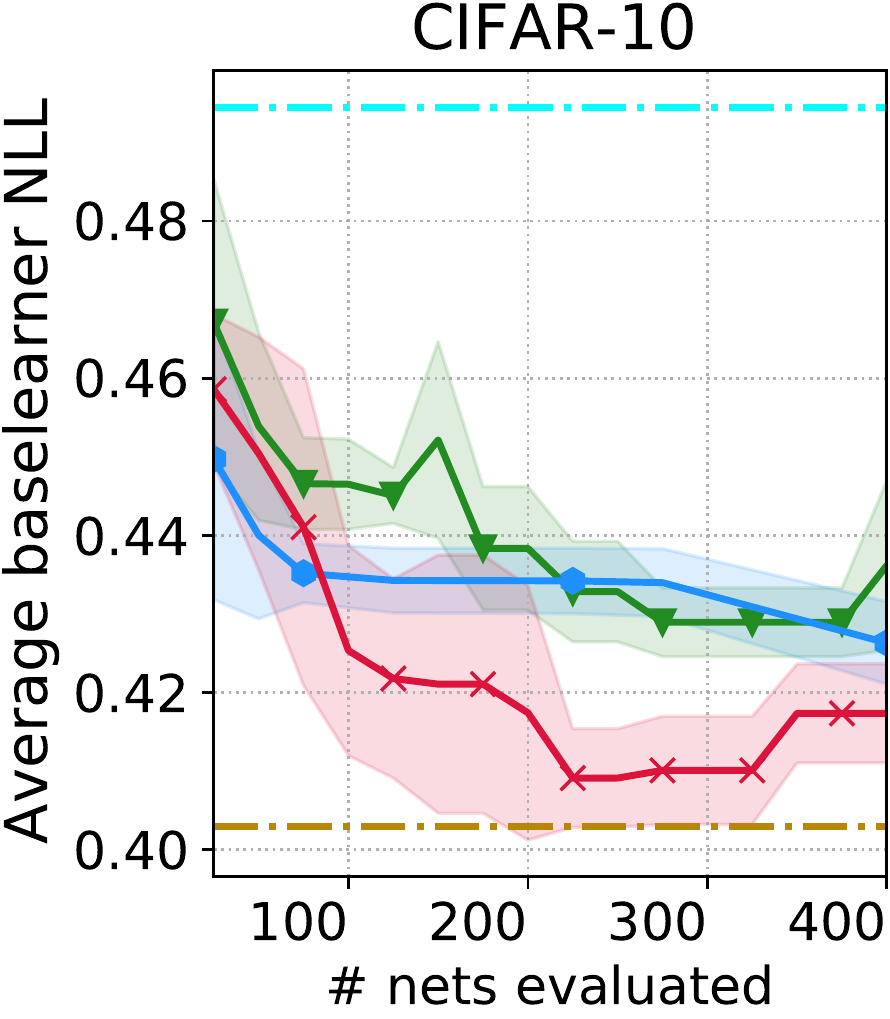}
        \includegraphics[width=.29\linewidth]{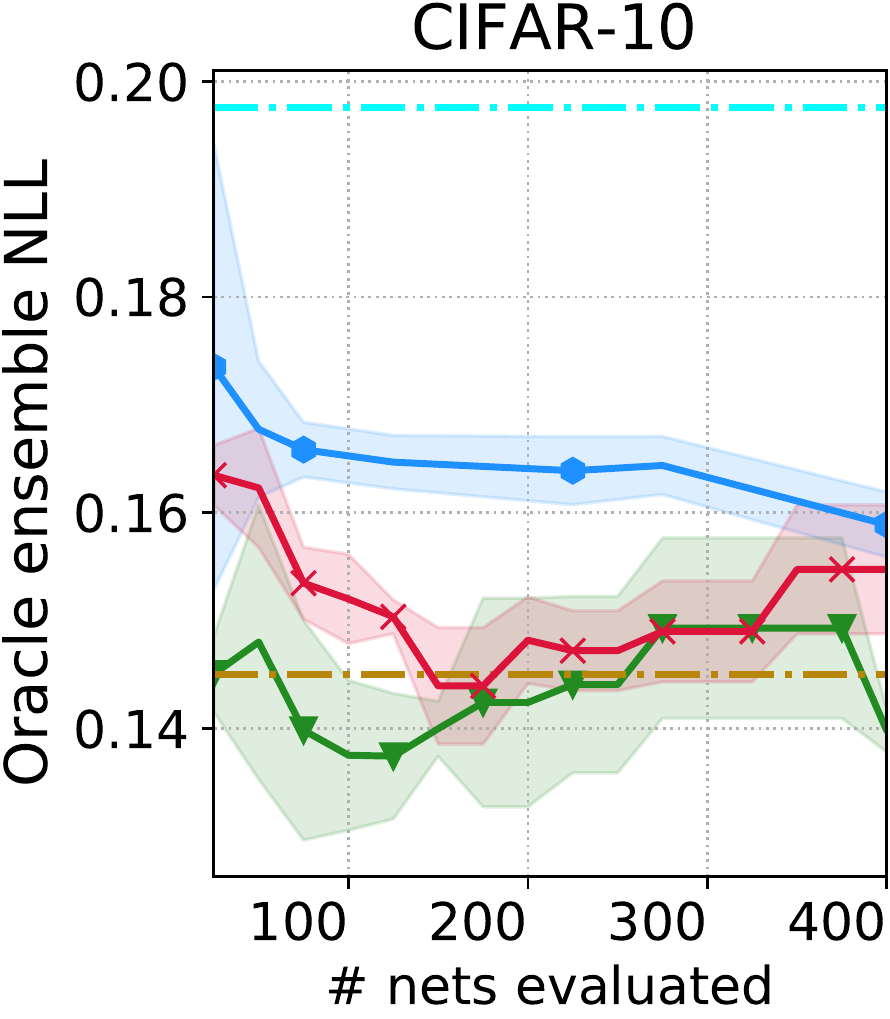}
        \subcaption{No data shift}
    \end{subfigure}%
    \begin{subfigure}[t]{0.49\textwidth}
        \centering
        \includegraphics[width=.29\linewidth]{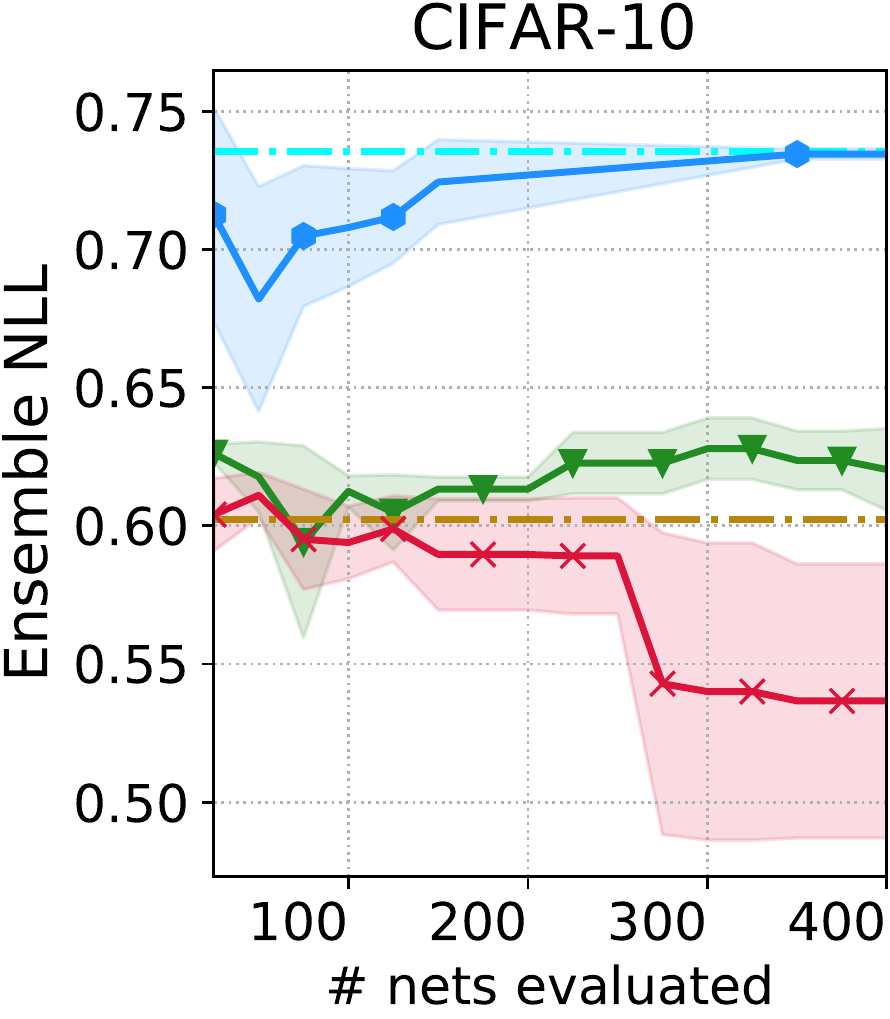}
        \includegraphics[width=.29\linewidth]{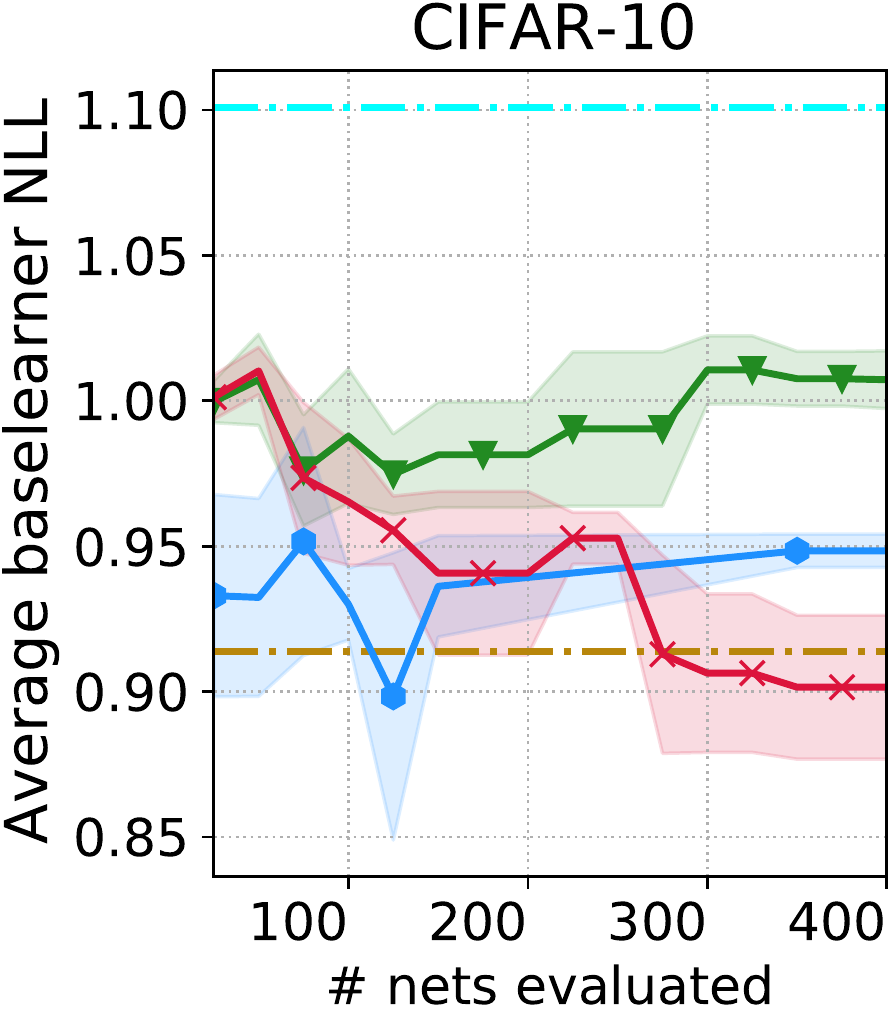}
        \includegraphics[width=.29\linewidth]{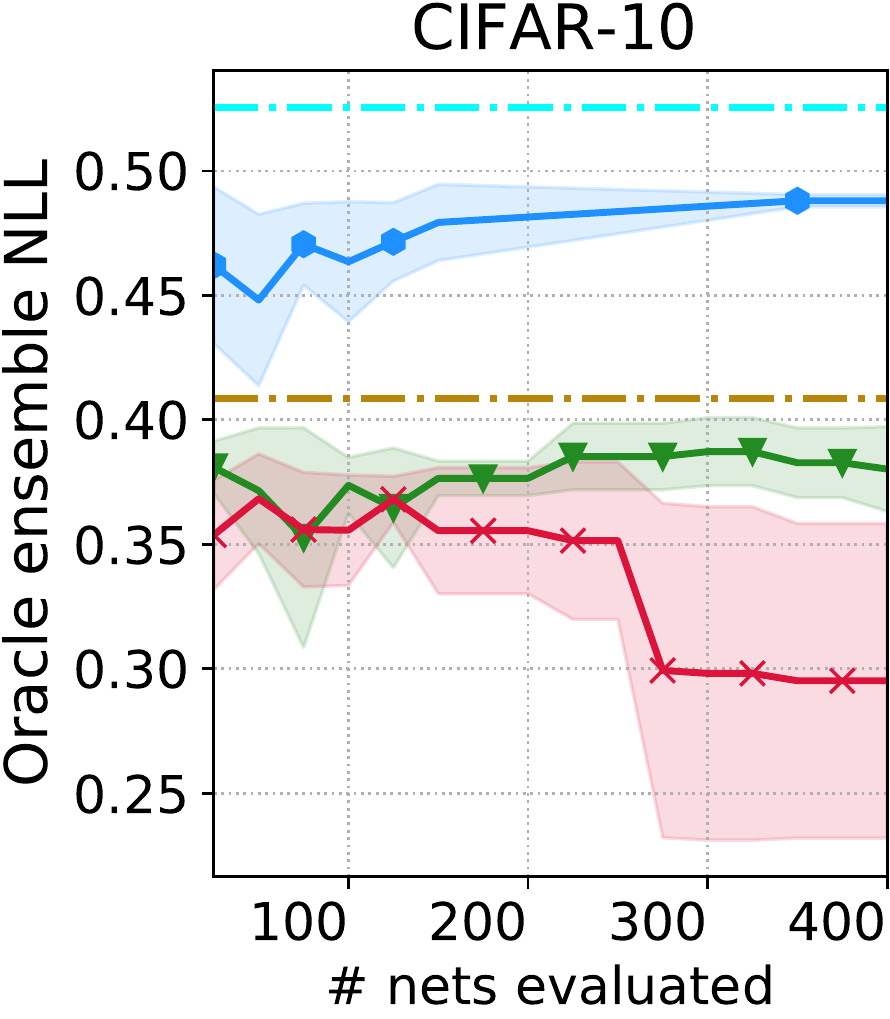}
        \subcaption{Data shift (severity 1)}
    \end{subfigure}\\ %
    \begin{subfigure}[t]{0.49\textwidth}
        \centering
        \includegraphics[width=.29\linewidth]{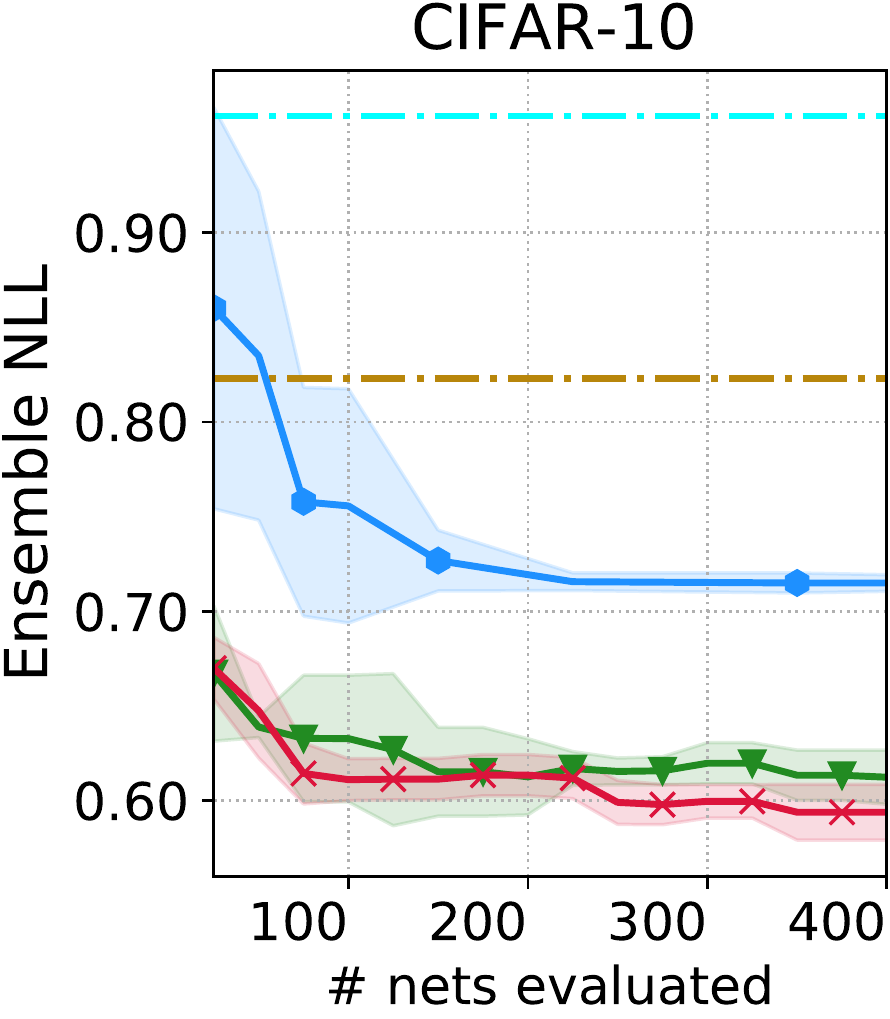}
        \includegraphics[width=.29\linewidth]{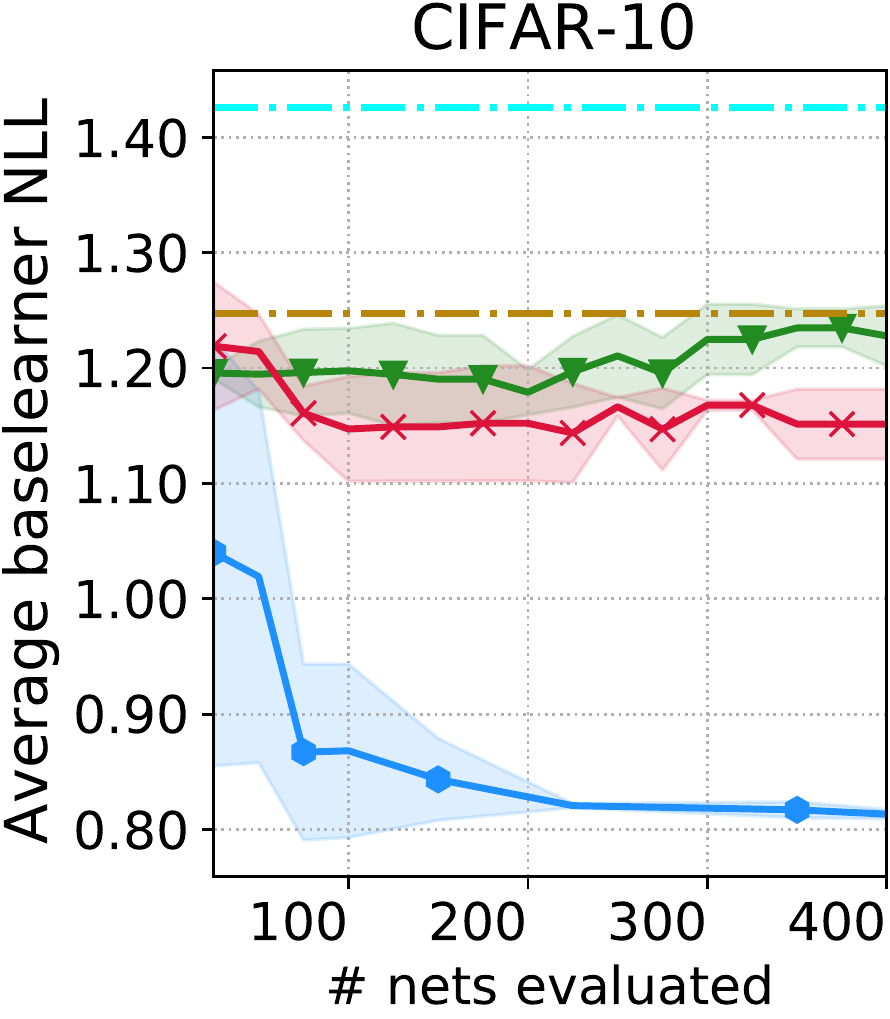}
        \includegraphics[width=.29\linewidth]{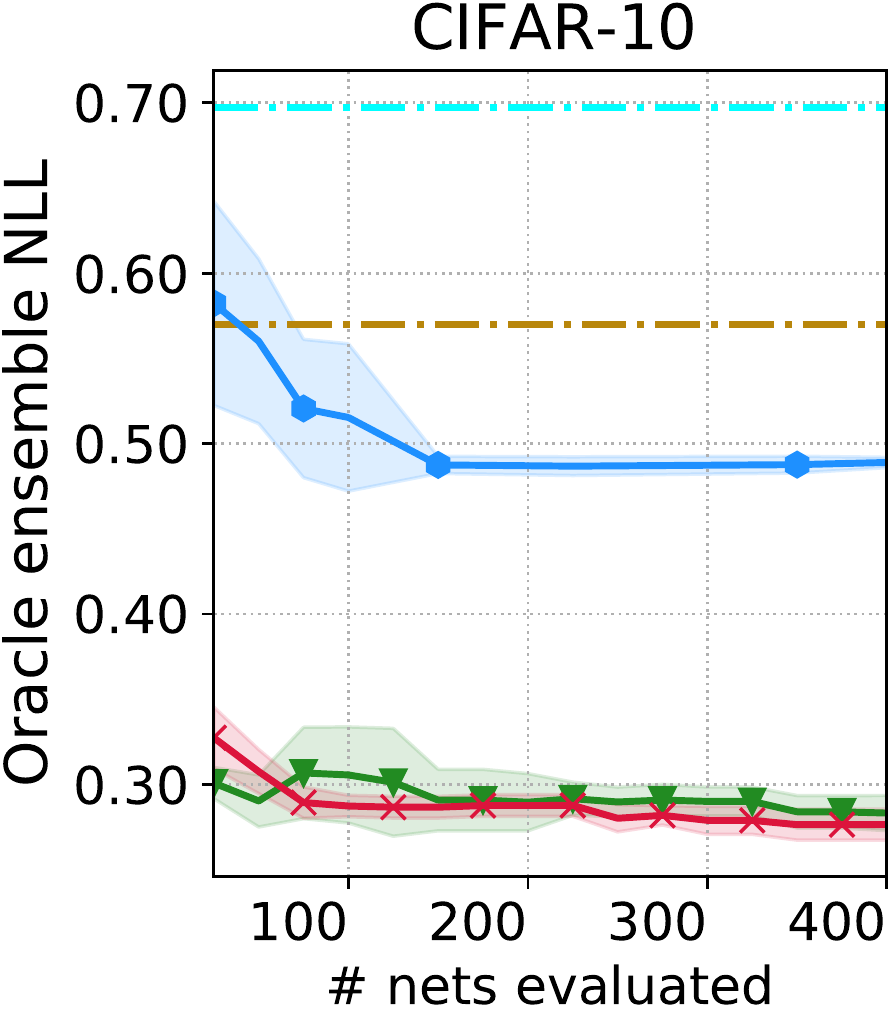}
        \subcaption{Data shift (severity 2)}
    \end{subfigure}%
    \begin{subfigure}[t]{0.49\textwidth}
        \centering
        \includegraphics[width=.29\linewidth]{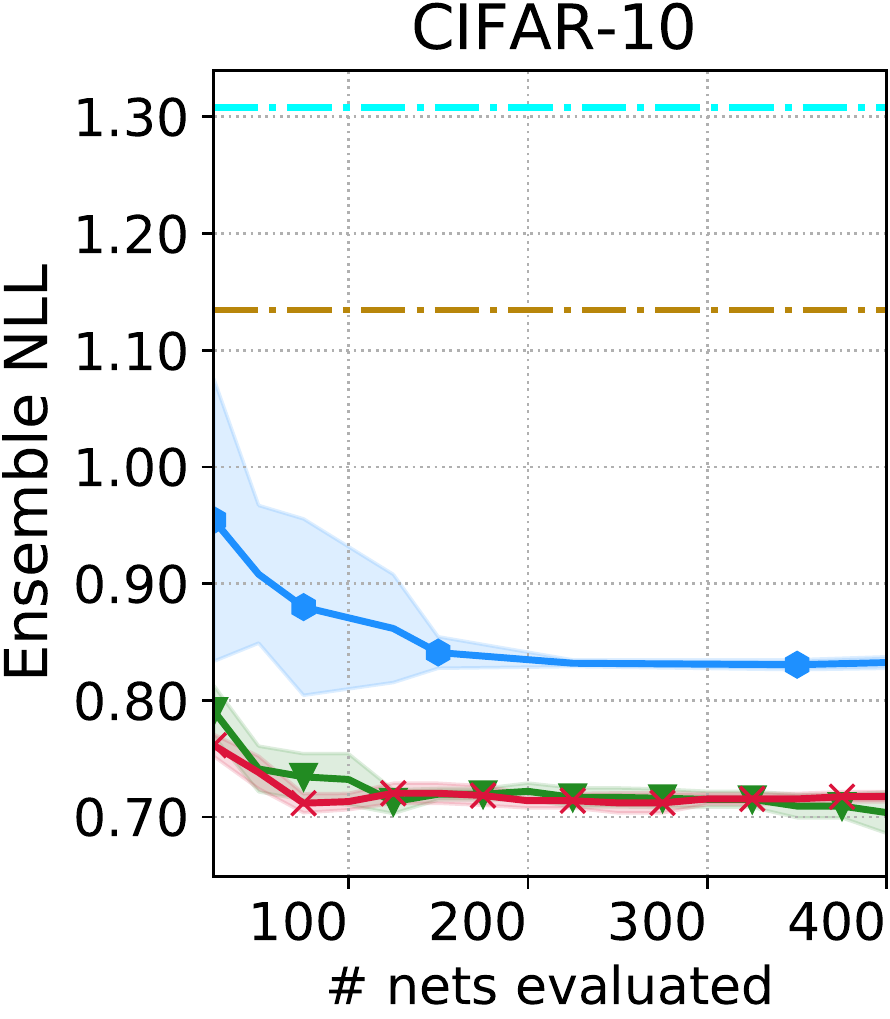}
        \includegraphics[width=.29\linewidth]{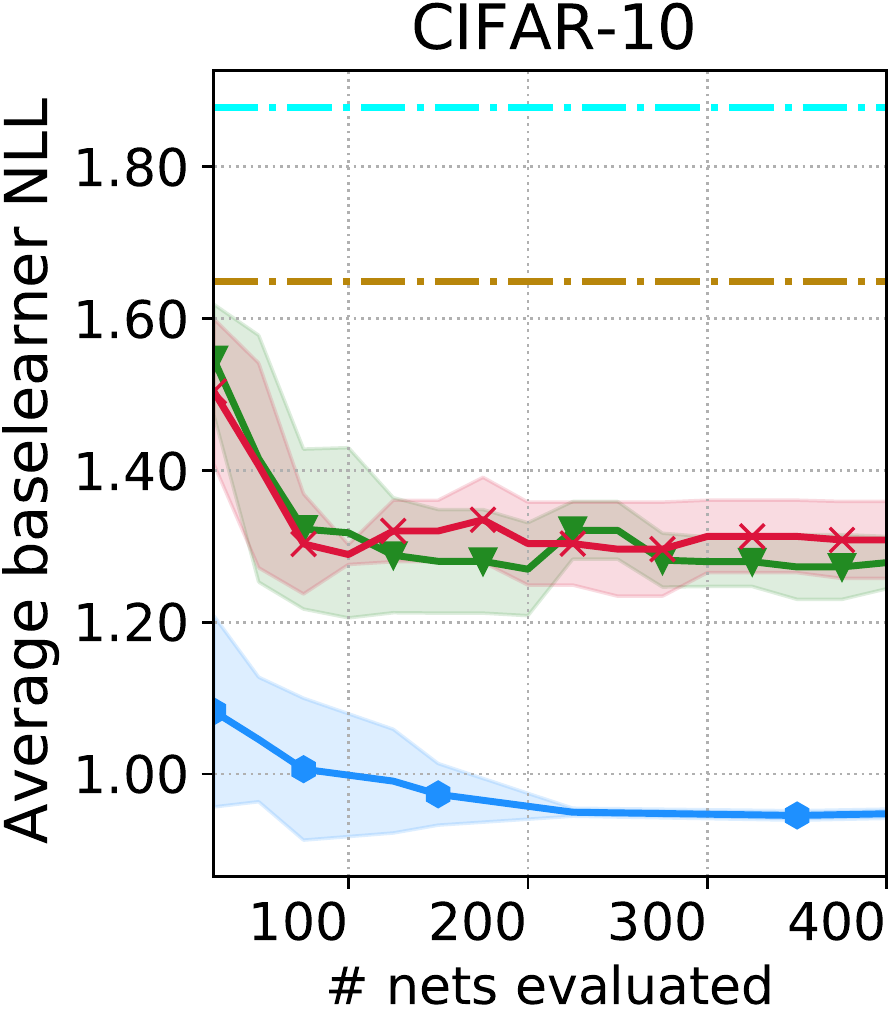}
        \includegraphics[width=.29\linewidth]{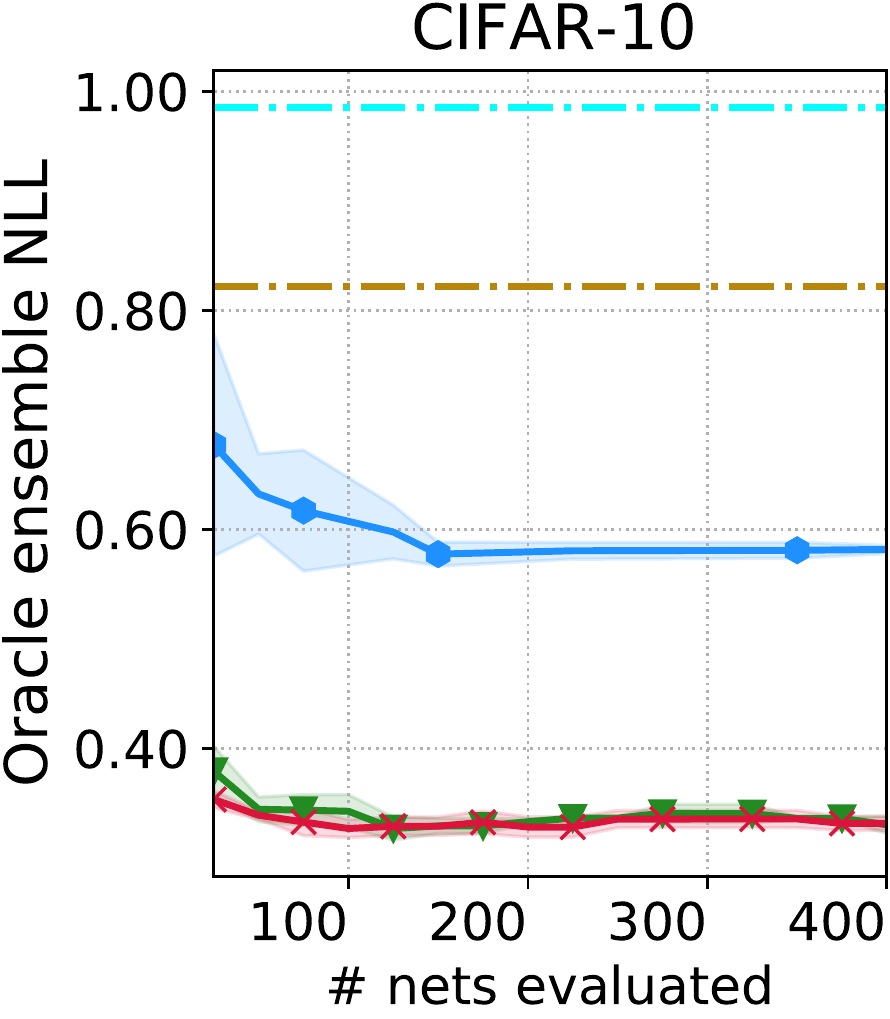}
        \subcaption{Data shift (severity 3)}
    \end{subfigure}\\ %
    \begin{subfigure}[t]{0.49\textwidth}
        \centering
        \includegraphics[width=.29\linewidth]{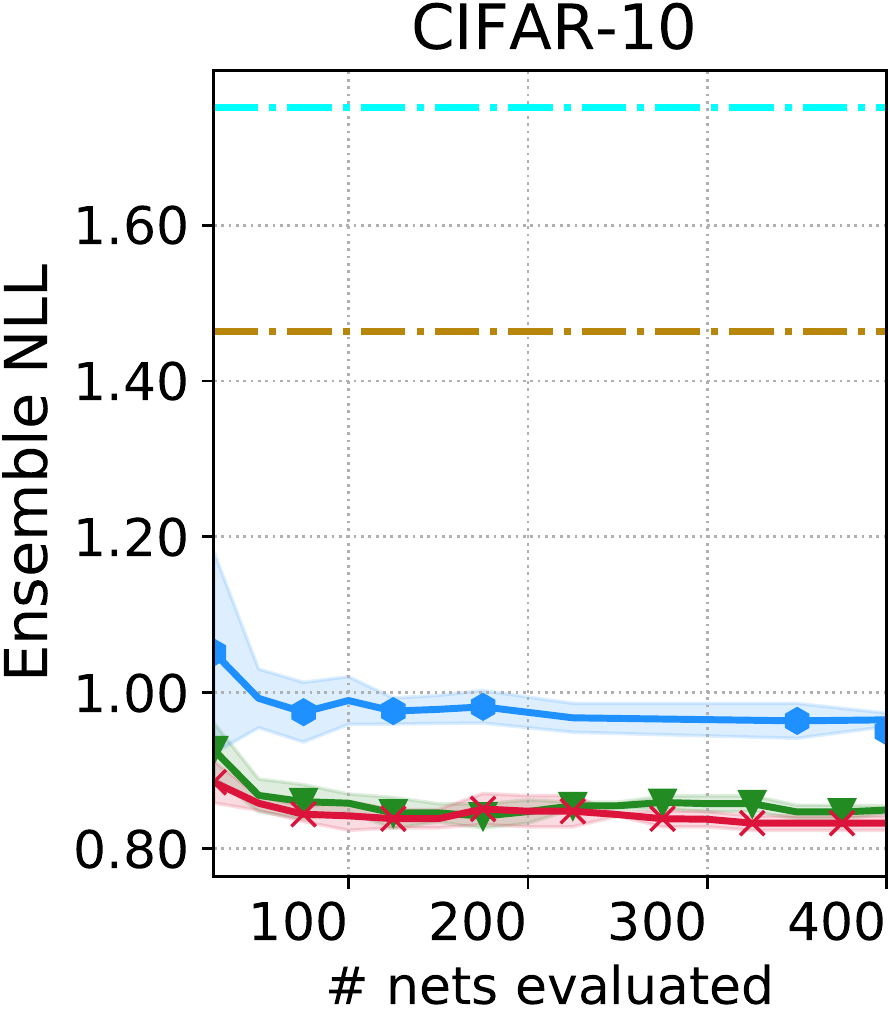}
        \includegraphics[width=.29\linewidth]{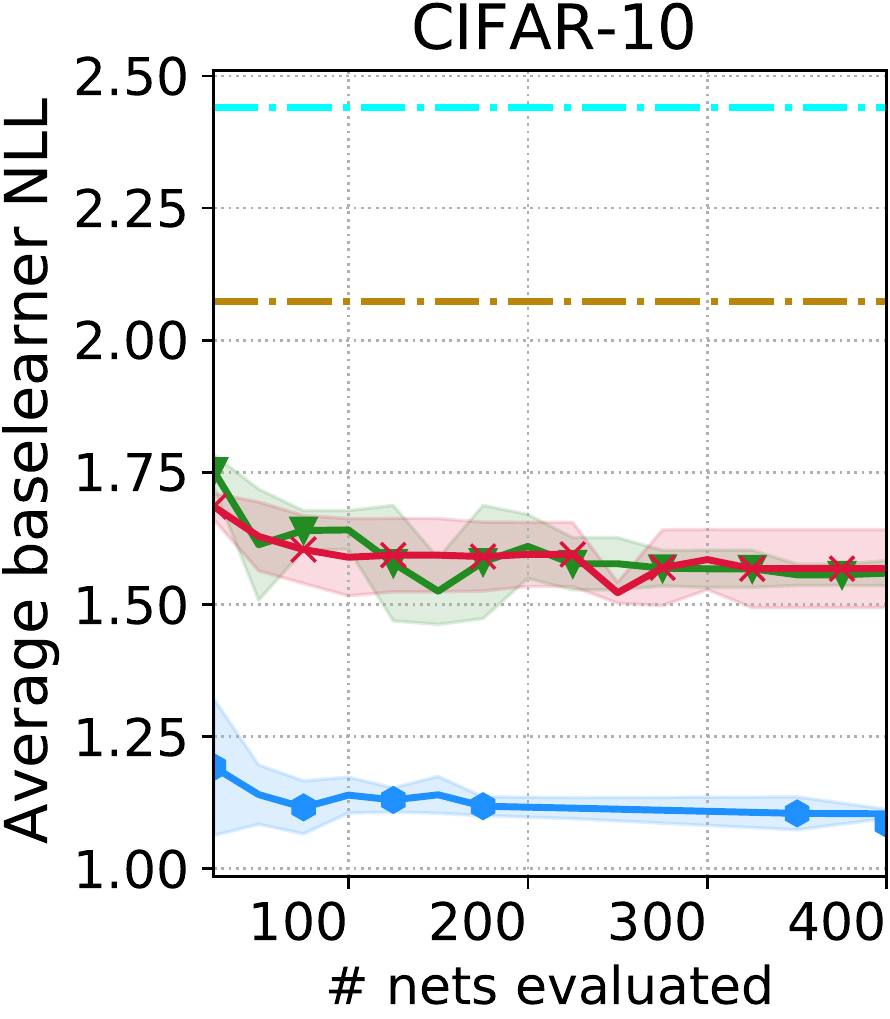}
        \includegraphics[width=.29\linewidth]{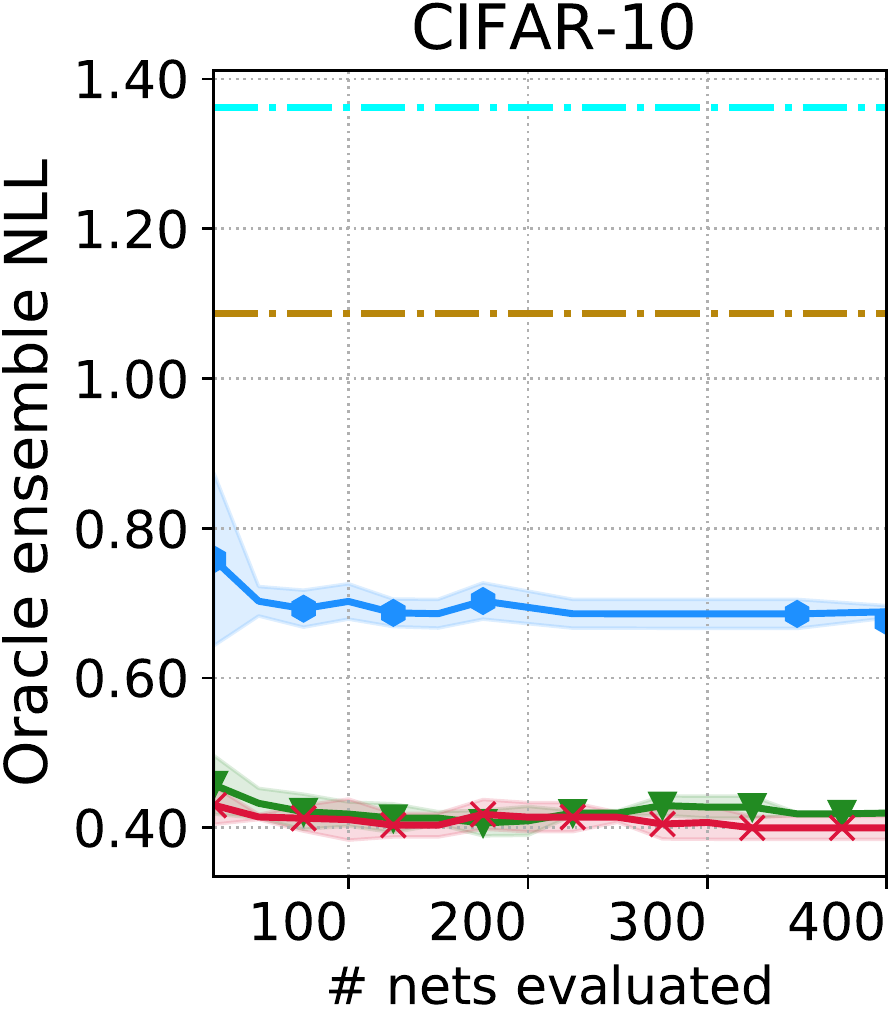}
        \subcaption{Data shift (severity 4)}
    \end{subfigure}%
    \begin{subfigure}[t]{0.49\textwidth}
        \centering
        \includegraphics[width=.29\linewidth]{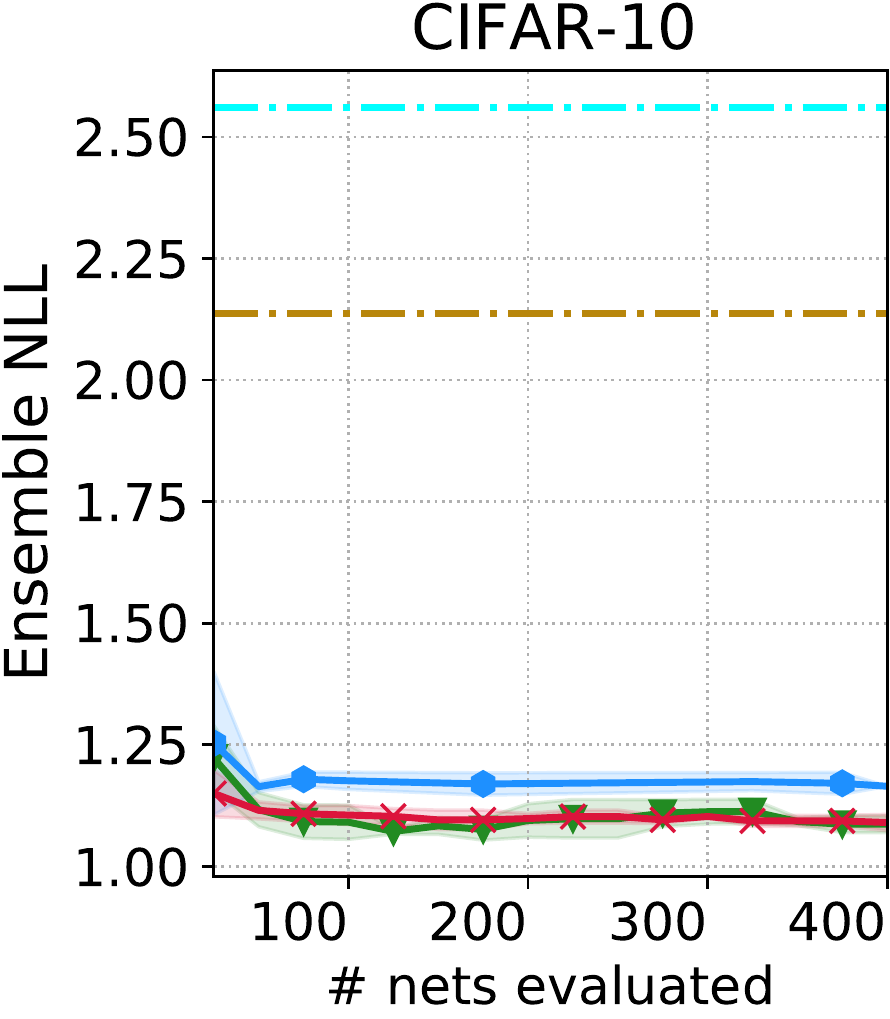}
        \includegraphics[width=.29\linewidth]{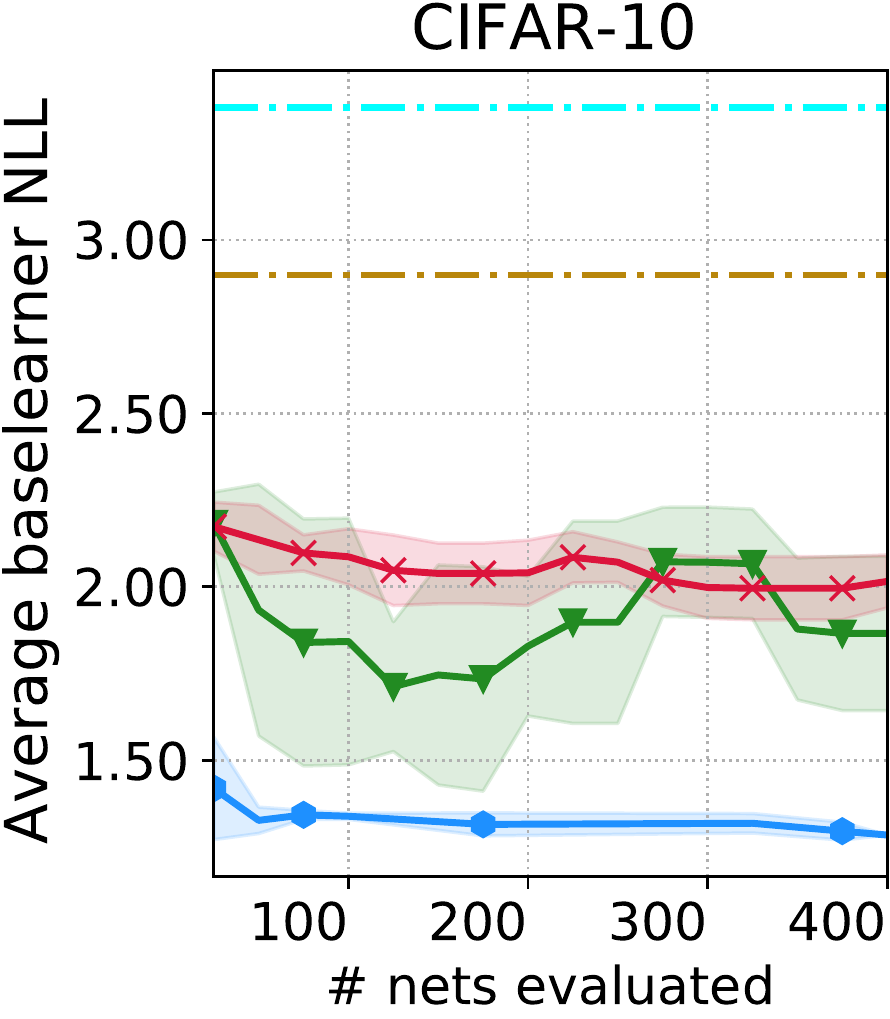}
        \includegraphics[width=.29\linewidth]{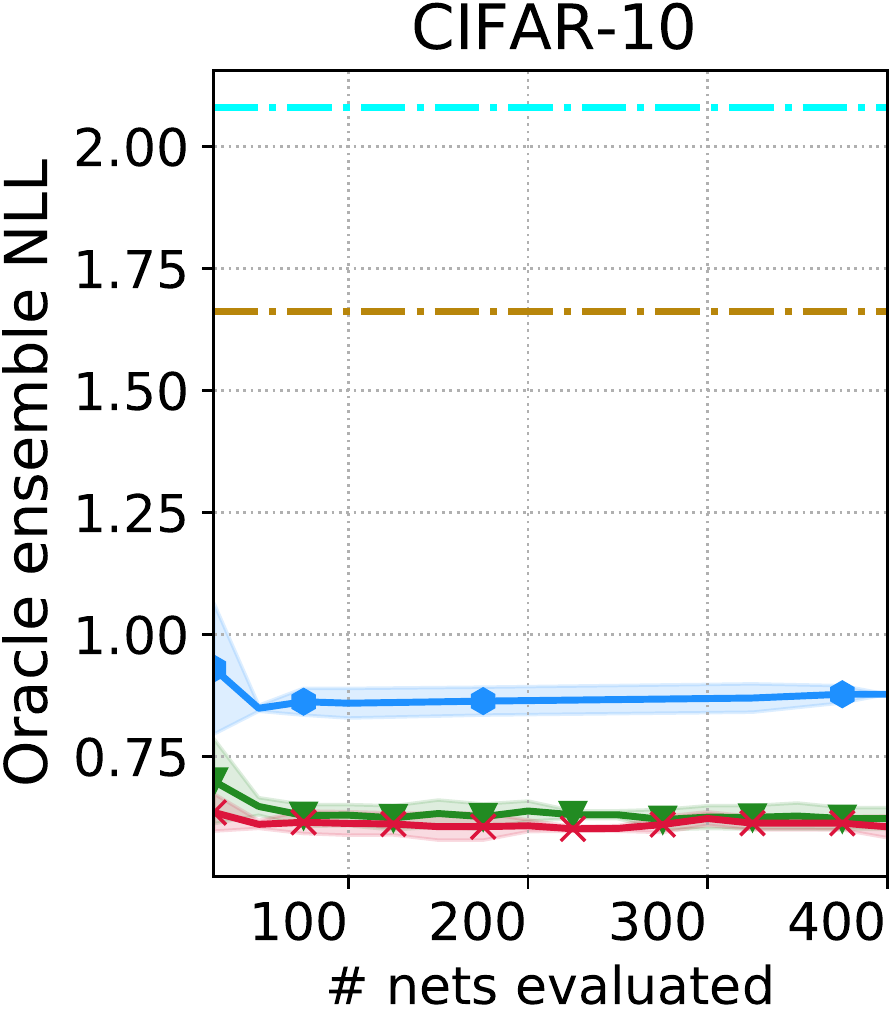}
        \subcaption{Data shift (severity 5)}
    \end{subfigure}
    
    \caption{NES and deep ensembles performance on the NAS-Bench-201 search space and CIFAR-10 dataset. Plots show ensemble NLL, average baselearner NLL and oracle ensemble NLL vs. budget $\budget$. Ensemble size fixed at $M = 3$.}
    \label{fig:nb201_c10_budget}
\end{figure*}

\begin{figure*}
    \centering
    \captionsetup[subfigure]{justification=centering}
    \begin{subfigure}[t]{0.49\textwidth}
        \centering
        \includegraphics[width=.29\linewidth]{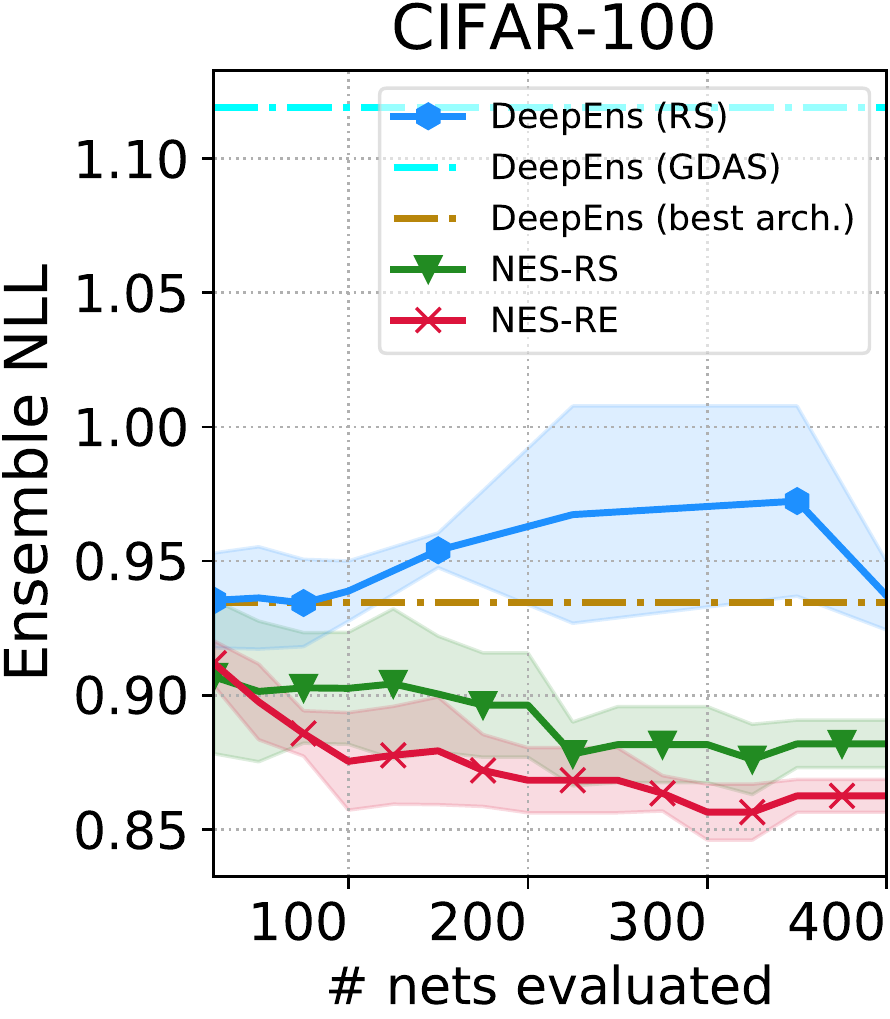}
        \includegraphics[width=.29\linewidth]{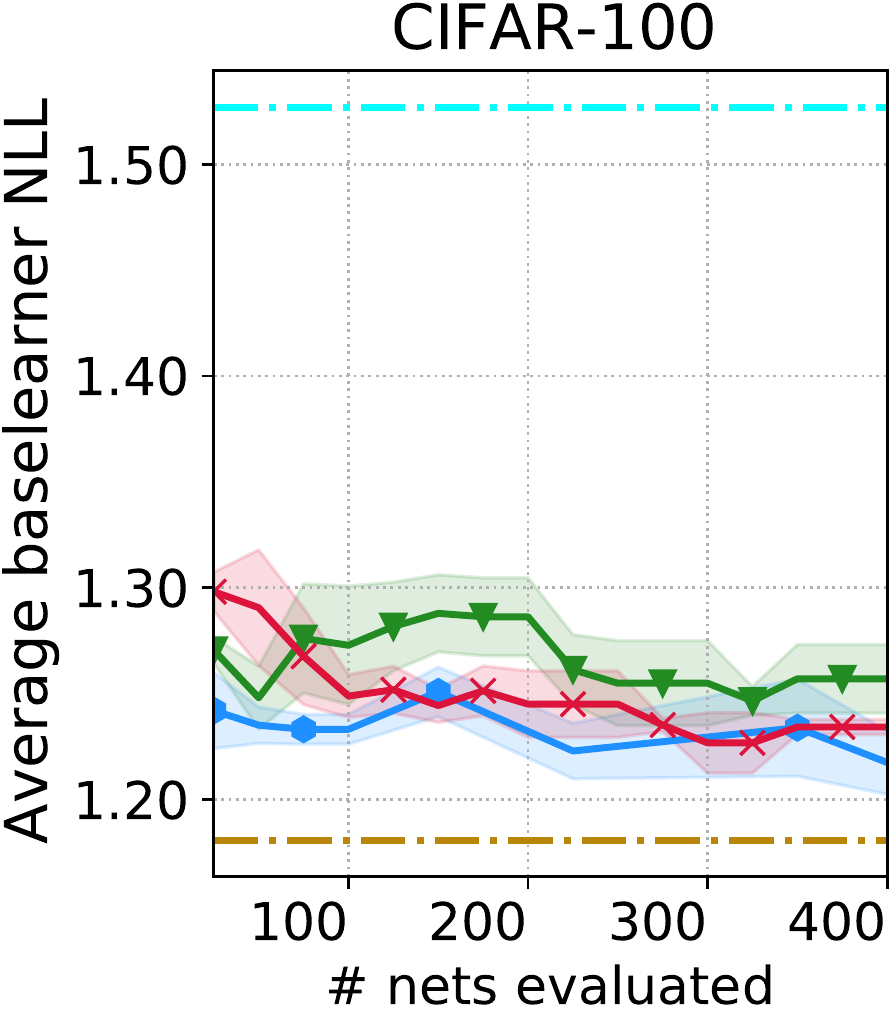}
        \includegraphics[width=.29\linewidth]{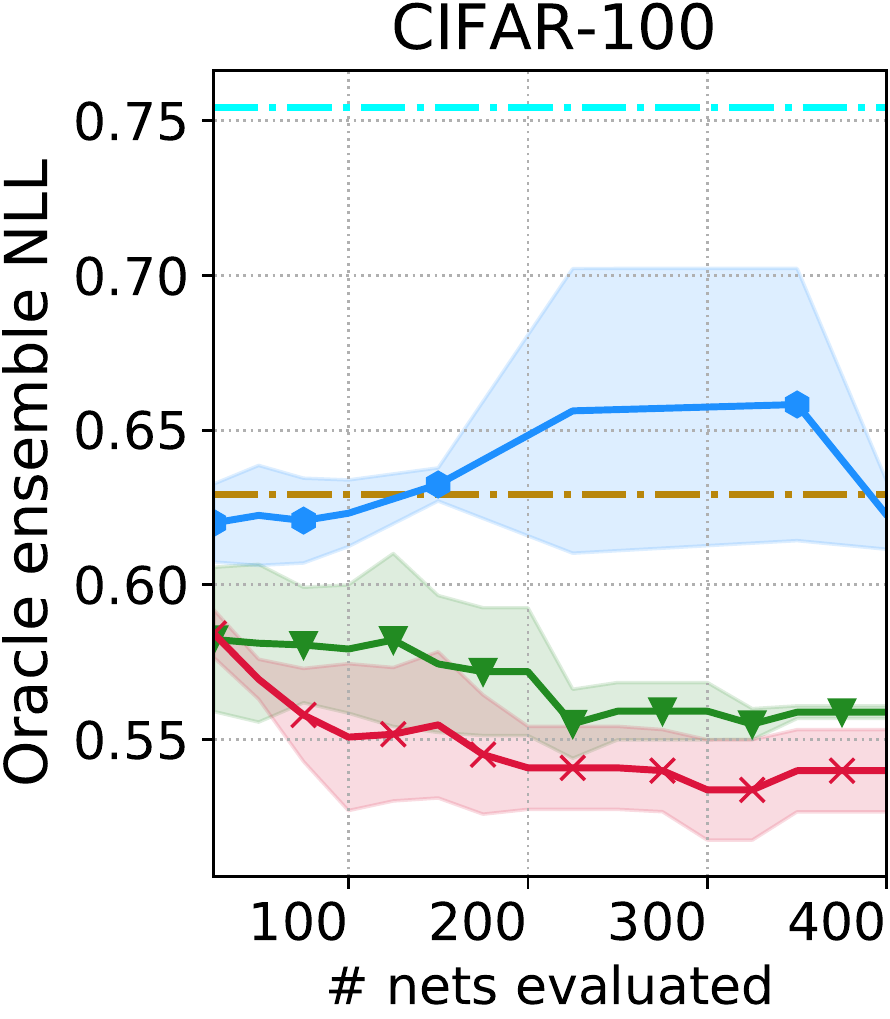}
        \subcaption{No data shift}
    \end{subfigure}%
    \begin{subfigure}[t]{0.49\textwidth}
        \centering
        \includegraphics[width=.29\linewidth]{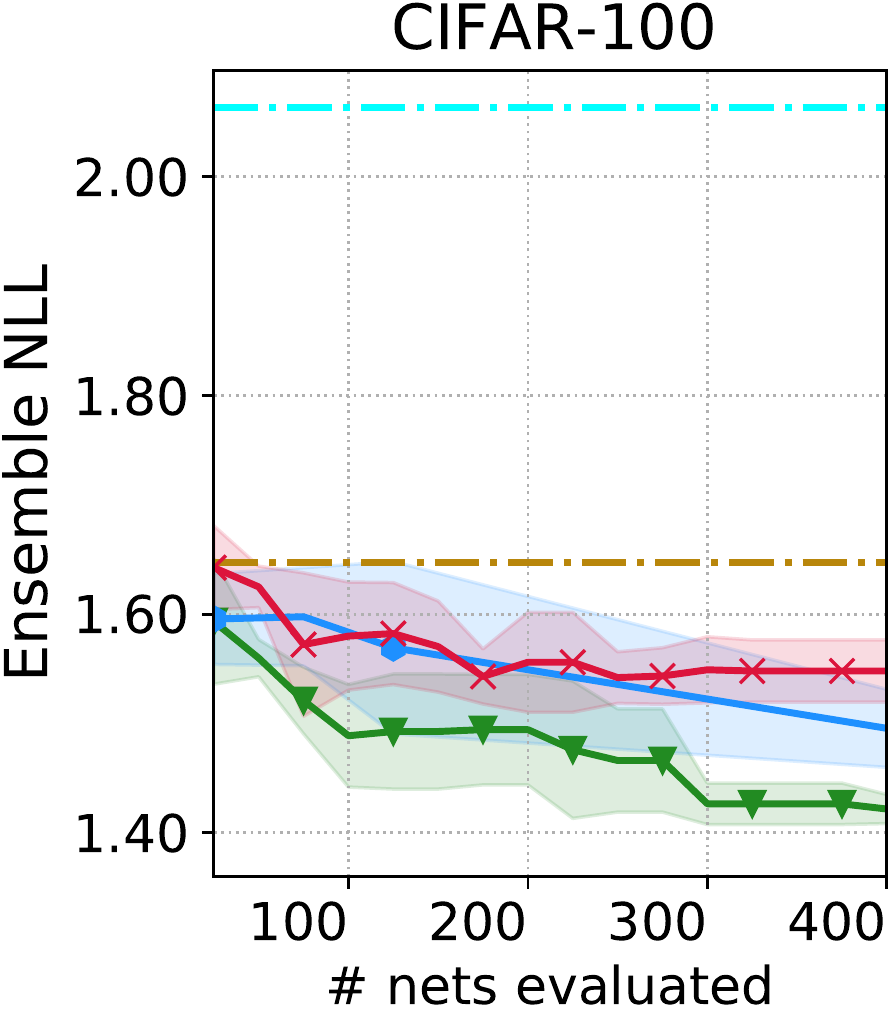}
        \includegraphics[width=.29\linewidth]{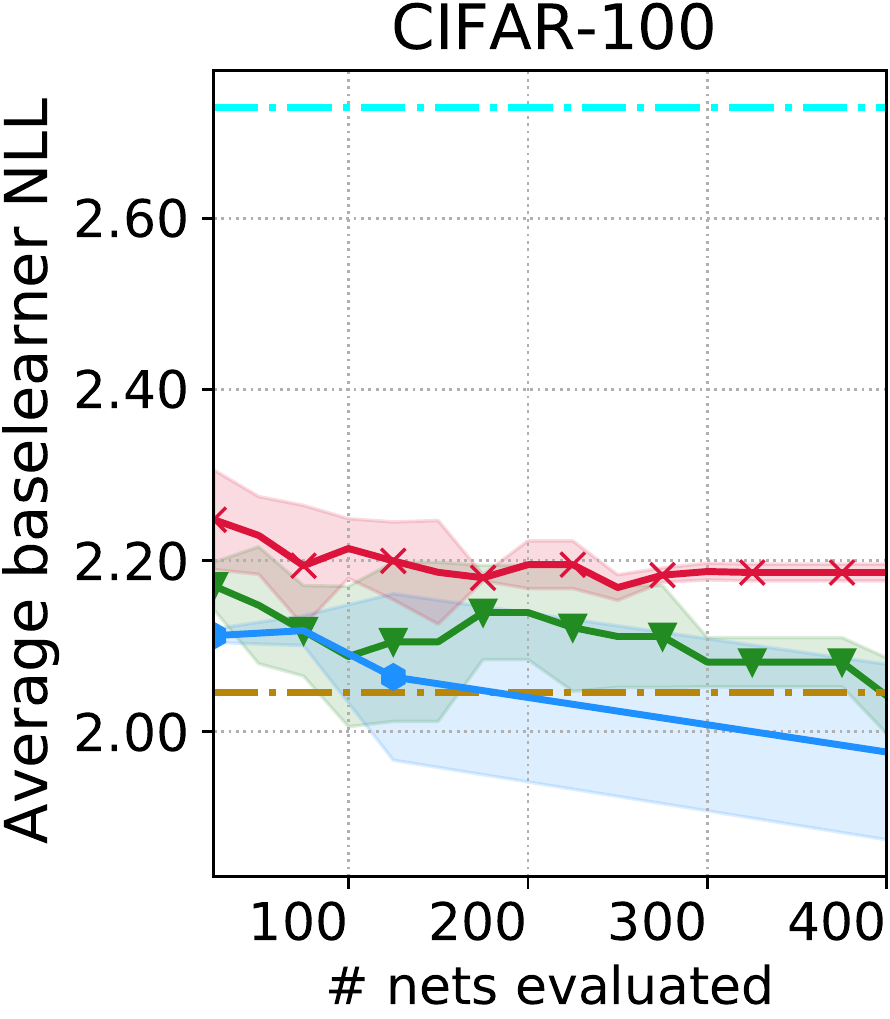}
        \includegraphics[width=.29\linewidth]{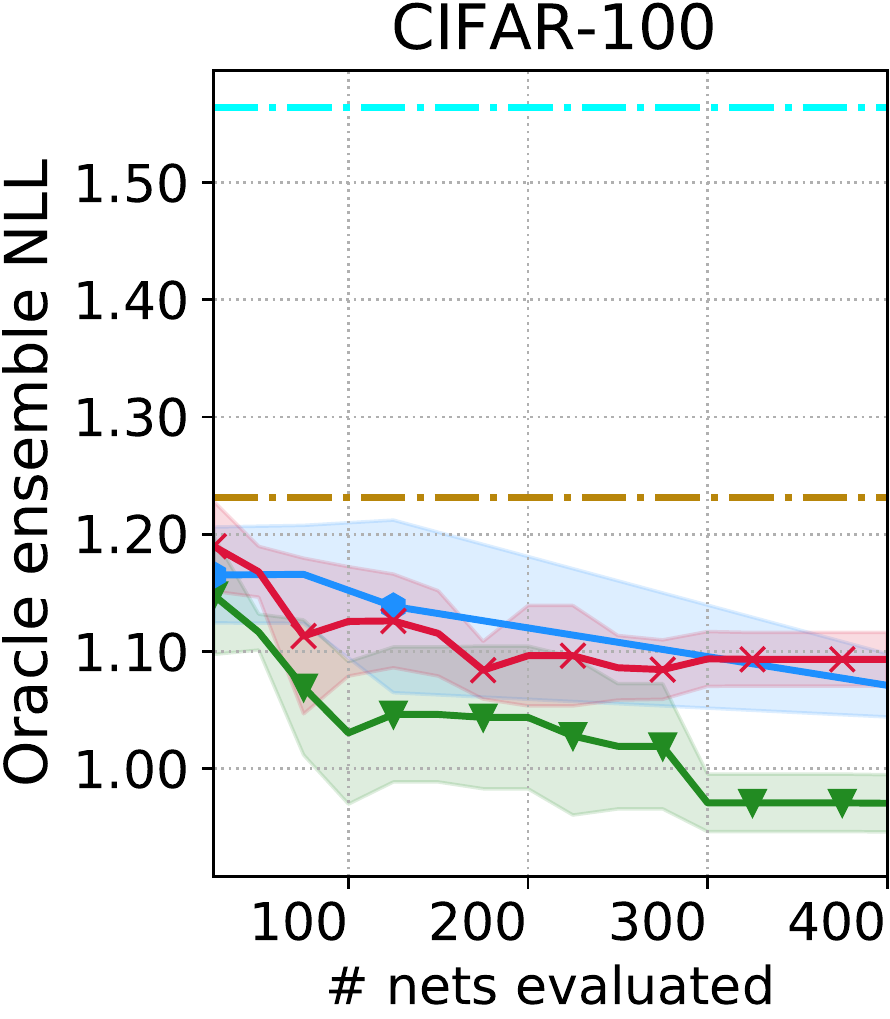}
        \subcaption{Data shift (severity 1)}
    \end{subfigure}\\ %
    \begin{subfigure}[t]{0.49\textwidth}
        \centering
        \includegraphics[width=.29\linewidth]{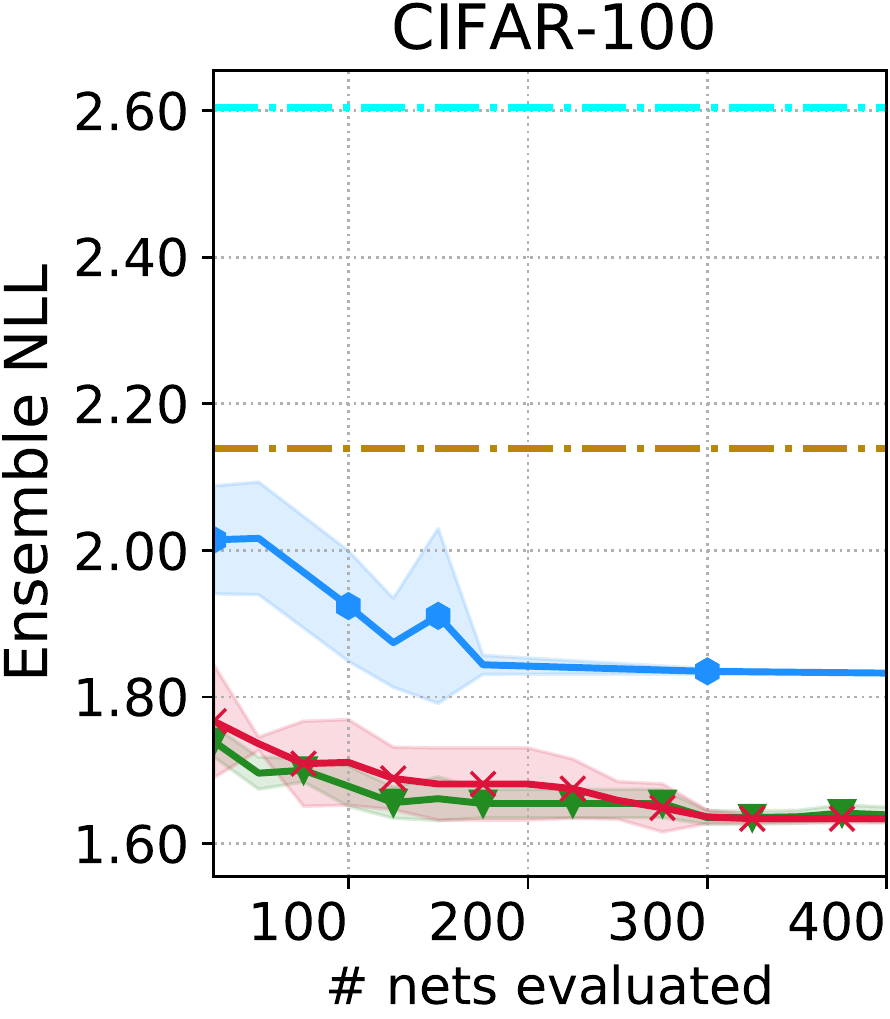}
        \includegraphics[width=.29\linewidth]{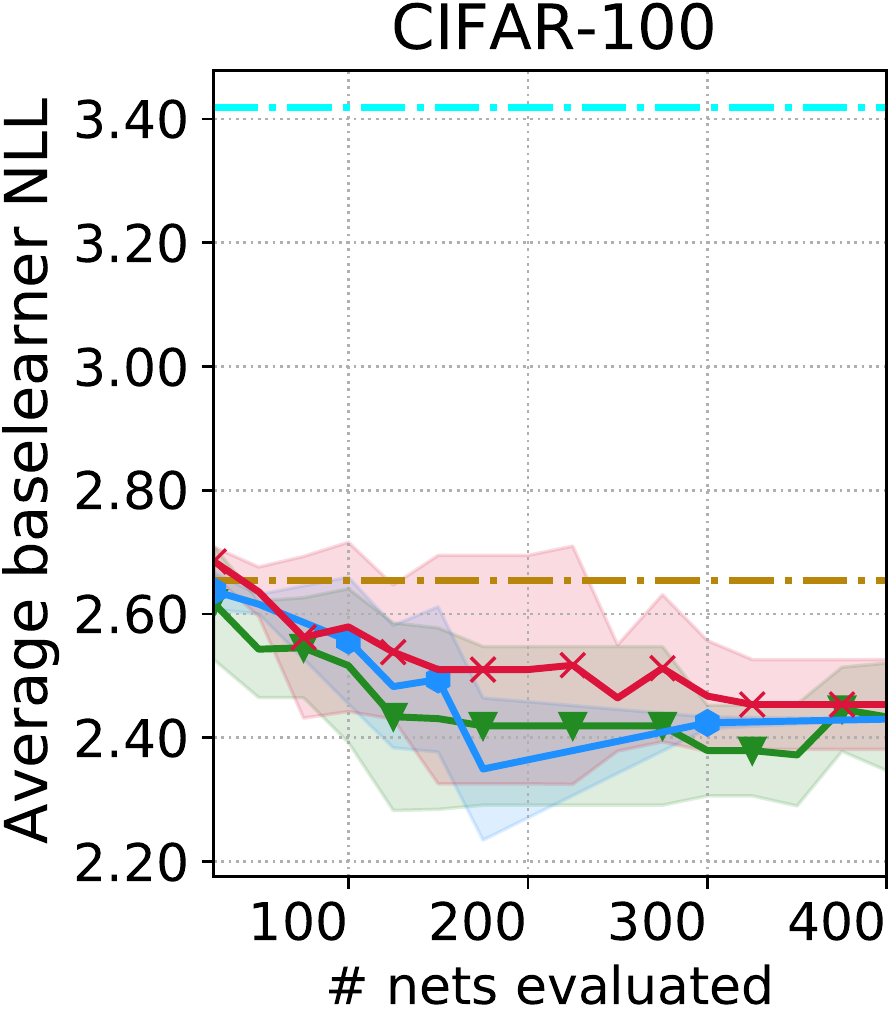}
        \includegraphics[width=.29\linewidth]{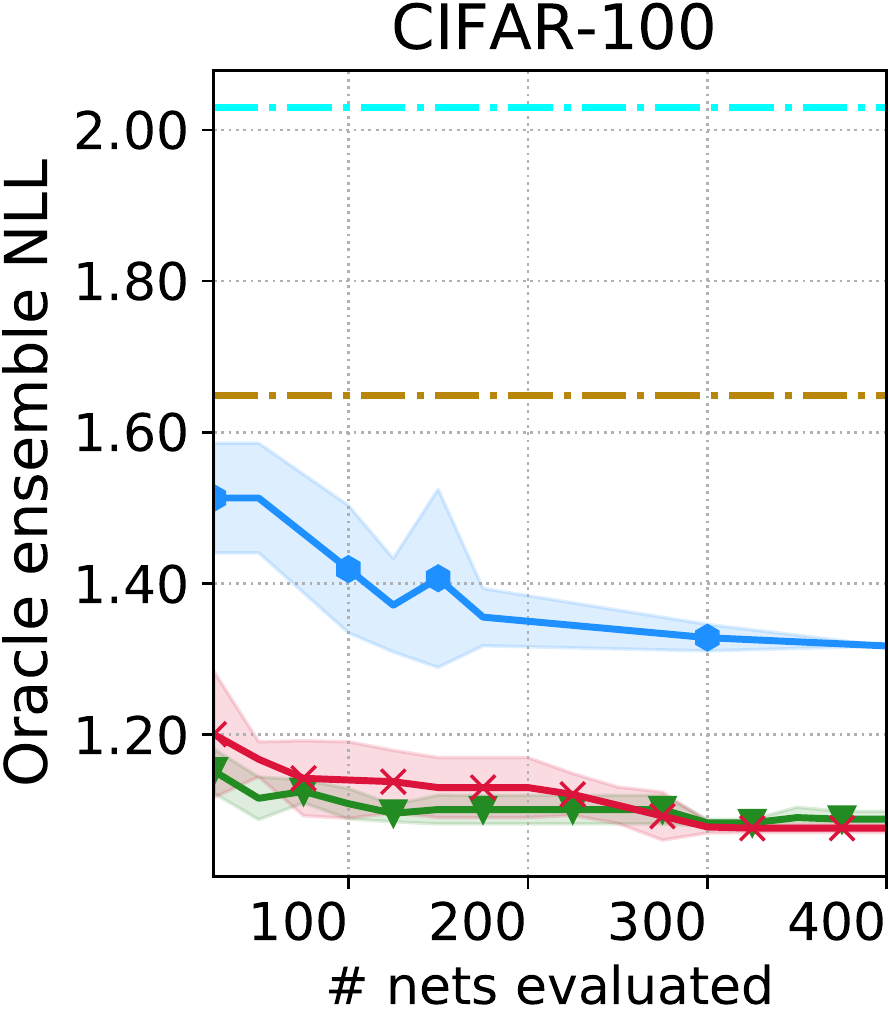}
        \subcaption{Data shift (severity 2)}
    \end{subfigure}%
    \begin{subfigure}[t]{0.49\textwidth}
        \centering
        \includegraphics[width=.29\linewidth]{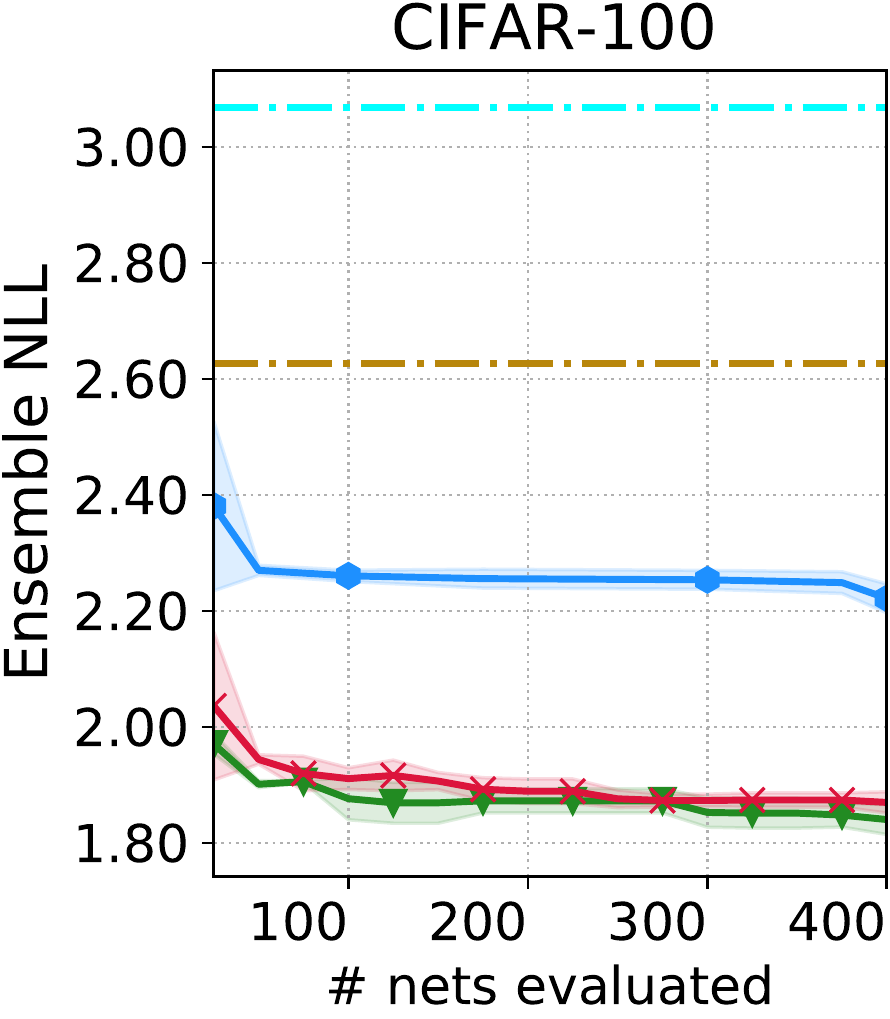}
        \includegraphics[width=.29\linewidth]{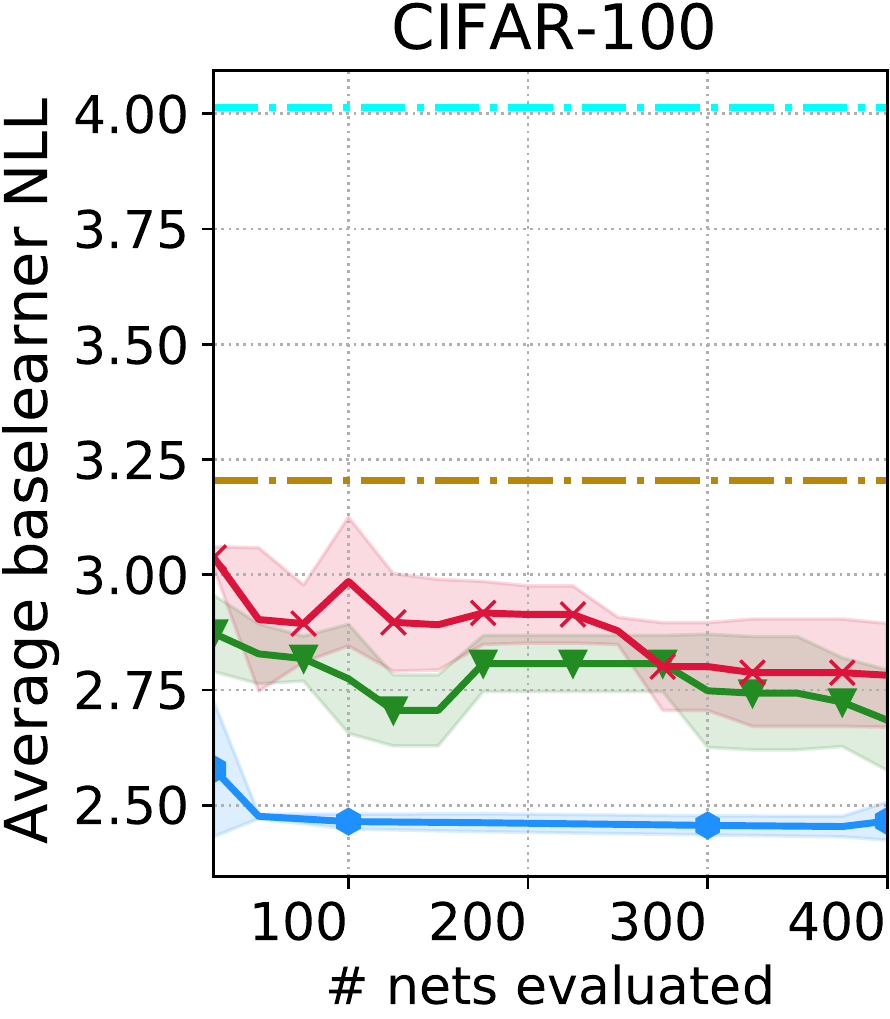}
        \includegraphics[width=.29\linewidth]{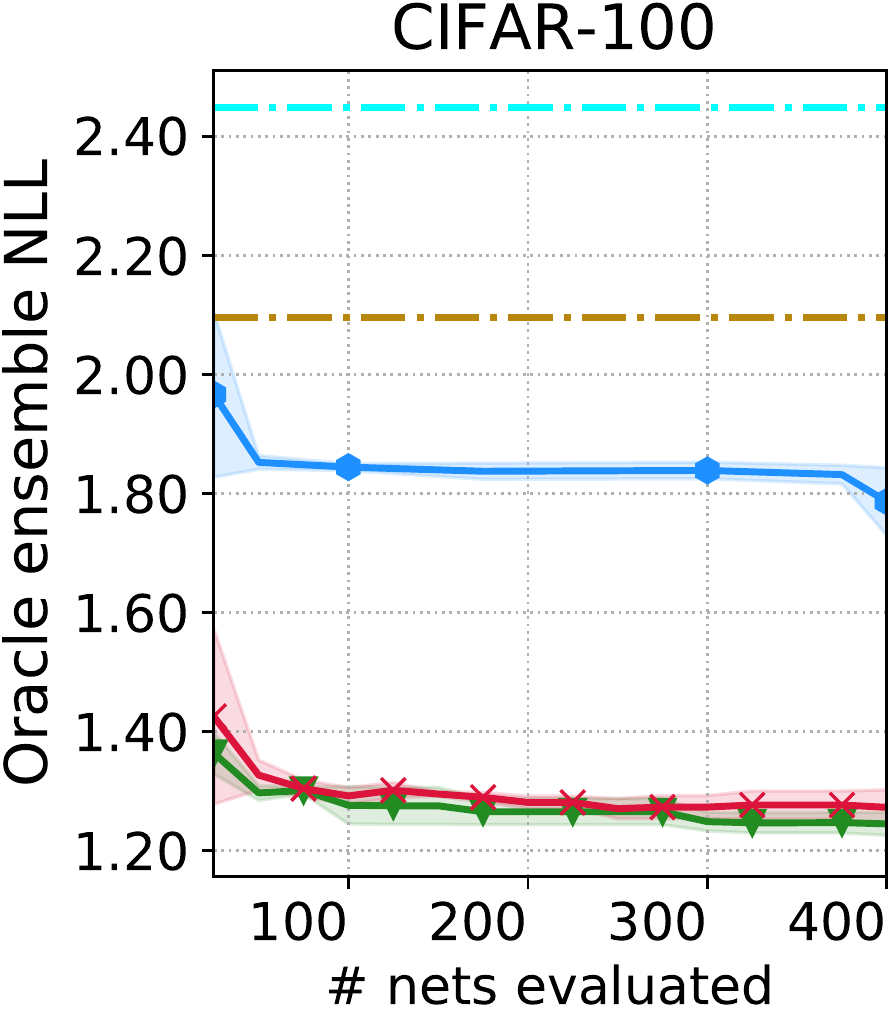}
        \subcaption{Data shift (severity 3)}
    \end{subfigure}\\ %
    \begin{subfigure}[t]{0.49\textwidth}
        \centering
        \includegraphics[width=.29\linewidth]{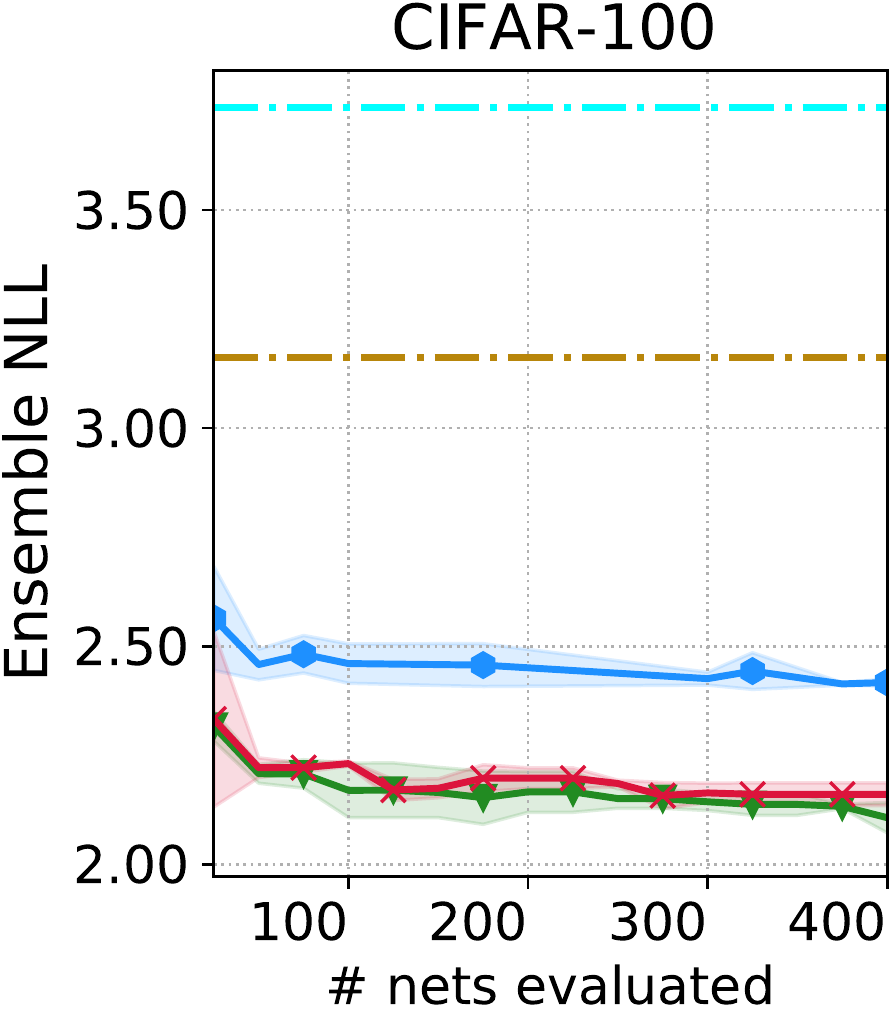}
        \includegraphics[width=.29\linewidth]{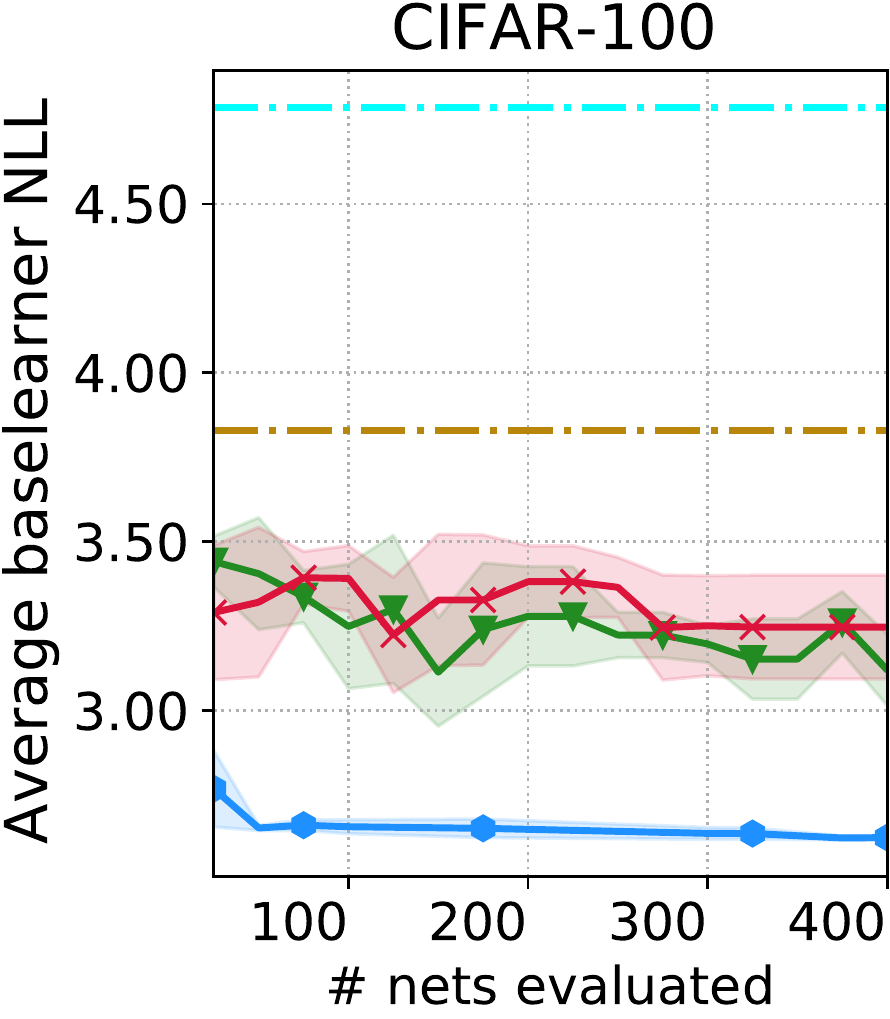}
        \includegraphics[width=.29\linewidth]{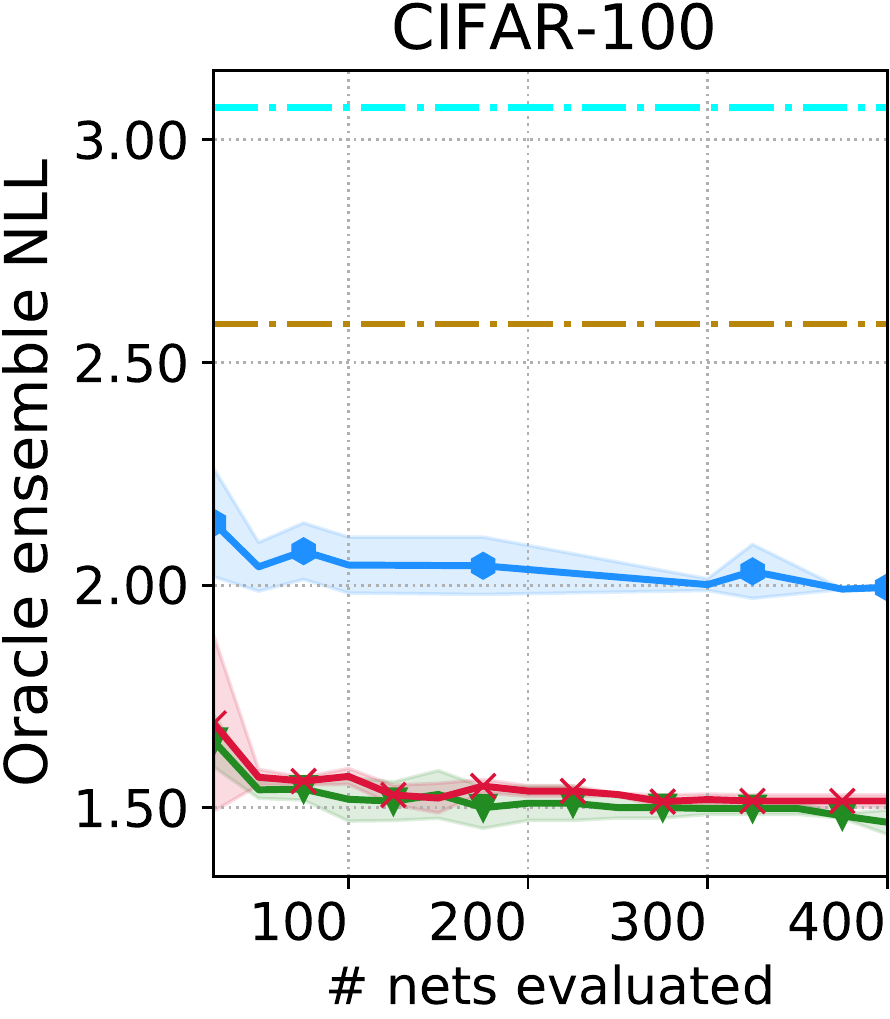}
        \subcaption{Data shift (severity 4)}
    \end{subfigure}%
    \begin{subfigure}[t]{0.49\textwidth}
        \centering
        \includegraphics[width=.29\linewidth]{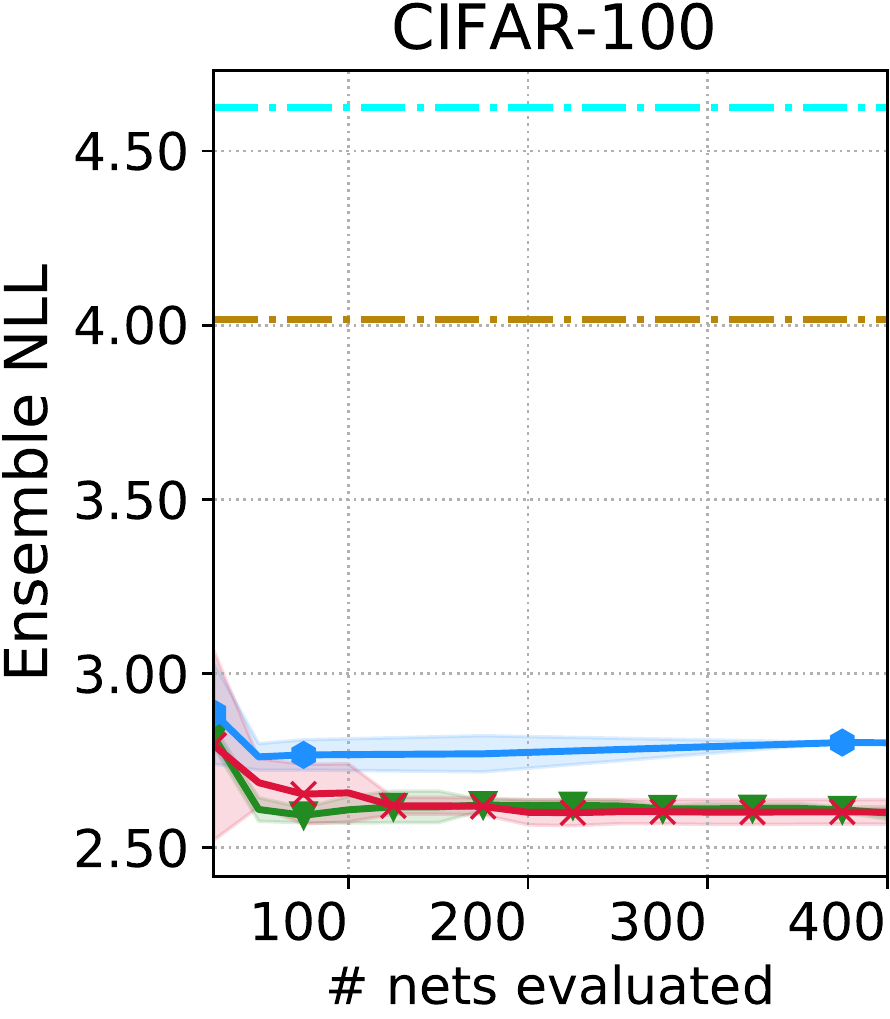}
        \includegraphics[width=.29\linewidth]{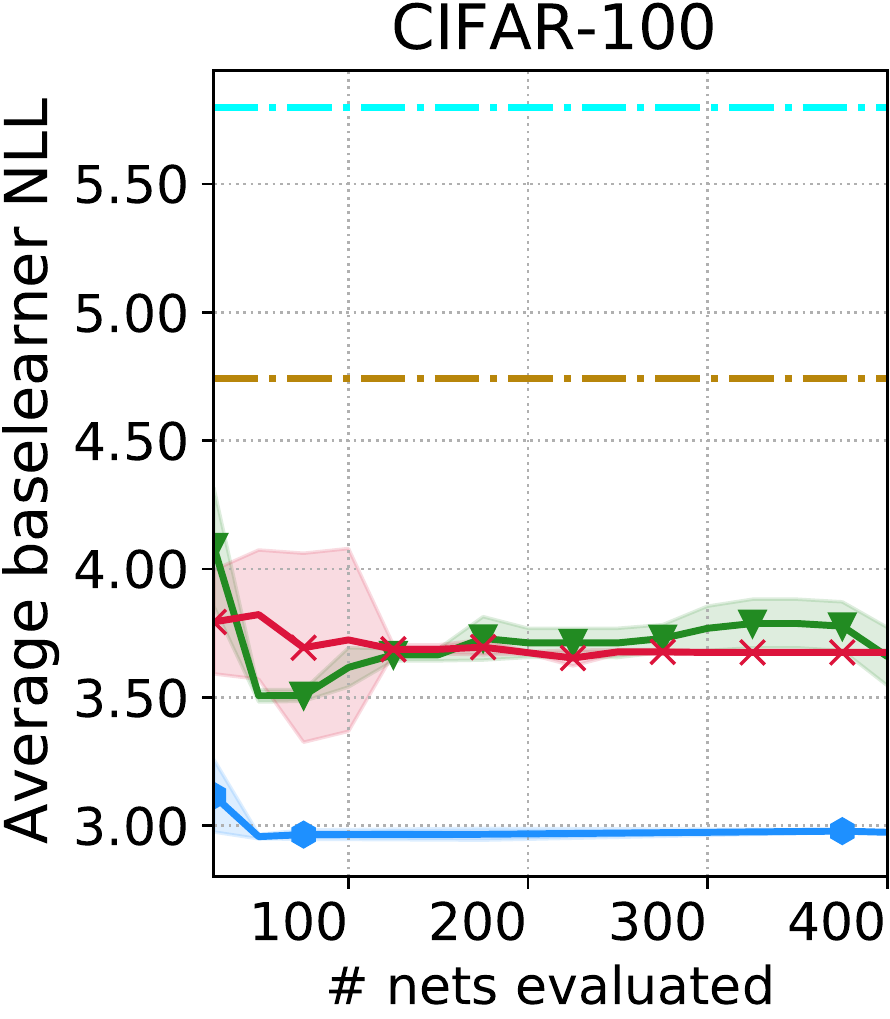}
        \includegraphics[width=.29\linewidth]{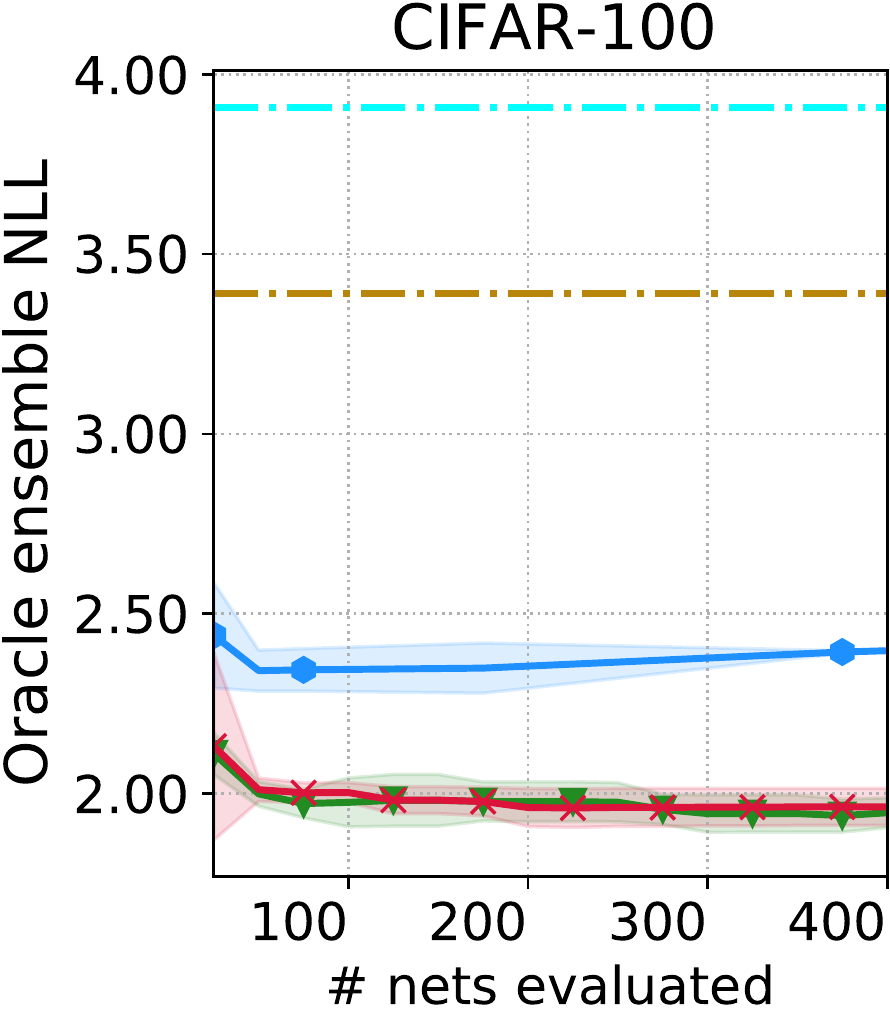}
        \subcaption{Data shift (severity 5)}
    \end{subfigure}
    
    \caption{CIFAR-100 analogous to Figure~\ref{fig:nb201_c10_budget}.}
    \label{fig:nb201_c100_budget}
\end{figure*}

\begin{figure*}
    \centering
    \captionsetup[subfigure]{justification=centering}
    \begin{subfigure}[t]{0.31\textwidth}
        \centering
        \includegraphics[width=.49\linewidth]{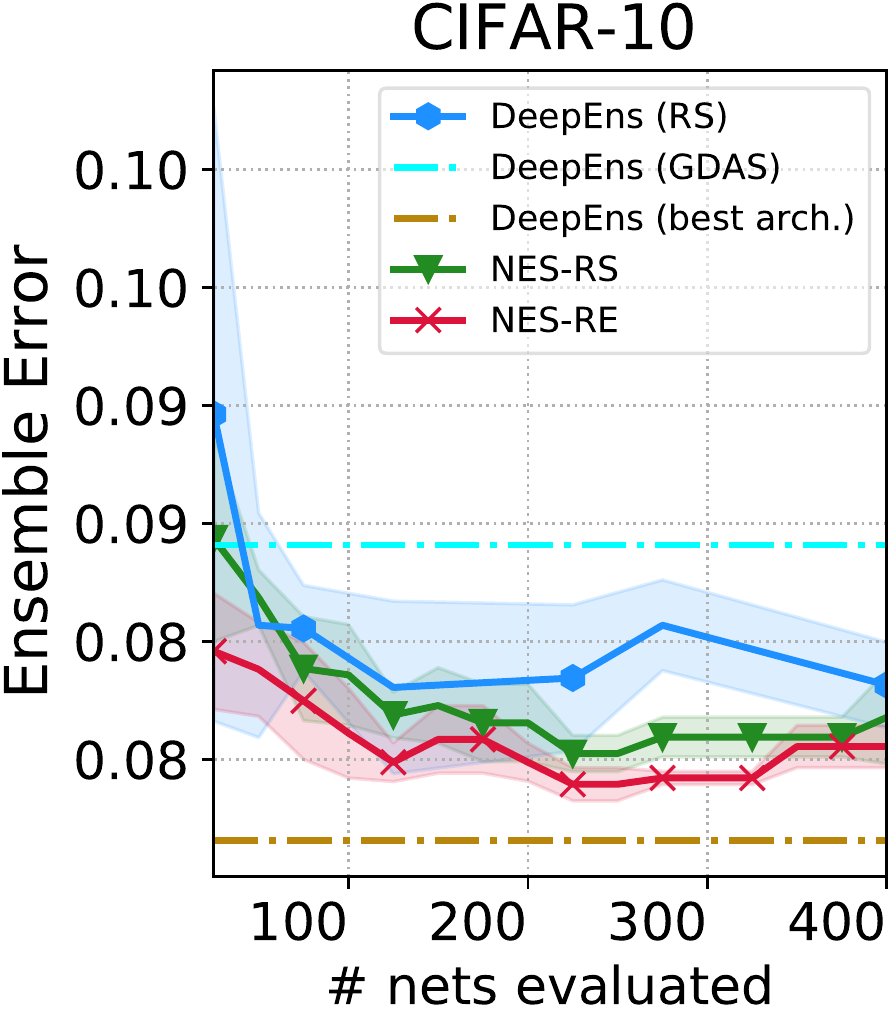}
        \includegraphics[width=.49\linewidth]{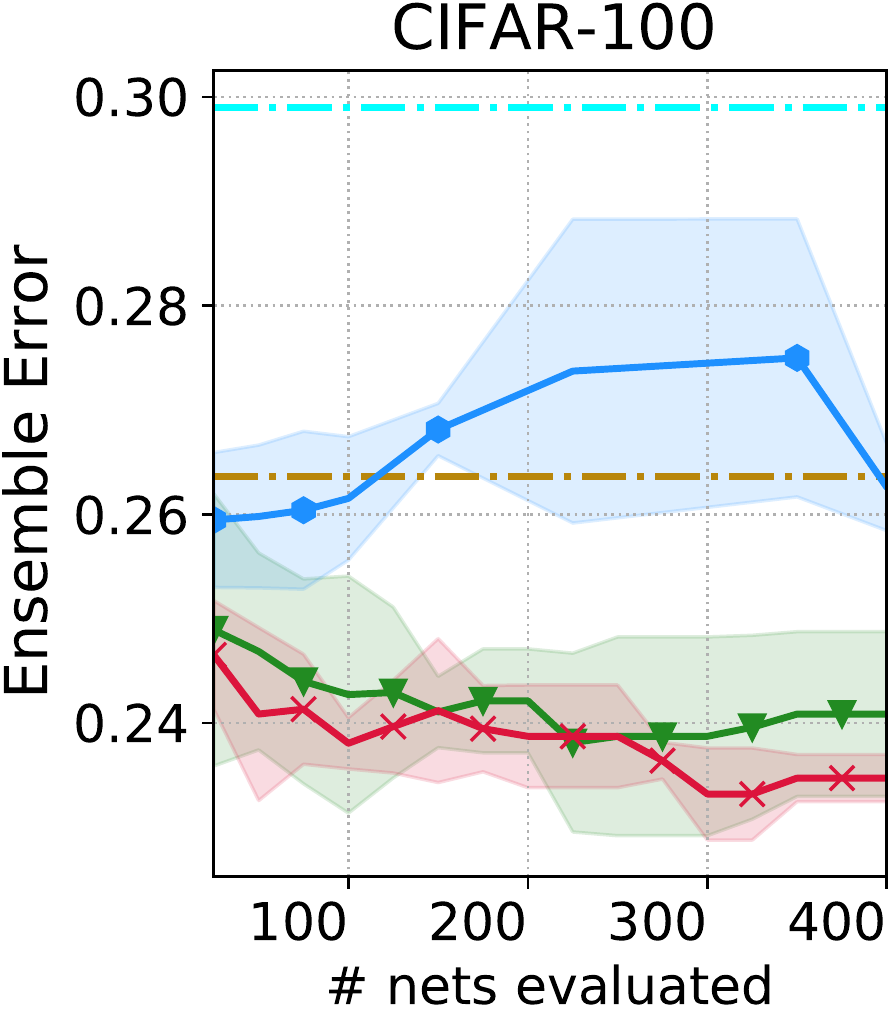}
        \subcaption{No data shift}
    \end{subfigure}%
    ~\hspace{.04cm} %
    \begin{subfigure}[t]{0.31\textwidth}
        \centering
        \includegraphics[width=.49\linewidth]{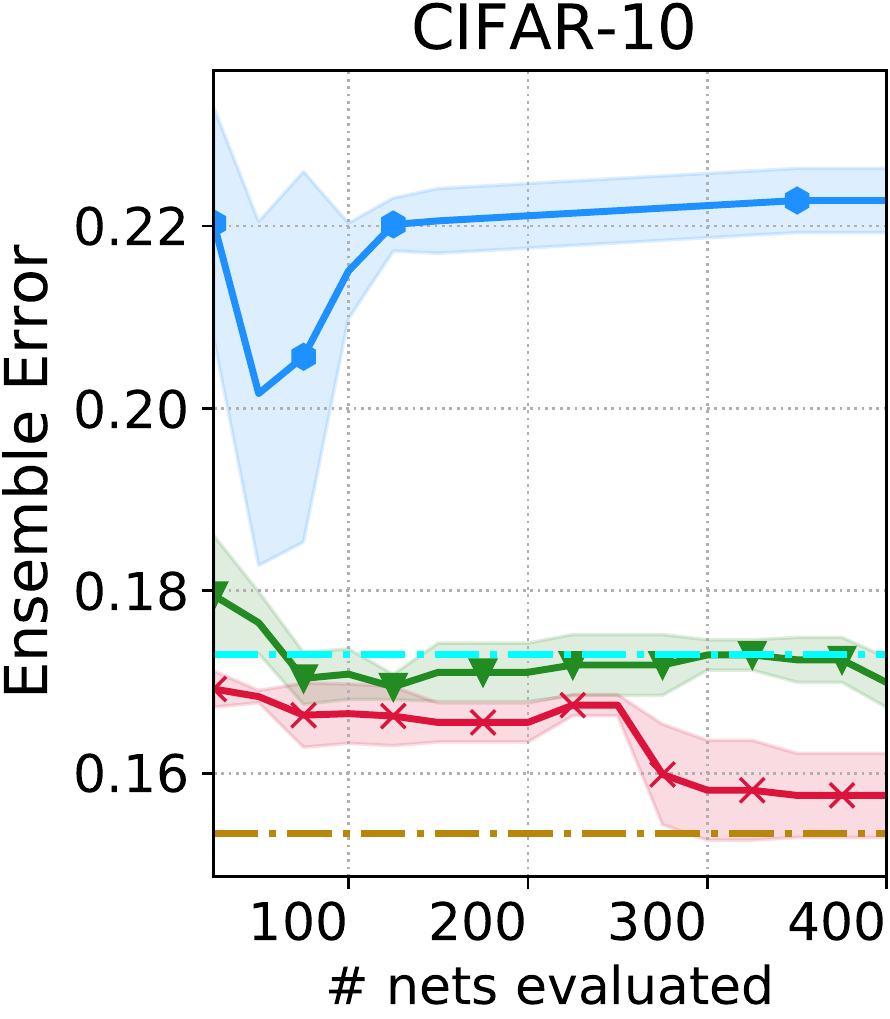}
        \includegraphics[width=.49\linewidth]{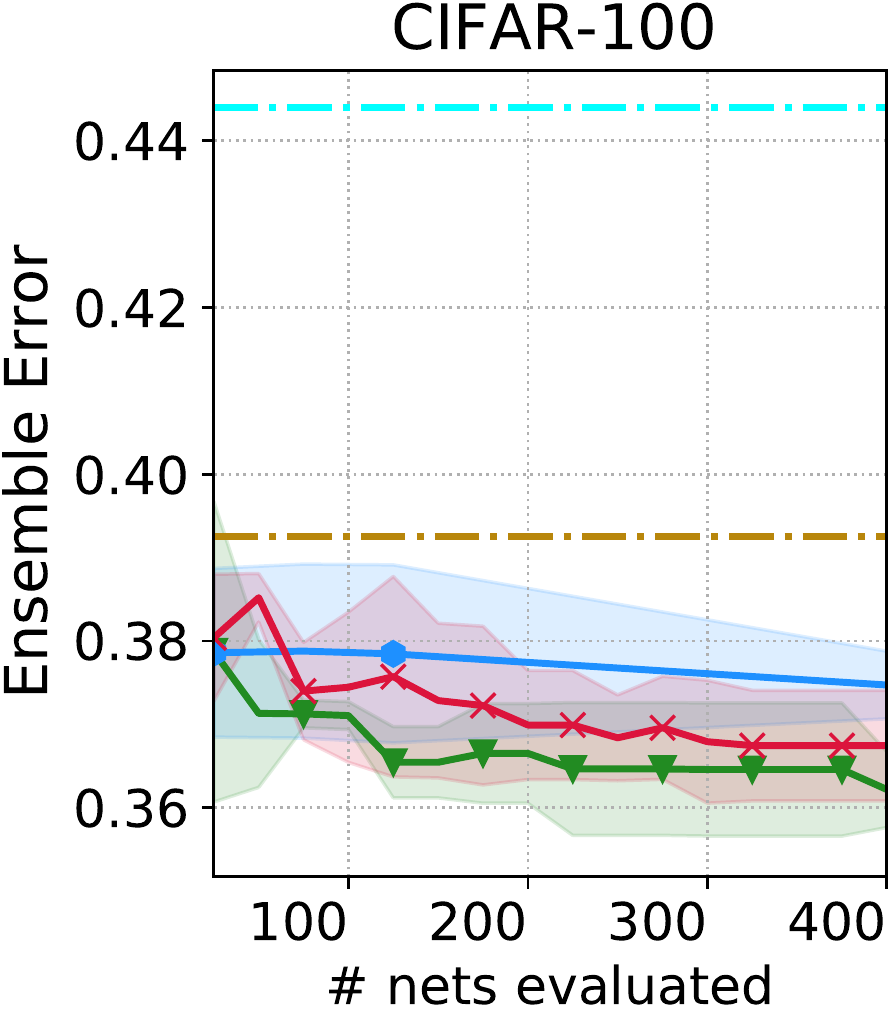}
        \subcaption{Data shift (severity 1)}
    \end{subfigure}
    ~\hspace{.04cm}
    \begin{subfigure}[t]{0.31\textwidth}
        \centering
        \includegraphics[width=.49\linewidth]{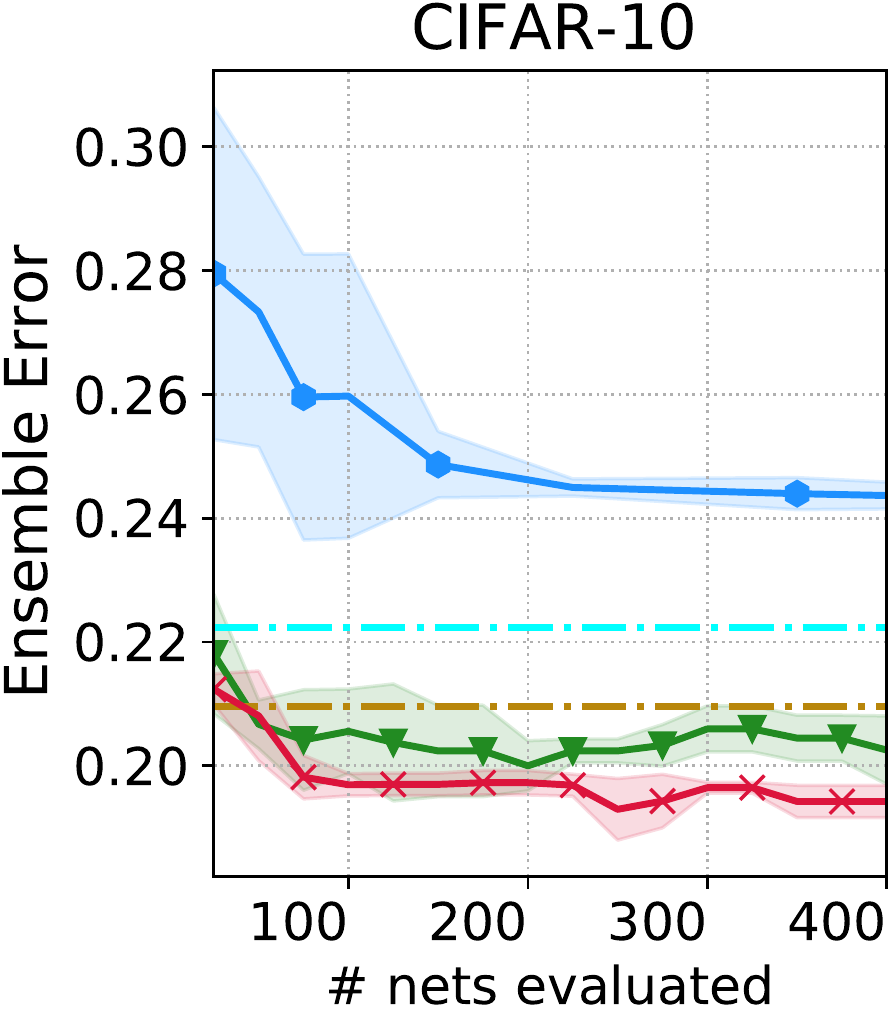}
        \includegraphics[width=.49\linewidth]{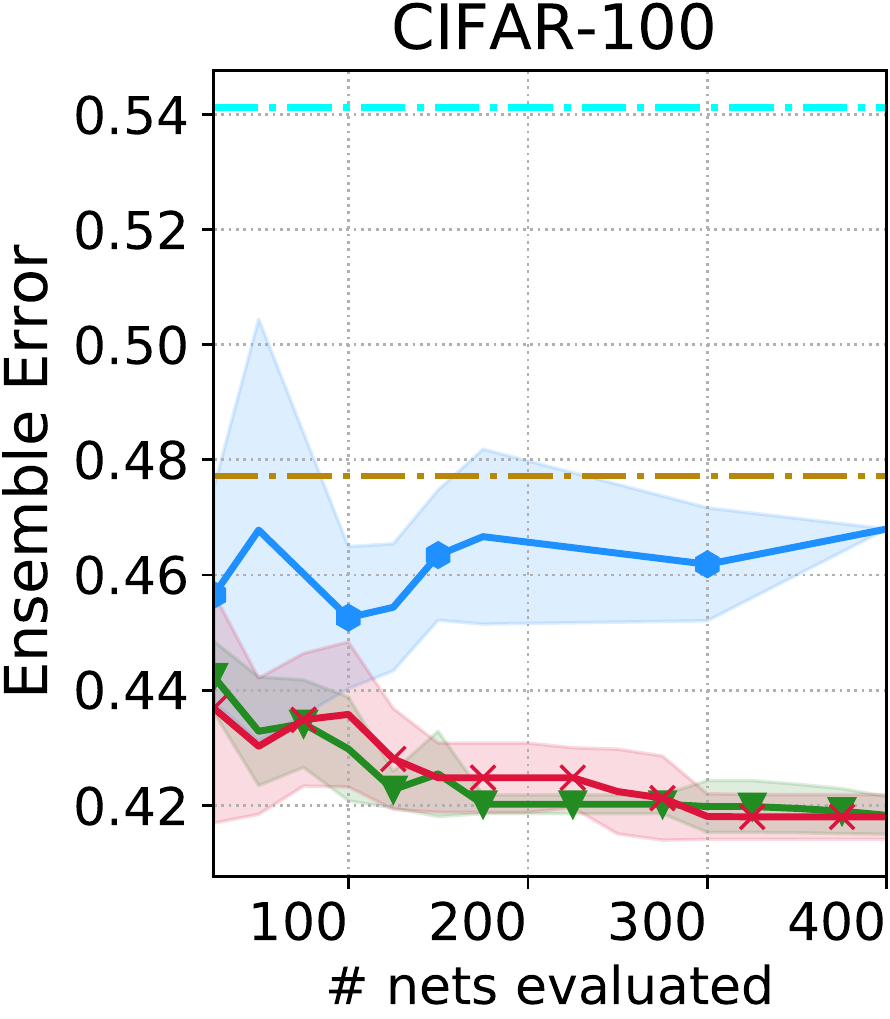}
        \subcaption{Data shift (severity 2)}
    \end{subfigure}\\ %
    \begin{subfigure}[t]{0.31\textwidth}
        \centering
        \includegraphics[width=.49\linewidth]{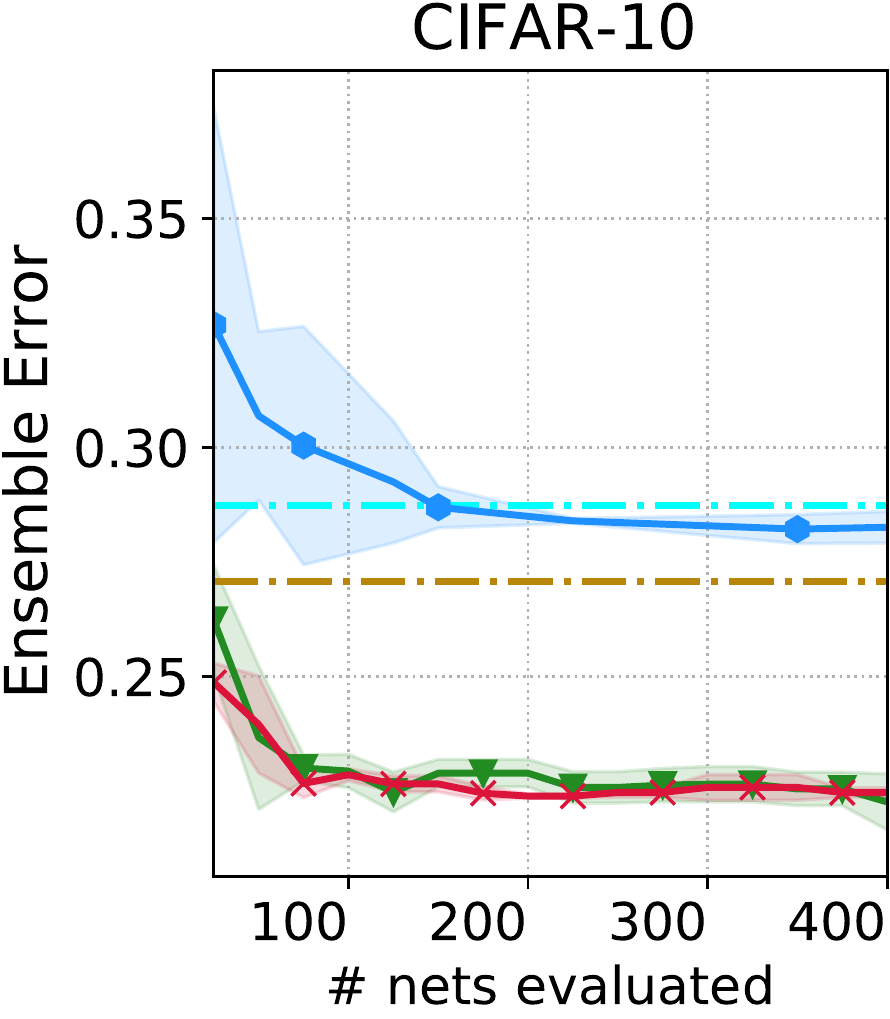}
        \includegraphics[width=.49\linewidth]{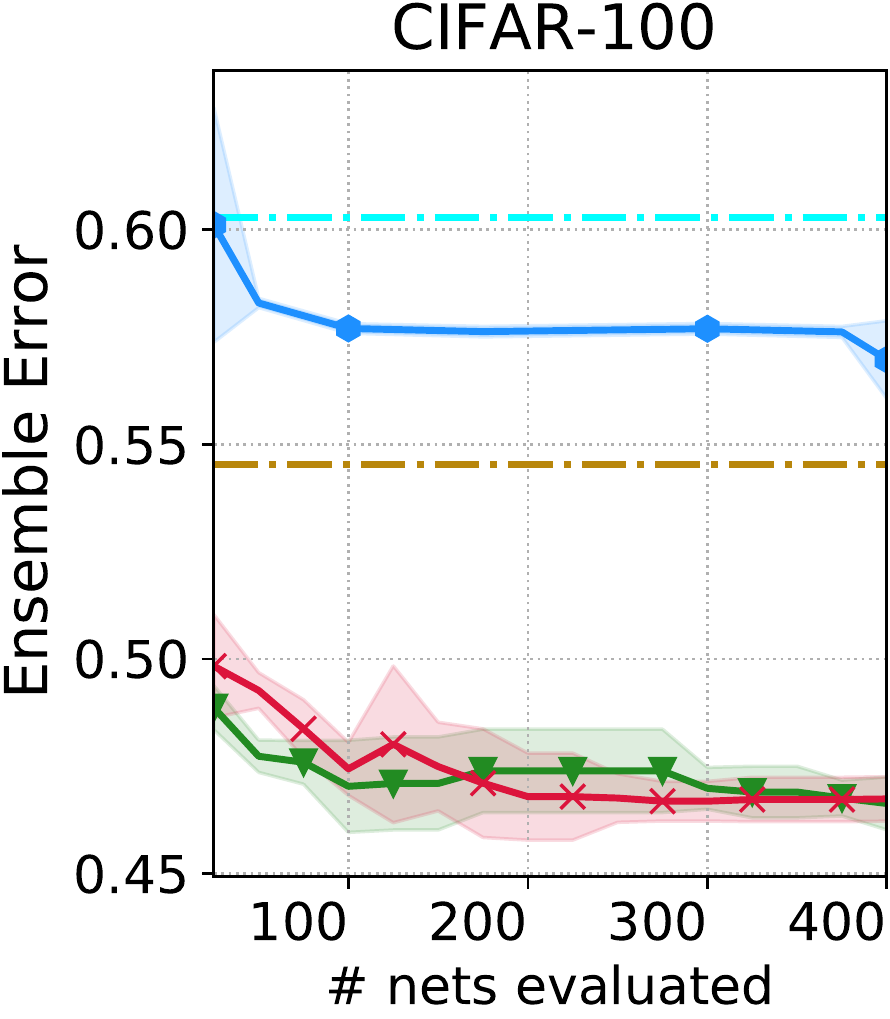}
        \subcaption{Data shift (severity 3)}
    \end{subfigure}
    ~\hspace{.04cm}
    \begin{subfigure}[t]{0.31\textwidth}
        \centering
        \includegraphics[width=.49\linewidth]{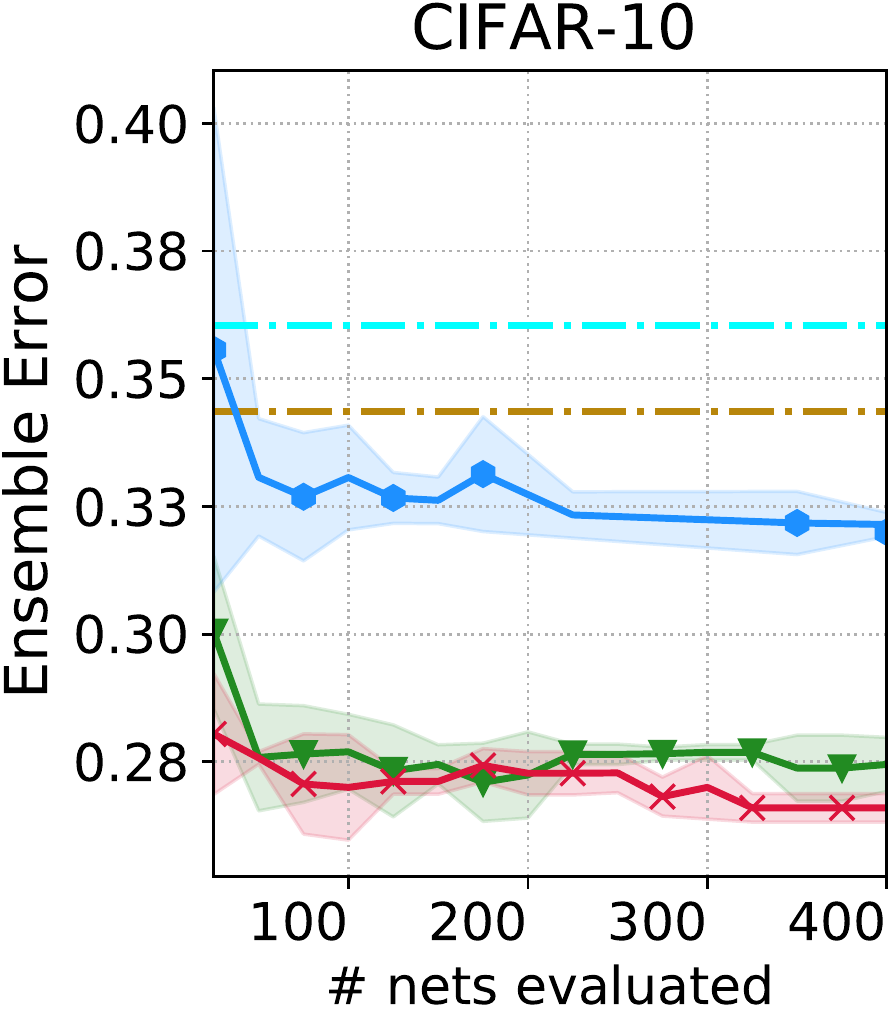}
        \includegraphics[width=.49\linewidth]{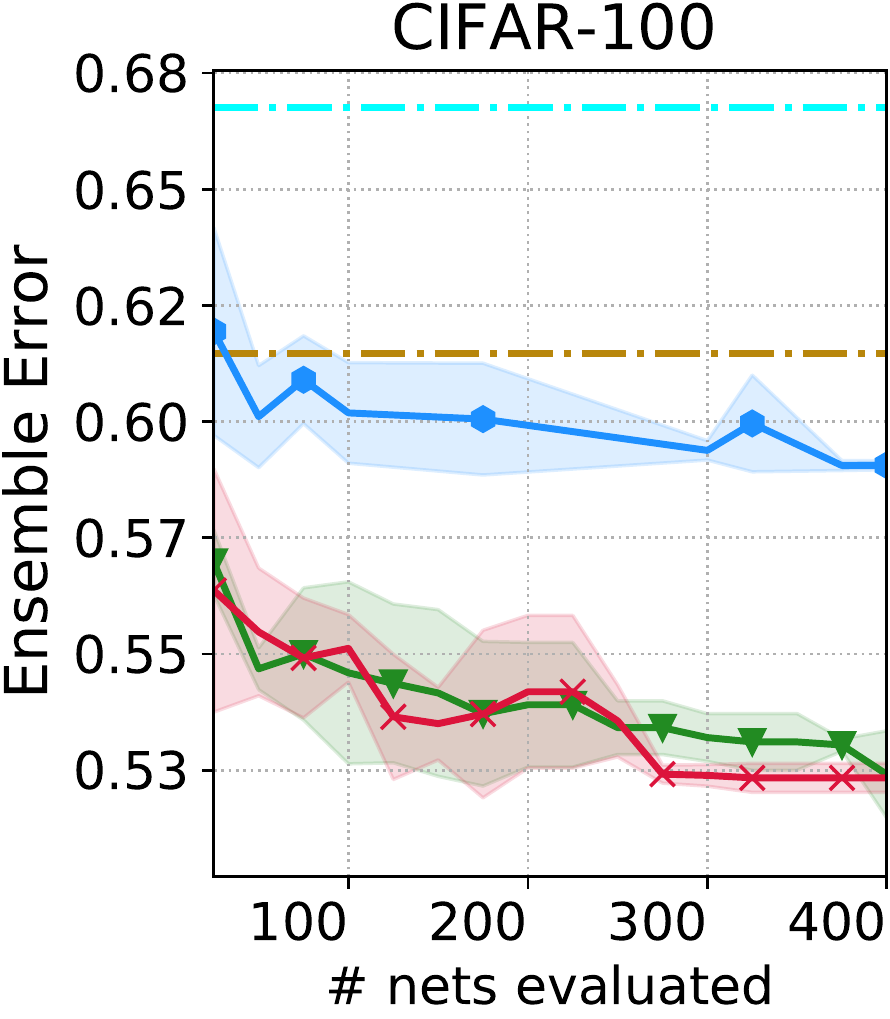}
        \subcaption{Data shift (severity 4)}
    \end{subfigure}%
    ~\hspace{.04cm} %
    \begin{subfigure}[t]{0.31\textwidth}
        \centering
        \includegraphics[width=.49\linewidth]{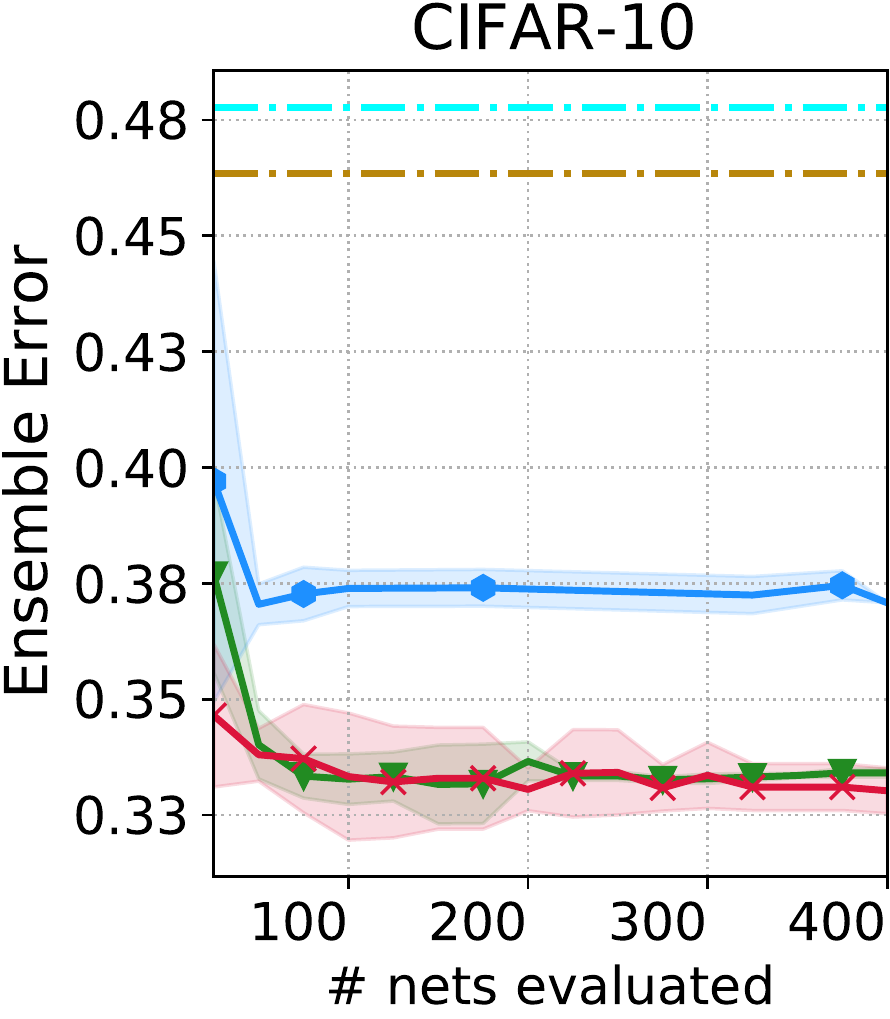}
        \includegraphics[width=.49\linewidth]{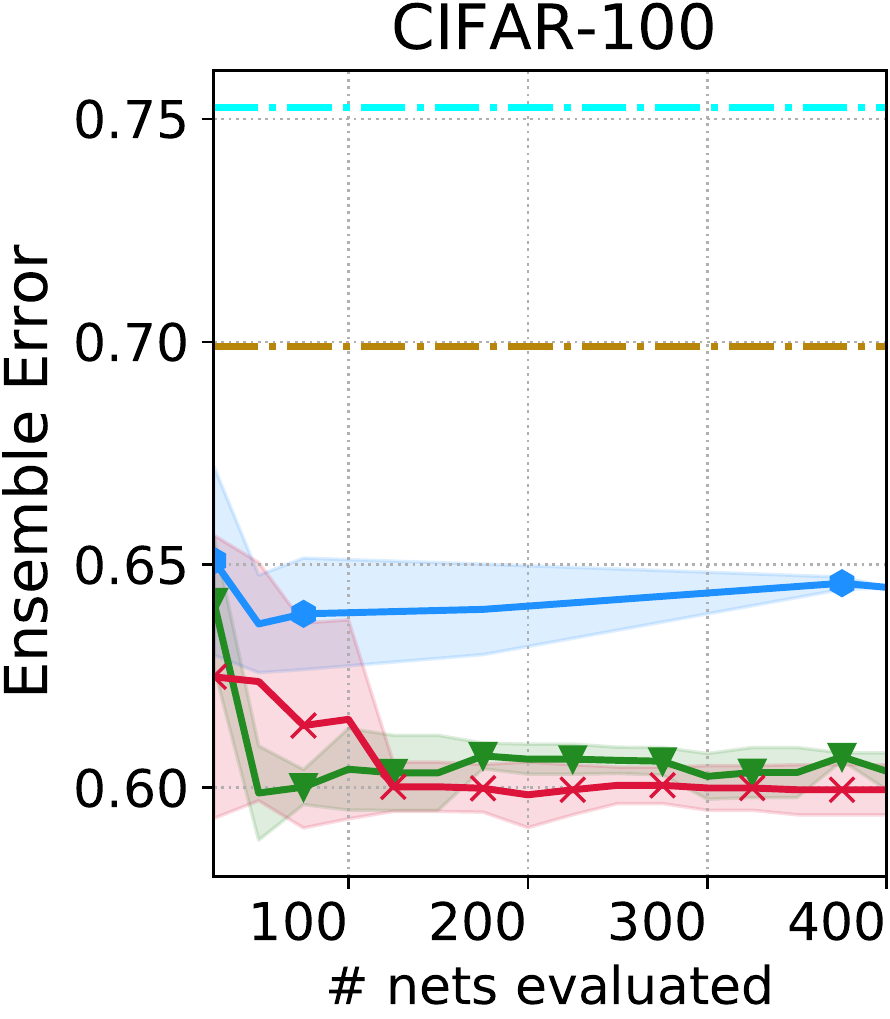}
        \subcaption{Data shift (severity 5)}
    \end{subfigure}
    
    \caption{Ensemble error vs. budget $\budget$ on the NAS-Bench-201 space (CIFAR-10 and CIFAR-100). Ensemble size fixed at $M = 3$.}
    \label{fig:nb201_error}
\end{figure*}

\begin{figure*}
    \centering
    \captionsetup[subfigure]{justification=centering}
    \begin{subfigure}[t]{0.31\textwidth}
        \centering
        \includegraphics[width=.48\linewidth]{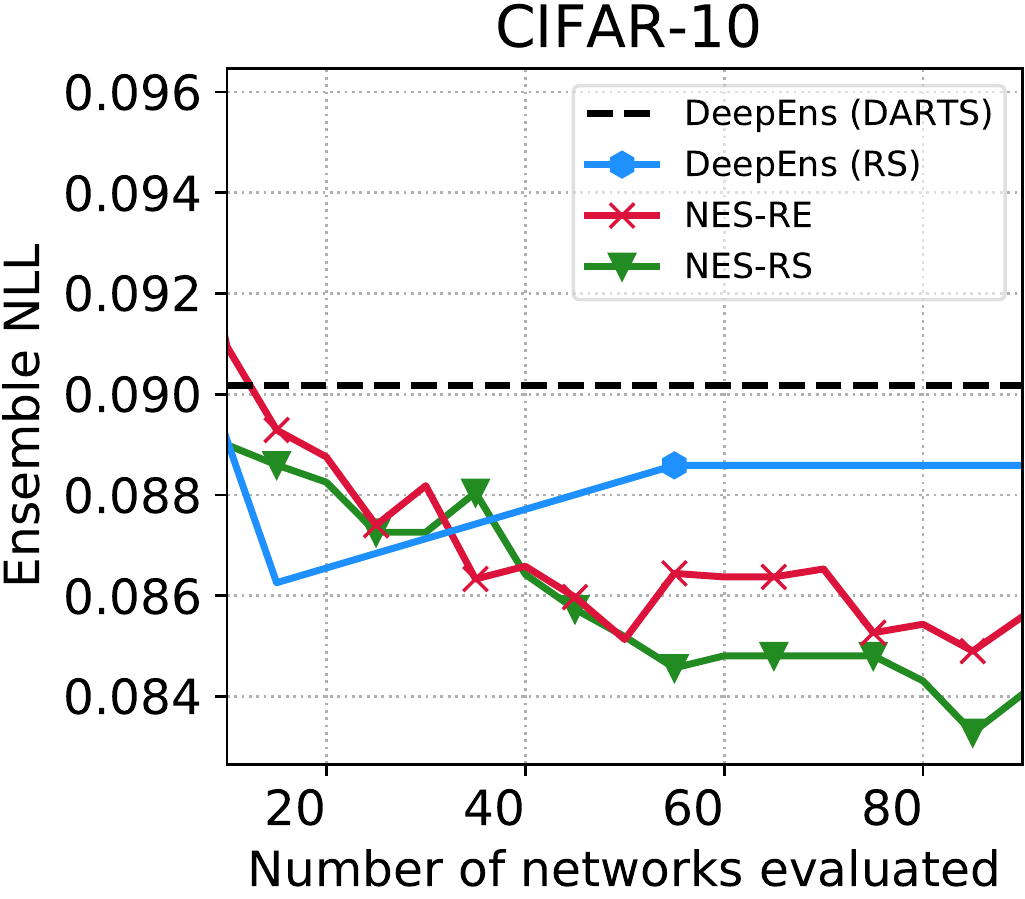}
        \includegraphics[width=.48\linewidth]{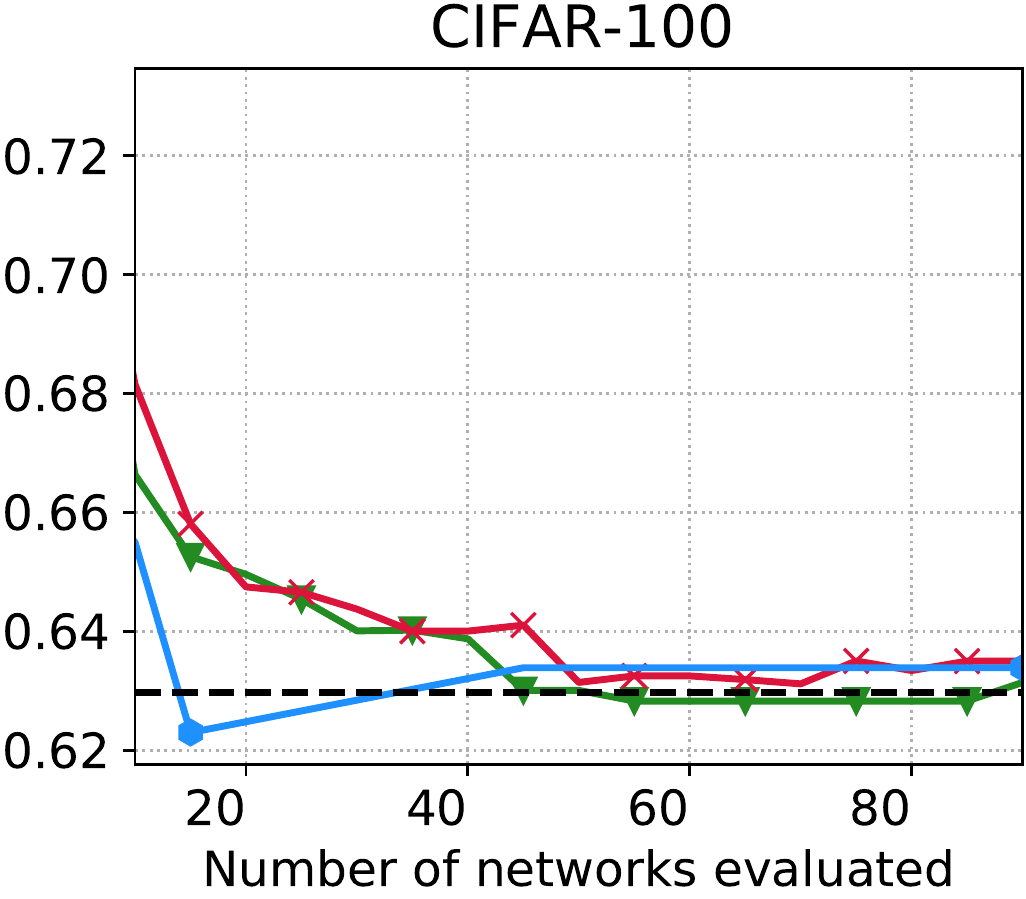}
        \subcaption{No data shift}
        \label{fig:test_loss_full_fidelity_0}
    \end{subfigure}%
    ~\hspace{.1cm}
    \begin{subfigure}[t]{0.31\textwidth}
        \centering
        \includegraphics[width=.48\linewidth]{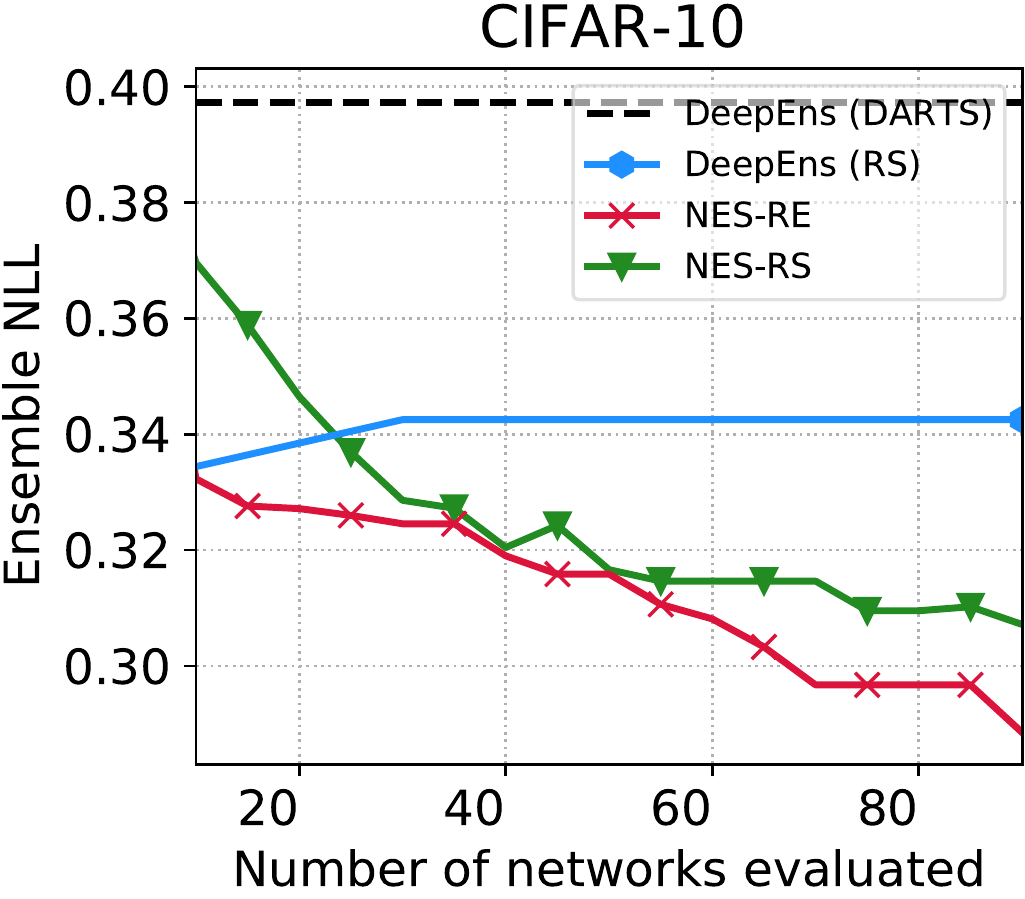}
        \includegraphics[width=.48\linewidth]{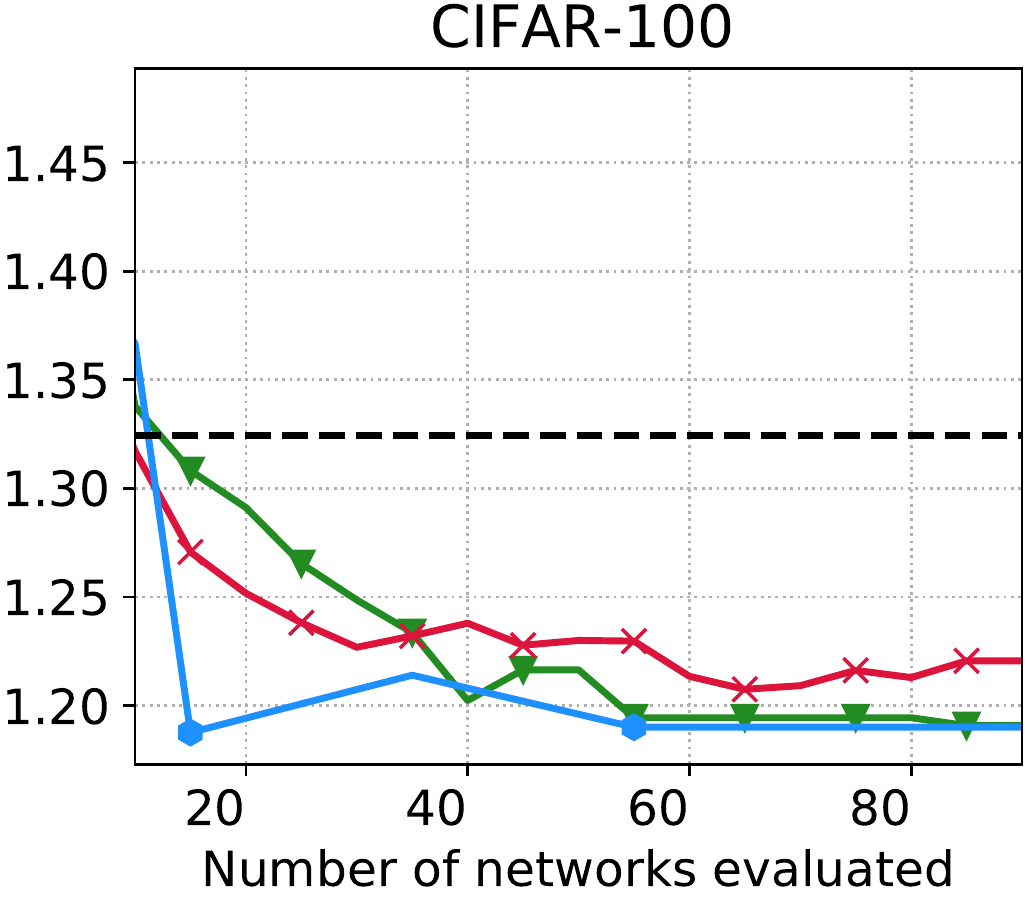}
        \subcaption{Data shift (severity 1)}
        \label{fig:test_loss_full_fidelity_1}
    \end{subfigure}
    ~\hspace{.1cm}
    \begin{subfigure}[t]{0.31\textwidth}
        \centering
        \includegraphics[width=.48\linewidth]{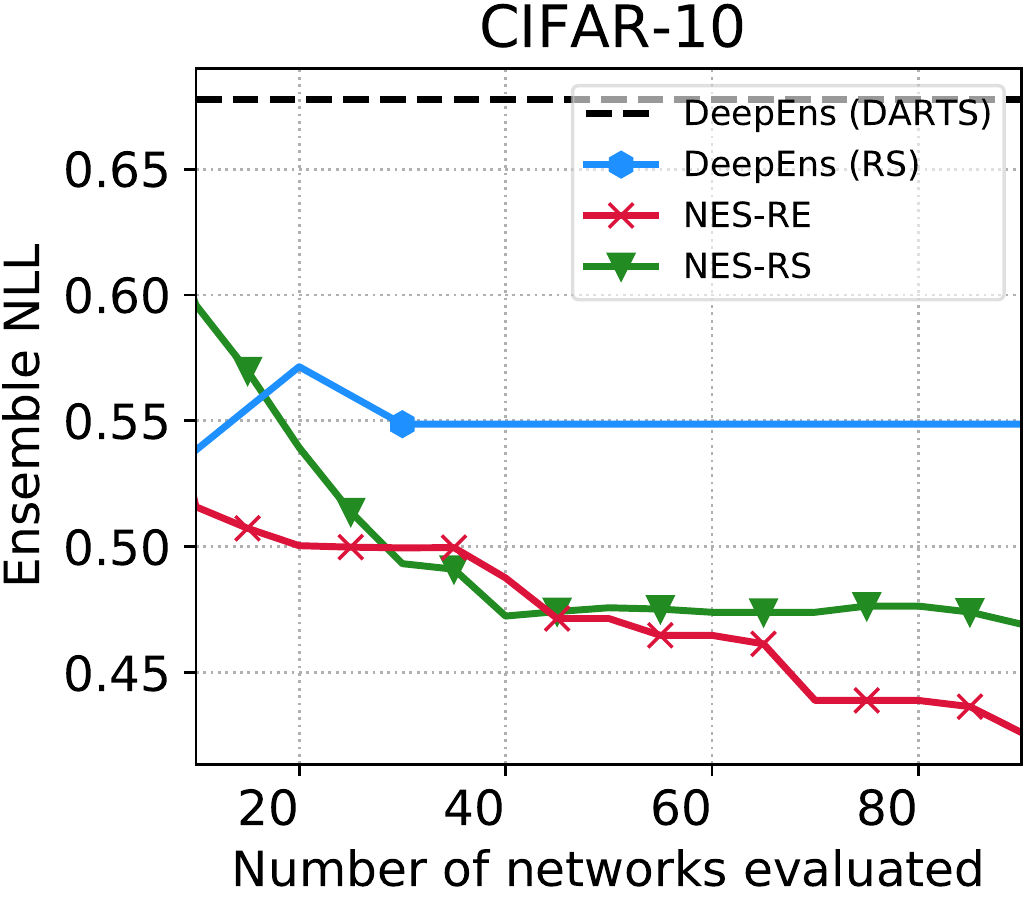}
        \includegraphics[width=.48\linewidth]{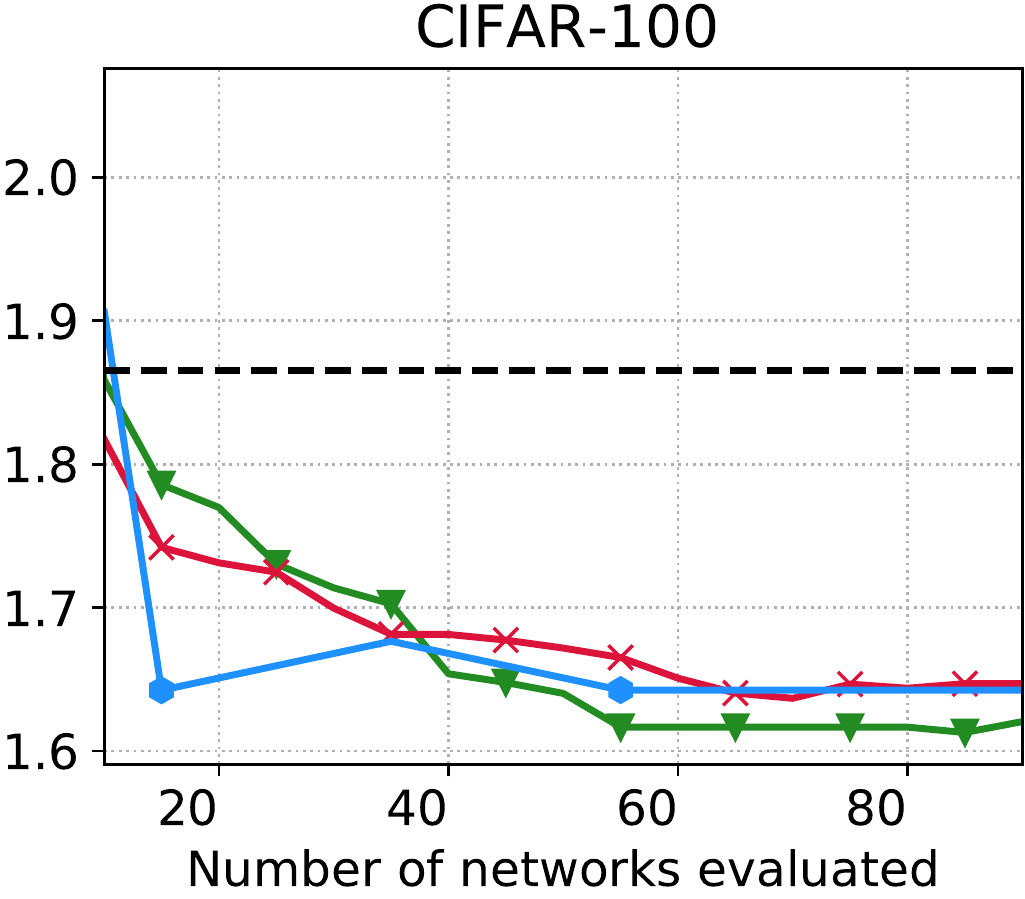}
        \subcaption{Data shift (severity 2)}
        \label{fig:test_loss_full_fidelity_2}
    \end{subfigure}\\ %
    \begin{subfigure}[t]{0.31\textwidth}
        \centering
        \includegraphics[width=.48\linewidth]{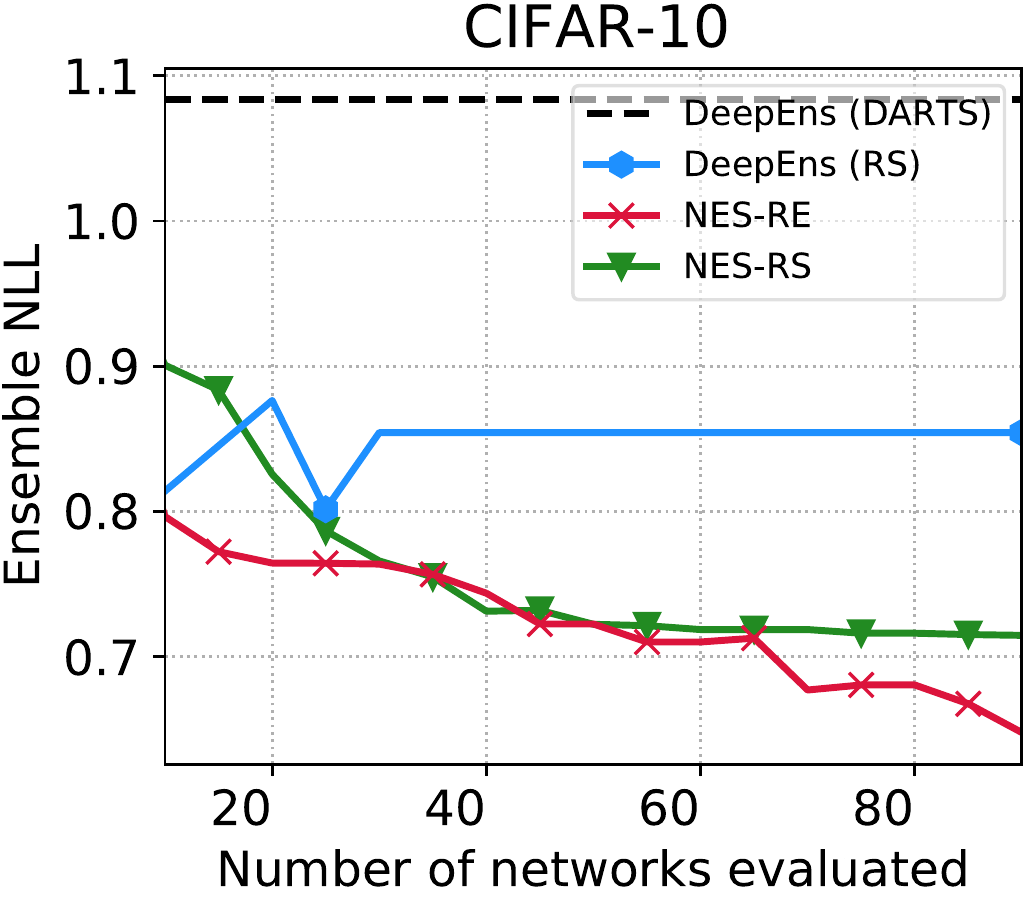}
        \includegraphics[width=.48\linewidth]{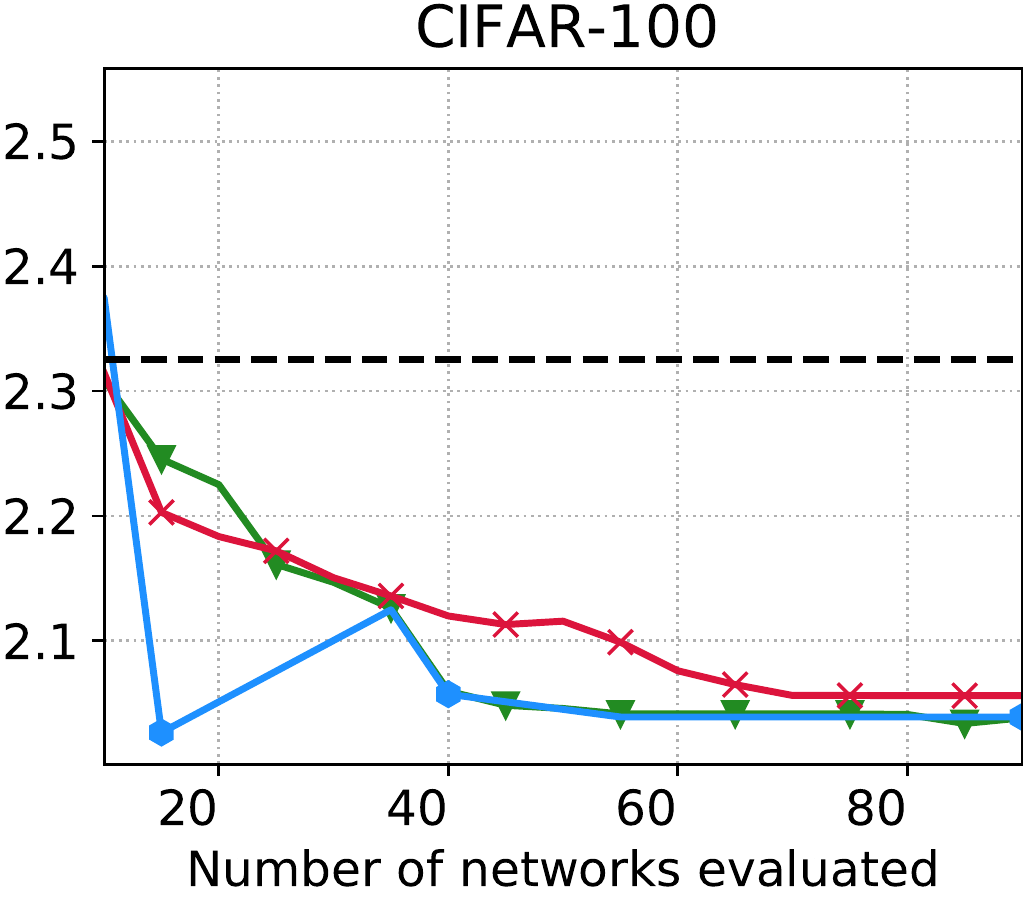}
        \subcaption{Data shift (severity 3)}
        \label{fig:test_loss_full_fidelity_3}
    \end{subfigure}%
    ~\hspace{.1cm}
    \begin{subfigure}[t]{0.31\textwidth}
        \centering
        \includegraphics[width=.48\linewidth]{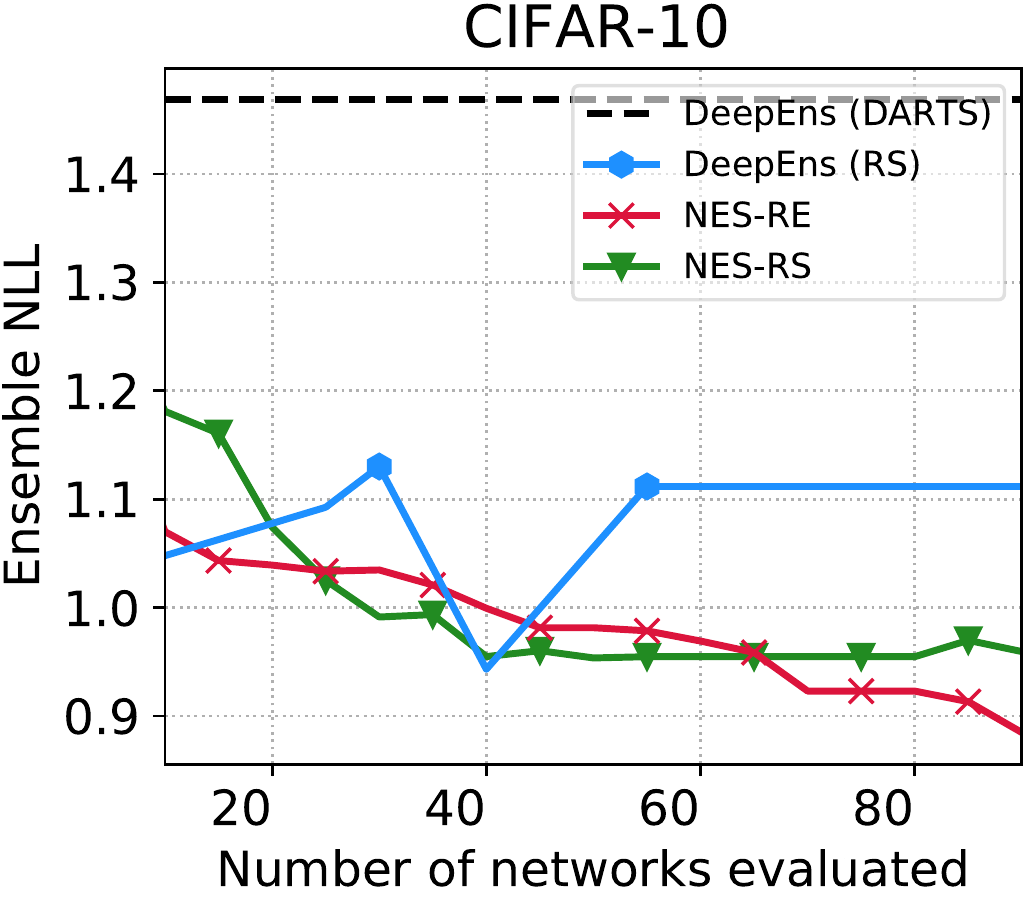}
        \includegraphics[width=.48\linewidth]{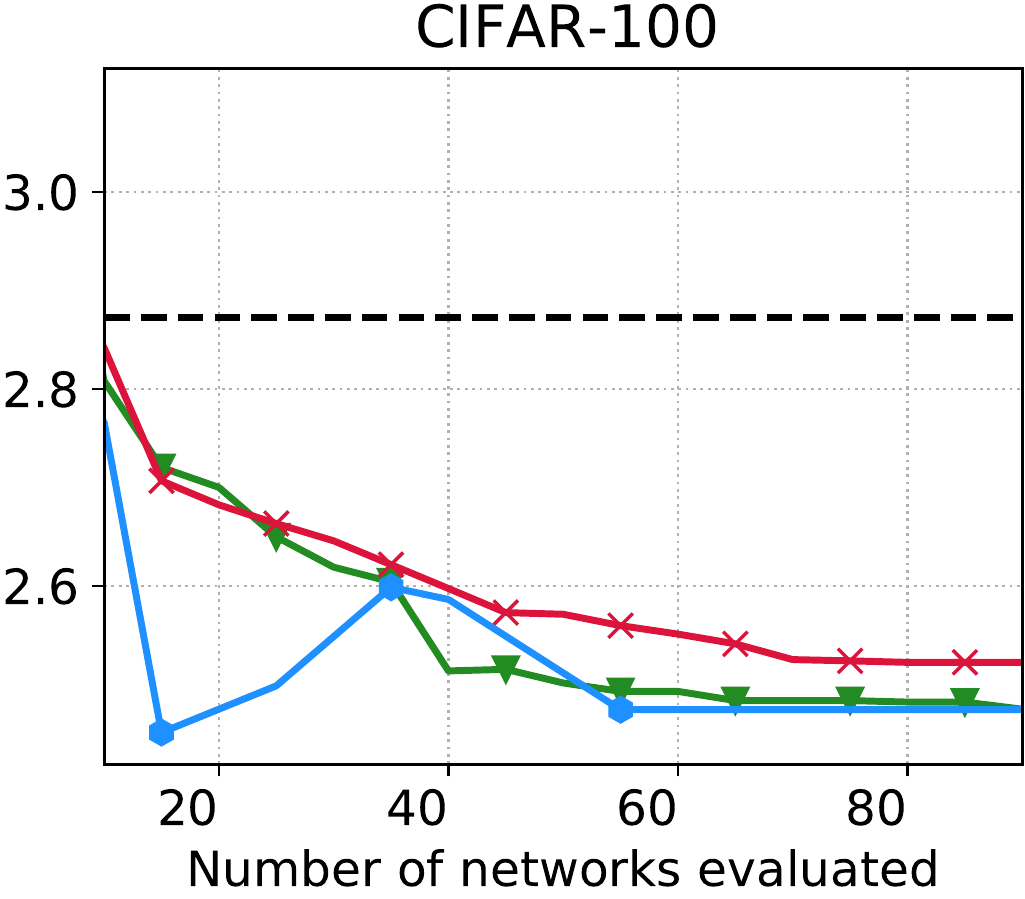}
        \subcaption{Data shift (severity 4)}
        \label{fig:test_loss_full_fidelity_4}
    \end{subfigure}
    ~\hspace{.1cm}
    \begin{subfigure}[t]{0.31\textwidth}
        \centering
        \includegraphics[width=.48\linewidth]{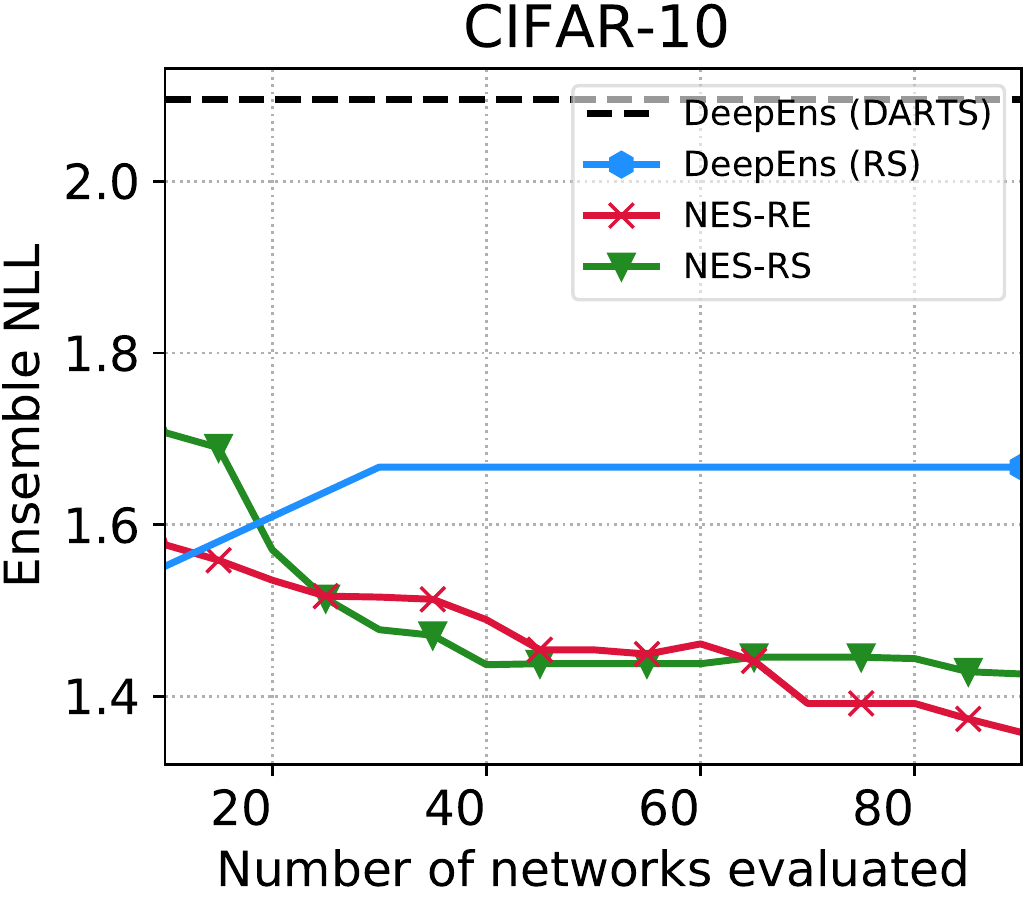}
        \includegraphics[width=.48\linewidth]{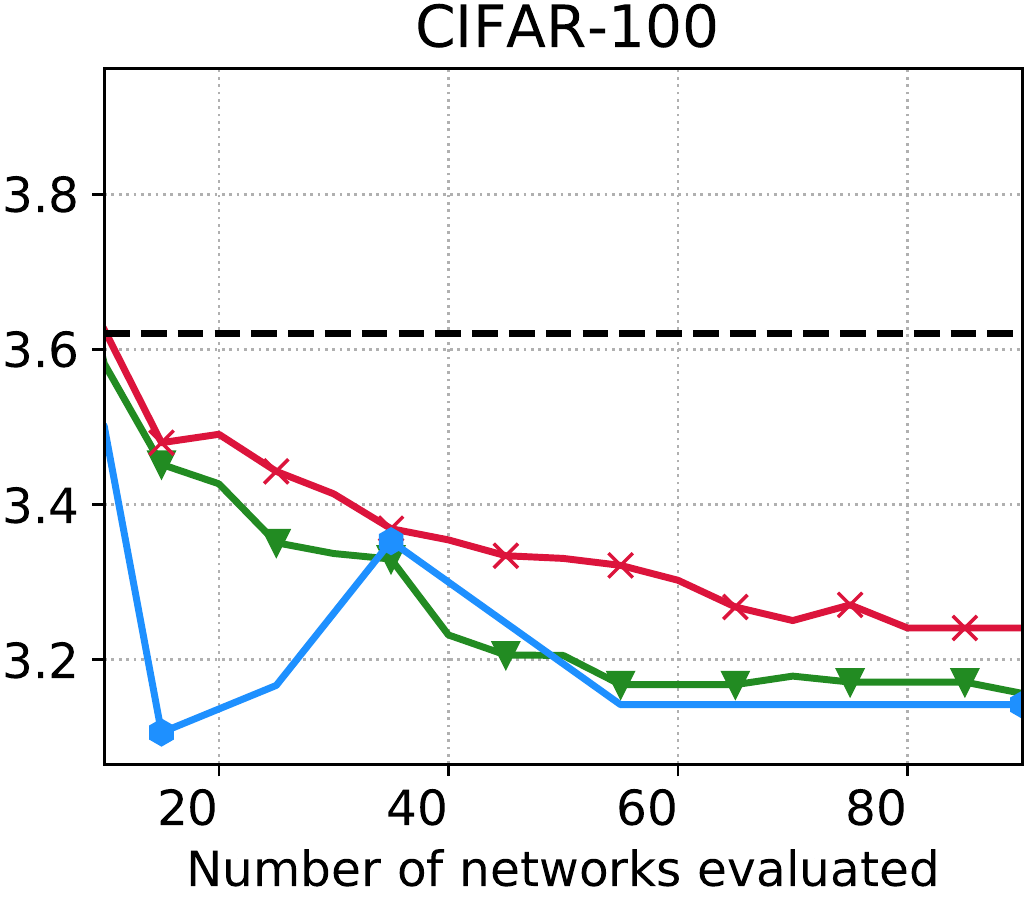}
        \subcaption{Data shift (severity 5)}
        \label{fig:test_loss_full_fidelity_5}
    \end{subfigure}
    
    \caption{High fidelity NLL vs. budget $\budget$ on CIFAR-10 and CIFAR-100 with and without respective dataset shifts over the DARTS search space. Ensemble size is fixed at $M = 10$.}
    \label{fig:test_loss_full_fid}
\end{figure*}

\begin{figure*}
    \centering
    \captionsetup[subfigure]{justification=centering}
    \begin{subfigure}[t]{0.31\textwidth}
        \centering
        \includegraphics[width=.48\linewidth]{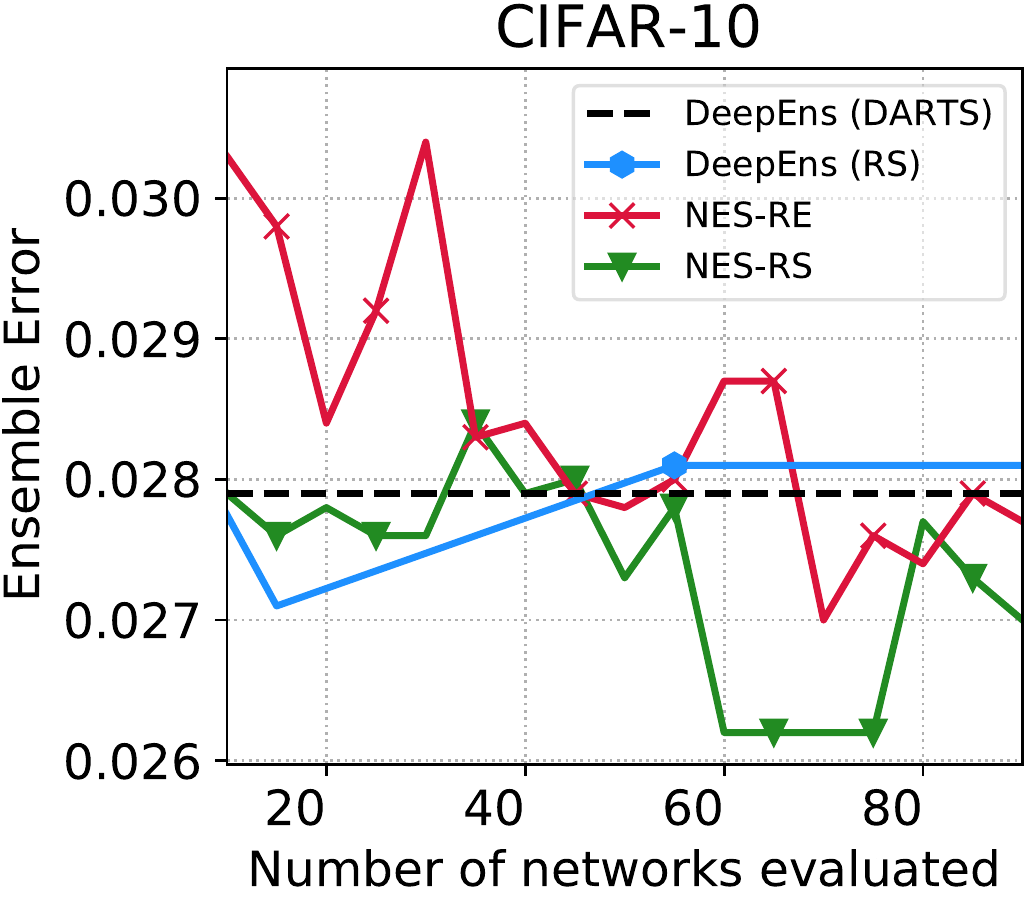}
        \includegraphics[width=.48\linewidth]{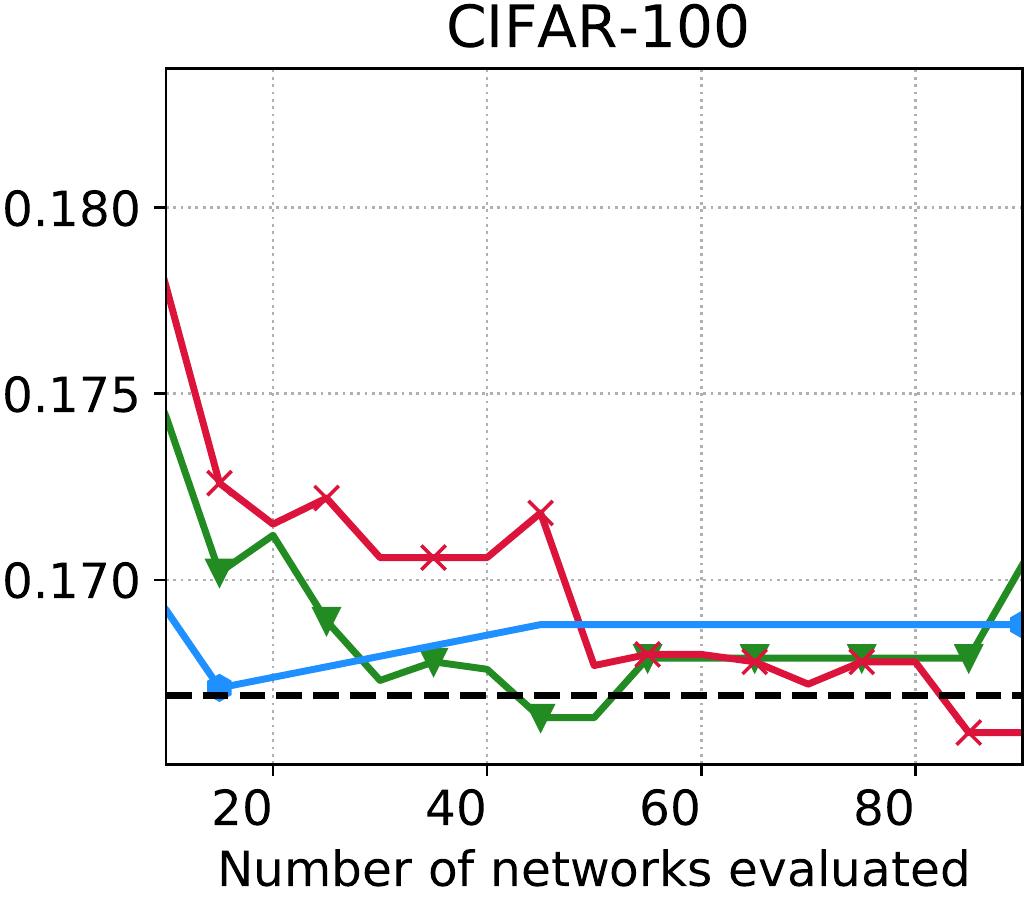}
        \subcaption{No data shift}
        \label{fig:test_error_full_fidelity_0}
    \end{subfigure}%
    ~\hspace{.1cm}
    \begin{subfigure}[t]{0.31\textwidth}
        \centering
        \includegraphics[width=.48\linewidth]{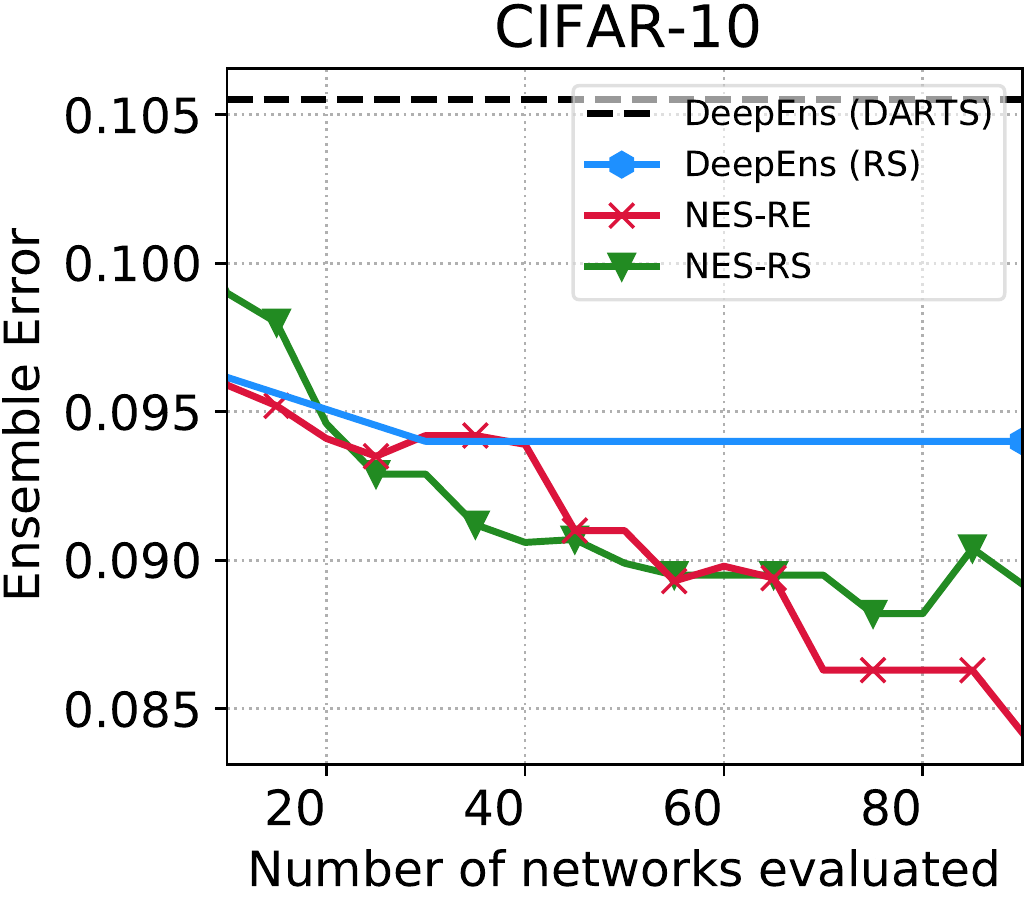}
        \includegraphics[width=.48\linewidth]{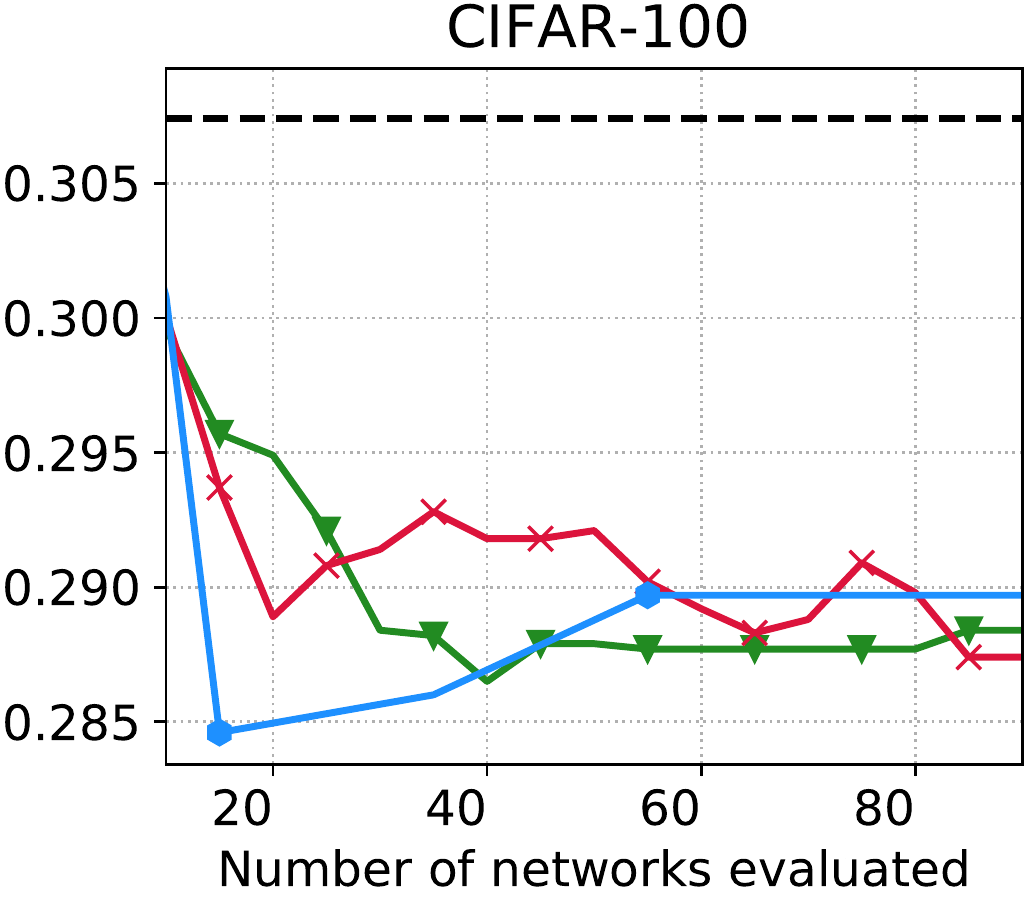}
        \subcaption{Data shift (severity 1)}
        \label{fig:test_error_full_fidelity_1}
    \end{subfigure}
    ~\hspace{.1cm}
    \begin{subfigure}[t]{0.31\textwidth}
        \centering
        \includegraphics[width=.48\linewidth]{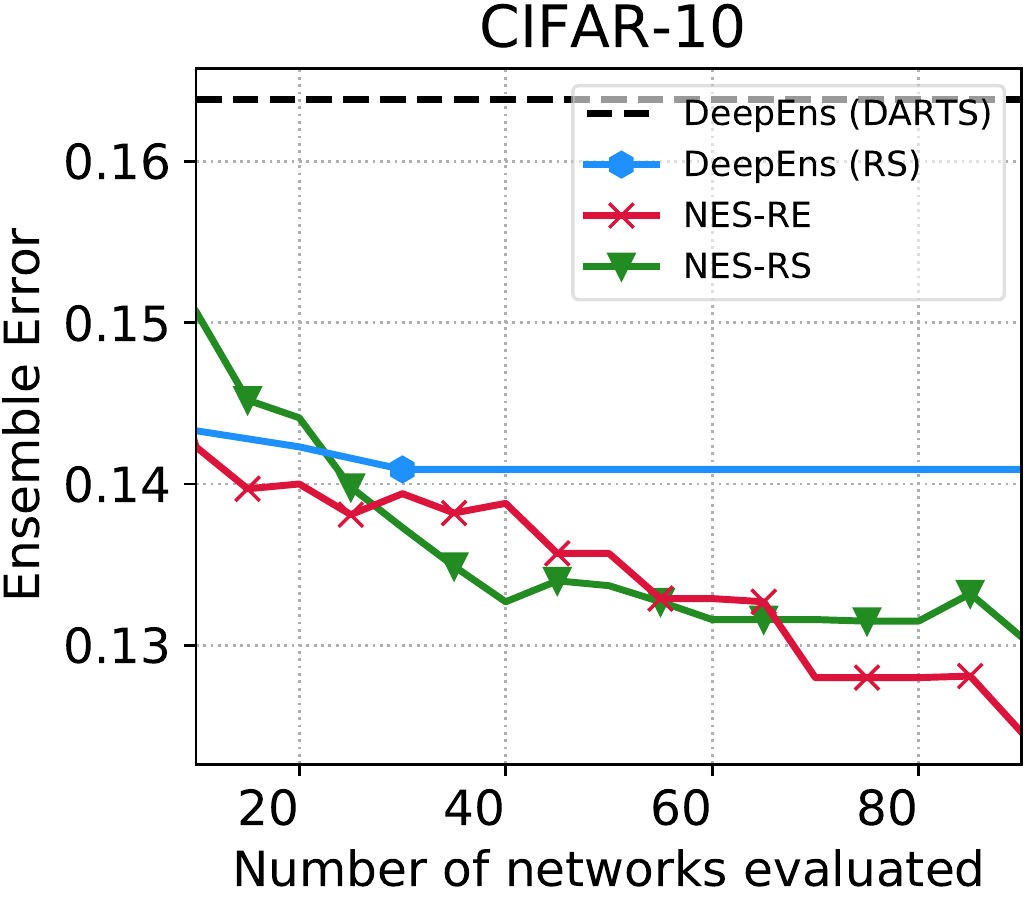}
        \includegraphics[width=.48\linewidth]{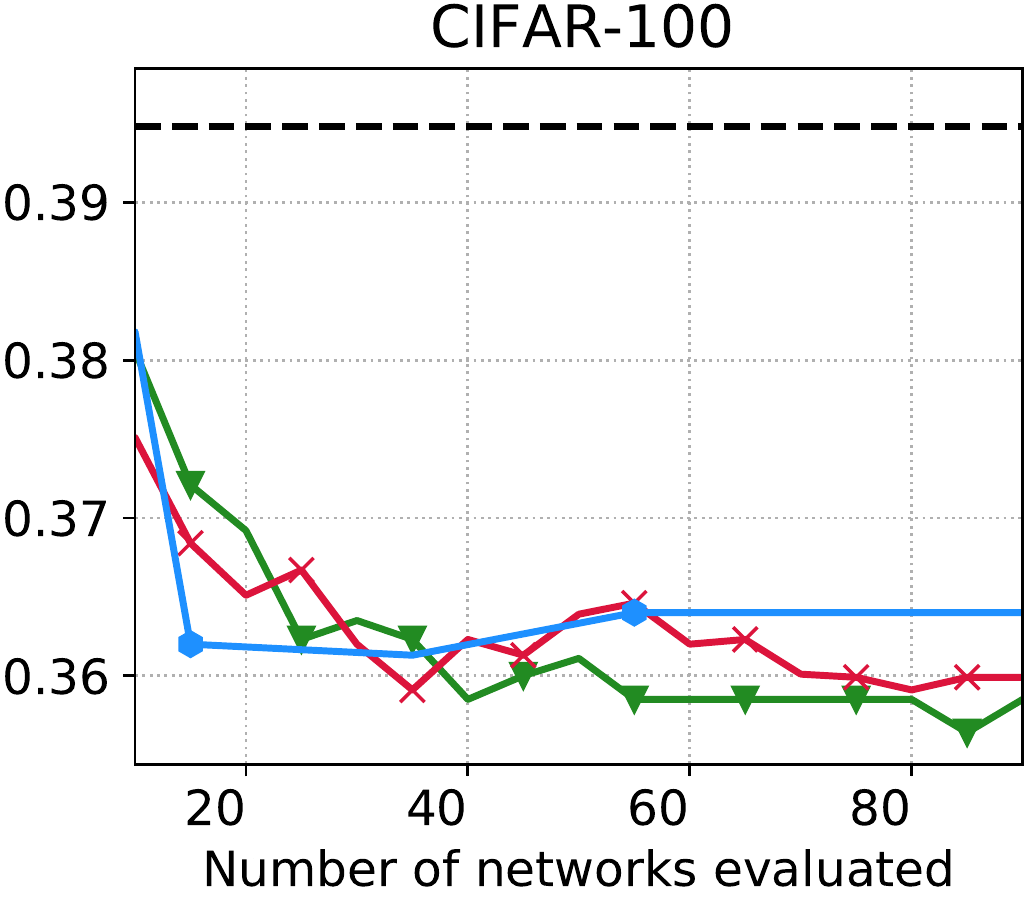}
        \subcaption{Data shift (severity 2)}
        \label{fig:test_error_full_fidelity_2}
    \end{subfigure}\\ %
    \begin{subfigure}[t]{0.31\textwidth}
        \centering
        \includegraphics[width=.48\linewidth]{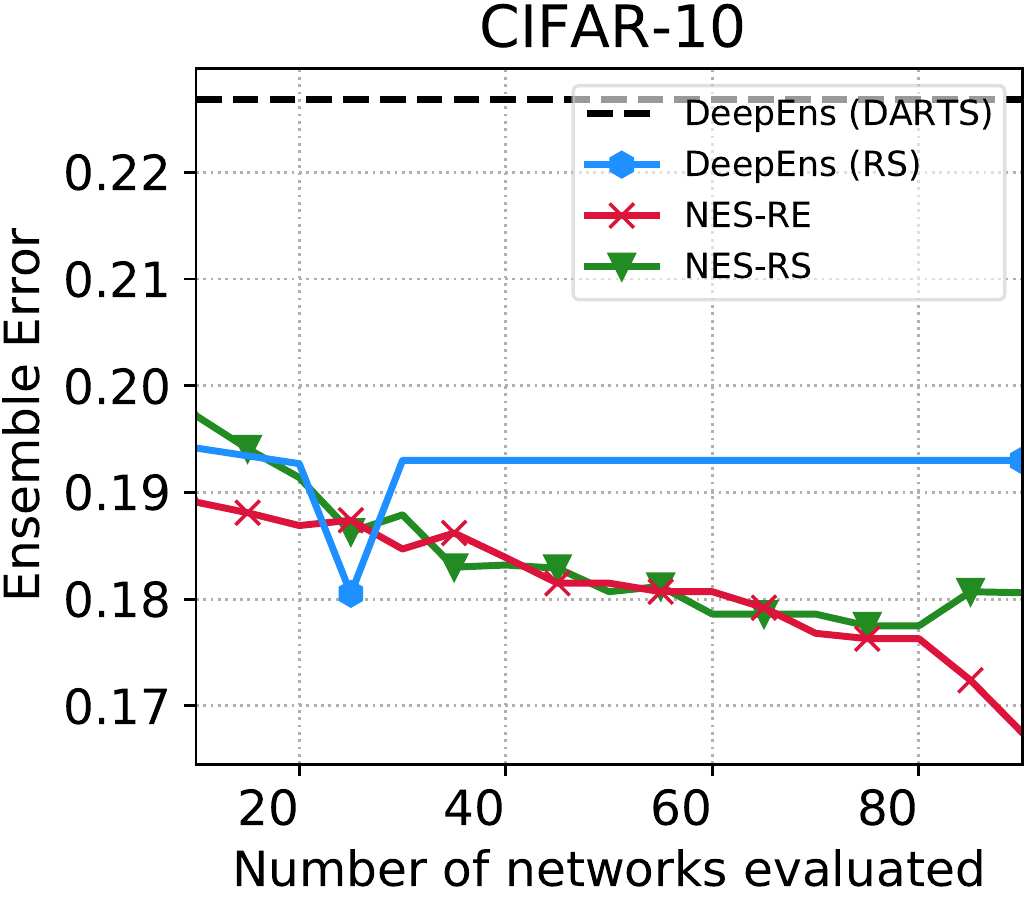}
        \includegraphics[width=.48\linewidth]{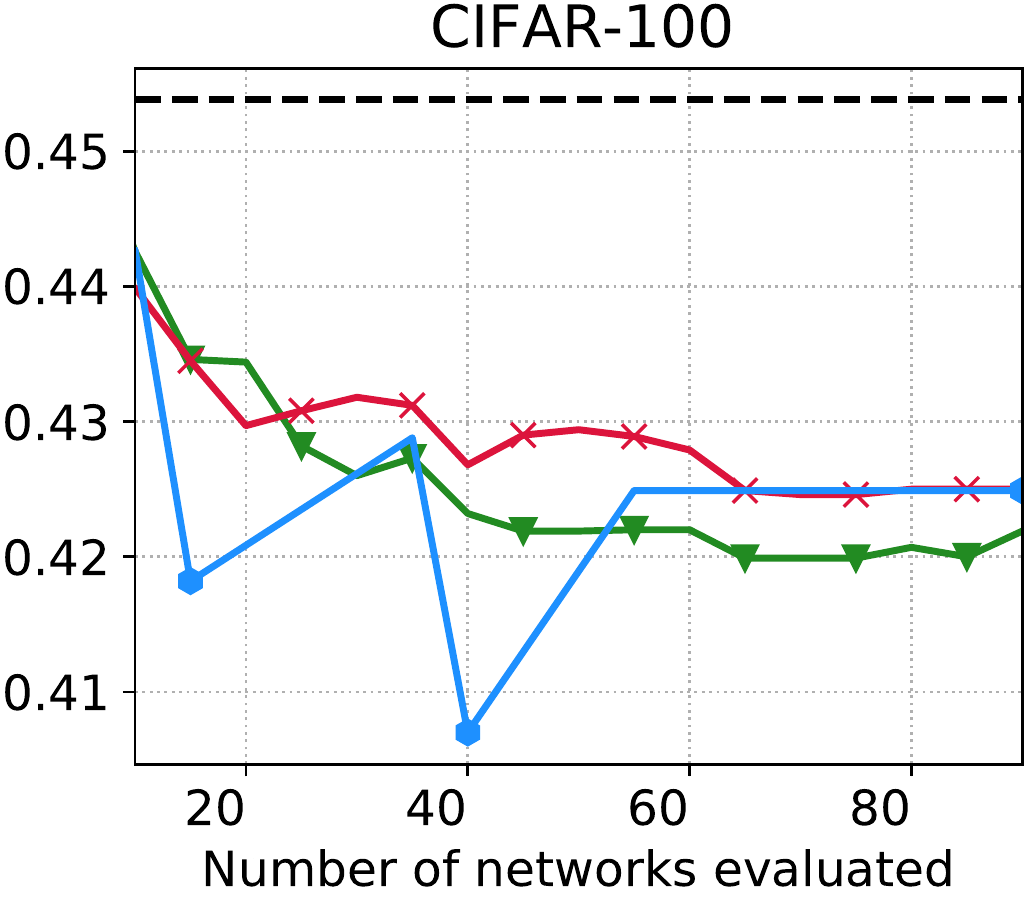}
        \subcaption{Data shift (severity 3)}
        \label{fig:test_error_full_fidelity_3}
    \end{subfigure}%
    ~\hspace{.1cm}
    \begin{subfigure}[t]{0.31\textwidth}
        \centering
        \includegraphics[width=.48\linewidth]{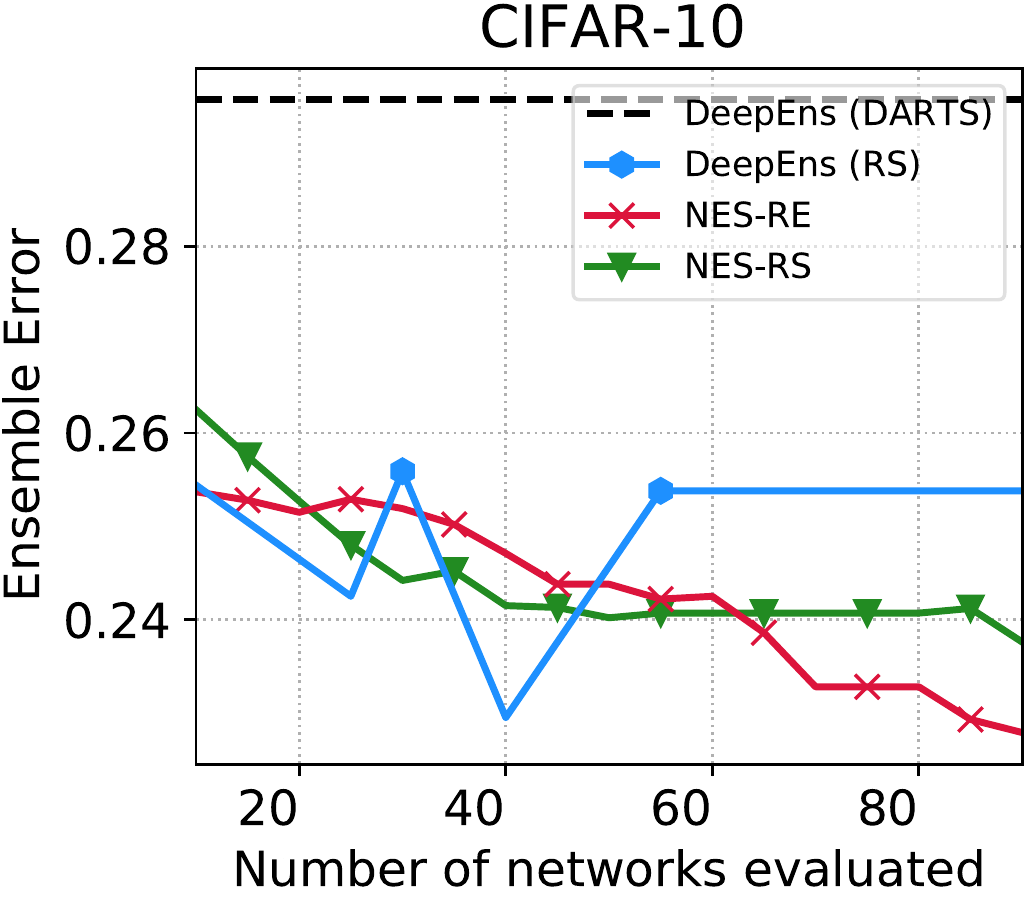}
        \includegraphics[width=.48\linewidth]{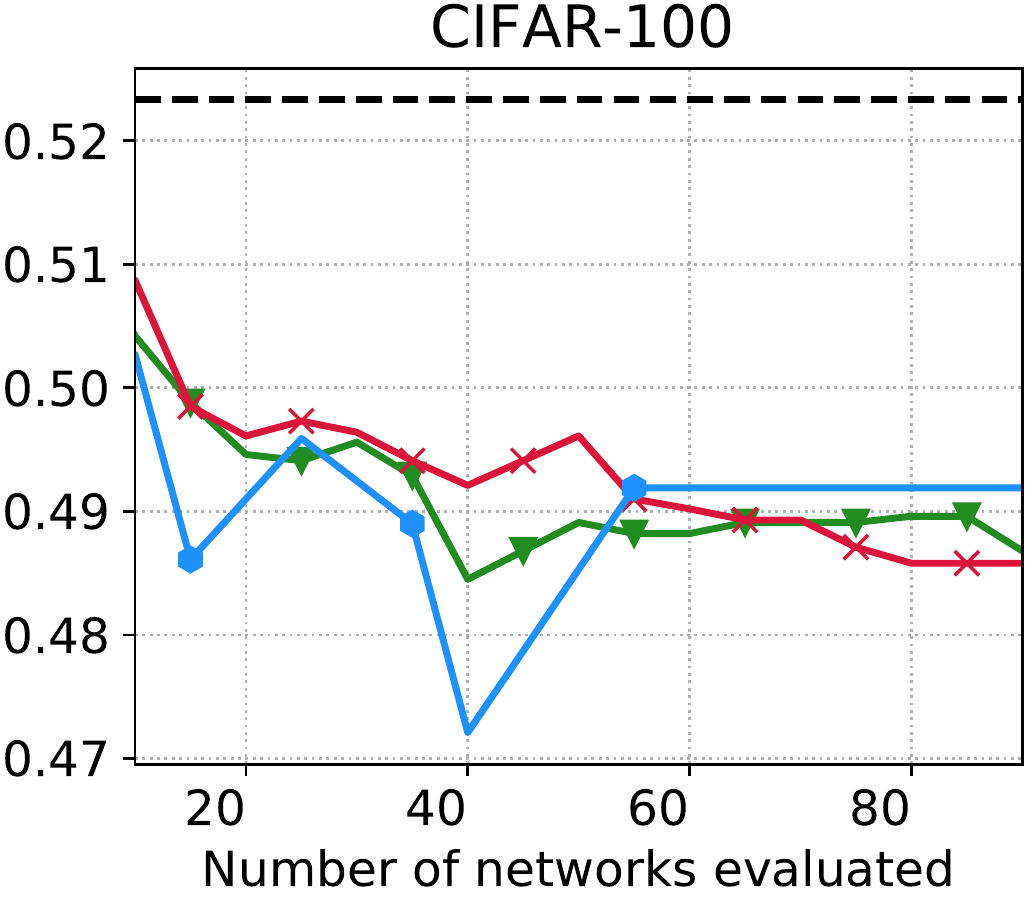}
        \subcaption{Data shift (severity 4)}
        \label{fig:test_error_full_fidelity_4}
    \end{subfigure}
    ~\hspace{.1cm}
    \begin{subfigure}[t]{0.31\textwidth}
        \centering
        \includegraphics[width=.48\linewidth]{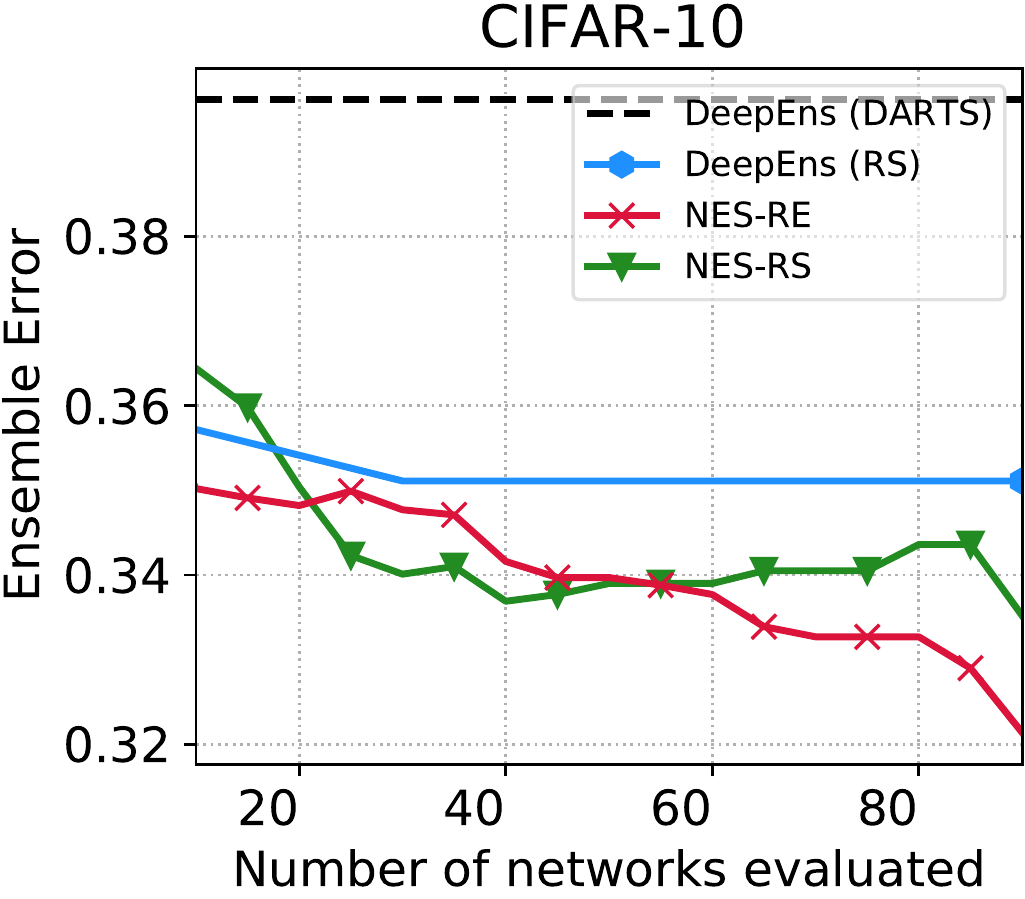}
        \includegraphics[width=.48\linewidth]{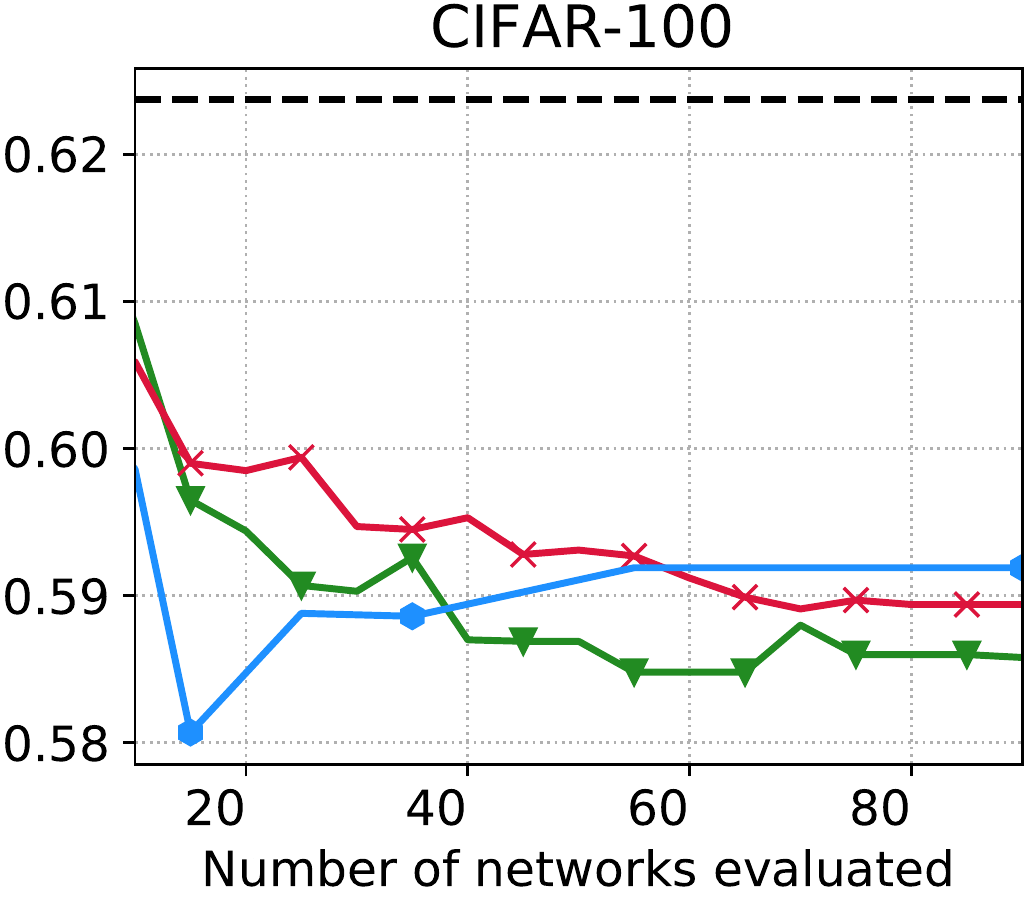}
        \subcaption{Data shift (severity 5)}
        \label{fig:test_error_full_fidelity_5}
    \end{subfigure}
    
    \caption{High fidelity classification error vs. budget $\budget$ on CIFAR-10 and CIFAR-100 with and without respective dataset shifts over the DARTS search space. Ensemble size is fixed at $M = 10$.}
    \label{fig:test_error_full_fid}
\end{figure*}